\documentclass[lettersize,journal]{IEEEtran}
\usepackage{amssymb,amsthm,amsmath}
\usepackage{bm,bbm}
\usepackage[nocompress]{cite}
\usepackage{float}
\usepackage{graphicx}
\usepackage{latexsym}
\usepackage{multirow}
\usepackage{newtxmath}
\usepackage{hyperref}
\usepackage{url}
\usepackage{booktabs}
\usepackage{array}
\usepackage{xcolor}
\hypersetup{bookmarksnumbered=true, bookmarksopen=true, colorlinks=true, linkcolor=blue, citecolor=blue, urlcolor=blue, }
\newcolumntype{C}[1]{>{\hfil}m{#1}<{\hfil}}
\newtheorem{definition}{Definition}
\newtheorem{theorem}{Theorem}
\newtheorem{corollary}{Corollary}

\newcommand{\tcr}[1]{{\textcolor{red}{#1}}}
\definecolor{midori}{rgb}{0.0, 0.5, 0.0}
\newcommand{\tcg}[1]{{\textcolor{midori}{#1}}}
\newcommand{\tcb}[1]{{\textcolor{blue}{#1}}}
\newcommand{\tcc}[1]{{\textcolor{cyan}{#1}}}
\definecolor{gold}{rgb}{1.0, 0.843137, 0.0}
\newcommand{\tcy}[1]{{\textcolor{gold}{#1}}}
\newcommand{\tcm}[1]{{\textcolor{magenta}{#1}}}
\definecolor{brown}{rgb}{0.54296875, 0.26953125, 0.07421875}
\newcommand{\tco}[1]{{\textcolor{brown}{#1}}}
\definecolor{gray}{rgb}{0.40, 0.40, 0.40}
\newcommand{\tca}[1]{{\textcolor{gray}{#1}}}
\renewcommand{\sl}{{\rm sl}}
\newcommand{\vsl}{{\rm vsl}}
\newcommand{\maul}{{\rm maul}}
\newcommand{\ul}{{\rm ul}}
\newcommand{\cl}{{\rm cl}}

\newcommand{\tot}{{\rm tot}}
\newcommand{\tra}{{\rm tra}}
\newcommand{\val}{{\rm val}}

\DeclareMathOperator{\argmin}{{arg\,min}}
\DeclareMathOperator{\argmax}{{arg\,max}}
\newcommand{\CPD}{{\mathbf{CPD}}}

\newcommand{\bg}{{\bm{g}}}

\newcommand{\bp}{{\bm{p}}}
\newcommand{\bq}{{\bm{q}}}

\newcommand{\bu}{{\bm{u}}}
\newcommand{\bv}{{\bm{v}}}

\newcommand{\bx}{{\bm{x}}}
\newcommand{\bX}{{\bm{X}}}

\newcommand{\rmH}{{\mathrm{H}}}

\newcommand{\calG}{{\mathcal{G}}}
\newcommand{\calR}{{\mathcal{R}}}

\DeclareMathOperator{\bbE}{{\mathbb{E}}}
\DeclareMathOperator{\bbI}{{\mathbbm{1}}}
\newcommand{\bbR}{{\mathbb{R}}}
\newcommand{\bbN}{{\mathbb{N}}}

\begin{document}
\title{Approximately Unimodal Likelihood Models\\for Ordinal Regression}
\author{Ryoya Yamasaki%
\IEEEcompsocitemizethanks{\IEEEcompsocthanksitem%
Ryoya Yamasaki is affiliated with 
Hitotsubashi Institute for Advanced Study, 
Hitotsubashi University, 
2-1 Naka, Kunitachi, Tokyo 186-8601 Japan. 
E-mail: ryoya.yamasaki@r.hit-u.ac.jp.
\copyright 2025 IEEE. 
Personal use of this material is permitted. 
Permission from IEEE must be obtained for all other uses, in any current or future media, including reprinting/republishing this material for advertising or promotional purposes, creating new collective works, for resale or redistribution to servers or lists, or reuse of any copyrighted component of this work in other works.}}
\markboth{Postprint version}
{Shell \MakeLowercase{\textit{et al.}}: Approximately Unimodal Likelihood Models for Ordinal Regression}
\maketitle
\begin{abstract}
Ordinal regression (OR, also called ordinal classification) 
is classification of ordinal data, in which the underlying 
target variable is categorical and considered to have a natural 
ordinal relation for the underlying explanatory variable.
A key to successful OR models is to find a data structure 
`natural ordinal relation' common to many ordinal data 
and reflect that structure into the design of those models.
A recent OR study found that many real-world ordinal data show a tendency 
that the conditional probability distribution (CPD) of the target variable 
given a value of the explanatory variable will often be unimodal.
Several previous studies thus developed unimodal likelihood models, 
in which a predicted CPD is guaranteed to become unimodal.
However, it was also observed experimentally that many 
real-world ordinal data partly have values of the explanatory 
variable where the underlying CPD will be non-unimodal, 
and hence unimodal likelihood models may suffer from a bias for such a CPD.
Therefore, motivated to mitigate such a bias, 
we propose approximately unimodal likelihood models, 
which can represent up to a unimodal CPD and a CPD that is close to be unimodal.
We also verify experimentally that a proposed model can be 
effective for statistical modeling of ordinal data and OR tasks.
\end{abstract}
\begin{IEEEkeywords}
Ordinal regression, ordinal data, unimodality hypothesis, likelihood model
\end{IEEEkeywords}
\section{Introduction}
\label{sec:Introduction}
\IEEEPARstart{O}{rdinal} regression (OR, also called ordinal classification) 
is classification of ordinal data in which the underlying 
target variable is labeled from a categorical label set 
(ordinal scale) that is considered to have a natural ordinal 
relation for the underlying explanatory variable.
The ordinal scale is typically formed as a graded (interval) 
summary of objective indicators like age groups \{`0 to 9 years old', 
`10 to 19 years old', \ldots, `90 to 99 years old', `over 100 years old'\} 
or graded evaluation of subjectivity like human rating 
\{`excellent', `good', `average', `bad', `terrible'\} 
(a sort of Likert scale \cite{likert1932technique}), 
and ordinal data appear in various applications such as face-age estimation 
\cite{niu2016ordinal, yamasaki2023optimal}, 
disease stage estimation \cite{medmnistv2},
monocular depth estimation \cite{fu2018deep},
information retrieval \cite{liu2009learning}, 
and analysis of rating data \cite{kim2012corporate, yu2006collaborative} and 
questionnaire survey in social research \cite{chen1995response, burkner2019ordinal}.

Consider an example of the questionnaire survey about rating a certain item 
that requires subjects to respond from \{`excellent', `good', `average', `bad', `terrible'\}.
There, subjects with features specific to those who typically respond `good'
might respond `average', but would be less likely to respond `terrible'.
Such a data structure can be rephrased as 
the unimodality of the conditional probability distribution (CPD) 
of the target variable given a value of the explanatory variable.
A recent work \cite{yamasaki2022unimodal} 
supposed the unimodality hypothesis, 
which states that the CPD of the target variable given 
a value of the explanatory variable will often be unimodal, 
as a characterization of the natural ordinal relation of ordinal data, 
and verified that this hypothesis holds well for many real-world data 
that previous OR studies have treated as ordinal data 
(see `UR' in Table~\ref{tab:Data-Prop} or further explanation in Section~\ref{sec:Data}).
Many OR users may often judge that the data have 
a natural ordinal relation and decide to treat them within the OR framework, 
with (unconsciously) expecting their unimodality.

\begin{table}[!t]
\centering%
\renewcommand{\tabcolsep}{3pt}%
\caption{%
Properties of real-world ordinal data: 
`dataset' shows the dataset name used in this paper, 
$n_\tot$ is the total number of used data points, 
$d$ is the dimension of the explanatory variable, 
$K$ is the number of classes of the target variable, 
and `UR' and `MHD' show the mean and standard deviation (STD) 
of 100-trial (for SB datasets) or 5-trial (for CV datasets) estimates of 
UR (see \eqref{eq:UR}) and MHD (indicator of UD; see \eqref{eq:MHD}) 
as `mean$\pm$STD'.
Also, `uniform on $\Delta_{K-1}$' is of 100-trial estimates 
for uniform random data on $\Delta_{K-1}$ 
(instead of the CPD $(\Pr(Y=y|\bX=\bx))_{y\in[K]}$ in \eqref{eq:UR} 
and \eqref{eq:MHD} for real-world ordinal data).
The larger UR or the smaller MHD, 
the stronger the tendency for the data to be unimodal.}
\label{tab:Data-Prop}
{\footnotesize\renewcommand{\arraystretch}{0.75}%
\begin{tabular}{cC{58pt}ccccc}\toprule
\multicolumn{2}{c}{dataset}& $n_\tot$ & $d$ & $K$ & UR & MHD \\\midrule
\multirow{21}{*}{\rotatebox{90}{SB~~~~~~}}
&BA5' & 8192 & 32 & 5 & $.9999\!\pm\!{.0006}$ & $.0000\!\pm\!{.0000}$ \\
&SWD & 1000 & 10 & 4 & $.9996\!\pm\!{.0031}$ & $.0000\!\pm\!{.0002}$ \\
&WQR & 1599 & 11 & 6 & $.9959\!\pm\!{.0186}$ & $.0000\!\pm\!{.0001}$ \\
&CO5' & 8192 & 21 & 5 & $.9958\!\pm\!{.0250}$ & $.0000\!\pm\!{.0000}$ \\
&CO5 & 8192 & 12 & 5 & $.9889\!\pm\!{.0459}$ & $.0001\!\pm\!{.0003}$ \\
&CE5' & 22784 & 16 & 5 & $.9814\!\pm\!{.0366}$ & $.0004\!\pm\!{.0007}$ \\
&BA5 & 8192 & 8 & 5 & $.9760\!\pm\!{.0894}$ & $.0000\!\pm\!{.0001}$ \\
&LEV & 1000 & 4 & 5 & $.9594\!\pm\!{.0570}$ & $.0003\!\pm\!{.0007}$ \\
&CAR & 1728 & 21 & 4 & $.9380\!\pm\!{.1693}$ & $.0003\!\pm\!{.0010}$ \\
&CH5 & 20640 & 8 & 5 & $.9334\!\pm\!{.0924}$ & $.0009\!\pm\!{.0017}$ \\
&CE5 & 22784 & 8 & 5 & $.8866\!\pm\!{.1102}$ & $.0020\!\pm\!{.0024}$ \\
&BA10 & 8192 & 8 & 10 & $.8831\!\pm\!{.2141}$ & $.0000\!\pm\!{.0001}$ \\
&AB5 & 4177 & 10 & 5 & $.8785\!\pm\!{.1024}$ & $.0017\!\pm\!{.0022}$ \\
&CO10' & 8192 & 21 & 10 & $.8605\!\pm\!{.1948}$ & $.0003\!\pm\!{.0009}$ \\
&CO10 & 8192 & 12 & 10 & $.8308\!\pm\!{.1917}$ & $.0009\!\pm\!{.0019}$ \\
&CE10' & 22784 & 16 & 10 & $.7239\!\pm\!{.2580}$ & $.0017\!\pm\!{.0023}$ \\
&ERA & 1000 & 4 & 9 & $.7122\!\pm\!{.1581}$ & $.0048\!\pm\!{.0049}$ \\
&BA10' & 8192 & 32 & 10 & $.6805\!\pm\!{.3600}$ & $.0012\!\pm\!{.0018}$ \\
&CH10 & 20640 & 8 & 10 & $.5086\!\pm\!{.3127}$ & $.0061\!\pm\!{.0072}$ \\
&CE10 & 22784 & 8 & 10 & $.4535\!\pm\!{.2263}$ & $.0058\!\pm\!{.0052}$ \\
&AB10 & 4177 & 10 & 10 & $.3232\!\pm\!{.1989}$ & $.0105\!\pm\!{.0086}$ \\
\midrule
\multirow{3}{*}{\rotatebox{90}{CV~\,}}
&AFAD & 164418 & $3\!\cdot\!128^2$ & 26 & $.0006\!\pm\!{.0009}$ & $.0148\!\pm\!{.0022}$ \\
&CACD & 159370 & $3\!\cdot\!128^2$ & 49 & $.0002\!\pm\!{.0001}$ & $.0106\!\pm\!{.0023}$ \\
&MORPH & 55013 & $3\!\cdot\!128^2$ & 55 & $.0075\!\pm\!{.0088}$ & $.0069\!\pm\!{.0015}$ \\
\midrule
\multirow{8}{*}{\rotatebox{90}{synthesis~~~~}}
&uniform on $\Delta_3$ & $1000$ & - & 4 & $.3326\!\pm\!{.0135}$ & $.0752\!\pm\!{.0026}$ \\
&\hphantom{unifh}''\hphantom{i on} $\Delta_4$ & '' & - & 5 & $.1314\!\pm\!{.0108}$ & $.1000\!\pm\!{.0023}$ \\
&\hphantom{unifh}''\hphantom{i on} $\Delta_5$ & '' & - & 6 & $.0443\!\pm\!{.0065}$ & $.1162\!\pm\!{.0023}$ \\
&\hphantom{unifh}''\hphantom{i on} $\Delta_8$ & '' & - & 9 & $.0009\!\pm\!{.0010}$ & $.1365\!\pm\!{.0016}$ \\
&\hphantom{unifh}''\hphantom{i on} $\Delta_9$ & '' & - & 10 & $.0001\!\pm\!{.0003}$ & $.1385\!\pm\!{.0014}$ \\
&\hphantom{unifh}''\hphantom{i on} $\Delta_{25}$ & '' & - & 26 & $.0000\!\pm\!{.0000}$ & $.1294\!\pm\!{.0007}$ \\
&\hphantom{unifh}''\hphantom{i on} $\Delta_{48}$ & '' & - & 49 & $.0000\!\pm\!{.0000}$ & $.1093\!\pm\!{.0004}$ \\
&\hphantom{unifh}''\hphantom{i on} $\Delta_{54}$ & '' & - & 55 & $.0000\!\pm\!{.0000}$ & $.1053\!\pm\!{.0004}$ \\
\bottomrule\end{tabular}}
\end{table}

\begin{figure}[!t]
\centering%
\includegraphics[height=4.25cm, bb=0 0 603 299]{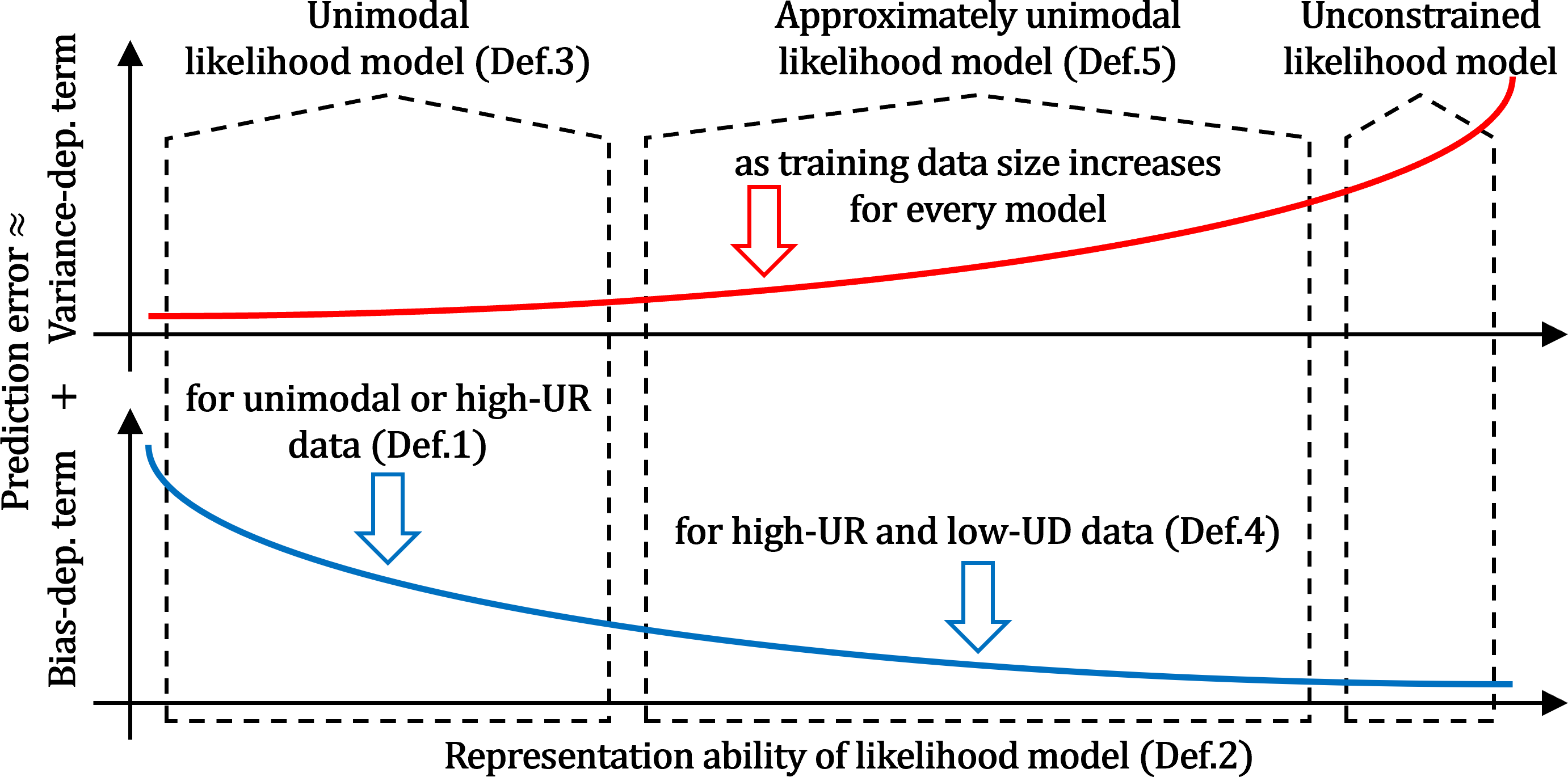}
\caption{%
Illustration of the idea of this study:
Under the supposition that many ordinal data 
are not only high-UR but also low-UD, 
we develop approximately unimodal likelihood models, 
which can represent low-UD data to mitigate 
a bias of unimodal likelihood models 
that can represent only unimodal data, 
and which have a lower representation ability than 
unconstrained likelihood models to decrease a variance 
that becomes relatively large within 
the prediction error for small-size training data.}
\label{fig:Idea}
\end{figure}

Generally, prediction performance of a statistical model depends on 
the underlying data distribution, size of training data, 
representation ability of that model, and so on.
Recall that prediction performance can be roughly 
decomposed into bias- and variance-dependent terms, 
that is, bias-variance trade-off \cite{bishop2006pattern}
(for simplicity, we ignore the optimization error \cite{Bottou}
in the discussion of this paper):
as Figure~\ref{fig:Idea} schematizes,
a likelihood model that has a strongly constrained representation ability 
will result in a large bias if it cannot represent the underlying data distribution,
and a model that is unnecessarily flexible to represent the underlying 
data distribution will result in a large variance especially when 
the size of training data is small.
Considering that unimodality constraint on the likelihood model can 
restrict the representation ability of the model while still allowing for 
good representation of ordinal data under the unimodality hypothesis,
previous studies \cite{da2008unimodal, iannario2011cub, beckham2017unimodal, 
yamasaki2022unimodal} developed unimodal likelihood models, 
in which a predicted CPD is guaranteed to become unimodal.
Especially, the previous study \cite{yamasaki2022unimodal} developed a variety of 
unimodal likelihood models with various levels of representation ability,
including the most representable one that can represent arbitrary unimodal CPD, 
and showed that those models can perform better than unconstrained models 
such as multinomial logistic regression (MLR) model
in modeling of real-world ordinal data 
and OR tasks with small-size training data.

However, it was also observed in experiments of \cite{yamasaki2022unimodal}
that many real-world ordinal data partly have values of the explanatory variable 
where the underlying CPD will not be unimodal 
(as `UR' in Table~\ref{tab:Data-Prop} is not strictly 1), 
and hence a unimodal likelihood model has a bias from such a CPD, 
which may be a performance catch.
One promising approach to mitigate 
such a bias is to strengthen the representation ability 
from that of unimodal likelihood models in a way appropriate for ordinal data, 
but what are properties of a non-unimodal CPD of ordinal data?
In this study, we suppose that 
the underlying CPD of ordinal data should be close to be unimodal 
even at points where the CPD is not unimodal strictly,
and verify this hypothesis experimentally 
(see `MHD' in Table~\ref{tab:Data-Prop},
or Figure~\ref{fig:Data-Unimodality} and further explanation in Section~\ref{sec:Proposed}).
Therefore, in order to mitigate a bias of unimodal likelihood models, 
we develop likelihood models that we call approximately unimodal likelihood models 
and that can represent up to a unimodal CPD and a CPD that is close to be unimodal:
we construct them as a mixture of a unimodal likelihood model 
and unconstrained likelihood model.
In this paper, we also present theorems 
(Theorem~\ref{def:MAUL} and Corollary~\ref{cor:MAUL})
on the representation ability of proposed models, 
and experimentally show the effectiveness of a proposed model 
for statistical modeling of real-world ordinal data and OR tasks
with small-size training data.

The rest of this paper is organized as follows:
Section~\ref{sec:Preparation} describes preparation for the discussion of this study,
including formulation of ordinal data, unimodality hypothesis,
and likelihood model and its representation ability.
In Section~\ref{sec:Related}, we review 
unimodal likelihood models developed in previous works,
particularly in \cite{yamasaki2022unimodal}.
In Section~\ref{sec:Proposed}, we describe proposed models,
mixture-based approximately unimodal likelihood models,
and show their theoretical properties.
We provide numerical experiments to show the effectiveness of 
a proposed model for statistical modeling of ordinal data and OR tasks 
in Section~\ref{sec:Experiments}.
Also, we give a discussion on an approach 
to leverage the unimodality of ordinal data that is different 
from our approach of devising a likelihood model
in Section~\ref{sec:Discussion}, 
and present numerical experiments to show relationships 
between that approach and the proposed model
in Section~\ref{sec:ExperimentsII}.
Refer to also supplement for supplemental experiments 
and \url{https://github.com/yamasakiryoya/AULM} for used program codes.
Finally, Section~\ref{sec:Conclusion} concludes this paper
and presents a future research direction.

\section{Preparation}
\label{sec:Preparation}
\subsection{Ordinal Data and Unimodality Hypothesis}
\label{sec:Data}
Suppose that one has data $(\bx_1,y_1),\cdots,(\bx_n,y_n)$ 
that are assumed to be drawn independently from 
an identical distribution of $(\bX,Y)\in\bbR^d\times[K]$, 
where $n,d,K\in\bbN$ such that $K\ge3$, 
and where $[K]\coloneq\{1,\ldots,K\}$.
In the OR framework, suppose further that 
the categorical target variable $Y$ is considered to have 
a natural ordinal relation for the underlying explanatory variable $\bX$
like examples described in the head of Section~\ref{sec:Introduction},
and such data are called ordinal data (or ordinal categorical data).
However, there is no more strict definition or unified understanding 
of the natural ordinal relation among previous OR studies, 
and each statistician and each practitioner have referred to data that 
they judged to have a natural ordinal relation as ordinal data.
Thus, in this paper, we refer to data, which previous OR 
studies have treated as ordinal data, as ordinal data.

However, for systematic discussion of the OR task, it would be 
important to formally define the natural ordinal relation that is likely 
to be common in many ordinal data and that we suppose in the work.
As a characterization of the natural ordinal relation of ordinal data,
several previous OR studies \cite{da2008unimodal, iannario2011cub, 
beckham2017unimodal, yamasaki2022unimodal, yamasaki2024remarks}
and this study also suppose the unimodality hypothesis, 
stating that many real-world ordinal data 
would be unimodal or high-UR (almost-unimodal in 
\cite{yamasaki2022unimodal}) according to the following definition:
\begin{definition}[{Unimodal/high-UR data}]
\label{def:unimodal}
For a probability mass function (PMF) $\bp=(p_k)_{k\in[K]}\in\Delta_{K-1}$, 
we call every $M\in\argmax_k(p_k)_{k\in[K]}$ a {mode} of $\bp$, 
and say that $\bp$ is {unimodal} if it satisfies
\begin{align}
\label{eq:unimodality}
	p_1\le\cdots\le p_M\text{~and~}p_M\ge\cdots\ge p_K
\end{align}
with any mode $M$, where $\Delta_{K-1}$ is the $(K-1)$-dimensional probability simplex 
$\{(p_k)_{k\in[K]}\in\bbR^K\mid\sum_{k=1}^K p_k=1,\,p_k\in[0,1]\text{ for all }k\in[K]\}$.
Also, we further introduce the set
$\hat{\Delta}_{K-1}\coloneq\{\bp\in\Delta_{K-1}\mid \bp\text{ is unimodal.}\}$
of unimodal PMFs.
Moreover, if the unimodality rate (UR) of the CPD, 
\begin{align}
\label{eq:UR}
	\bbE_{\bx\sim\bX}[\bbI\{(\Pr(Y=y|\bX=\bx))_{y\in[K]}\in\hat{\Delta}_{K-1}\}],
\end{align}
is 1 or high,
we say that the data is unimodal or high-UR, 
where $\bbE_{\bx\sim\bX}[\cdot]$ is the expectation over $\bx\sim\bX$,
and where$\bbI\{c\}$ is the indicator function that 
values 1 if the condition $c$ is true or 0 otherwise.
\end{definition}

The OR study \cite{yamasaki2022unimodal} verified that 
the unimodality hypothesis holds well for many real-world data 
that previous OR studies have treated as ordinal data.
For numerical experiments, it used 21 real-world datasets 
(which we call small-size benchmark (SB) datasets) with 
the total data size $n_\tot$ that is 1000 or more among those used in 
experiments by the previous OR study \cite{gutierrez2015ordinal}.
{AB5}, \ldots, {CO5'} (resp.\ {AB10}, \ldots, {CO10'}) are datasets 
generated by discretizing a real-valued target of datasets, 
which are often used to benchmark regression methods, 
by 5 (resp.\,10) different bins with equal proportions.
{CAR}, {ERA}, {LEV}, {SWD}, and {WQR} originally have a categorical target, 
and the authors of \cite{gutierrez2015ordinal} judged 
that these data have a natural ordinal relation.
We are also interested in the unimodality of 
ordinal data used for modern applications,
so we further tackled the problem of predicting the age 
from a facial image using AFAD \cite{niu2016ordinal}, CACD 
\cite{chen2014cross}, and MORPH \cite{ricanek2006morph} datasets
(we refer to these 3 datasets as computer vision (CV) datasets).%
\footnote{%
One can get SB datasets from a researchers'\,site 
(\url{https://www.uco.es/grupos/ayrna/orreview}) 
of \cite{gutierrez2015ordinal} or GitHub repository 
(\url{https://github.com/yamasakiryoya/AULM}) for our study, 
and sources of CV datasets were
\url{https://github.com/afad-dataset/tarball},
\url{https://bcsiriuschen.github.io/CARC}, 
and \url{https://ebill.uncw.edu/C20231_ustores/web}.}
See supplement for more detailed explanation of SB and CV datasets.
As in \cite{yamasaki2022unimodal}, we gave sample-based estimation 
of the UR \eqref{eq:UR} in Table~\ref{tab:Data-Prop}, 
where we used an unconstrained likelihood model 
(SL model in Section~\ref{sec:Settings} with $n_\tra=800$ for SB datasets 
or $n_\tra=40000$ for CV datasets) for estimation of the CPD.
Considering that, 
for standard classification data with no natural ordinal relation,
it would be permissible to assign labels $1,\ldots,K$ 
to observations of the target variable in any order,
we also calculated a sample-based estimate of the UR for
the uniform random variable on $\Delta_{K-1}$ 
as a simulation of the CPD of standard classification data.%
\footnote{%
We sampled uniform random data 
with \texttt{numpy.random.dirichlet}
in Python and NumPy, which utilizes the relation that 
$(Z_1,\ldots,Z_K)^\top/\sum_{k=1}^KZ_k$ is 
a Dirichlet random variable on $\Delta_{K-1}$ of 
concentration parameters $\alpha_1$, $\ldots,\alpha_K$
(which is a uniform random variable when $\alpha_1=\cdots=\alpha_K=1$)
for gamma random variables $Z_1,\ldots,Z_K$ of 
shape parameters $\alpha_1,\ldots,\alpha_K$ \cite{Devroye1986}.}
Comparing the UR of real-world ordinal data and 
that of uniform random data of the same $K$, one can find that 
many real-world ordinal data enjoy strong tendency of the unimodality.
On the ground of this observation, it can be considered that
many OR users may often judge that the data have 
a natural ordinal relation and decide to treat them within the OR framework, 
with (unconsciously) expecting their unimodality.

\subsection{Ordinal Regression Methods and Representation Ability}
\label{sec:Task}
In this paper, we consider statistical OR methods.
A likelihood model $(P,\calG)$ is employed to estimate 
the underlying conditional probability $\Pr(Y=y|\bX=\bx)$ as 
$\hat{\Pr}(Y=y|\bX=\bx)=P(y;\bg(\bx))$ with a fixed part $P$ 
(say, the softmax function in MLR model) 
and learnable part $\bg\in\calG$ (say, a certain neural network model).
Here, we call $P$ the {link function} and 
$\bg$ the {learner model} in the {learner class} $\calG$.
In a conditional probability estimation task,
we typically train a learner model $\bg$ from a learner class $\calG$ 
by minimizing a distribution-to-distribution divergence-based criterion between 
the empirical distribution based on the data $(\bx_1,y_1),\ldots,(\bx_n,y_n)$
and the likelihood $(P(y;\bg(\cdot)))_{y\in[K]}$.
A popular divergence-based criterion is the negative log likelihood (NLL) 
$-\frac{1}{n}\sum_{i=1}^n\log P(y_i;\bg(\bx_i))$.

OR tasks are typically formulated as searching for a classifier
$f:\bbR^d\to[K]$ that is good in the sense that the task risk 
$\bbE[\ell(f(\bX),Y)]=\bbE_{\bx\sim\bX}[\sum_{y=1}^K\Pr(Y=y|\bX=\bx)\ell(f(\bx),y)]$ 
becomes small for a specified task loss $\ell:[K]^2\to[0,\infty)$.
For OR tasks, popular instances of the task loss 
are $\ell(u,v)=\bbI\{u\neq v\}, |u-v|, (u-v)^2$:
correspondingly, the empirical task risk is called mean zero-one error (MZE),
mean absolute error (MAE), and mean squared error (MSE).
Statistical OR methods typically adopt a likelihood-based classifier 
$f(\bx)\in\argmin_{k\in[K]}\sum_{y=1}^KP(y;\bg(\bx))\ell(k,y)$,
trusting that the likelihood model is learned to be a good estimate of the conditional probability.

Good estimation of the conditional probability will lead to 
good performance of a statistical OR method in OR tasks as well. 
Therefore, we focus our discussion on the conditional probability estimation task.
For the conditional probability estimation task,
a key notion in the discussion of this paper is 
the representation ability of the likelihood model.
The representation ability of the likelihood model we discuss is defined as 
the size of the set (function space) of every CPD that that model can represent:
\begin{definition}[{Representation ability}]
\label{def:RepAbi}
We say that a likelihood model $(P_b,\calG_b)$ 
has a higher representation ability (or is more representable) 
than a likelihood model $(P_a,\calG_a)$ if 
\begin{align}
	\calR(P_a,\calG_a)\subseteq\calR(P_b,\calG_b),
\end{align}
where $\calR(P,\calG)$ for a likelihood model $(P,\calG)$ is defined by
\begin{align}
\label{eq:RA}
	\calR(P,\calG)\coloneq\{(P(y;\bg(\cdot)))_{y\in[K]}\mid \bg\in\calG\}.
\end{align}
\end{definition}

Recall that prediction performance can be roughly 
decomposed into bias- and variance-dependent terms:
a likelihood model that has a strongly constrained representation ability will 
result in a large bias if it cannot represent the underlying data distribution,
and a model that is unnecessarily flexible to represent the underlying data 
distribution will result in a large variance especially for small-size training data.
This relation is known as the bias-variance trade-off 
\cite{bishop2006pattern,Bottou} and suggests that 
likelihood models, which can adequately represent the data 
but have as low a representation ability as possible,
and associated OR methods
are promising for better prediction performance.

\section{Related Works: Unimodal Likelihood Models}
\label{sec:Related}
Under the unimodality hypothesis,
unimodality constraint on the likelihood model 
is thought to restrict the representation ability of the model 
while still allowing for a good representation 
of ordinal data that are high-UR.
Accordingly, several previous studies 
\cite{da2008unimodal, iannario2011cub, beckham2017unimodal, 
yamasaki2022unimodal} developed unimodal likelihood models,
in which a predicted CPD is guaranteed to become unimodal:
\begin{definition}[Unimodal likelihood model]
\label{def:Uni}
We say that a likelihood model $(P,\calG)$ is 
a unimodal likelihood model if 
\begin{align}
	\calR(P,\calG)\subseteq\{\bp:\bbR^d\to\hat{\Delta}_{K-1}\}.
\end{align}
\end{definition}

Especially, the previous work \cite{yamasaki2022unimodal} developed 
more representable unimodal likelihood models than those by 
\cite{da2008unimodal, iannario2011cub, beckham2017unimodal};
those models include VSL model 
(which he called V-shaped stereotype logit (VS-SL) model) 
that has the highest representation ability 
among all those of the unimodal likelihood models.
In the following, we review the VSL model
as a representative unimodal likelihood model.

The VSL model was developed as a modification of 
{SL model} (which is similar to MLR model; refer to OR studies, 
\cite{anderson1984regression} and \cite[Section 4.3]{agresti2010analysis},
for the SL model):
the SL model attempts to model the conditional probabilities of 
multiple categories paired with a certain fixed (stereotype) category,
$\frac{\Pr(Y=1|\bX=\bx)}{\Pr(Y=1|\bX=\bx)+\Pr(Y=y|\bX=\bx)}$, by
\begin{align}
	\frac{\hat{\Pr}(Y=1|\bX=\bx)}{\hat{\Pr}(Y=1|\bX=\bx)+\hat{\Pr}(Y=y|\bX=\bx)}
	=\frac{1}{1+e^{-g_y(\bx)}}
\end{align}
for $y\in[K]$
with a learner model $\bg\in\calG_K$,
where $\calG_k\coloneq\{\bg:\bbR^d\to\bbR^k\}$ for every $k\in\bbN$.
Namely, the SL model is based on the {SL link function} 
\begin{align}
\label{eq:SLFunc}
	P_\sl(y;\bu)
	\coloneq\frac{e^{-u_y}}{\sum_{k=1}^K e^{-u_k}}
	\text{~for~}y\in[K],\bu\in\bbR^K
\end{align}
and a learner class $\calG\subseteq\calG_K$
(or a constrained class in $\{(g_k)_{k\in[K]}\in\calG_K\mid g_1\equiv0\}$).
The SL model is an unconstrained model in that 
it can represent (almost) every CPD as follows:
\begin{theorem}[{Representation ability of the SL model; \cite[Theorem 5]{yamasaki2022unimodal}}]
\label{thm:SL-REP}
It holds that
\begin{align}
\label{eq:SLEQ}
	\calR(P_\sl,\calG_K)=\{\bp:\bbR^d\to\Delta_{K-1}^\emptyset\},
\end{align}
where $S^\emptyset\coloneq \{(p_k)_{k\in[K]}\in S\mid p_k\neq0\text{ for all }k\in[K]\}$
for a set $S\subseteq\Delta_{K-1}$.
\end{theorem}

Note that exclusion of the border 
$\{(p_k)_{k\in[K]}\in\Delta_{K-1}\mid p_k=0\text{ for some }k\in[K]\}$
of $\Delta_{K-1}$ (which we call the border issue)
comes from using the SL link function, 
but has the advantage of making the NLL less likely to diverge (or take NaN).

\begin{figure}[!t]
\centering%
\includegraphics[height=4.25cm, bb=0 0 907 434]{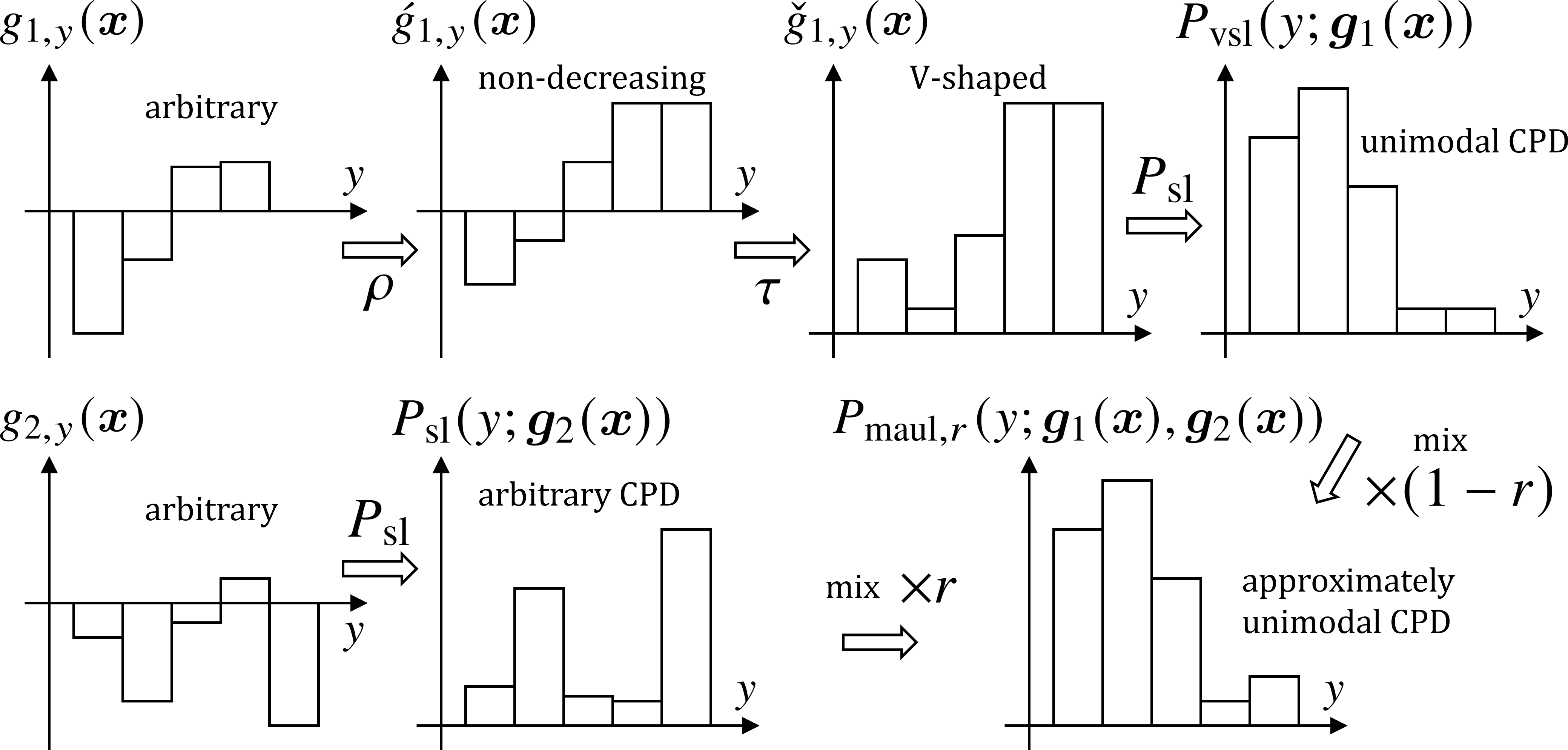}
\caption{%
Illustration of models:
The upper row shows the output flow of a unimodal VSL
model \eqref{eq:VSLLink}, the left side of the lower row 
shows that of an unconstrained SL model \eqref{eq:SLFunc},
and a proposed MAUL model \eqref{eq:MAUL} mixes their outputs
to represent an approximately unimodal CPD.}
\label{fig:MAUL}
\end{figure}

The OR work \cite{yamasaki2022unimodal} realized a unimodal likelihood model
imposing V-shaped constraint on the learner under the use of 
the SL link function $P_\sl$, as summarized also in Figure~\ref{fig:MAUL}.
First, with a user-specified non-negative function $\rho$ satisfying
\begin{align}
\label{eq:RHO}
	\{\rho(u)\mid u\in\bbR\}=[0,\infty)
\end{align}
such as $\rho(u)=u^2$,
it transforms an arbitrary model $\bg\in\calG_K$ to an ordered model $\acute{\bg}$ 
such that $\acute{g}_1(\bx)\le\cdots\le\acute{g}_K(\bx)$ for any $\bx\in\bbR^d$ by
\begin{align}
\label{eq:RHOFunc}
	\acute{g}_k(\bx)
	=\begin{cases}
	g_1(\bx)&\text{for }k=1,\\
	\acute{g}_{k-1}(\bx)+\rho(g_k(\bx))&\text{for }k=2,\ldots,K,
	\end{cases}
\end{align}
where we denote this transformation as $\acute{\bg}(\bx)=\rho[\bg(\bx)]$.
Next, with a user-specified V-shaped function $\tau$ satisfying
\begin{align}
	&\tau(u)\text{~is non-increasing in~}u<0\text{~and non-decreasing in~}u>0,
\nonumber\\
\label{eq:TAU}
	&\text{and~}\{\tau(u)\mid u\le0\}=\{\tau(u)\mid u\ge0\}=[\tau(0),\infty),
\end{align}
such as $\tau(u)=|u|$ and $\tau(u)=u^2$,
it transforms an ordered model $\acute{\bg}$ to a V-shaped model $\check{\bg}$ 
such that there exists $m_{\bx}\in[K]$ such that $\check{g}_1(\bx)\ge\cdots\ge\check{g}_{m_\bx}(\bx)$ 
and $\check{g}_{m_\bx}(\bx)\le\cdots\le\check{g}_K(\bx)$ for each $\bx\in\bbR^d$ by
\begin{align}
	\check{g}_k(\bx)=\tau(\acute{g}_k(\bx))\text{ for }k\in[K],
\end{align}
where we denote this transformation as $\check{\bg}(\bx)=\tau(\acute{\bg}(\bx))$.
Finally, the SL link function transforms $\check{\bg}$ to a unimodal likelihood model.
In other words, the VSL model is based on the VSL link function
\begin{align}
\label{eq:VSLLink}
	P_\vsl(y;\bu)=P_\sl(y;\tau(\rho[\bu]))
	\text{~for~}y\in[K],\bu\in\bbR^K
\end{align}
and a learner class $\calG\subseteq\calG_K$.
The VSL model is ensured to be unimodal and further can 
represent arbitrary unimodal CPD ignoring the border issue:
\begin{theorem}[{Representation ability of the VSL model; \cite[Theorem 7]{yamasaki2022unimodal}}]
\label{thm:VSL-REP}
For any functions $\rho$ satisfying \eqref{eq:RHO} and $\tau$ satisfying \eqref{eq:TAU}, 
it holds that
\begin{align}
\label{eq:VSL-REP}
	\calR(P_\vsl,\calG_K)=\{\bp:\bbR^d\to\hat{\Delta}_{K-1}^\emptyset\}.
\end{align}
\end{theorem}

The previous work \cite{yamasaki2022unimodal} showed 
experimentally that unimodal likelihood models performed 
better than unconstrained likelihood models
in statistical modeling of real-world ordinal data 
and OR tasks with small-size training data.

\section{Proposed Models: Mixture-Based Approximately Unimodal Likelihood Models}
\label{sec:Proposed}
Although many real-world ordinal data enjoy strong tendency of the unimodality,
their underlying CPD $(\Pr(Y=y|\bX=\bx))_{y\in[K]}$ 
would be non-unimodal at some points $\bx$
as the UR in Table~\ref{tab:Data-Prop} is not strictly 1.
Therefore, a unimodal likelihood model has a bias from such a CPD.
One promising approach to mitigate such a bias 
is to strengthen the representation ability of 
a likelihood model to use in a way appropriate for ordinal data.
What are properties of a non-unimodal CPD of ordinal data?
We here consider that 
the underlying CPD of ordinal data should be close to be unimodal 
even at points where the CPD is not unimodal strictly.
Formally, we suppose that many real-world ordinal data 
are not only high-UR but also low-UD
according to the following definition:
\begin{definition}[{Low-UD data}]
\label{def:AULM1}
We say that the data is low-UD,
if deviation (which we call unimodality deviation; UD) 
$D(\CPD(\bx))$ of the CPD $(\Pr(Y=y|\bX=\bx))_{y\in[K]}$ 
from the set $\hat{\Delta}_{K-1}$ of unimodal PMFs tends to 
be low at $\bx$ in whole the support of the distribution of $\bX$, 
where the UD should satisfy that
$D(\bp)=\min_{\bq\in\Delta_{K-1}}D(\bq)$ for $\bp\in\hat{\Delta}_{K-1}$ and 
$D(\bp)>\min_{\bq\in\Delta_{K-1}}D(\bq)$ for $\bp\not\in\hat{\Delta}_{K-1}$.
In this paper, we measure the UD by $D_\rmH(\cdot,\hat{\Delta}_{K-1})$, 
where $D_\rmH(\bu,S)\coloneq\min_{\bv\in S}\|\bu-\bv\|$ with the Euclidean norm $\|\cdot\|$
is the Hausdorff distance (HD) between $\bu\in\bbR^K$ and $S\subseteq\bbR^K$.
\end{definition}

\begin{figure}[!t]
\centering%
\renewcommand{\tabcolsep}{0pt}%
\begin{tabular}{C{2.95cm}C{2.95cm}C{2.95cm}}%
\includegraphics[height=1.54cm, bb=0 0 596 327]{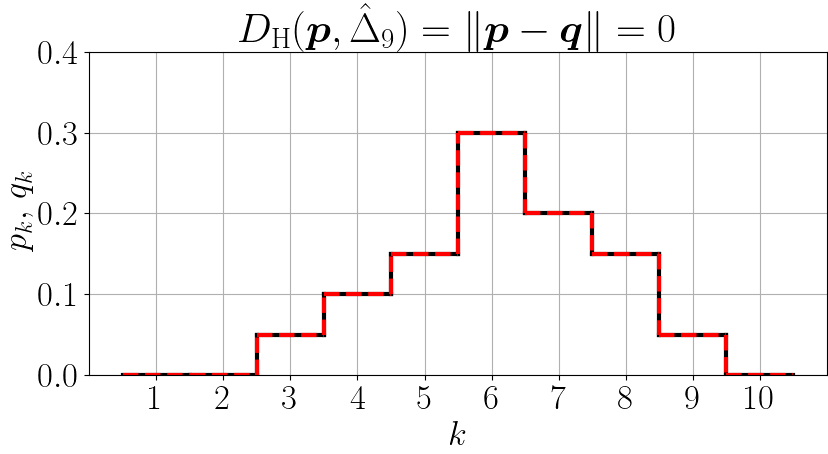}&
\includegraphics[height=1.54cm, bb=0 0 596 327]{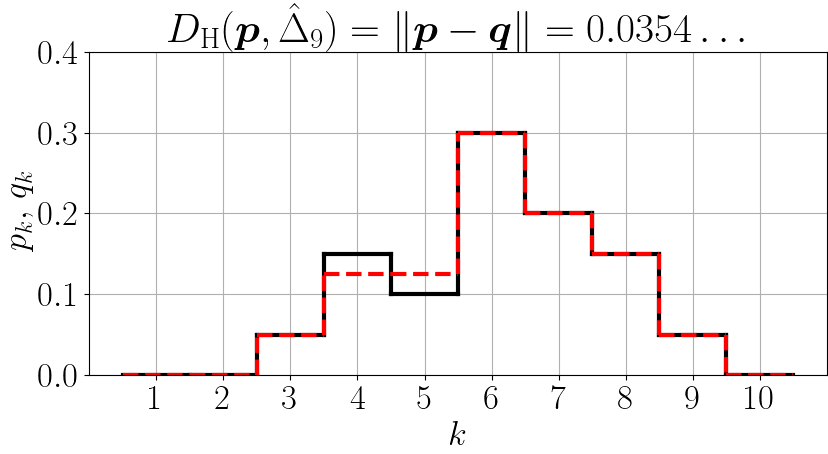}&
\includegraphics[height=1.54cm, bb=0 0 596 327]{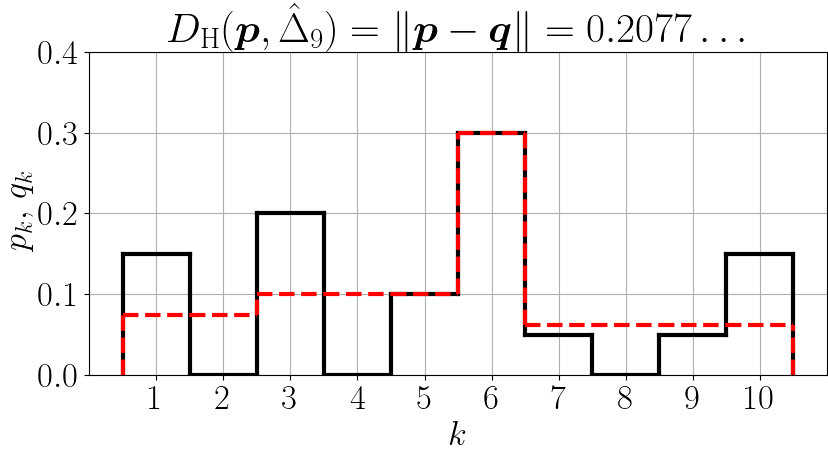}
\end{tabular}
\caption{%
Instance of $D_\rmH(\bp,\hat{\Delta}_{K-1})$ with $K=10$ and
$\bp=\,$(0, 0, .05, .1, .15, .3, .2, .15, .05, 0$)^\top$ (left),
(0, 0, .05, .15, .1, .3, .2, .15, .05, 0$)^\top$ (center), 
(.15, 0, .2, 0, .1, .3, .05, 0, .05, .15$)^\top$ (right),
where we show $\bp$ by a black solid polyline and 
$\bq=\argmin_{\bv\in\hat{\Delta}_{K-1}}\|\bp-\bv\|$ by a red dotted polyline.}
\label{fig:Instance-Unimodality}
\end{figure}

Note that $D_\rmH(\cdot,\hat{\Delta}_{K-1})$ can be 
rephrased as the $L_2$-distance to the nearest unimodal PMF
(see instances in Figure~\ref{fig:Instance-Unimodality}) 
and calculated with convex quadratic programming.

\begin{figure}[!t]
\centering%
\renewcommand{\arraystretch}{0.1}%
\renewcommand{\tabcolsep}{0pt}%
\begin{tabular}{C{2.95cm}C{2.95cm}C{2.95cm}}%
\includegraphics[height=1.54cm, bb=0 0 597 327]{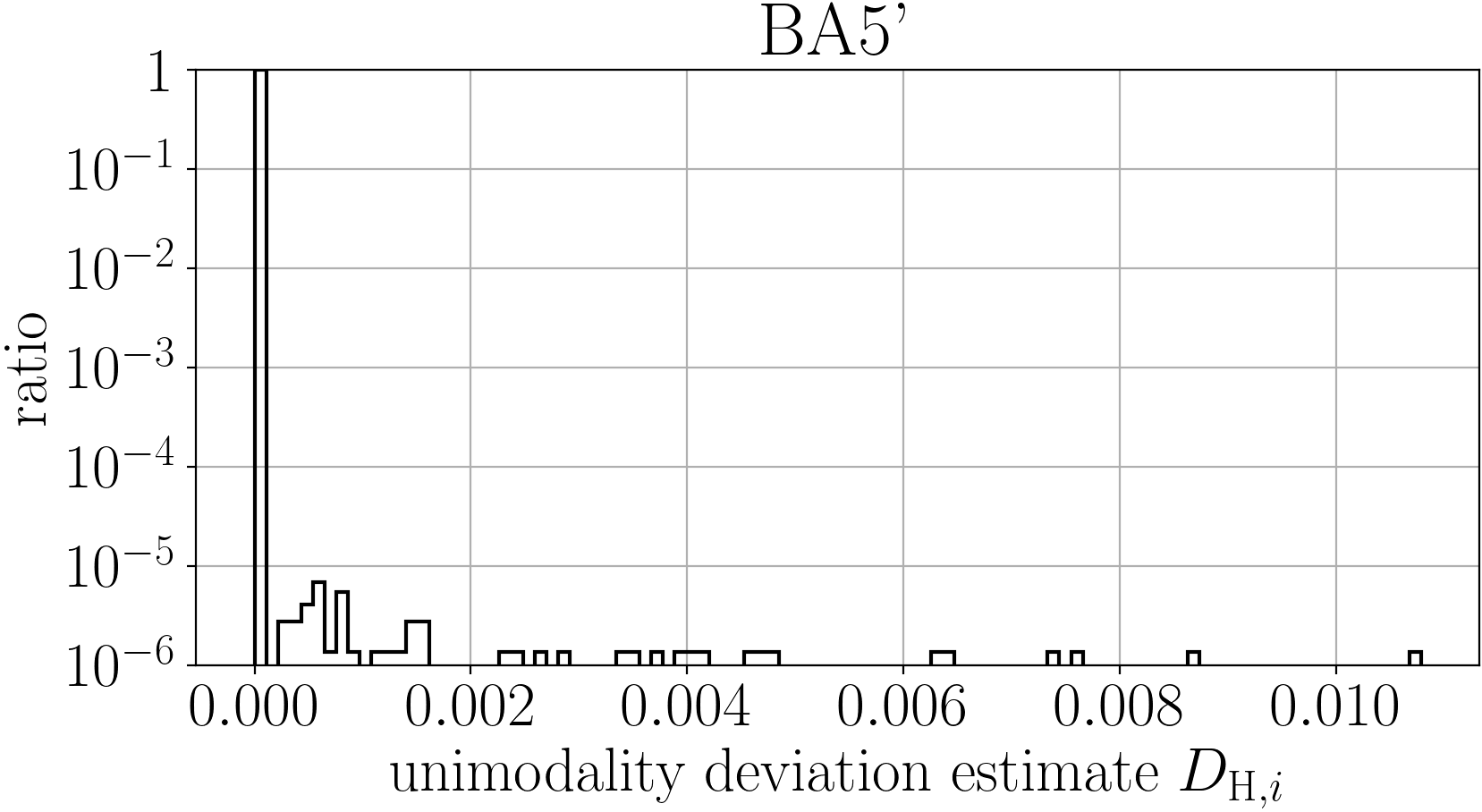}&
\includegraphics[height=1.54cm, bb=0 0 597 327]{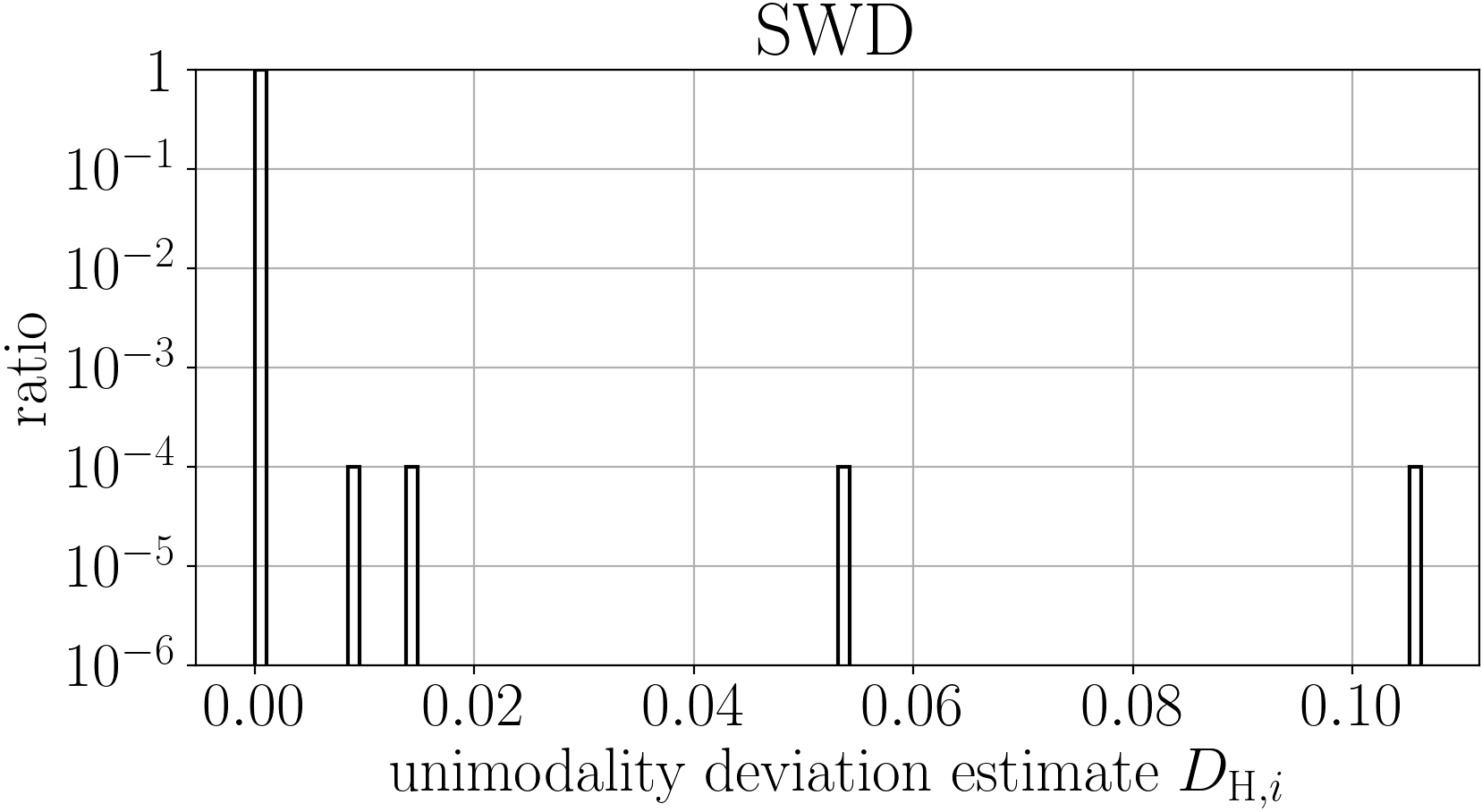}&
\includegraphics[height=1.54cm, bb=0 0 597 327]{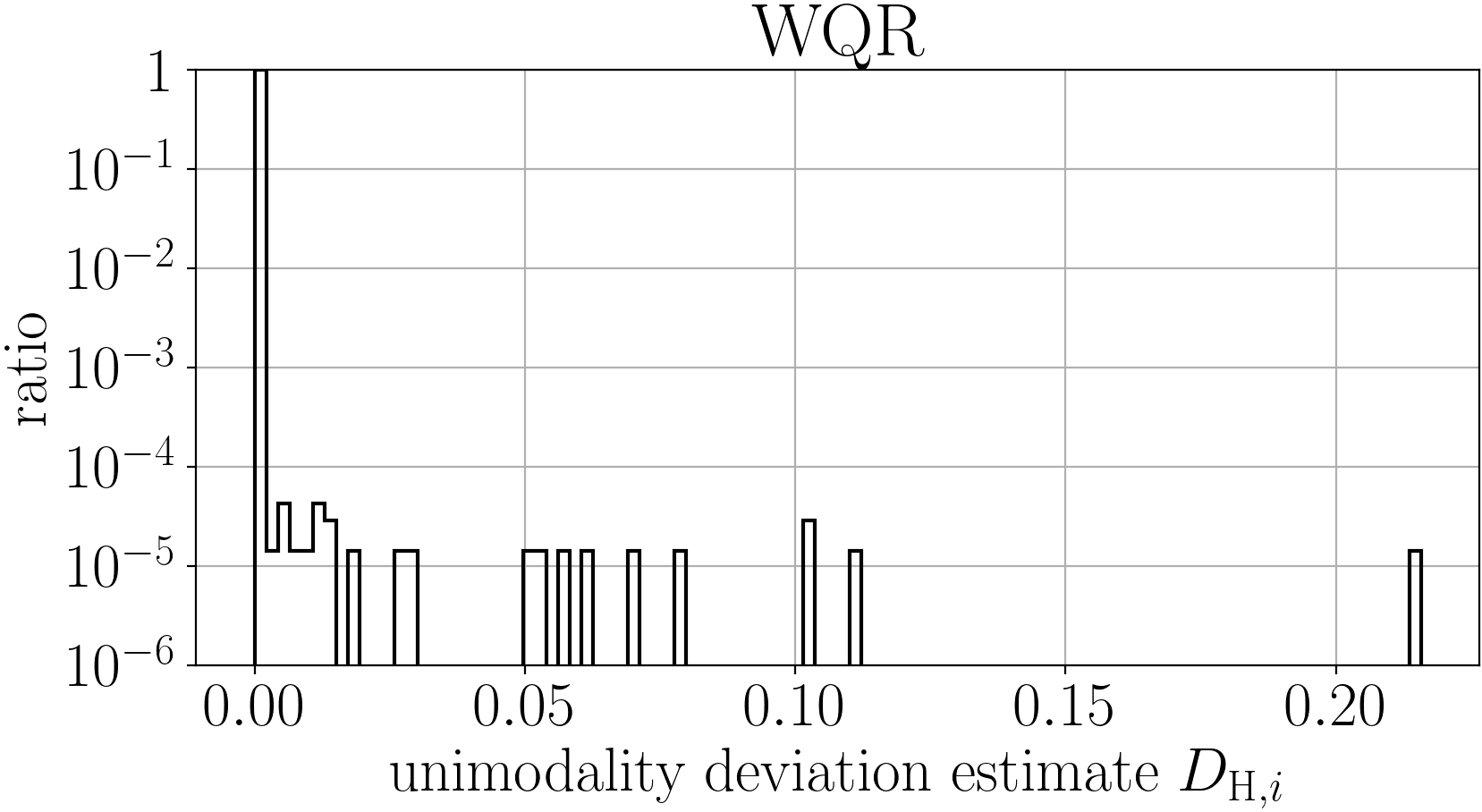}\\
\includegraphics[height=1.54cm, bb=0 0 597 327]{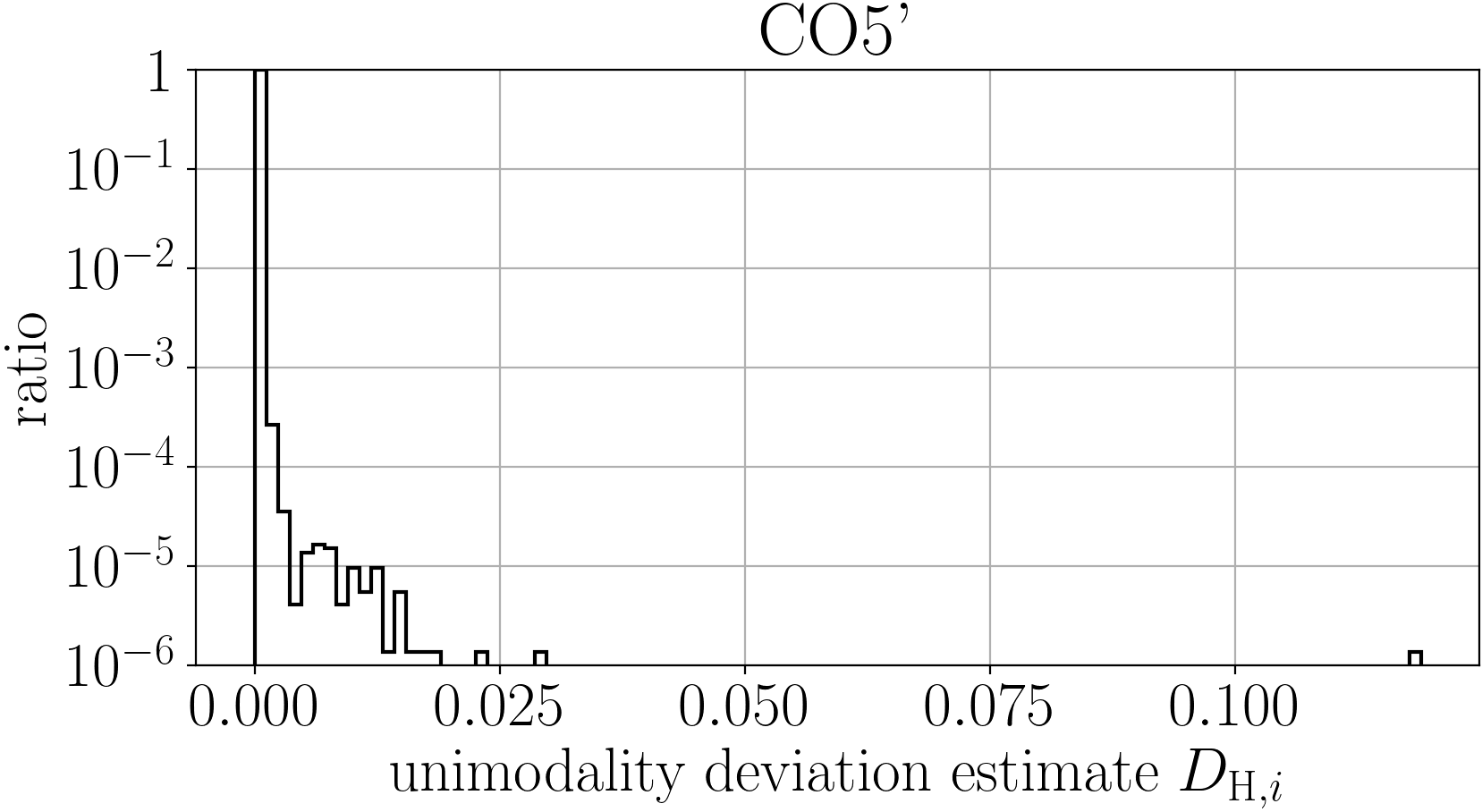}&
\includegraphics[height=1.54cm, bb=0 0 597 327]{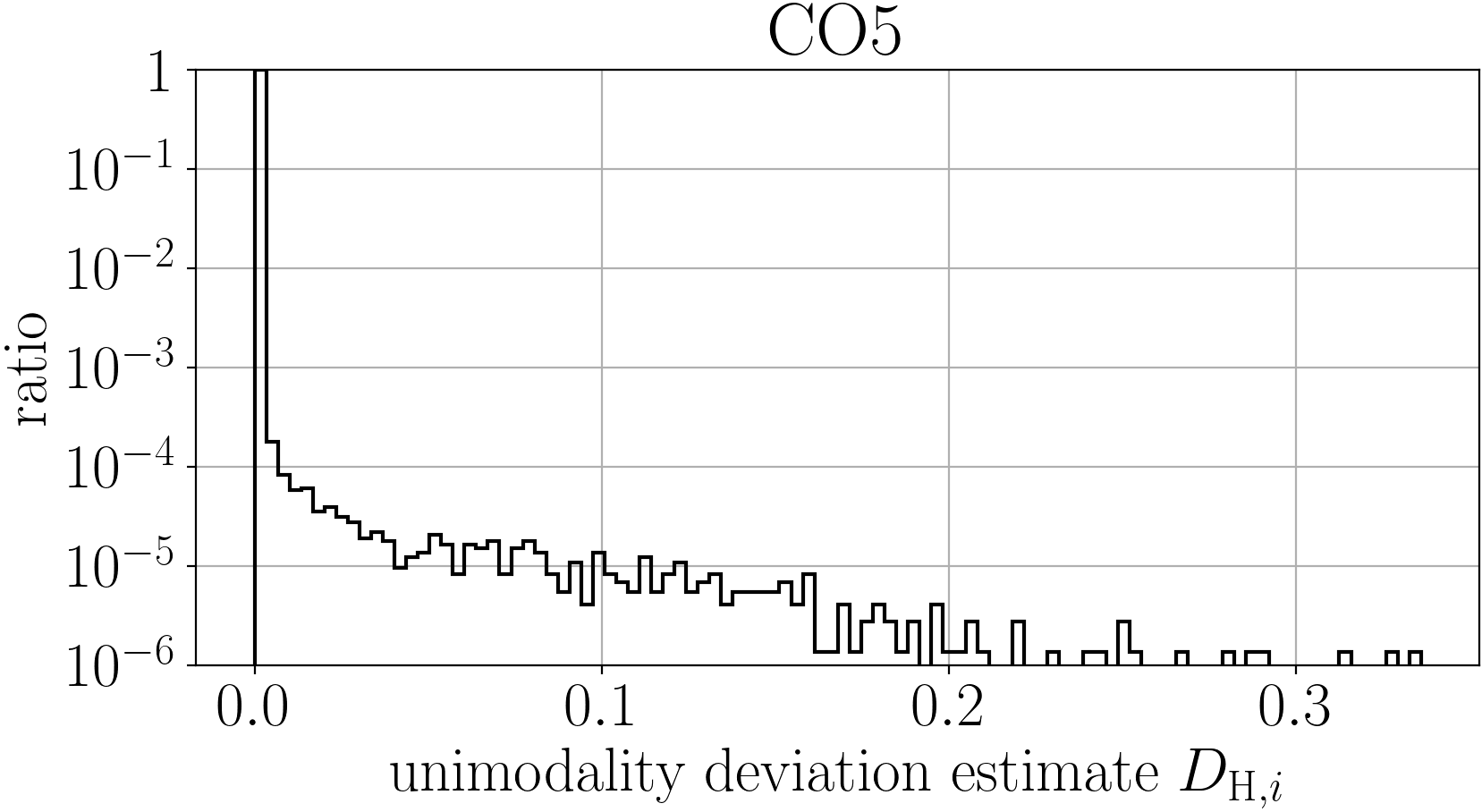}&
\includegraphics[height=1.54cm, bb=0 0 597 327]{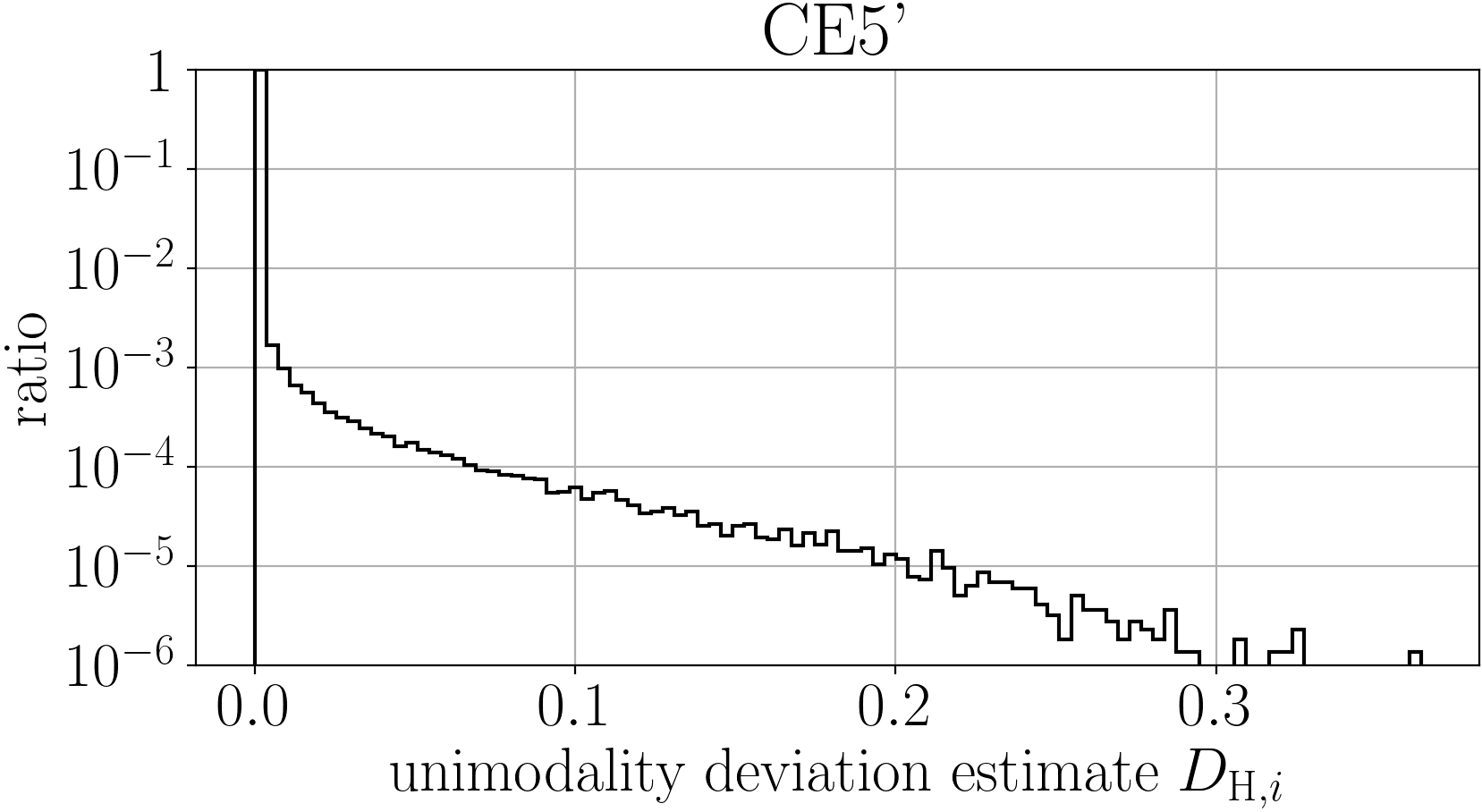}\\
\includegraphics[height=1.54cm, bb=0 0 597 327]{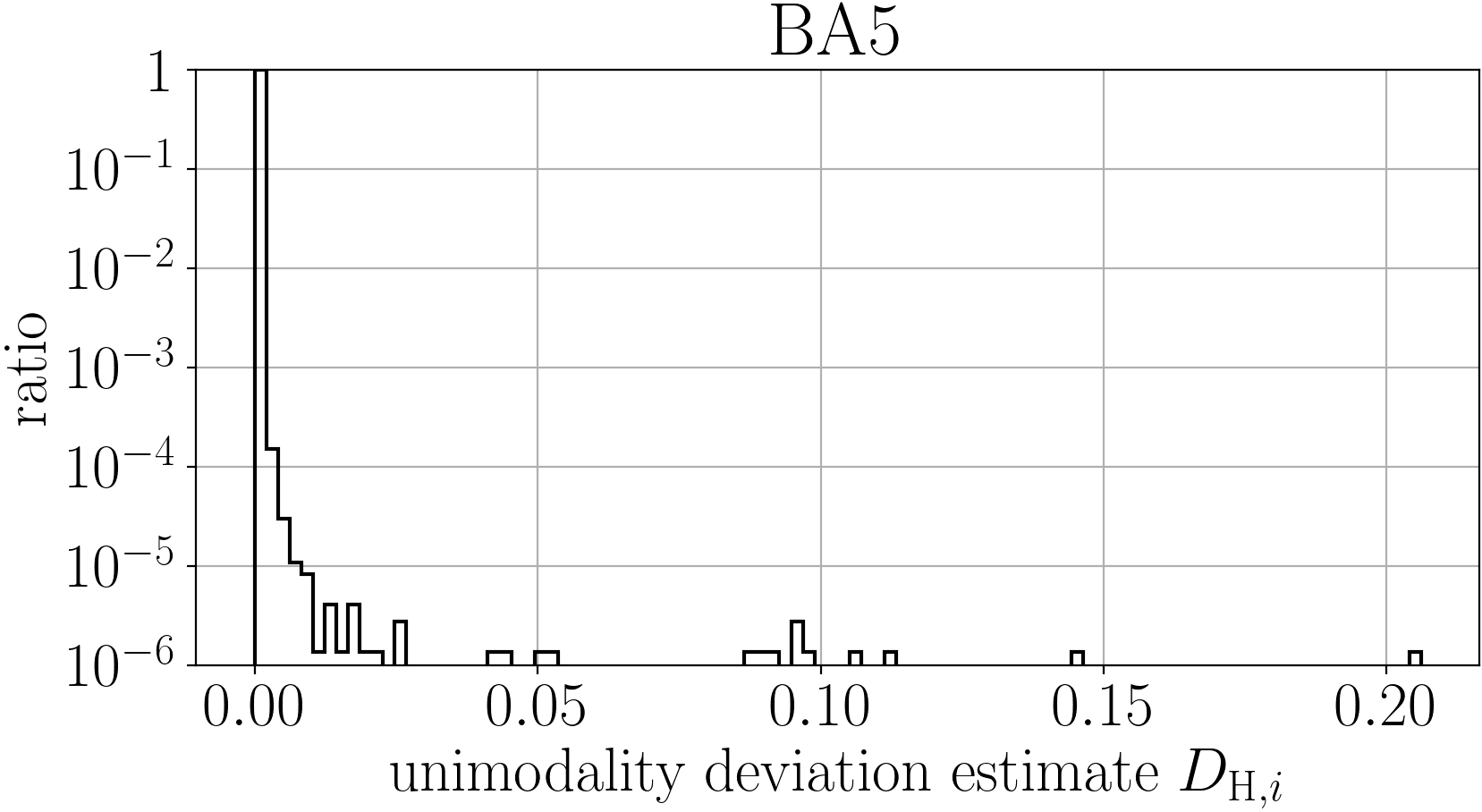}&
\includegraphics[height=1.54cm, bb=0 0 597 327]{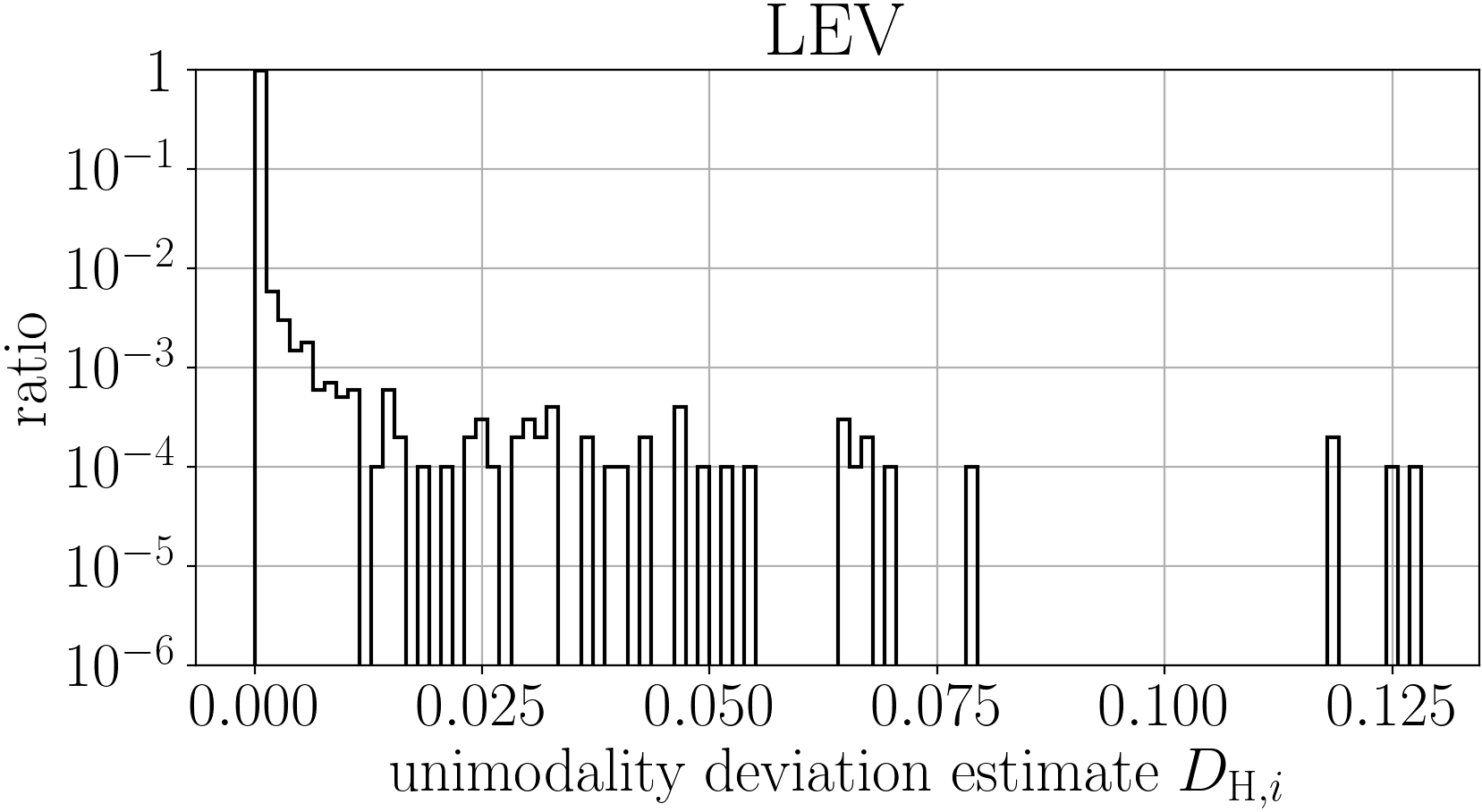}&
\includegraphics[height=1.54cm, bb=0 0 597 327]{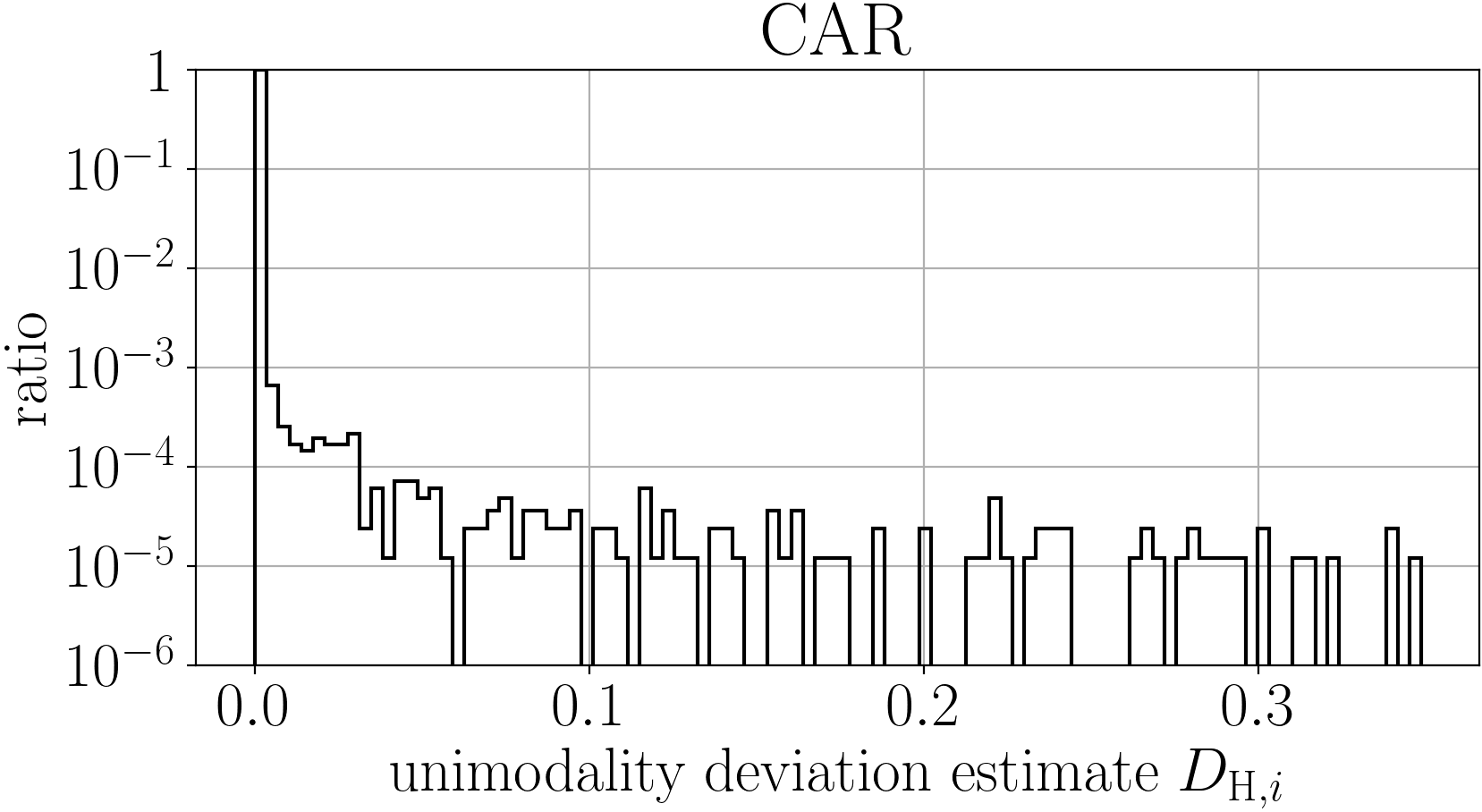}\\
\includegraphics[height=1.54cm, bb=0 0 597 327]{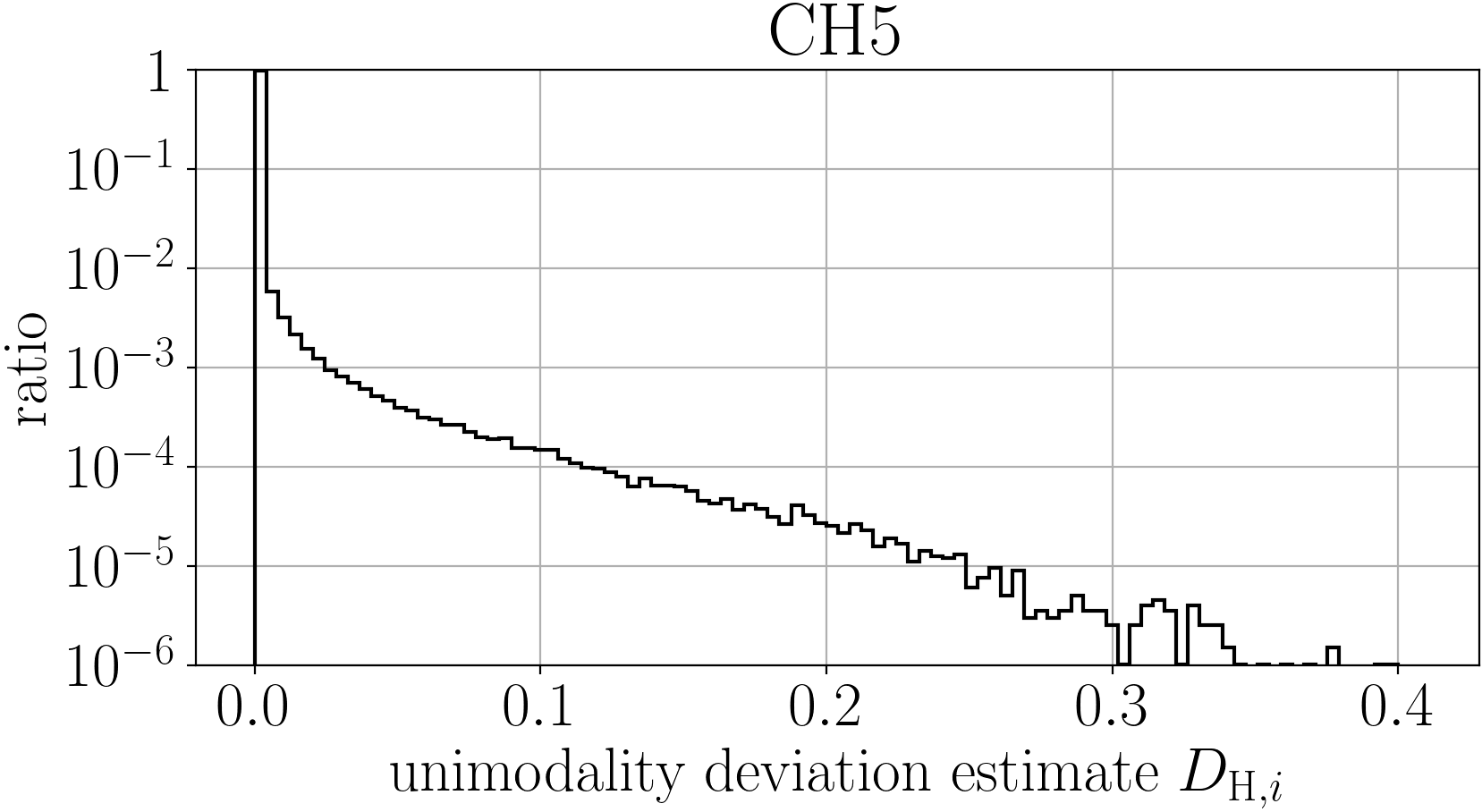}&
\includegraphics[height=1.54cm, bb=0 0 597 327]{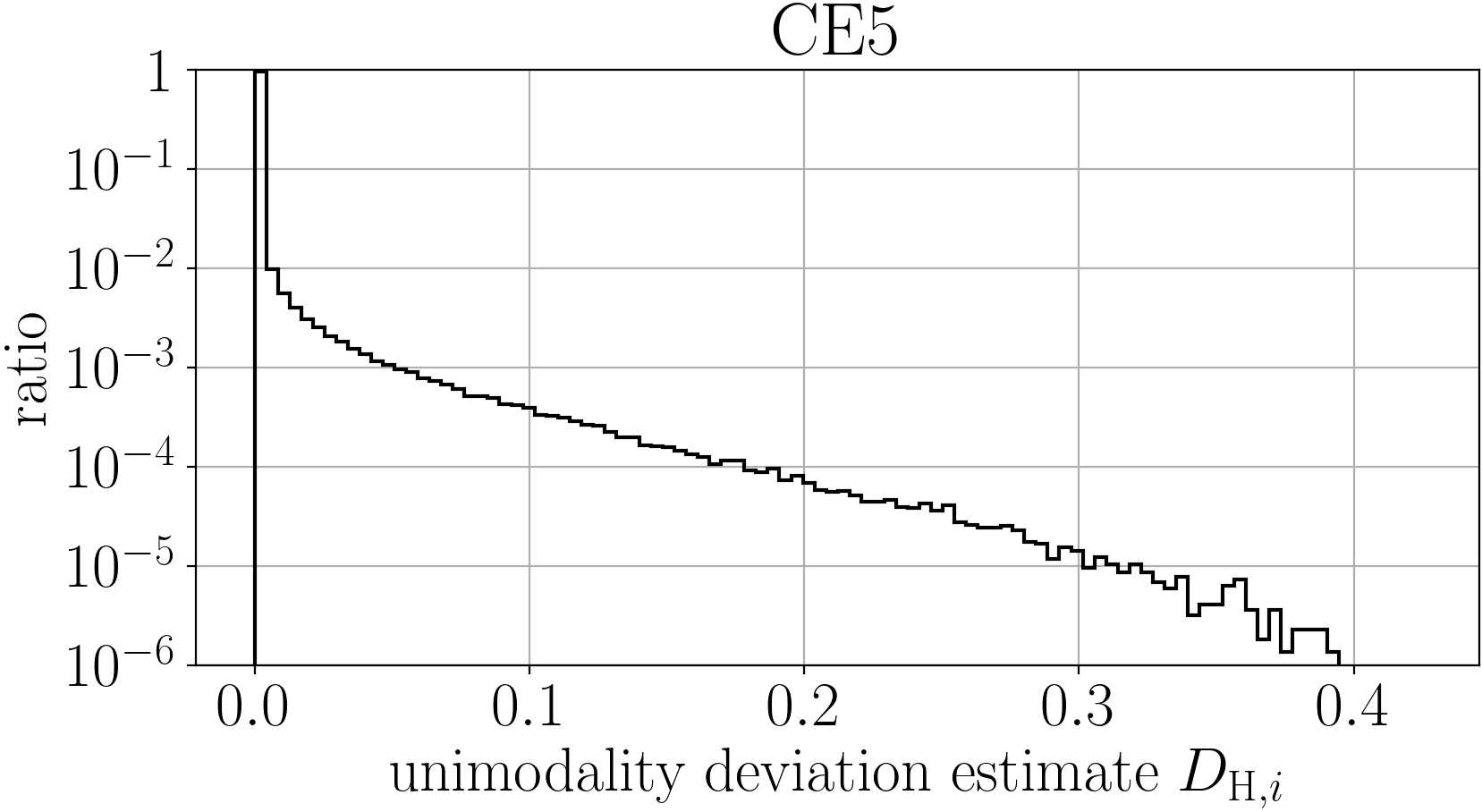}&
\includegraphics[height=1.54cm, bb=0 0 597 327]{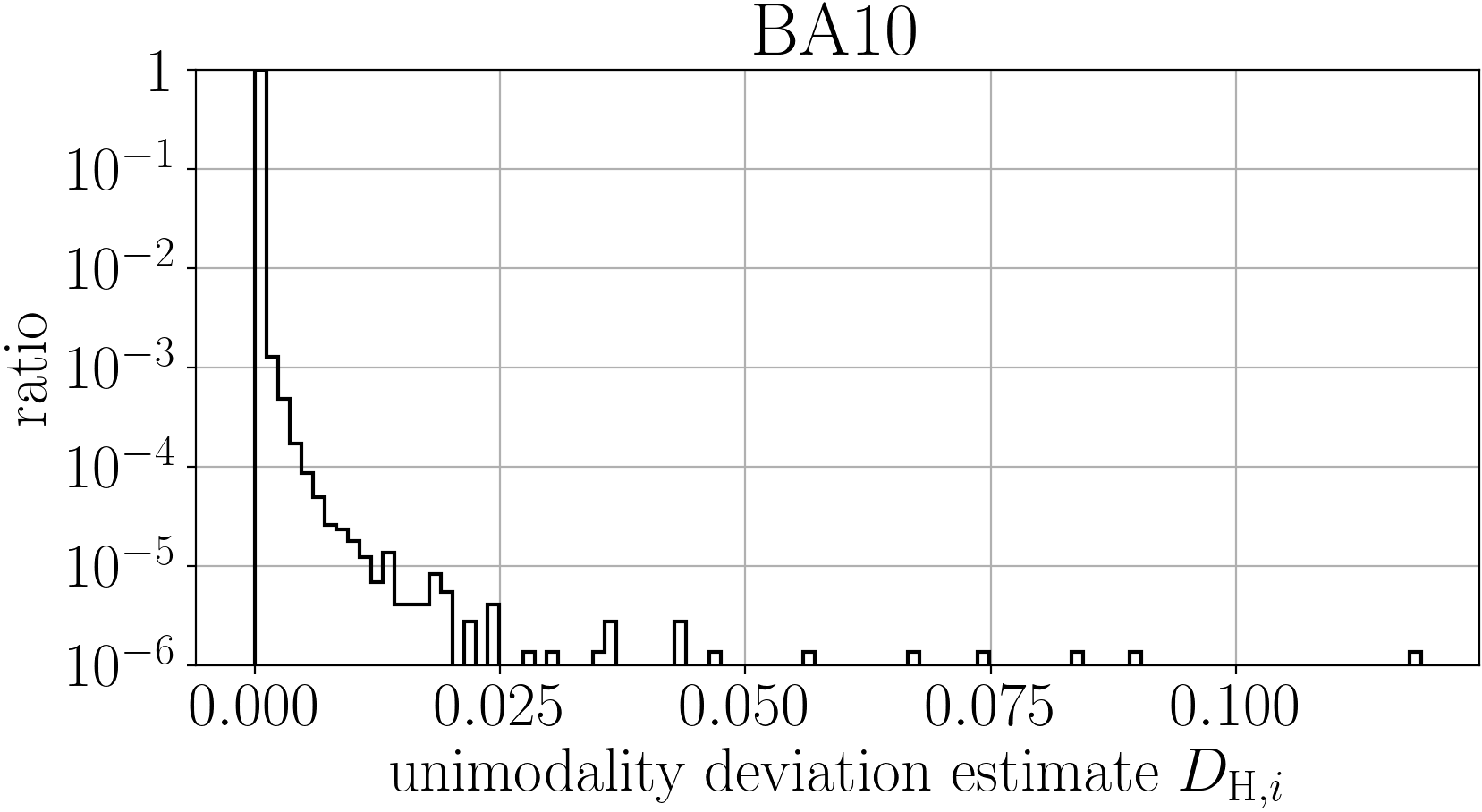}\\
\includegraphics[height=1.54cm, bb=0 0 597 327]{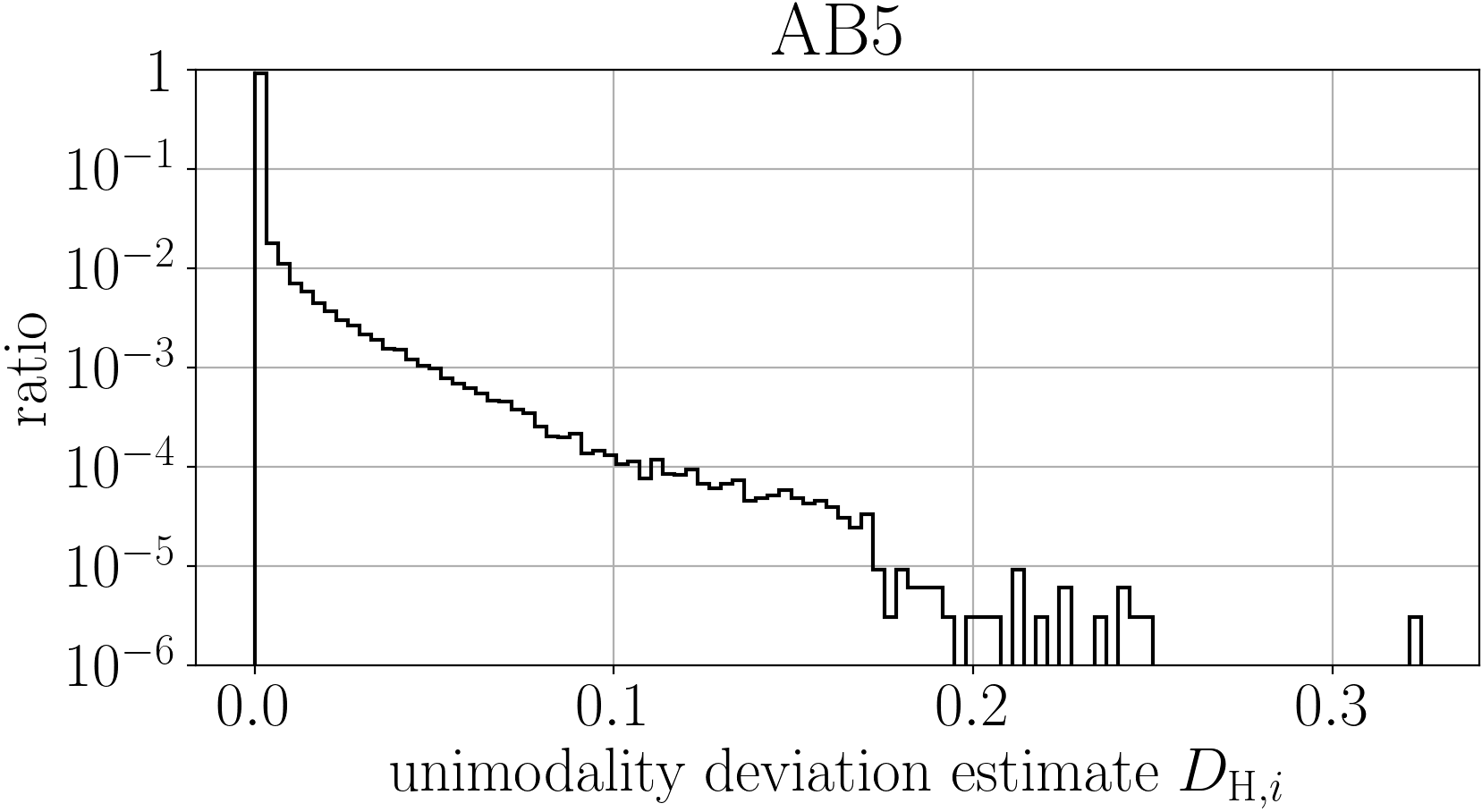}&
\includegraphics[height=1.54cm, bb=0 0 597 327]{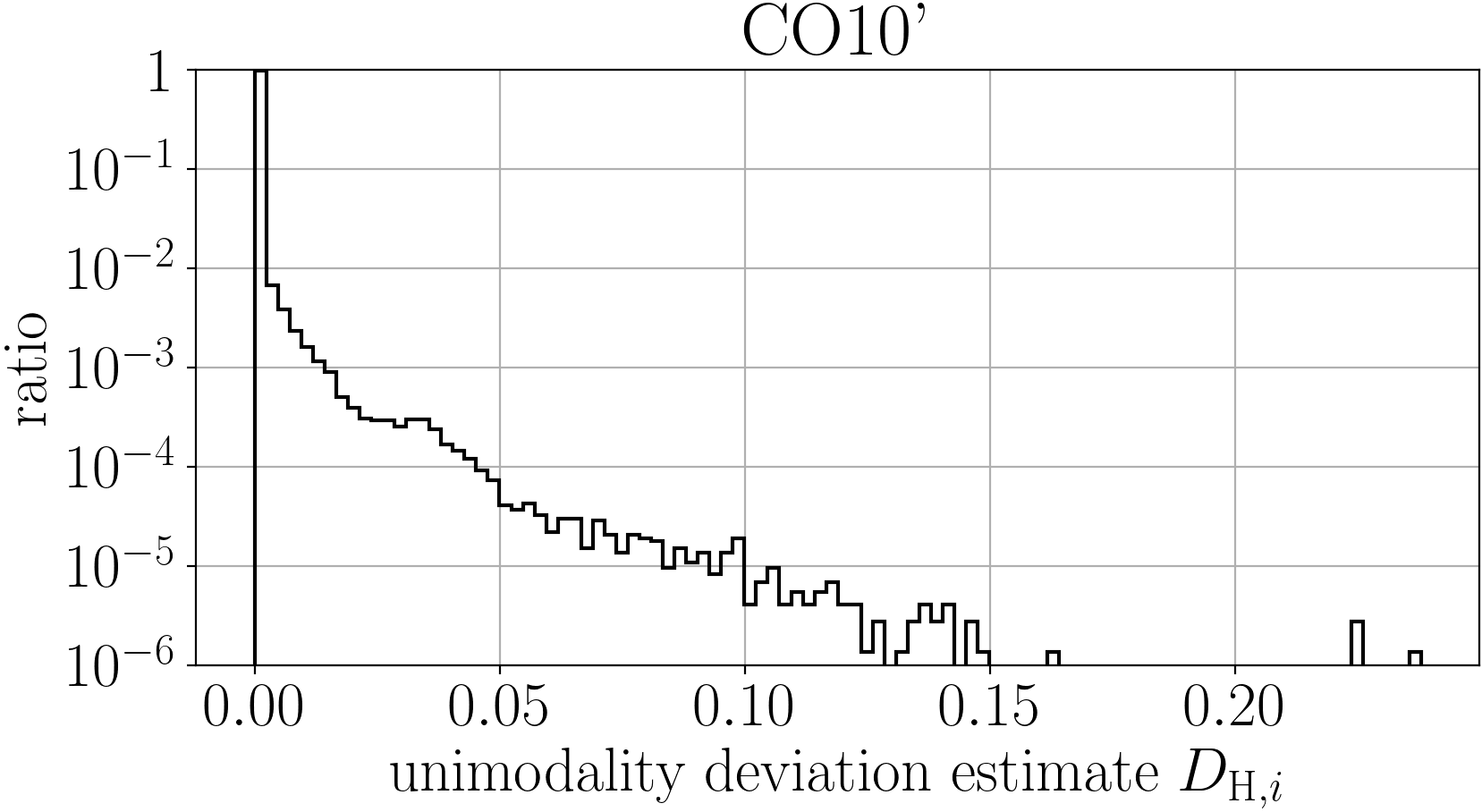}&
\includegraphics[height=1.54cm, bb=0 0 597 327]{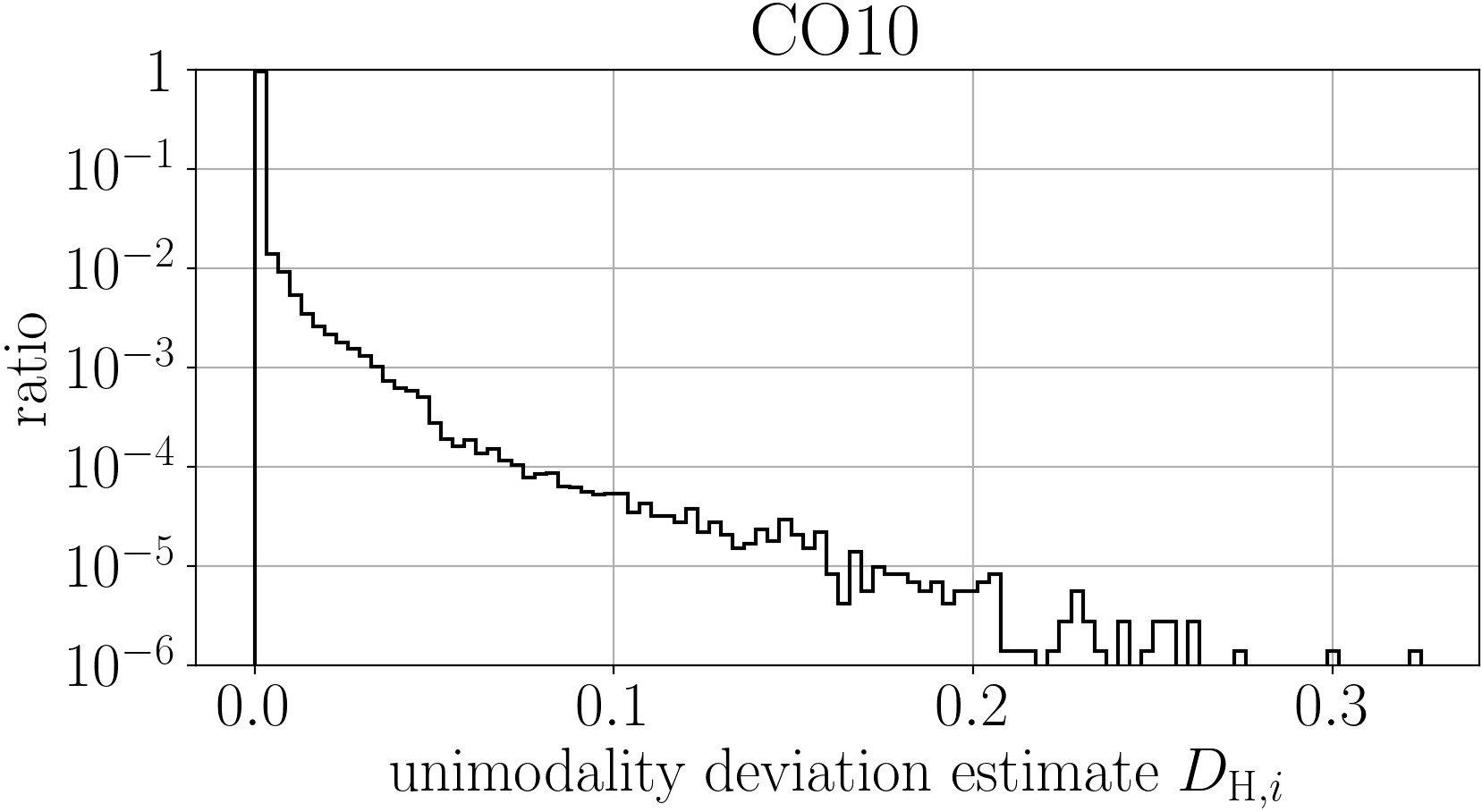}\\
\includegraphics[height=1.54cm, bb=0 0 597 327]{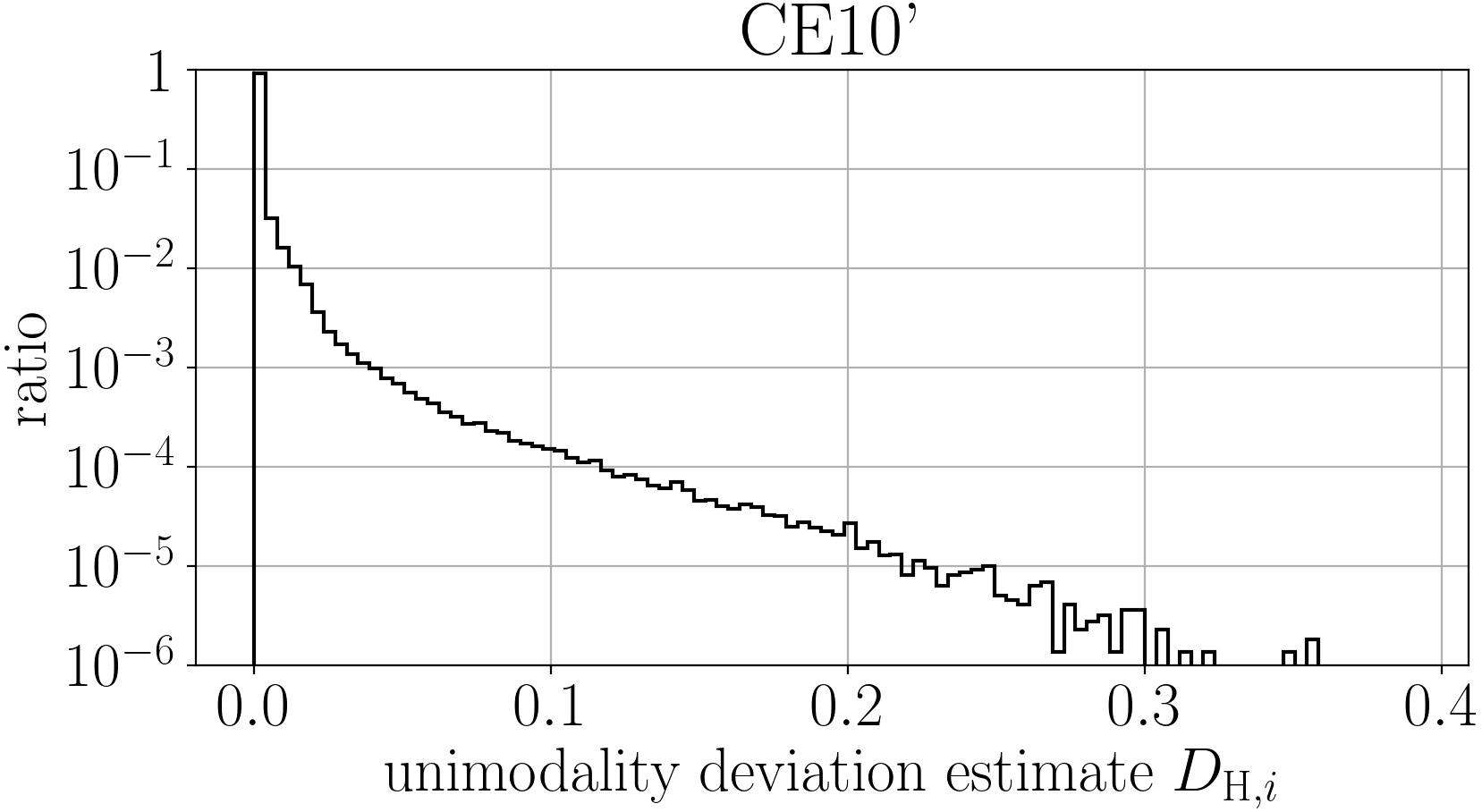}&
\includegraphics[height=1.54cm, bb=0 0 597 327]{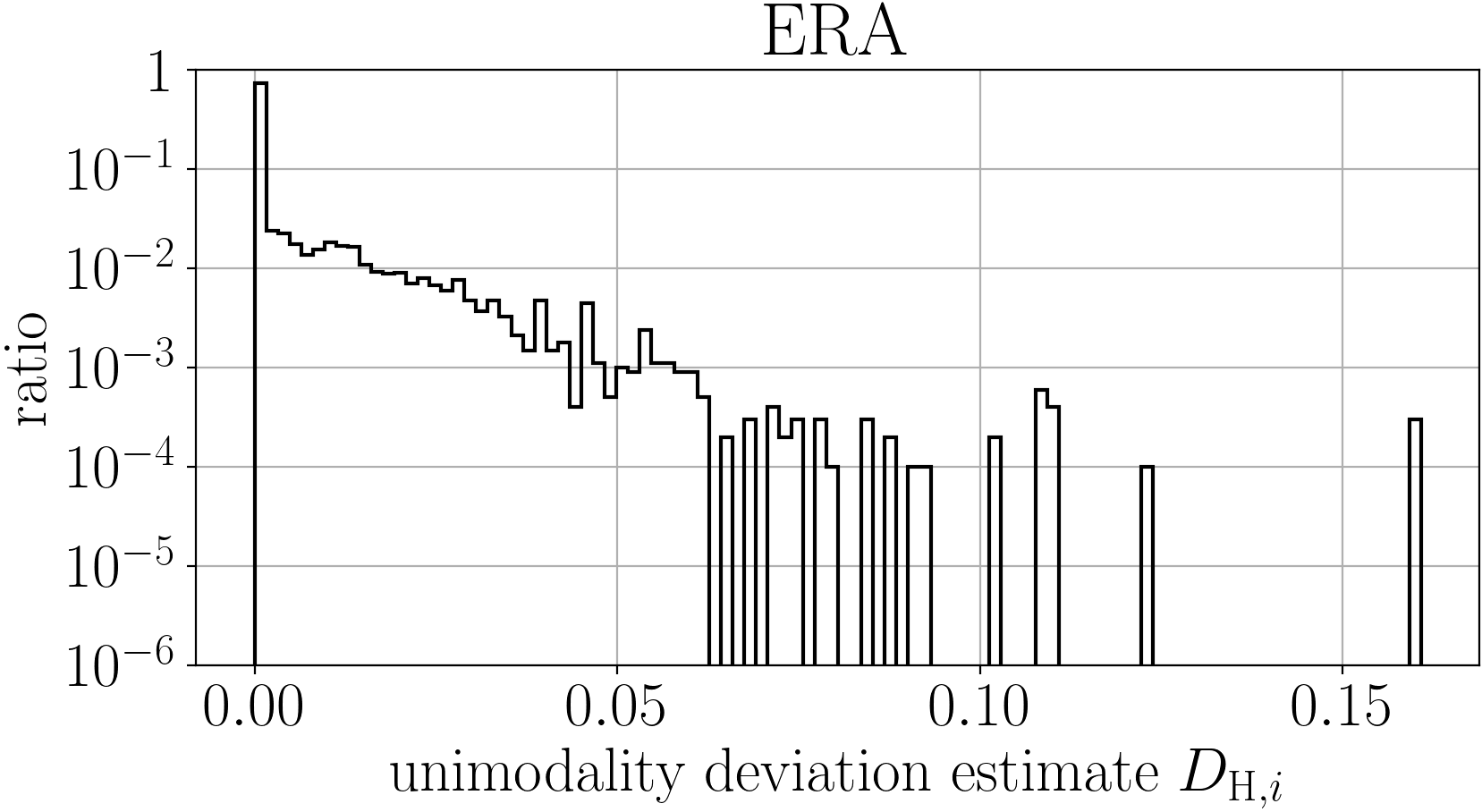}&
\includegraphics[height=1.54cm, bb=0 0 597 327]{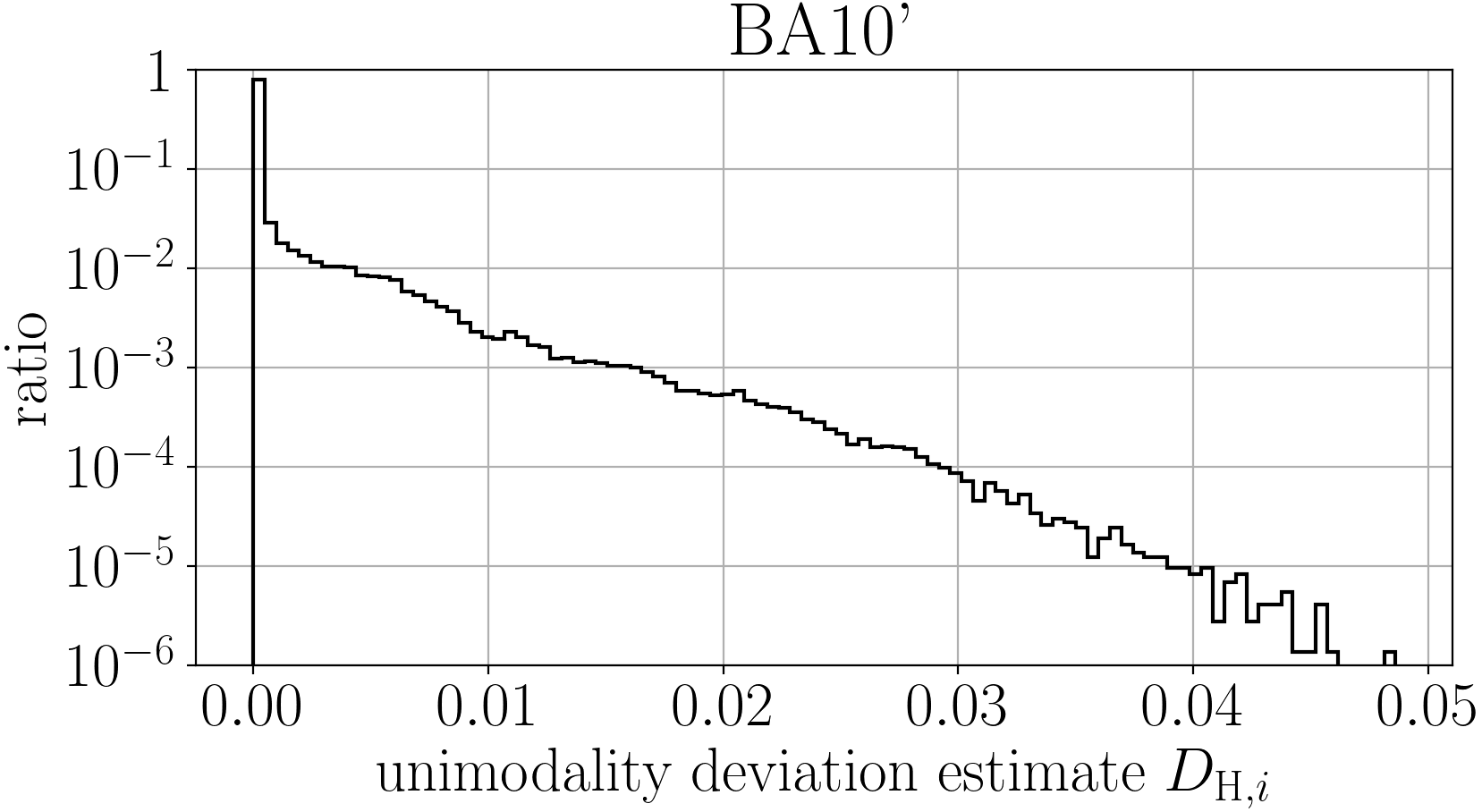}\\
\includegraphics[height=1.54cm, bb=0 0 597 327]{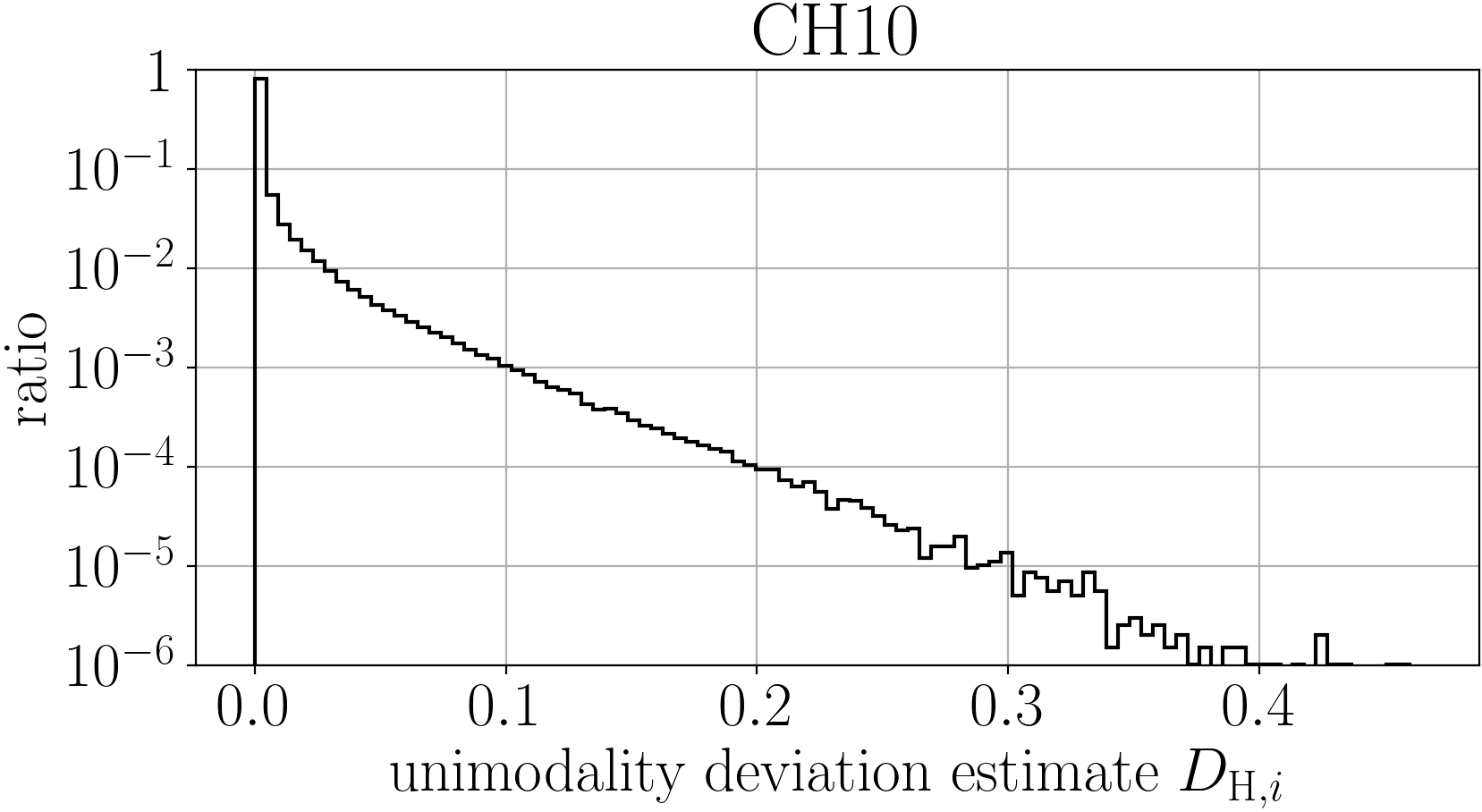}&
\includegraphics[height=1.54cm, bb=0 0 597 327]{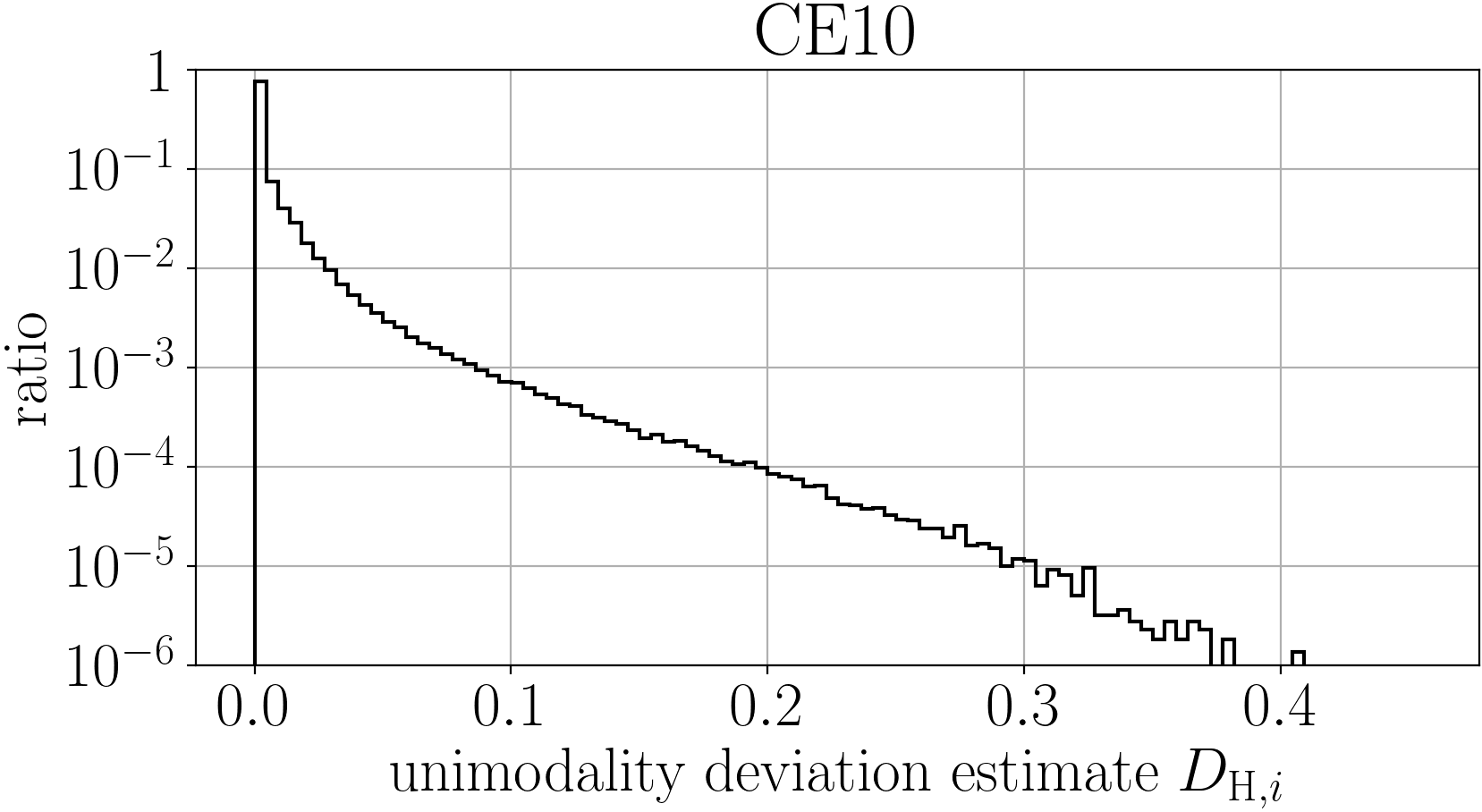}&
\includegraphics[height=1.54cm, bb=0 0 597 327]{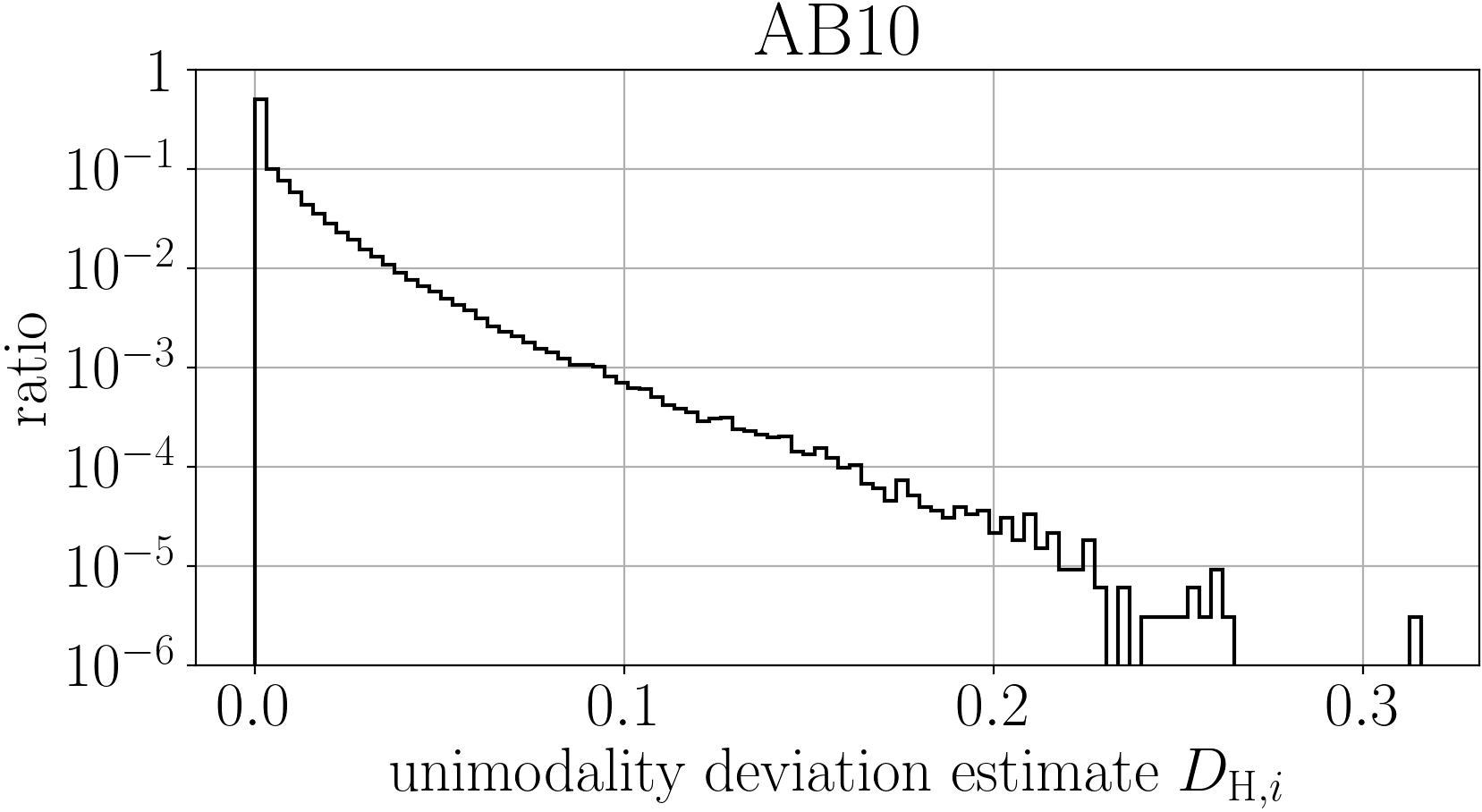}\\
\includegraphics[height=1.54cm, bb=0 0 597 327]{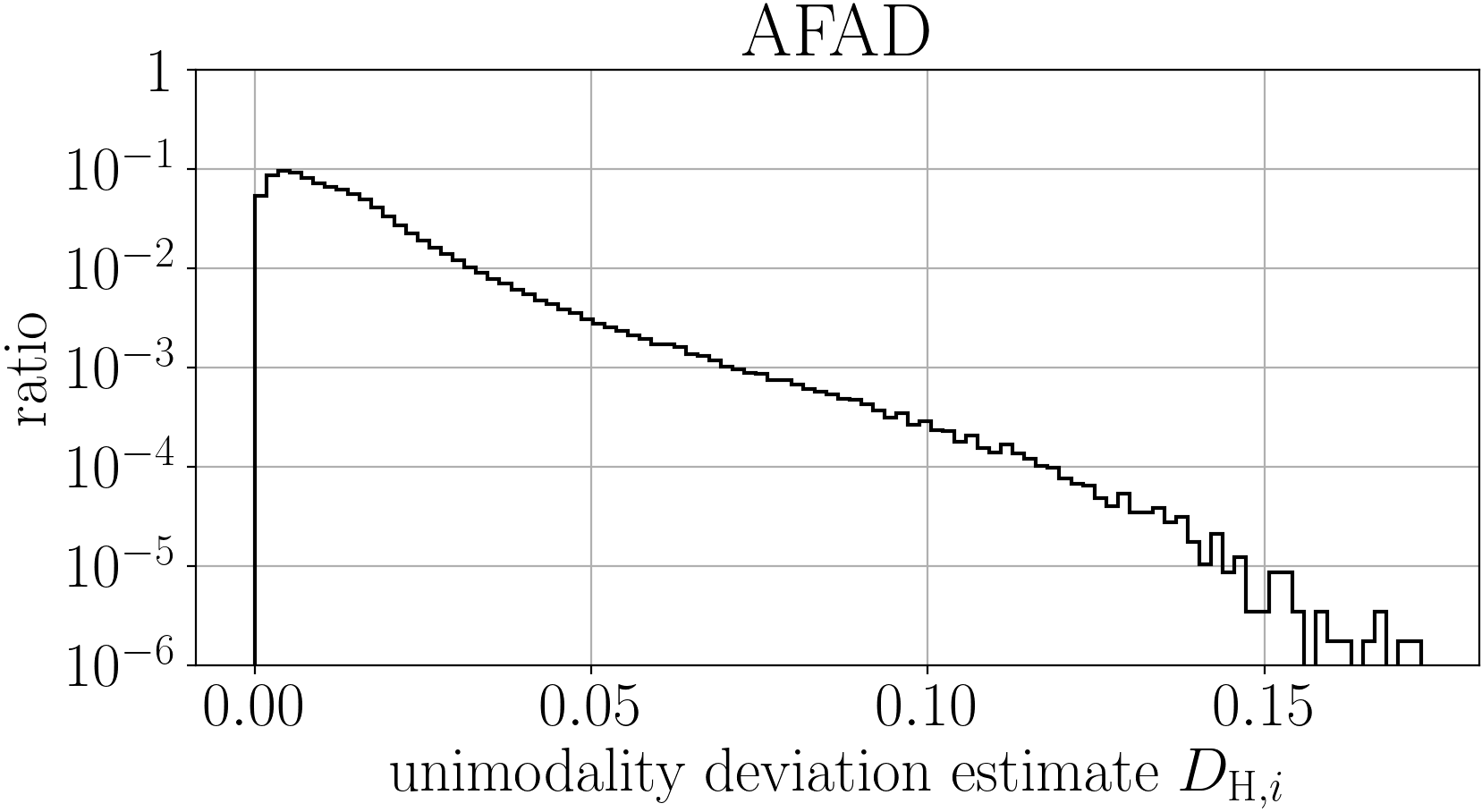}&
\includegraphics[height=1.54cm, bb=0 0 597 327]{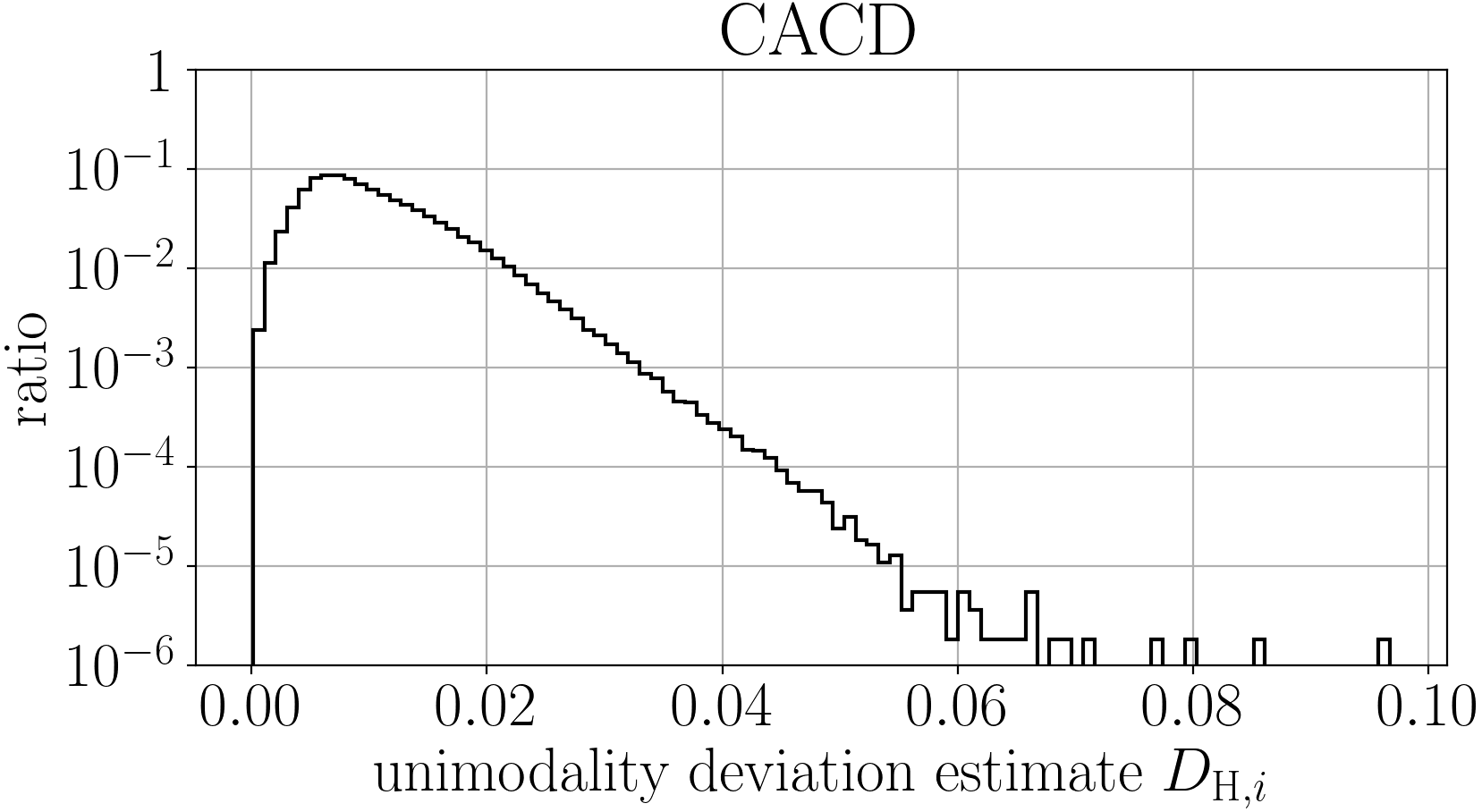}&
\includegraphics[height=1.54cm, bb=0 0 597 327]{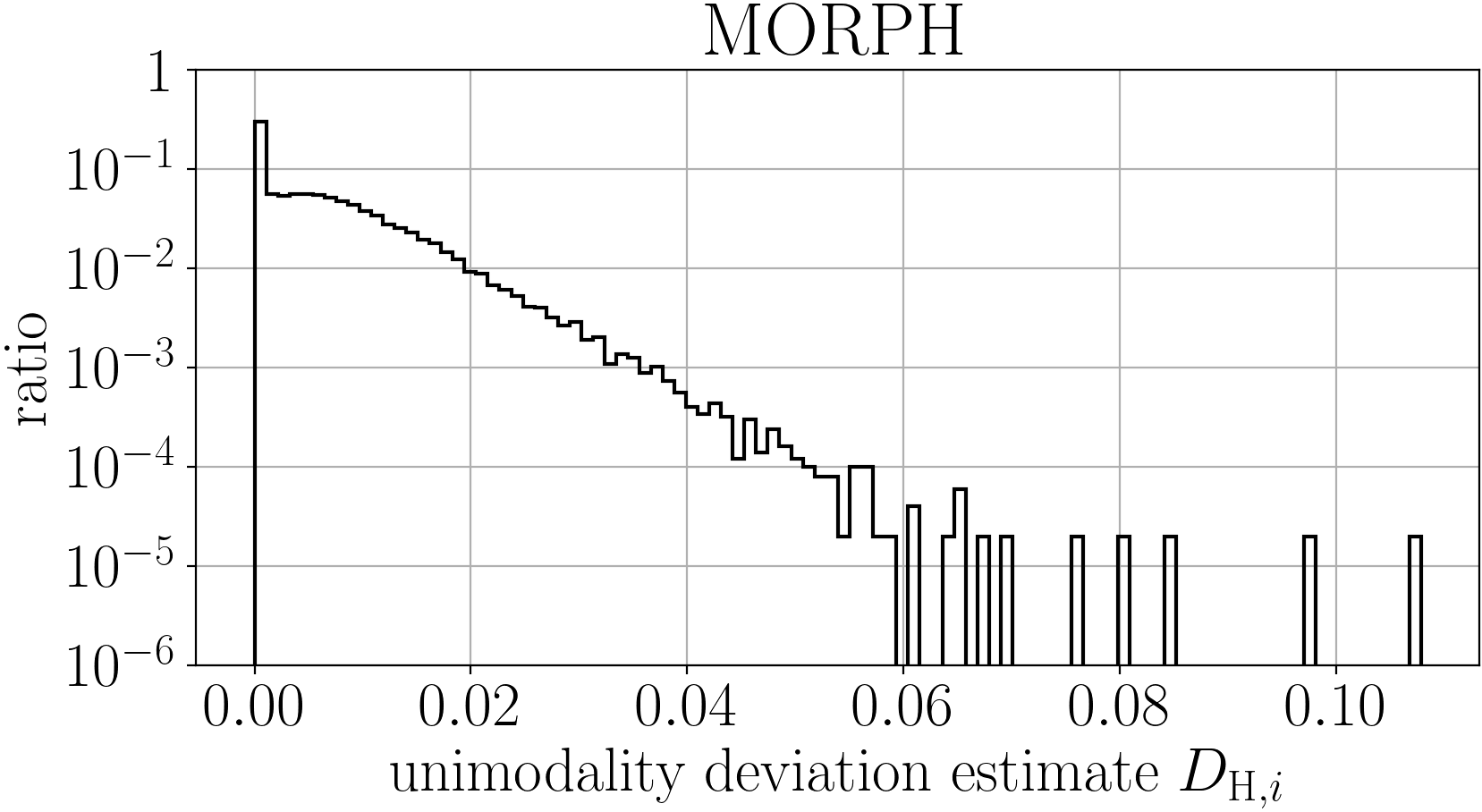}\\
\includegraphics[height=1.54cm, bb=0 0 597 327]{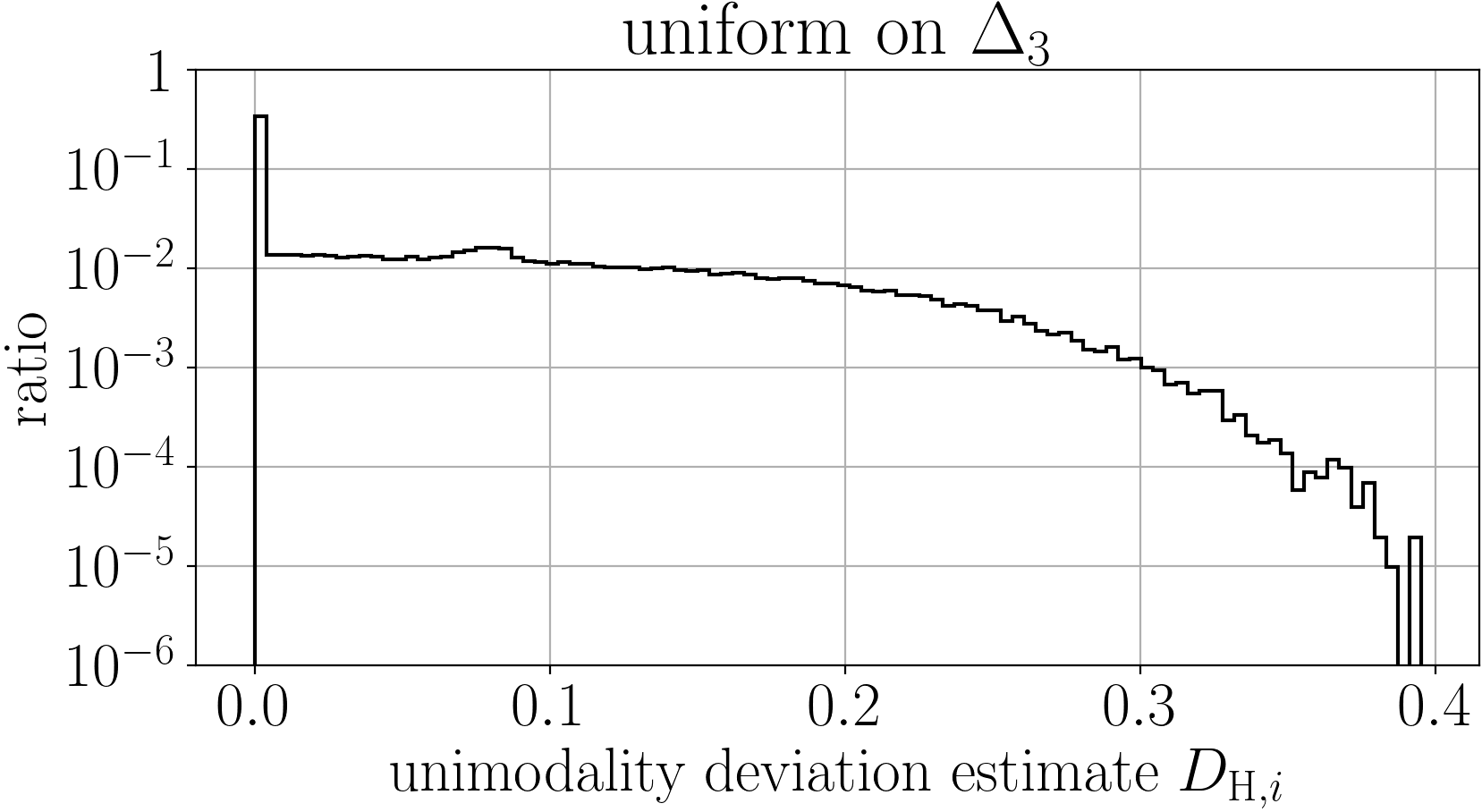}&
\includegraphics[height=1.54cm, bb=0 0 597 327]{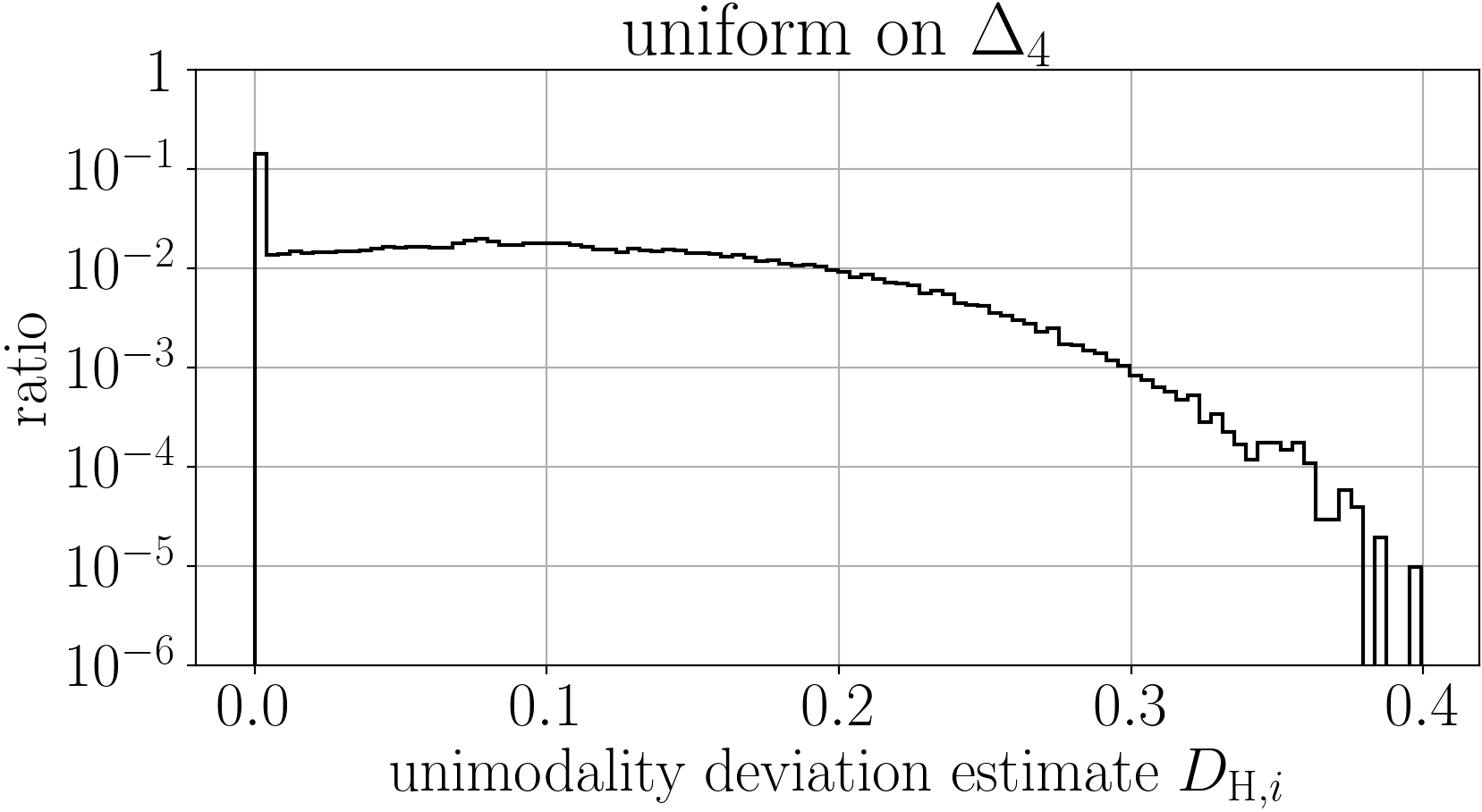}&
\includegraphics[height=1.54cm, bb=0 0 597 327]{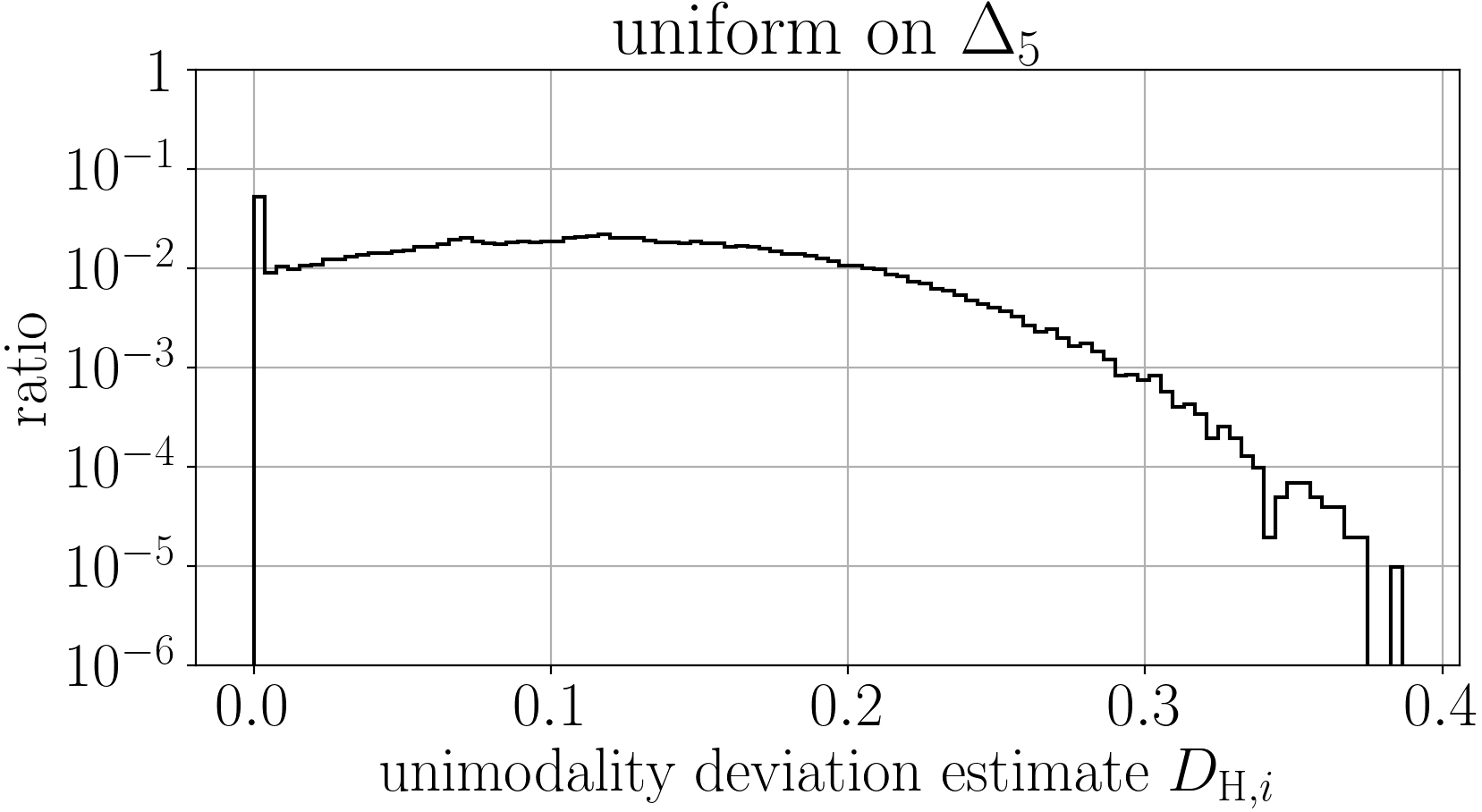}\\
\includegraphics[height=1.54cm, bb=0 0 597 327]{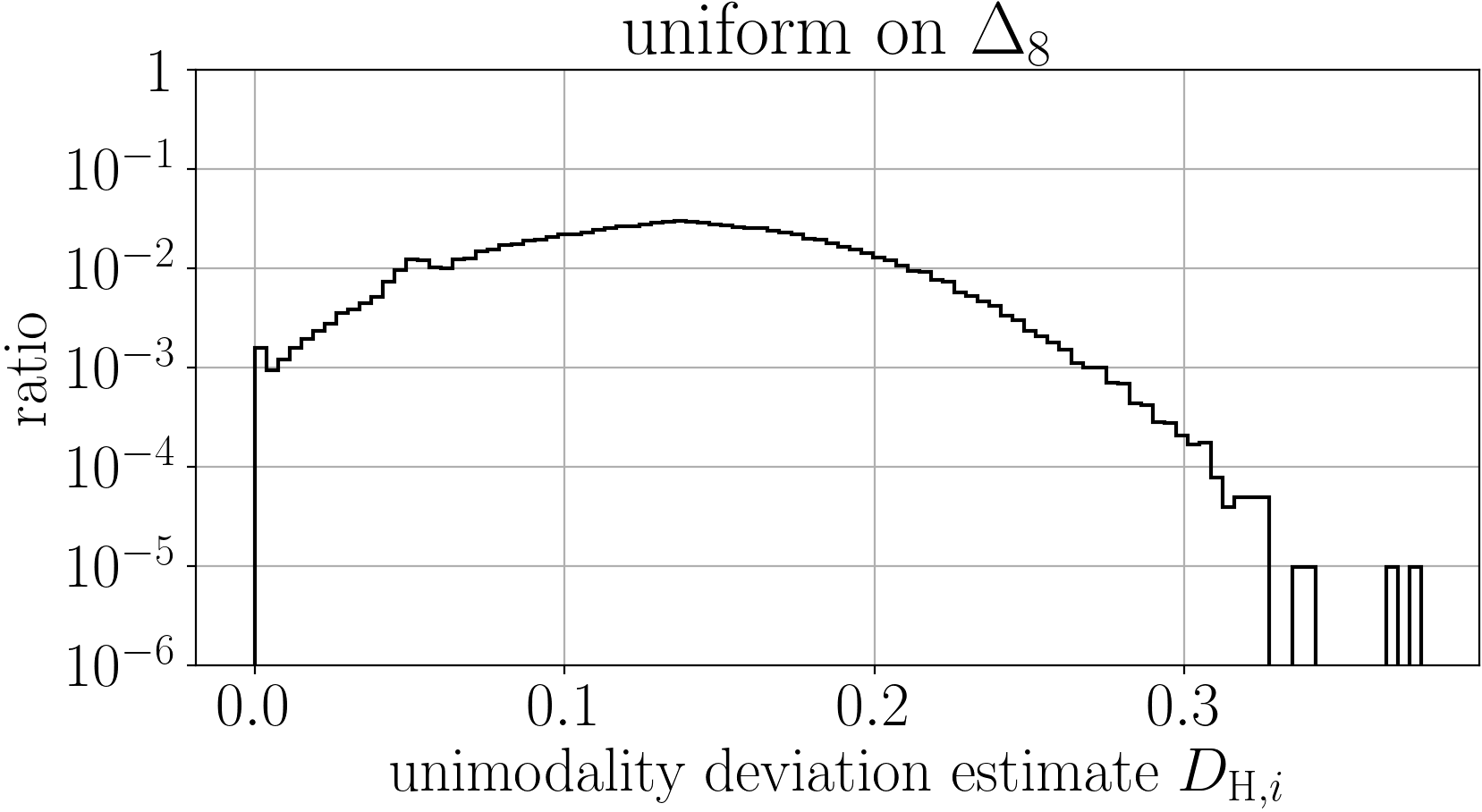}&
\includegraphics[height=1.54cm, bb=0 0 597 327]{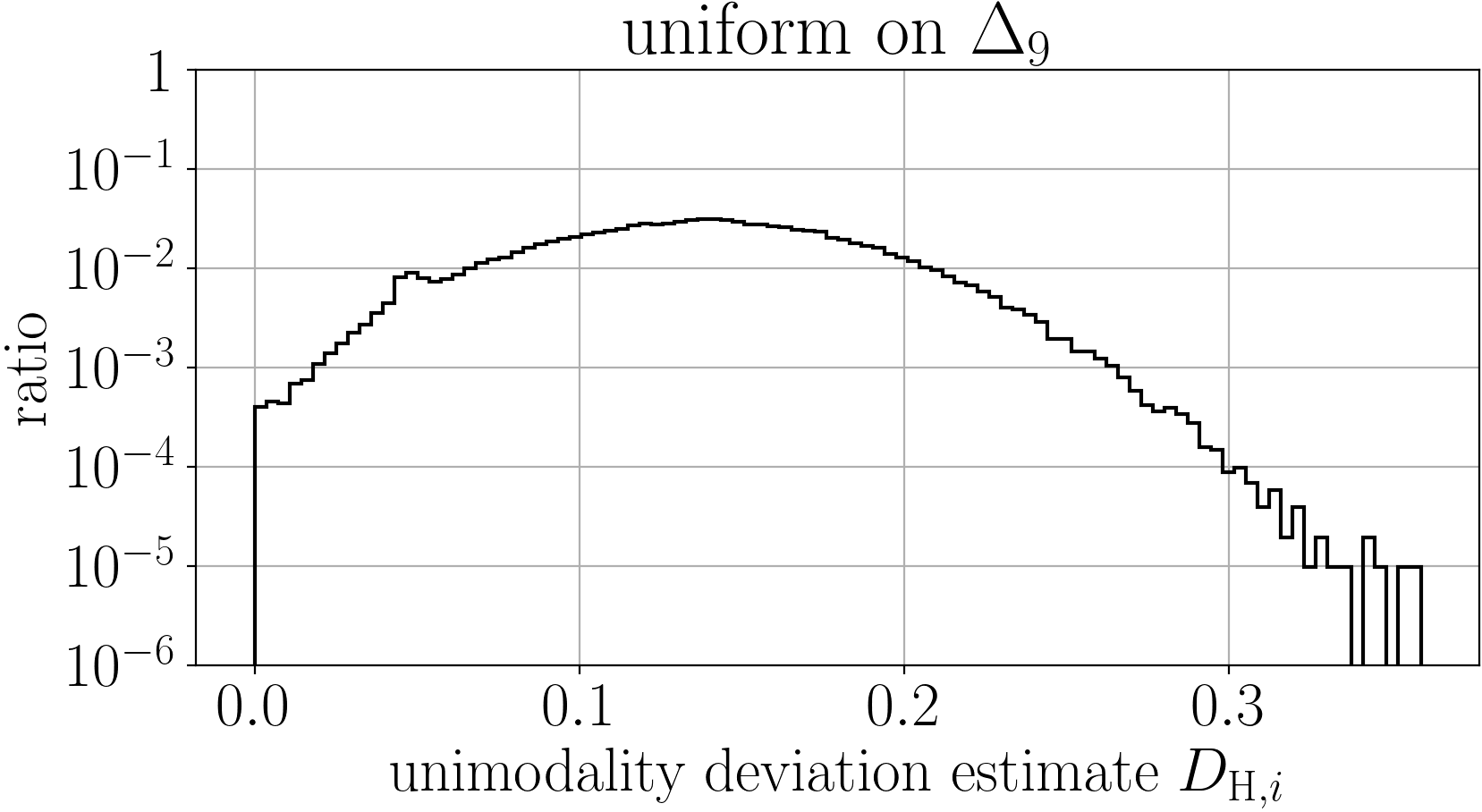}&
\includegraphics[height=1.54cm, bb=0 0 597 327]{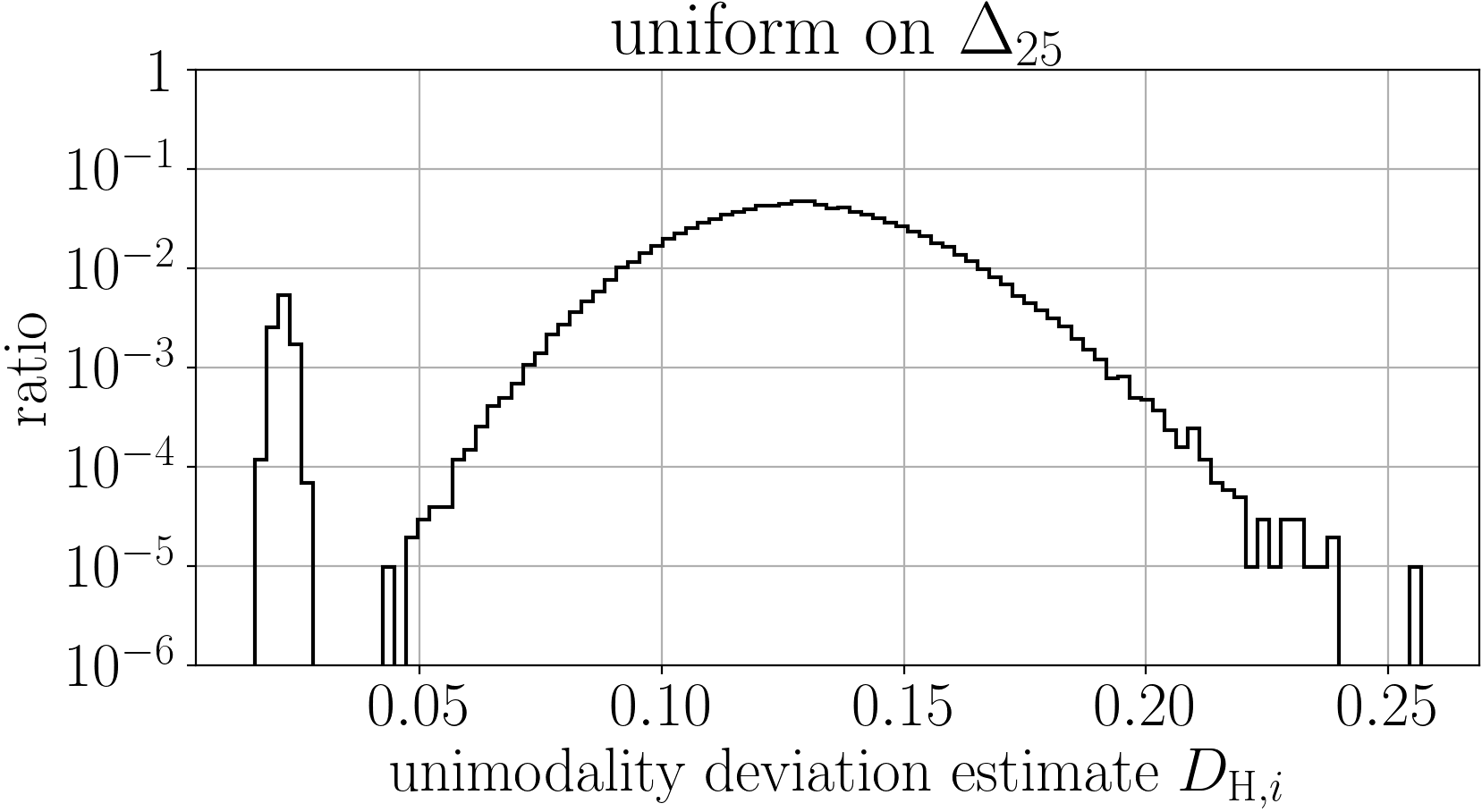}\\
\includegraphics[height=1.54cm, bb=0 0 597 327]{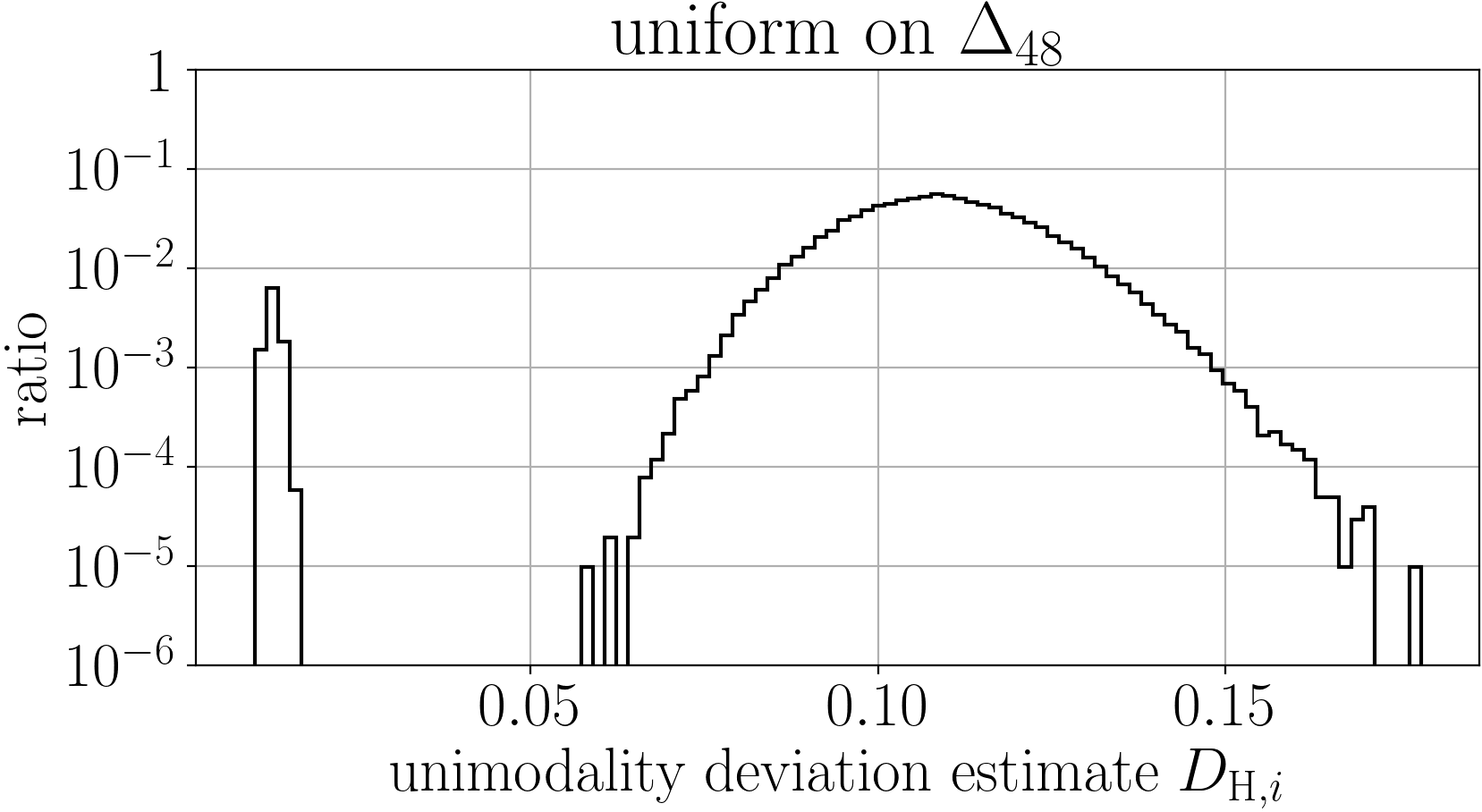}&
\includegraphics[height=1.54cm, bb=0 0 597 327]{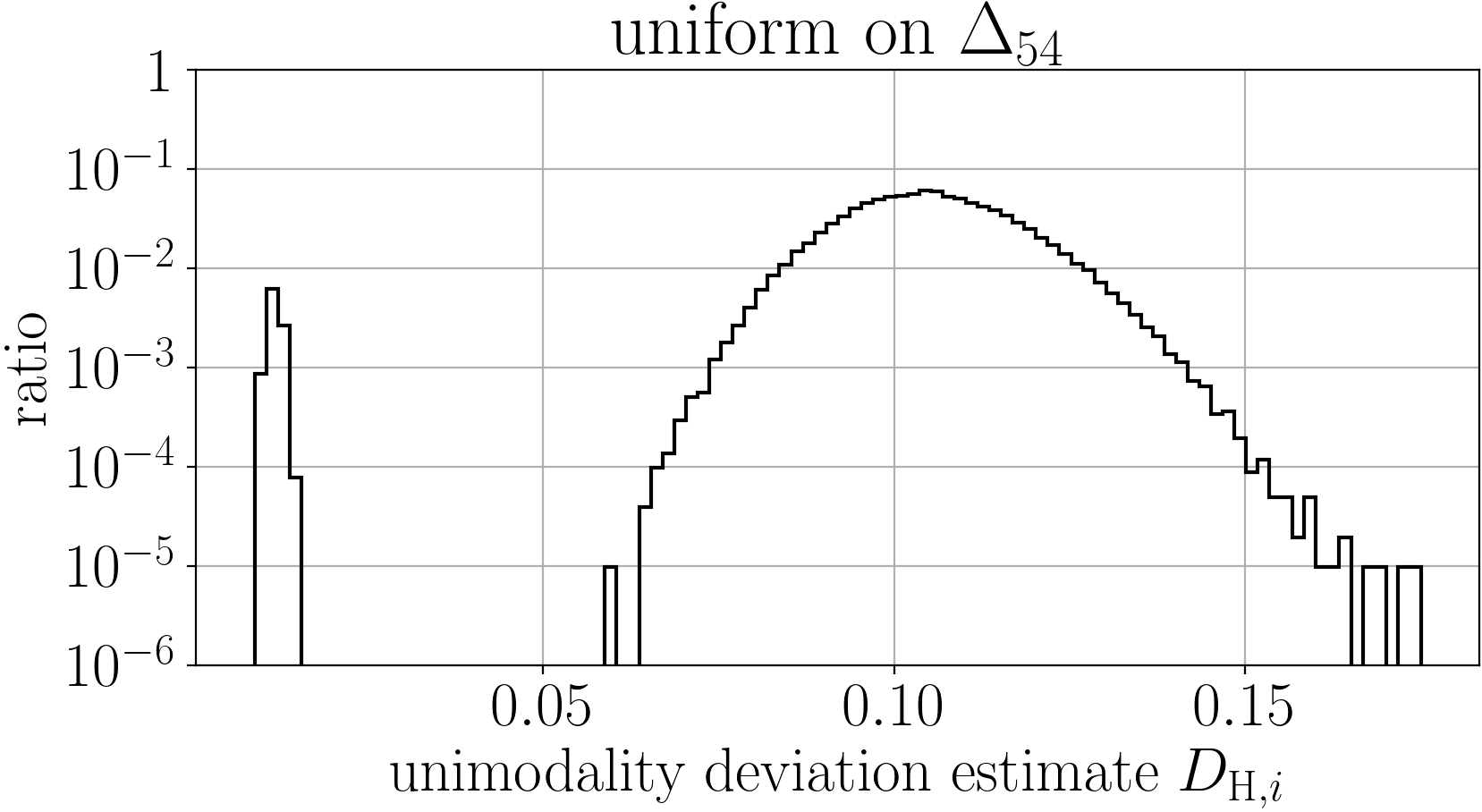}&
\end{tabular}
\caption{%
Log-scaled histogram (ratio of $D_{\rmH,i}$'s included in each of 100 bins)
of 100-trial (for SB datasets) or 5-trial 
(for CV datasets) aggregation of estimates $D_{\rmH,i}$ 
of $D_\rmH((\Pr(Y=y|\bX=\bx_i))_{y\in[K]},\hat{\Delta}_{K-1})$ 
for real-world ordinal data, and 100-trial aggregation 
for uniform random data on $\Delta_{K-1}$.}
\label{fig:Data-Unimodality}
\end{figure}

Our experiments also support this hypothesis:
see indicators of the UD,
sample-based estimation of mean HD (MHD)
\begin{align}
\label{eq:MHD}
	\bbE_{\bx\sim\bX}[D_\rmH((\Pr(Y=y|\bX=\bx))_{y\in[K]},\hat{\Delta}_{K-1})]
\end{align}
in Table~\ref{tab:Data-Prop}
and sample-based estimation of the probability distribution of HD
$D_\rmH((\Pr(Y=y|\bX=\bx))_{y\in[K]},\hat{\Delta}_{K-1})$
in Figure~\ref{fig:Data-Unimodality},
where we used an unconstrained likelihood model 
(the same SL model used to calculate the UR in 
Section~\ref{sec:Data}) for estimation of the CPD.%
\footnote{%
The maximum value (adopted in Definition~\ref{def:AULM2} 
below) of sample-based estimates of 
$D_\rmH((\Pr(Y=y|\bX=\bx))_{y\in[K]},\hat{\Delta}_{K-1})$
tends to increase with respect to the size of test data,
and hence is difficult to use for simple comparison.}

Therefore, for better modeling of low-UD data,
we propose an approximately unimodal likelihood model:
\begin{definition}[{$\epsilon$-approximately unimodal likelihood model}]
\label{def:AULM2}
For a PMF $\bp=(p_k)_{k\in[K]}\in\Delta_{K-1}$, 
we say that $\bp$ is {$\epsilon$-approximately unimodal} 
if it satisfies $\bp\in\hat{\Delta}_{K-1}^\epsilon$ with 
$\hat{\Delta}_{K-1}^\epsilon\coloneq\{\bp\in\Delta_{K-1}\mid D_\rmH(\bp,\hat{\Delta}_{K-1})\le\epsilon\}$.
Also, we say that a likelihood model $(P,\calG)$ is 
an ($\epsilon$-)approximately unimodal likelihood model 
if it satisfies
\begin{align}
	\calR(P,\calG)\subseteq
	\{\bp:\bbR^d\to\hat{\Delta}_{K-1}^\epsilon\}.
\end{align}
\end{definition}

In this paper, as an implementation of the approximately unimodal likelihood model,
we further propose a mixture-based approximately unimodal likelihood (MAUL) model:
\begin{align}
\label{eq:MAUL}
	\begin{split}
	&P_{\maul,r}(y;(\bg_1(\bx),\bg_2(\bx)))
\\
	&=(1-r)P_\ul(y;\bg_1(\bx))+r P_\sl(y;\bg_2(\bx))
	\end{split}
\end{align}
with $\bg_1\in\calG_\ul$, $\bg_2\in\calG_K$,
and a hyper-parameter $r\in[0,1]$, which we call the mixture rate,
for a certain unimodal likelihood model $(P_\ul,\calG_\ul)$
(here, the SL model can be replaced by another unconstrained likelihood model, 
but for simplicity we adopt the above formulation).
Figure~\ref{fig:MAUL} schematizes the construction of the MAUL model.
The representation ability of this model 
satisfies the following relationship:
\begin{theorem}[Representation ability of the MAUL model]
\label{def:MAUL}
Assume that a unimodal likelihood model $(P_\ul,\calG_\ul)$ satisfies 
$\calR(P_\ul,\calG_\ul)\subseteq\{\bp:\bbR^d\to\hat{\Delta}_{K-1}^\emptyset\}$,
and let $\calG_{\ul,K}=\{(\bg_1,\bg_2)\mid\bg_1\in\calG_\ul,\bg_2\in\calG_K\}$.
Then, the likelihood model $(P_{\maul,r},\calG_{\ul,K})$ 
defined in \eqref{eq:MAUL} with the mixture rate $r\in[0,1]$ satisfies
\begin{align}
\label{eq:MAULrep}
\begin{split}
	\calR(P_\ul,\calG_\ul)
	&\subseteq
	\calR(P_{\maul,r},\calG_{\ul,K})
\\
	&\subseteq
	\{\bp:\bbR^d\to(\hat{\Delta}_{K-1}^{\sqrt{2}r})^\emptyset\}.
\end{split}
\end{align}
Also, it holds that if $0\le r_1\le r_2\le 1$ then
\begin{align}
\label{eq:MAULrep2}
	\calR(P_{\maul,r_1},\calG_{\ul,K})
	\subseteq
	\calR(P_{\maul,r_2},\calG_{\ul,K}).
\end{align}
\end{theorem}
Note that we here introduced the assumption 
$\calR(P_\ul,\calG_\ul)\subseteq\{\bp:\bbR^d\to\hat{\Delta}_{K-1}^\emptyset\}$
on a unimodal likelihood model considering the simplicity of 
the statement and the convention of models using the SL link function.
\begin{proof}
The assumption ($(P_\ul(y;\bg_1(\bx)))_{y\in[K]}\in
\hat{\Delta}_{K-1}^\emptyset\subseteq\Delta_{K-1}^\emptyset$) 
and Theorem~\ref{thm:SL-REP} show that, 
for any $\bg_1\in\calG_\ul$, there exists $\bg_2\in\calG_K$ 
such that $P_\sl(y;\bg_2(\bx))=P_\ul(y;\bg_1(\bx))$ 
for any $\bx\in\bbR^d$ and $y\in[K]$.
For $\bg_1$ and such $\bg_2$, one has that
$P_{\maul,r}(y;(\bg_1(\bx),\bg_2(\bx)))=P_\ul(y;\bg_1(\bx))$ 
for any $(\bx,y)$.
This result shows the former part of \eqref{eq:MAULrep}.

Also, the latter part of \eqref{eq:MAULrep} can be proved from
\begin{align}
	&D_\rmH((P_{\maul,r}(y;(\bg_1(\bx),\bg_2(\bx))))_{y\in[K]},\hat{\Delta}_{K-1})
\nonumber\\
	&=\min_{\bq\in\hat{\Delta}_{K-1}}\|(P_{\maul,r}(y;(\bg_1(\bx),\bg_2(\bx))))_{y\in[K]}-\bq\|
\nonumber\\
	&(\because \text{definition $D_\rmH(\bu,S)\coloneq\min_{\bv\in S}\|\bu-\bv\|$})
\nonumber\\
	&=\min_{\bq\in\hat{\Delta}_{K-1}}\|((1-r)P_\ul(y;\bg_1(\bx))+r P_\sl(y;\bg_2(\bx)))_{y\in[K]}-\bq\|
\nonumber\\
	&(\because \text{definition of $P_{\maul,r}$})
\nonumber\\
	&\le\|((1-r)P_\ul(y;\bg_1(\bx))+r P_\sl(y;\bg_2(\bx)))_{y\in[K]}
\nonumber\\
	&\hphantom{\le\|}
	-(P_\ul(y;\bg_1(\bx)))_{y\in[K]}\|
\label{eq:The1}
\\
	&(\because \text{property of $\min$ and substitution}
\nonumber\\
	&\hphantom{(\because\;}
	\text{$\bq=(P_\ul(y;\bg_1(\bx)))_{y\in[K]}\in\hat{\Delta}_{K-1}$})
\nonumber\\
	&=r\|(P_\sl(y;\bg_2(\bx)))_{y\in[K]}-(P_\ul(y;\bg_1(\bx)))_{y\in[K]}\|
\nonumber\\
	&\le r\max_{\bu,\bv\in\Delta_{K-1}}\|\bu-\bv\|
\nonumber\\
	&(\because \text{property of $\max$, $(P_\sl(y;\bg_2(\bx)))_{y\in[K]}\in\Delta_{K-1}$ (as $\bu$),}
\nonumber\\
	&\hphantom{(\because\;}
	\text{and $(P_\ul(y;\bg_1(\bx)))_{y\in[K]}\in\Delta_{K-1}$ (as $\bv$)})
\nonumber\\
	&=\sqrt{2}r,
\nonumber
\end{align}
where $\max_{\bu,\bv\in\Delta_{K-1}}\|\bu-\bv\|=\sqrt{2}$ 
applied in the last equality of \eqref{eq:The1} is shown by
\begin{align}
	2
	&=\|(1,0,\ldots,0)^\top-(0,\ldots,0,1)^\top\|^2
\nonumber\\
	&\le
	(\max_{\bu,\bv\in\Delta_{K-1}}\|\bu-\bv\|)^2
	=\max_{\bu,\bv\in\Delta_{K-1}}\|\bu-\bv\|^2
\nonumber\\
	&(\because \text{property of $\max$, $(1,0,\ldots,0)^\top\in\Delta_{K-1}$ (as $\bu$),} 
\nonumber\\
	&\hphantom{(\because\;}
	\text{and $(0,\ldots,0,1)^\top\in\Delta_{K-1}$ (as $\bv$)})
\nonumber\\
	&=\max_{\bu,\bv\in\Delta_{K-1}}\left(\sum_{k=1}^Ku_k^2+v_k^2-2u_kv_k\right)
\nonumber\\
	&\le\max_{\bu,\bv\in\Delta_{K-1}}\left(\sum_{k=1}^Ku_k+v_k-2u_kv_k\right)
\\
	&(\because \text{$u^2\le u$ for $u\in[0,1]$})
\nonumber\\
	&=2-2\left(\min_{\bu,\bv\in\Delta_{K-1}}\sum_{k=1}^K u_kv_k\right)
\nonumber\\
	&(\because \text{$\textstyle\sum_{k=1}^Ku_k=\sum_{k=1}^Kv_k=1$ for $\bu,\bv\in\Delta_{K-1}$})
\nonumber\\
	&=2
\nonumber\\
	&(\because\text{property of $\min$ and $\textstyle\sum_{k=1}^K u_kv_k=0$ for }
\nonumber\\
	&\hphantom{(\because\;}
	\text{$\bu=(1,0,\ldots,0)^\top,\bv=(0,\ldots,0,1)^\top\in\Delta_{K-1}$}).
\nonumber
\end{align}

One has that $(r_2-r_1)P_\ul(y;\bg_{1,1}(\bx))+
r_1P_\sl(y;\bg_{2,1}(\bx))\in\{r_2\bp\mid\bp\in\Delta_{K-1}^\emptyset\}$
since $r_1\le r_2$.
Thus, for any $\bg_{1,1}\in\calG_\ul$ and $\bg_{2,1}\in\calG_K$, 
there exists $\bg_{2,2}\in\calG_K$ such that 
$(r_2-r_1)P_\ul(y;\bg_{1,1}(\bx))+r_1 P_\sl(y;\bg_{2,1}(\bx))=r_2 P_\sl(y;\bg_{2,2}(\bx))$.
For such $\bg_{2,2}$, one has that 
$P_{\maul,r_1}(y;(\bg_{1,1}(\bx),\bg_{2,1}(\bx)))
=P_{\maul,r_2}(y;(\bg_{1,1}(\bx),\bg_{2,2}(\bx)))$ at any $(\bx,y)$.
This result proves the relation \eqref{eq:MAULrep2}.

These conclude the proof of this theorem.
\end{proof}
As a representative MAUL model,
we propose a model that is consisted of the VSL and SL models.
For this model,
the relation \eqref{eq:MAULrep} reduces to the following one:
\begin{corollary}[{Corollary of Theorem~\ref{def:MAUL}}]
\label{cor:MAUL}
The likelihood model $(P_{\maul,r},\calG_{2K})$ defined in 
\eqref{eq:MAUL} with $(P_\ul,\calG_\ul)=(P_\vsl,\calG_K)$ for any functions 
$\rho$ satisfying \eqref{eq:RHO} and $\tau$ satisfying \eqref{eq:TAU}
and the mixture rate $r\in[0,1]$ satisfies
\begin{align}
\label{eq:MAULrep3}
	\begin{split}
	\{\bp:\bbR^d\to\hat{\Delta}_{K-1}^\emptyset\}
	&\subseteq
	\calR(P_{\maul,r},\calG_{2K})
\\
	&\subseteq
	\{\bp:\bbR^d\to(\hat{\Delta}_{K-1}^{\sqrt{2}r})^\emptyset\}.
\end{split}
\end{align}
\end{corollary}

The former relation implies that the likelihood model \eqref{eq:MAUL}
with $(P_\ul,\calG_\ul)=(P_\vsl,\calG_K)$ can represent 
any unimodal CPD ignoring the border issue,
and the latter relation implies that the model is 
a $\sqrt{2}r$-approximately unimodal likelihood model.%
\footnote{%
We are interested in whether $\calR(P_{\maul,r},\calG_{2K})=
\{\bp:\bbR^d\to(\hat{\Delta}_{K-1}^{\sqrt{2}r})^\emptyset\}$ 
(not \eqref{eq:MAULrep3}) holds, but we could not solve this question.}
According to the relation \eqref{eq:MAULrep2} of Theorem~\ref{def:MAUL},
the mixture rate $r\in[0,1]$ controls the tolerance of 
the UD of the likelihood model $(P_{\maul,r},\calG_{\ul,K})$,
and using larger $r$ makes the representation ability of the model higher:
the setting $r=0$ enforces the likelihood model to be unimodal,
and the setting $r=1$ tolerates the likelihood model to represent any CPD.
Considering the bias-variance trade-off,
it would be better to choose an appropriate value
of $r$ for the data through a validation process.

\section{Discussion: Relation between Other Ordinal Regression Methods and Unimodality}
\label{sec:Discussion}
\subsection{Monotonic Classification}
\label{sec:Monotonic}
The terminology `unimodality' appears in the context of 
`shape-constrained models or inference' in the machine 
learning literature \cite{brunk1955maximum,ZHU2017205,
dembczynski2009learning,pya2015shape,dumbgen2024shape}.
The unimodal or approximately unimodal likelihood models
would be seen as instances of shape-constrained models.

A representative instance of shape-constrained models is 
the `monotonic' model (also called the isotonic model) 
\cite{brunk1955maximum,ZHU2017205,dembczynski2009learning}.
For example, monotonic classification 
\cite{ZHU2017205,dembczynski2009learning} 
learns a learner model $\bg$ so that each element of $\bg$ 
becomes monotonic with respect to a pre-specified ordering 
of the values of the explanatory variable $\bX$.
On the other hand, the unimodal or approximately unimodal likelihood models 
do not impose any ordering on the explanatory variable $\bX$.
For complex data such as image data, it would be 
hard to impose an appropriate ordering on $\bX$, 
but the unimodal or approximately unimodal likelihood models 
do not suffer from such a difficulty.

\subsection{Ordinal Decomposition Approach}
\label{sec:OD}
The ordinal decomposition (OD) approach decomposes 
an original OR problem into several sub-classification 
problems for contiguous class groups, and predicts 
a class label by integrating model outputs for sub-problems 
\cite[Section 3.2]{gutierrez2015ordinal}. 
Sub-classification problems following the OD approach 
typically take a format of binary classification problems, 
`$k$ or less' vs.\,`more than $k$', $k$ vs.\,$(k+1)$, 
and $k$ vs.\,`more than $k$'.

For example, the previous works \cite{frank2001simple,
cardoso2007learning,cheng2008neural,niu2016ordinal} 
decompose the original OR problem into sub-problems 
`$k$ or less' vs.\,`more than $k$', $k=1,\ldots,K-1$, and employ 
a binary logistic regression model to each sub-problem.
Their model is called cumulative logits (CL) 
model in \cite{agresti2010analysis} and can be 
formalized as using the CL link function
\begin{align}
\label{eq:CL}
	P_{\cl}(y;\bu)\coloneq
	\begin{cases}
	\frac{1}{1+e^{-u_1}}&\text{for }y=1,\\
	1-\frac{1}{1+e^{-u_{K-1}}}&\text{for }y=K,\\
	\frac{1}{1+e^{-u_y}}-\frac{1}{1+e^{-u_{y-1}}}&\text{for }y=2,\ldots,K-1,
	\end{cases}
\end{align}
and a learner class in $\{\rho[\bg]\mid\bg\in\calG_{K-1}\}$ 
with a non-negative function $\rho$ satisfying 
\eqref{eq:RHO} to ensure $P_{\cl}(y;\bu)\in[0,1]$.

Although this model may appear 
to make good use of the ordinal relation, 
this model is an unconstrained model, and 
the representation ability \eqref{eq:RA} of this model is 
the same as the SL model (see Theorem~\ref{thm:SL-REP}) 
when $\{\rho(u)\mid u\in\bbR\}=(0,\infty)$.
Also, likelihood models that apply a binary logistic regression 
model to the decomposition 
$k$ vs.\,$(k+1)$ or $k$ vs.\,`more than $k$' are known as 
the adjacent-categories or continuation-ratio logits model 
\cite{agresti2010analysis,yamasaki2022unimodal,LIU202034}, 
and their situation regarding the representation ability 
is similar to that for the CL model.
Thus, the OD approach cannot be said to exploit 
the unimodality of ordinal data in the likelihood model.

\subsection{One-Dimensional Transformation Approach}
\label{sec:ODT}
One popular approach for modeling ordinal data is
the one-dimensional transformation (OT) approach:
it assumes a one-dimensional structure underlying the data
and implements a learner model with a one-dimensional 
transformation $a(\cdot)$ of the explanatory variable $\bX$.
This approach has been studied as the proportional odds (PO) 
constraint to the likelihood model in statistical literature
\cite{mccullagh1980regression, agresti2010analysis,cao2020rank,yamasaki2022unimodal}, 
or under the name of `threshold models or methods' in the machine learning literature
\cite{yamasaki2023optimal,yamasaki2024parallel,yamasaki2024remarks}.

\begin{figure}[!t]
\centering%
\renewcommand{\tabcolsep}{1.pt}%
\begin{tabular}{cc}
\includegraphics[height=2.35cm, bb=0 0 633 348]{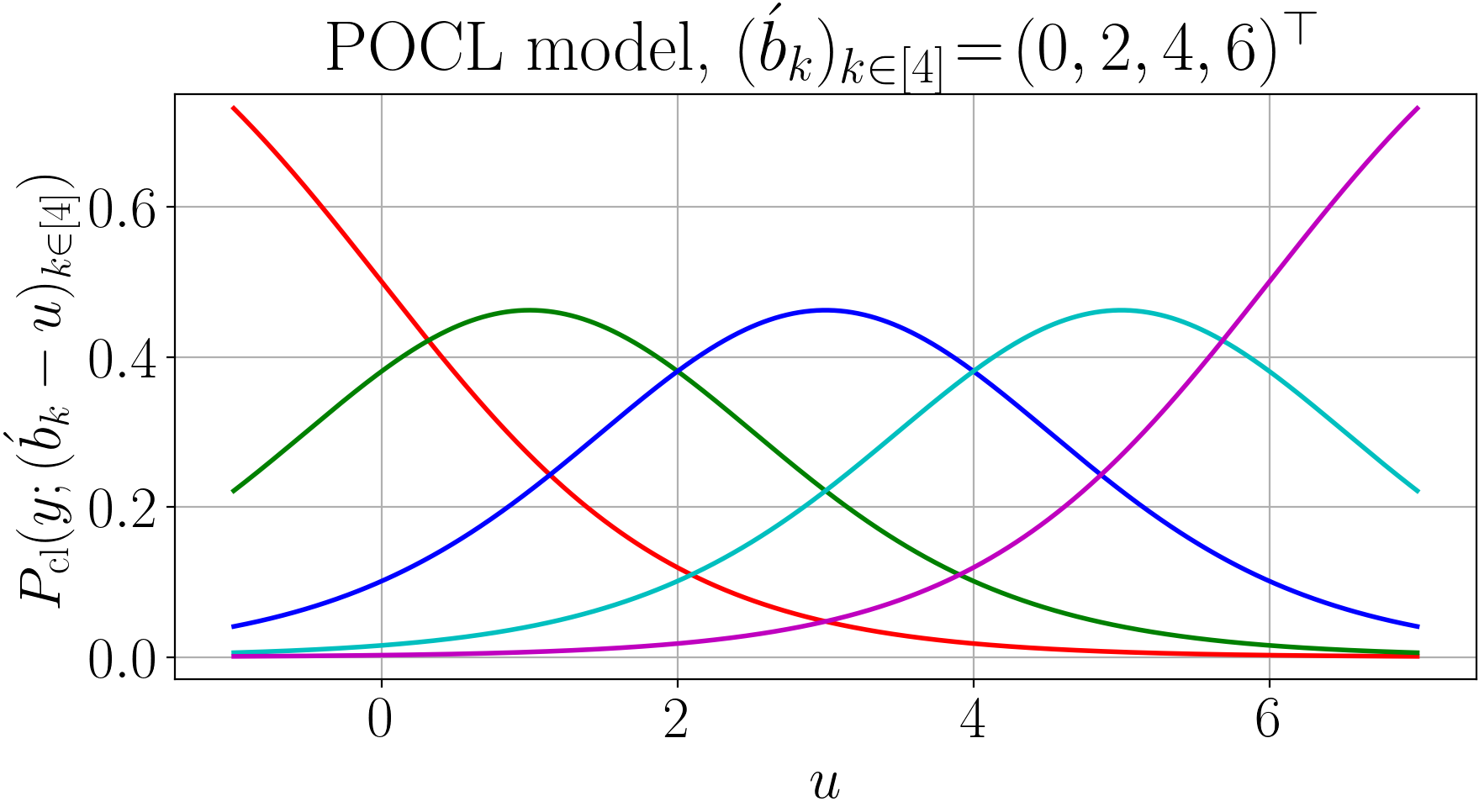}&
\includegraphics[height=2.35cm, bb=0 0 633 348]{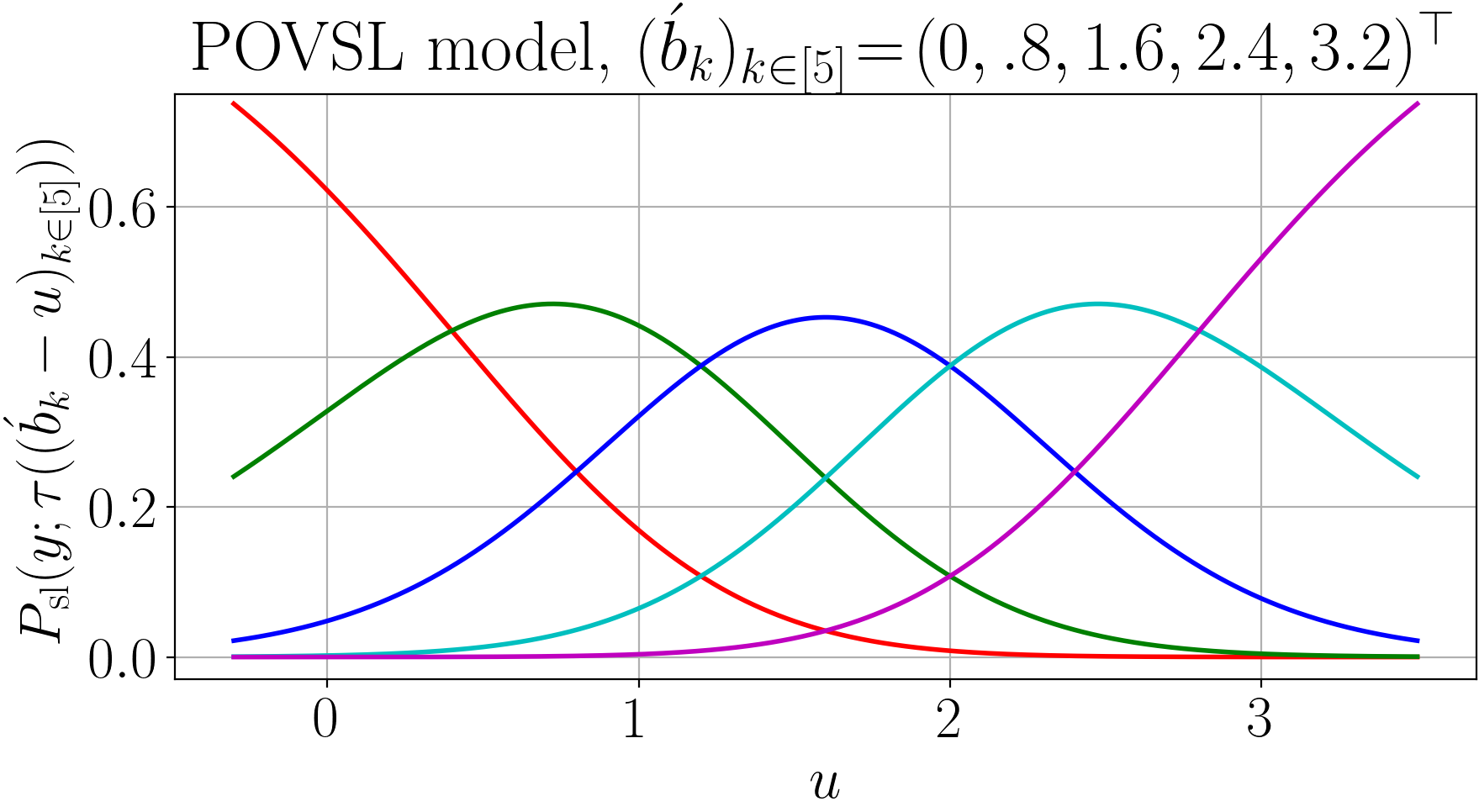}
\end{tabular}
\caption{%
Illustration of the POCL and POVSL models:
$P_{\rm cl}(y;(\acute{b}_k-u)_{k\in[K-1]})$ and 
$P_{\rm sl}(y;\tau((\acute{b}_k-u)_{k\in[K]}))$ with $\tau(u)=u^2$
for $y=1,\ldots,K$ (red, green, blue, cyan, magenta) with $K=5$.}
\label{fig:PO}
\end{figure}

For example, PO-constrained CL (POCL or ordinal logistic regression) model
\cite{mccullagh1980regression, agresti2010analysis,cao2020rank}
is representative, and it adopts the CL link function \eqref{eq:CL} 
and an ordered PO-constrained learner class in 
$\calG_{\text{ord-po},K-1}\coloneq\{(\acute{b}_k-a(\cdot))_{k\in[K-1]}
\mid a:\bbR^d\to\bbR,\,(b_k)_{k\in[K-1]}\in\bbR^{K-1},\,
(\acute{b}_k)_{k\in[K-1]}=\rho[(b_k)_{k\in[K-1]}]\}$.
In addition, the previous work \cite{yamasaki2022unimodal}
studied various PO-constrained models, such as 
PO-constrained VSL (POVSL) model
$(P_\sl(y;\tau(\cdot)),\calG_{\text{ord-po},K})$.
The POVSL model is a unimodal likelihood model, 
while the POCL model has no guarantee 
that a predicted CPD becomes unimodal.
However, one can see from the similarity between 
the POCL and POVSL models in Figure~\ref{fig:PO} that 
a learned POCL model often outputs an unimodal or approximately unimodal CPD.
In this way, many of existing PO-constrained models 
would be roughly understood as likelihood models that can 
represent a part of unimodal or approximately unimodal CPDs.

The PO constraint strongly restricts 
the representation ability of the likelihood model.
Therefore, the success or failure of PO-constrained models depends 
heavily on their compatibility with the data distribution, 
and it is known empirically that 
they often can achieve better prediction performance 
than unconstrained likelihood models 
only when the size of training data is small
\cite{yamasaki2022unimodal,gutierrez2015ordinal}.

\subsection{Unimodality-Promoting Regularized Learning}
\label{sec:UPRL}
Another promising approach to leverage the unimodality of ordinal data, 
other than devising a likelihood model, is to devise a learning procedure
\cite{belharbi2020,albuquerque2021ordinal,albuquerque2022quasi}.
Such a device is called regularization, and typically addressed 
by changing the objective function used for learning a model:
it adds a regularization term $\lambda\Omega(\bg)$
to the conventional objective function,
where $\lambda>0$ and $\Omega$ are called 
regularization parameter and regularizer, respectively.
In order to reflect the unimodality of ordinal data to the learning result,
it would be better that the regularizer $\Omega(\bg)$ takes a smaller value 
when a likelihood $(P(y;\bg(\cdot)))_{y\in[K]}$ is closer to be unimodal;
we call such methods unimodality-promoting regularized learning (UPRL) 
methods in this paper.

With the aim of penalizing the non-unimodality of a likelihood 
$(P(y;\bg(\cdot)))_{y\in[K]}$, the previous works 
\cite{belharbi2020,albuquerque2021ordinal} 
developed a UPRL method that employs the regularizer
\begin{align}
	&\frac{1}{n}\sum_{i=1}^n\sum_{k=1}^{K-1}
	\Bigl(\bbI(k<y_i)[\delta+P(k;\bg(\bx_i))-P(k+1;\bg(\bx_i))]_+
\nonumber\\
\label{eq:alb21}
	&+\bbI(k\ge y_i)[\delta+P(k+1;\bg(\bx_i))-P(k;\bg(\bx_i))]_+\Bigr),
\end{align}
where $[\cdot]_+\coloneq\max\{0,\cdot\}$,
and where \cite{albuquerque2021ordinal} 
recommended to use $\delta=0.05$.
The succeeding work \cite{albuquerque2022quasi} 
developed similar variants.
Note that they considered only an unconstrained 
(SL) model as a likelihood model $(P,\calG)$.

The authors of these works also explained behaviors 
of these methods by the bias-variance trade-off;
increasing the regularization parameter $\lambda$ 
increases a bias-dependent term of the prediction performance, 
and decreases a variance-dependent term.
Although these previous works did not examine the dependence of 
the prediction performance of UPRL methods on the size of training data,
the effectiveness will be limited to small-size training data 
as in approximately unimodal likelihood models
(as we will verify in Section~\ref{sec:ExperimentsII})
if the explanation based on the bias-variance trade-off is valid.
Therefore, UPRL is competitive with approximately unimodal 
likelihood models, but they can also be combined.

\section{Experiments I}
\label{sec:Experiments}
\subsection{Experimental Purposes}
\label{sec:Purposes}
From the bias-variance trade-off, we expect that 
likelihood models that can adequately represent the data but 
have as low a representation ability as possible will yield good 
prediction performance in the conditional probability estimation task.
More specifically, in statistical modeling of ordinal data that are high-UR and low-UD, 
we expect that the MAUL model may lead to better prediction 
performance depending on the size of the training data:
unconstrained likelihood models
are promising when the training data size is large enough,
while unimodal likelihood models or the MAUL model 
are promising when the training data size is small.
We performed numerical experiments (Experiments I)
comparing different likelihood models
with the purposes to verify this expectation and 
the practical effectiveness of the MAUL model.

\subsection{Experimental Settings}
\label{sec:Settings}
In the experiments, we used 24 datasets in total:
21 SB datasets with the total data size $n_\tot\ge1000$ among those 
used in experiments of the previous OR study \cite{gutierrez2015ordinal},
and 3 CV datasets, AFAD \cite{niu2016ordinal}, 
CACD \cite{chen2014cross}, and MORPH \cite{ricanek2006morph};
see Section~\ref{sec:Data} and supplement for detailed explanation of the datasets.
The data of these datasets have been confirmed to be high-UR and low-UD 
as shown in Table~\ref{tab:Data-Prop} and Figure~\ref{fig:Data-Unimodality}.

We experimented with 6 settings of the training data size, 
$n_\tra=25,50,100,200,400,800$ for SB datasets or
$n_\tra=1250,2500,5000,10000,20000,40000$ for CV datasets, 
to see the dependence of behaviors of each model on the training data size.

We compared 6 models in Experiments I:
the SL model as an unconstrained model, 
VSL model as a unimodal model, 
MAUL model based on the VSL and SL models 
(which we denote $\text{Mix}_{\mbox{\tiny\text{VSL,SL}}}$ model)
with the mixture rate $r=0.05,0.1,\ldots,0.95$,
CL model in the OD approach (an unconstrained model),
and POCL \cite{mccullagh1980regression,cao2020rank} and 
POVSL \cite{yamasaki2022unimodal} models in the OT approach 
(the POVSL model is a unimodal model).
For the VSL, $\text{Mix}_{\mbox{\tiny\text{VSL,SL}}}$, 
CL, POCL, and POVSL models, 
we used $\rho(u)=e^u$ and $\tau(u)=u^2$ following 
\cite{yamasaki2022unimodal} (supplement shows 
similar results for other $\rho$ and $\tau$).

\begin{figure}[!t]
\centering%
\renewcommand{\tabcolsep}{1.pt}%
\begin{tabular}{ccc}%
\includegraphics[height=2.75cm, bb=0 0 577 251]{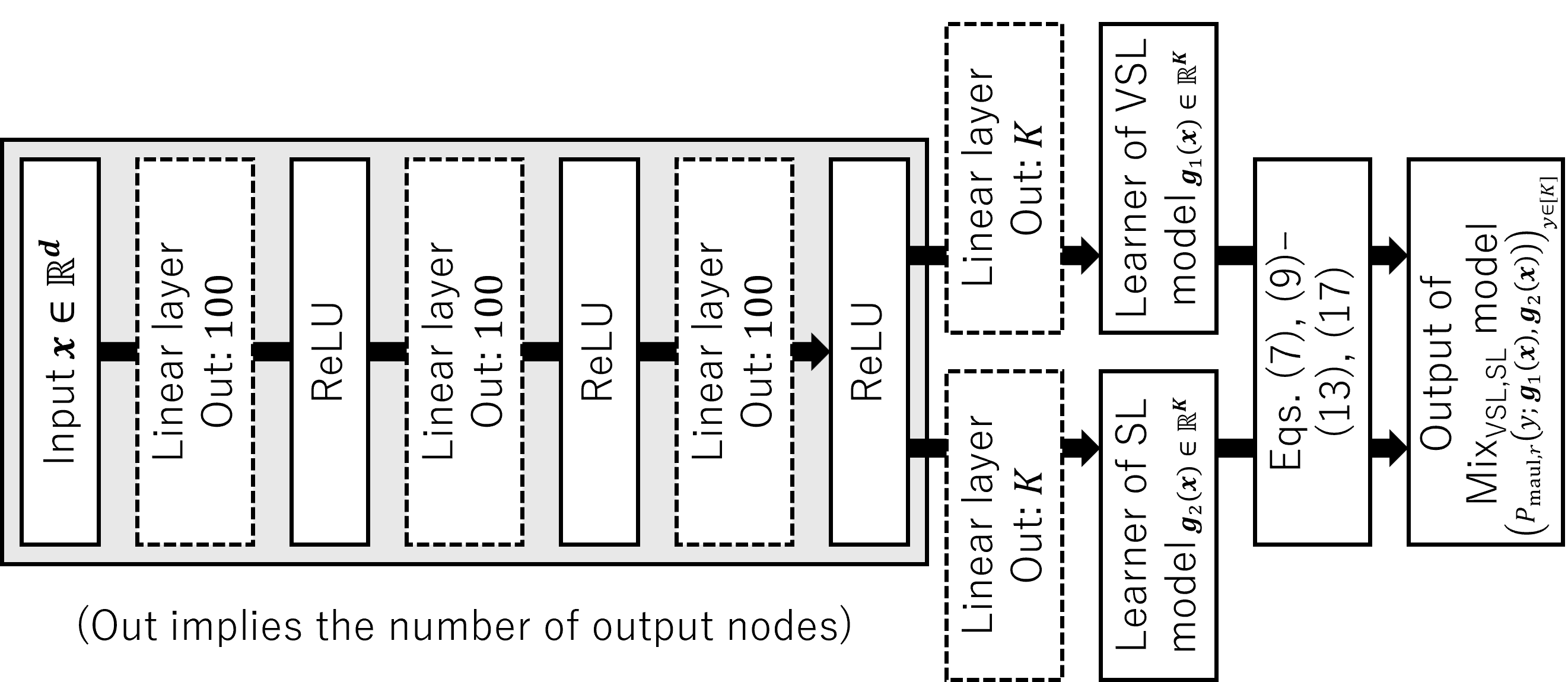}&
~&
\raisebox{13pt}{\includegraphics[height=1.75cm, bb=0 0 186 158]{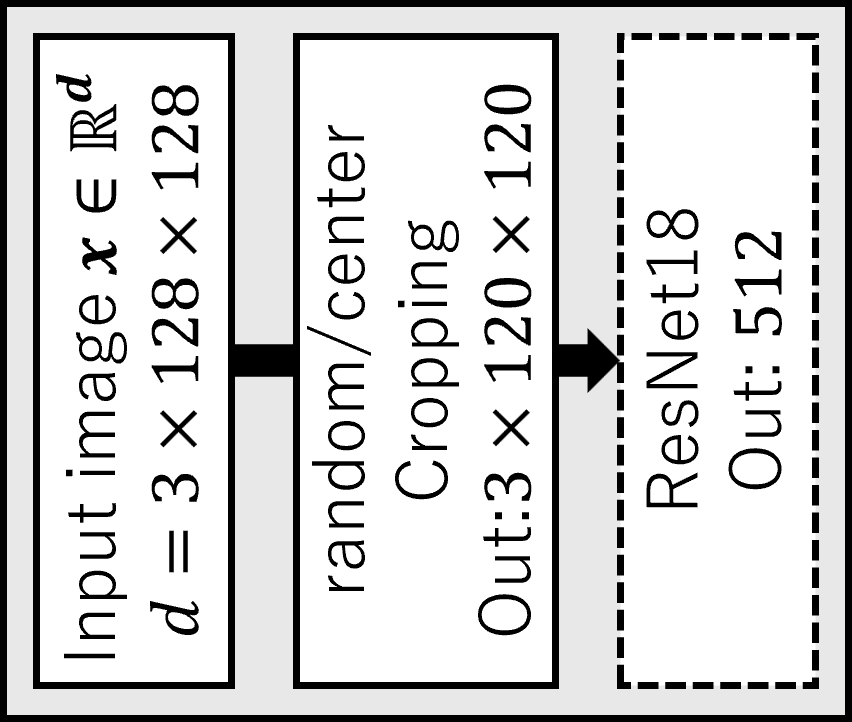}}
\end{tabular}
\caption{%
Illustration of the model implementation in experiments:
The left one is for the $\text{Mix}_{\mbox{\tiny\text{VSL,SL}}}$ model for SB datasets.
The gray-boxed part is common for other models for SB datasets,
and dotted boxes include parameters to be learned.
For CV datasets, the gray-boxed part was replaced with the right one.}
\label{fig:Implementation}
\end{figure}

We implemented all the learner models with a neural network 
shown in Figure~\ref{fig:Implementation}
(see supplement for comparison with other implementations).
For SB datasets, we used a 4-layer fully-connected neural network model 
in which every hidden layer has 100 nodes activated with the ReLU function 
in addition to bias nodes.
For CV datasets, we used ResNet18 \cite{he2016deep},
and all the images with $128\times128\times3$ pixels are 
randomly cropped with $120\times120\times3$ pixels 
as inputs to ResNet18 in a training phase so as to improve 
the stability of the model against the difference of the position of images
and center-cropped to the same size in validation and test phases,
following experimental procedures by \cite{cao2020rank}.

We trained a model by Adam optimization with the NLL as the objective function,
mini-batch size 16 for SB datasets or 256 for CV datasets,
and ascending learning rate $10^{2t/300-4}$ at $t$-th epoch during 300 epochs.
At the end of each epoch, we evaluated the errors, 
the NLL for a conditional probability estimation task 
and the MZE, MAE, and MSE with a likelihood-based classifier for OR tasks, 
with validation data of the size $n_\val=100$ for SB datasets 
or $n_\val=5000$ for CV datasets.
We then adopted a model at the epoch when the validation error 
got minimum (and simultaneously selected the mixture rate $r$ 
for the $\text{Mix}_{\mbox{\tiny\text{VSL,SL}}}$ model as $\bar{r}$), 
and calculated the error with remaining test data of 
the size $(n_\tot-n_\tra-n_\val)$ for that model.

We repeated the above procedure 100 trials 
for SB datasets or 5 trials for CV datasets 
with a random settings of the data split and 
initial parameters to obtain test errors.

\subsection{Experimental Results}
\label{sec:Results}
\begin{figure}[!t]
\centering%
\renewcommand{\arraystretch}{0.1}%
\renewcommand{\tabcolsep}{0pt}%
\begin{tabular}{C{2.95cm}C{2.95cm}C{2.95cm}}%
\includegraphics[height=1.54cm, bb=0 0 597 327]{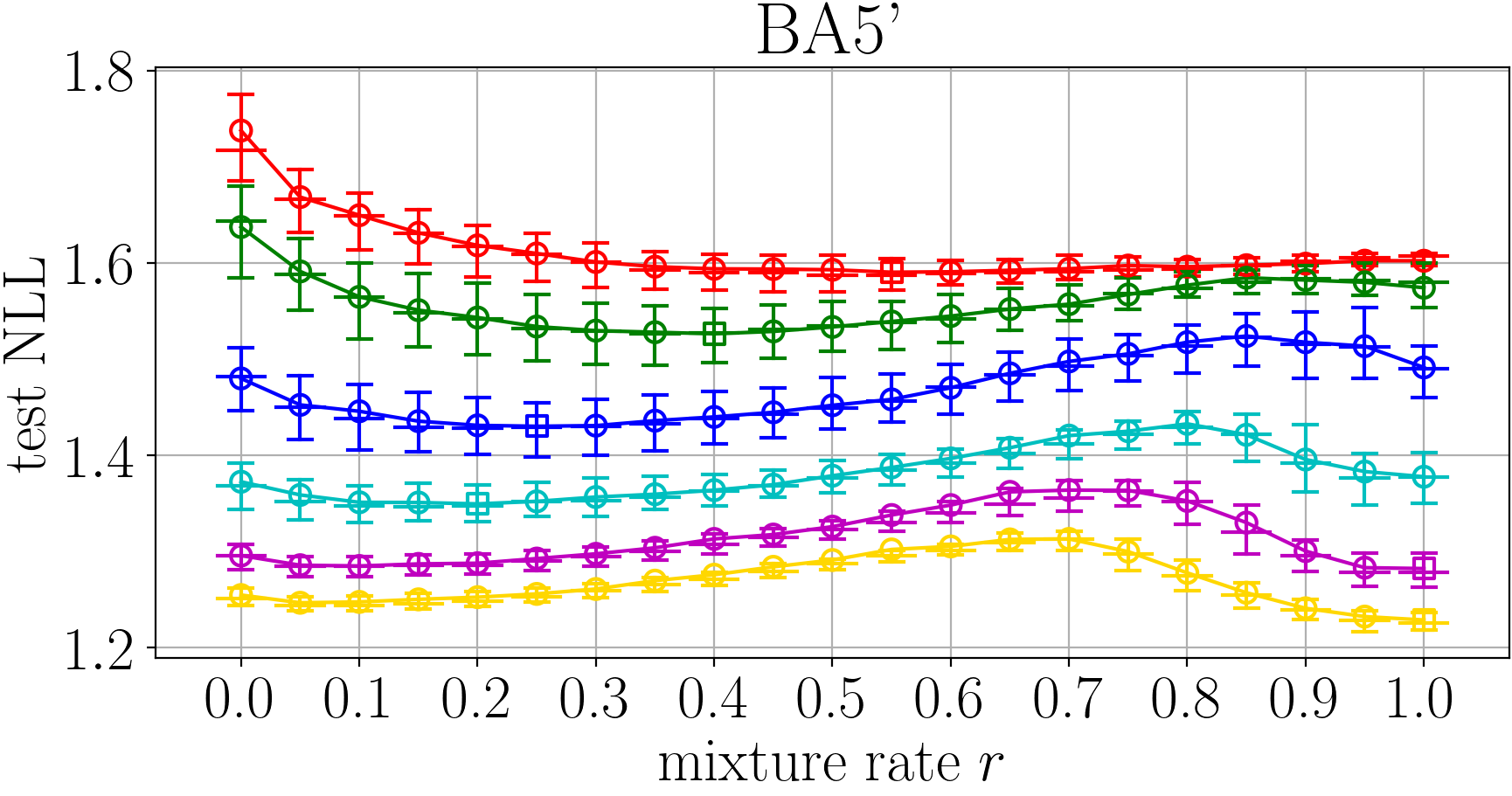}&
\includegraphics[height=1.54cm, bb=0 0 597 327]{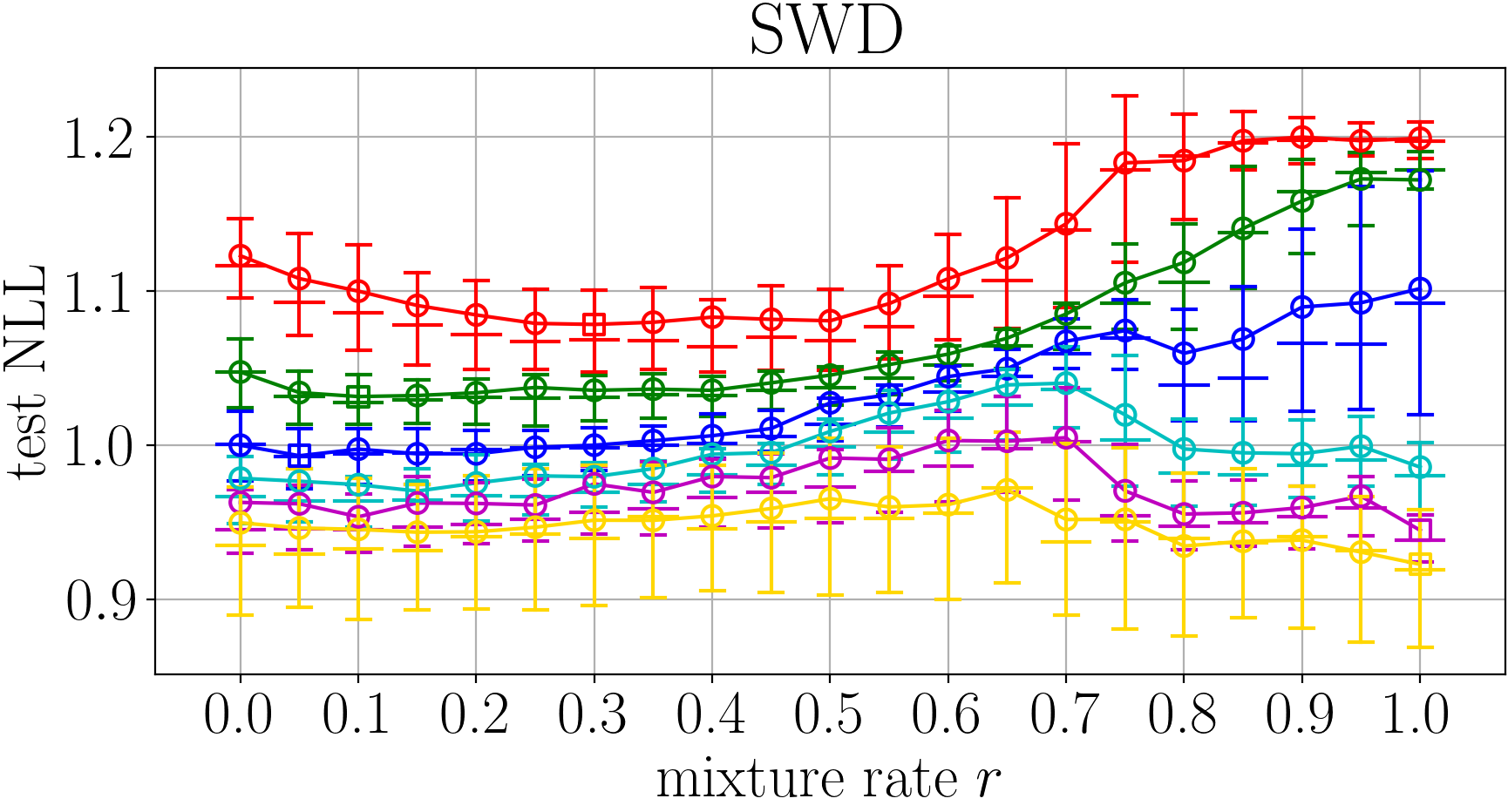}&
\includegraphics[height=1.54cm, bb=0 0 597 327]{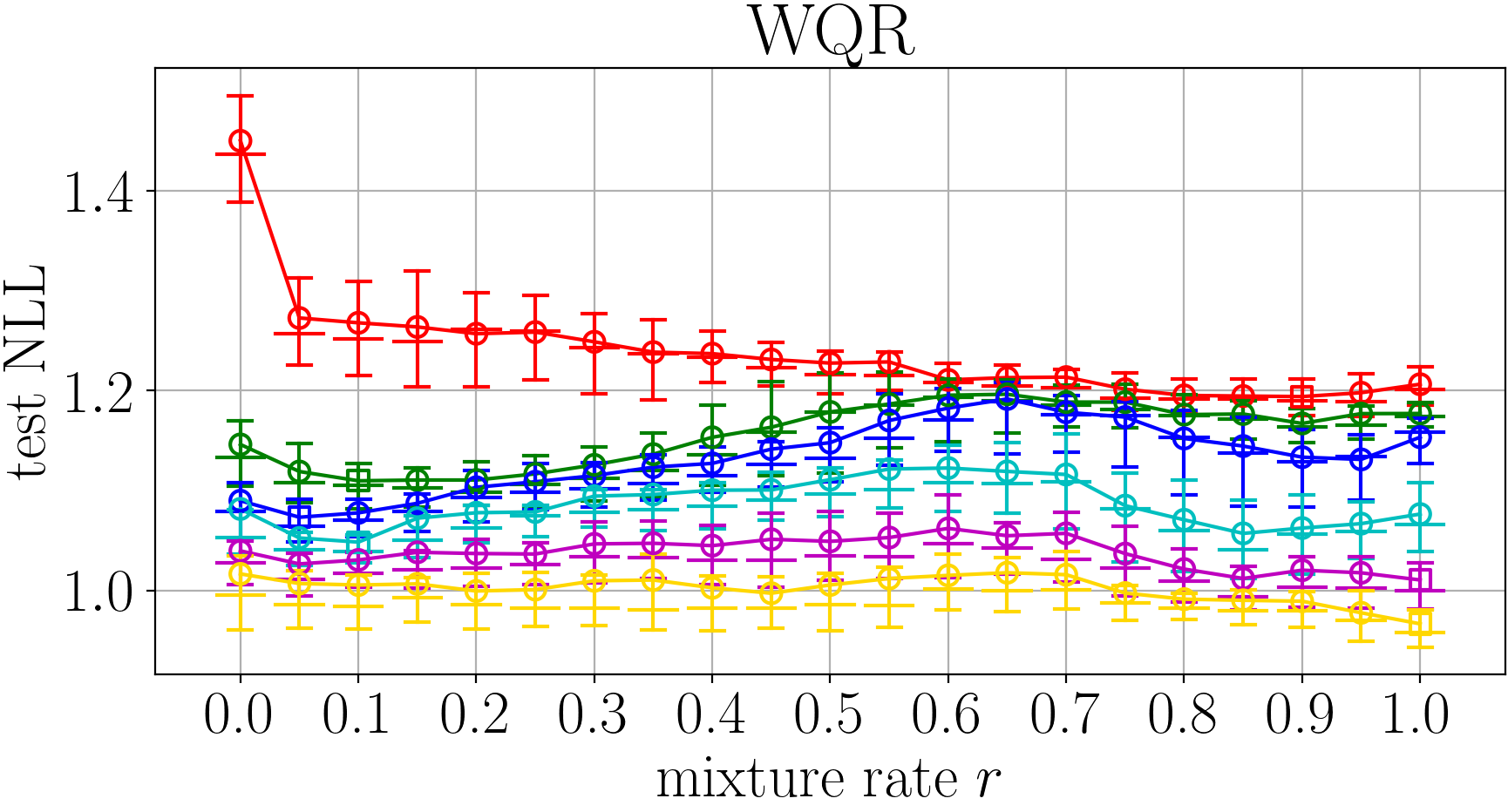}\\
\includegraphics[height=1.54cm, bb=0 0 597 327]{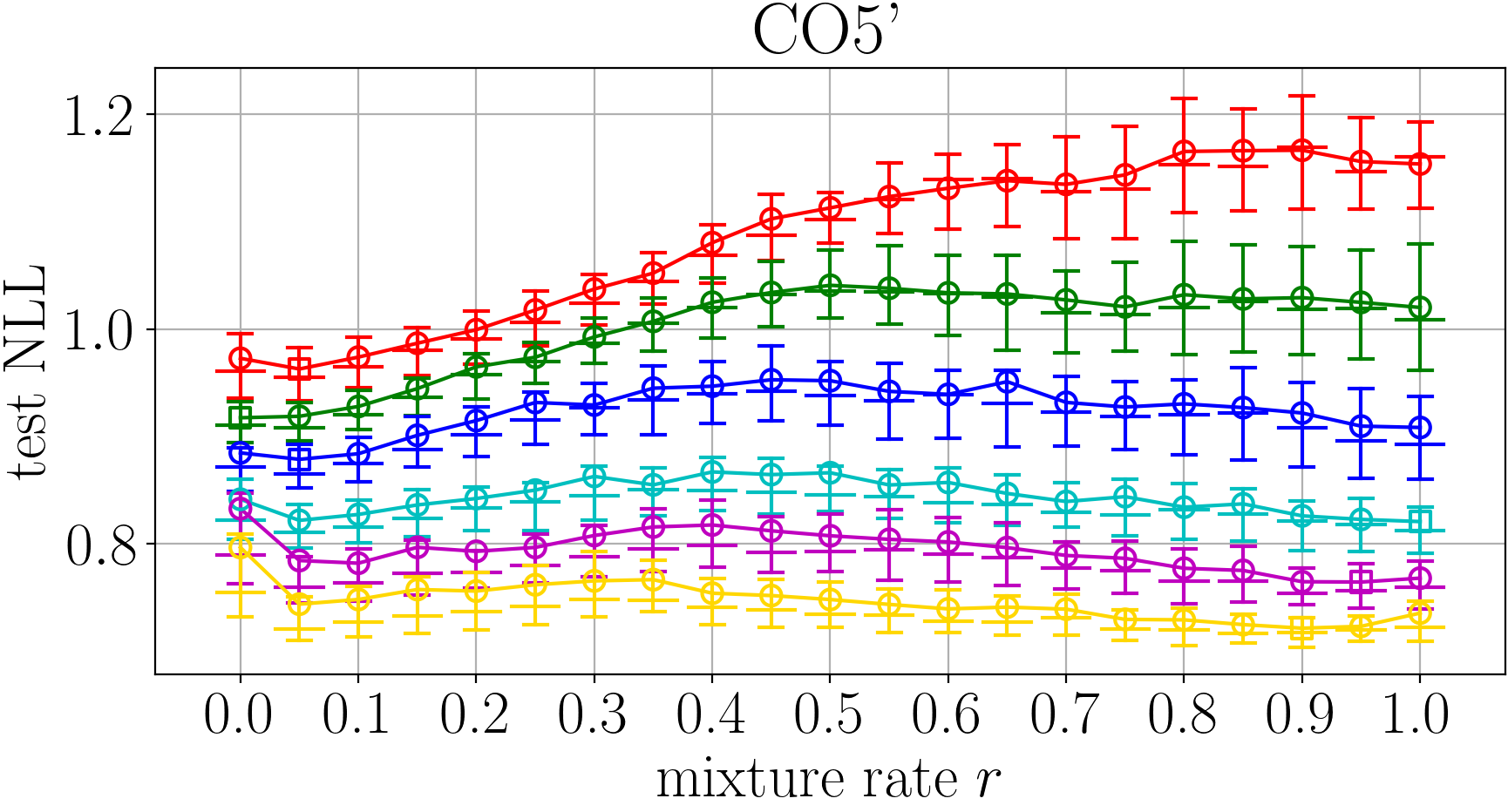}&
\includegraphics[height=1.54cm, bb=0 0 597 327]{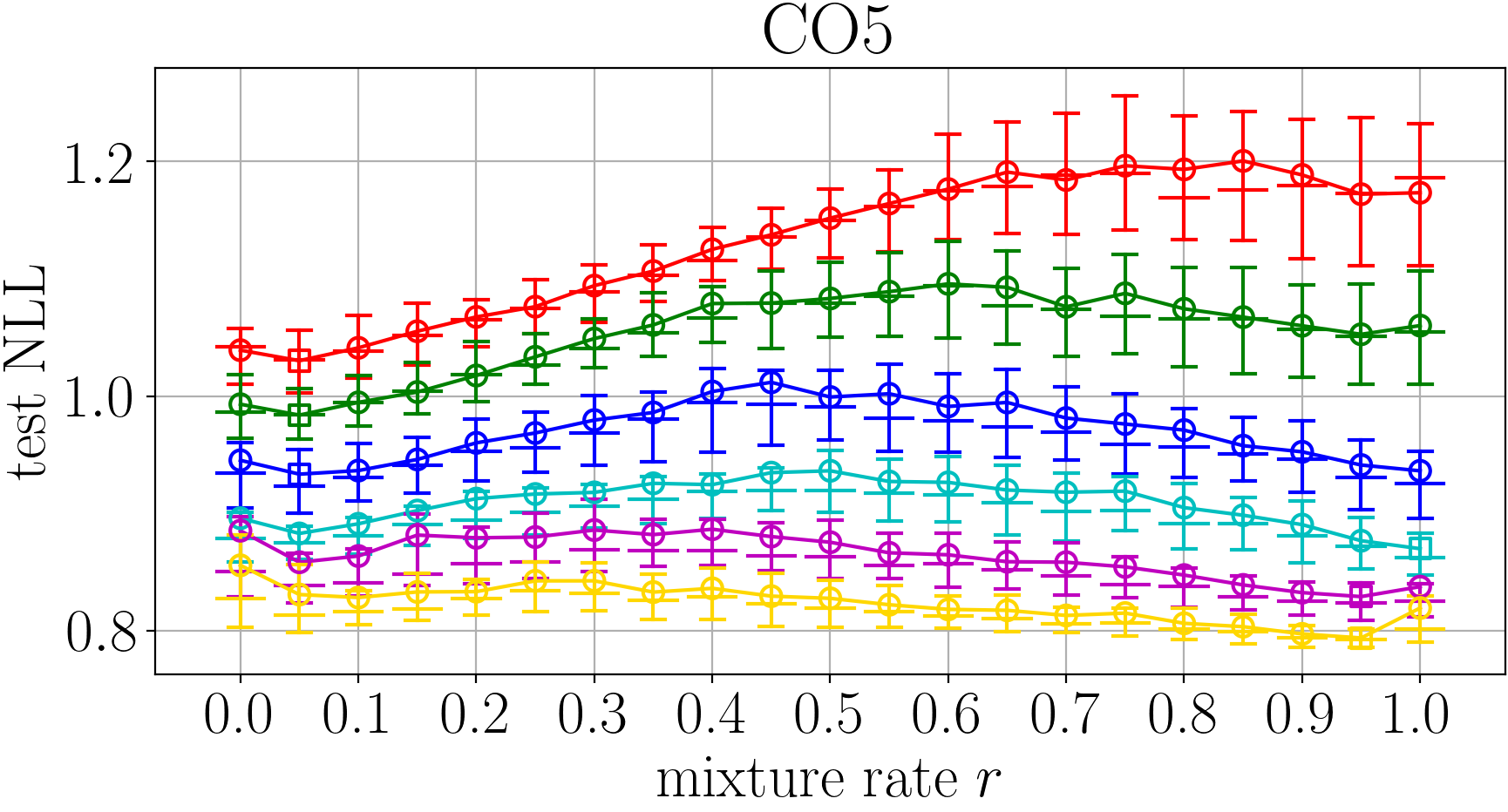}&
\includegraphics[height=1.54cm, bb=0 0 597 327]{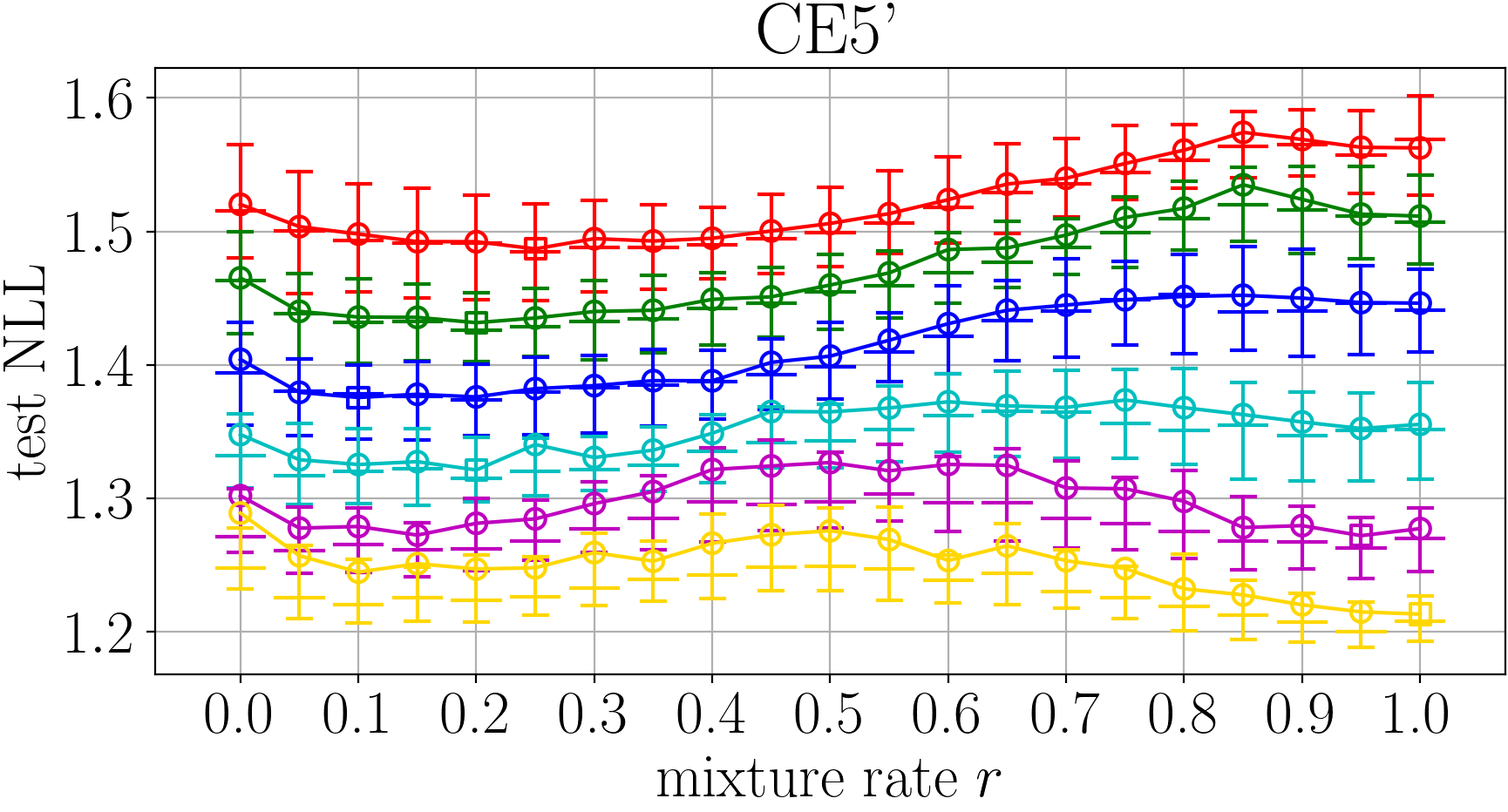}\\
\includegraphics[height=1.54cm, bb=0 0 597 327]{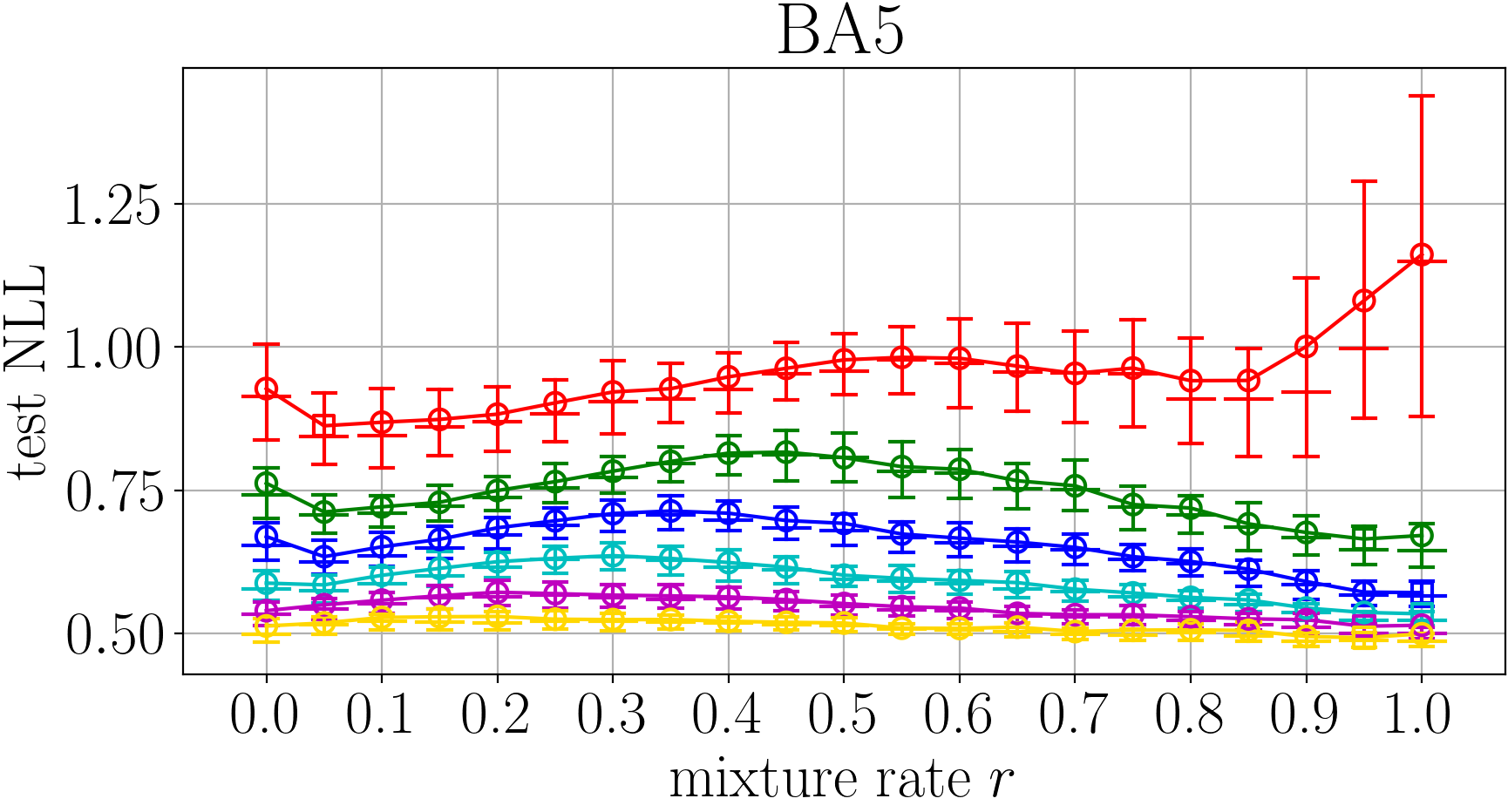}&
\includegraphics[height=1.54cm, bb=0 0 597 327]{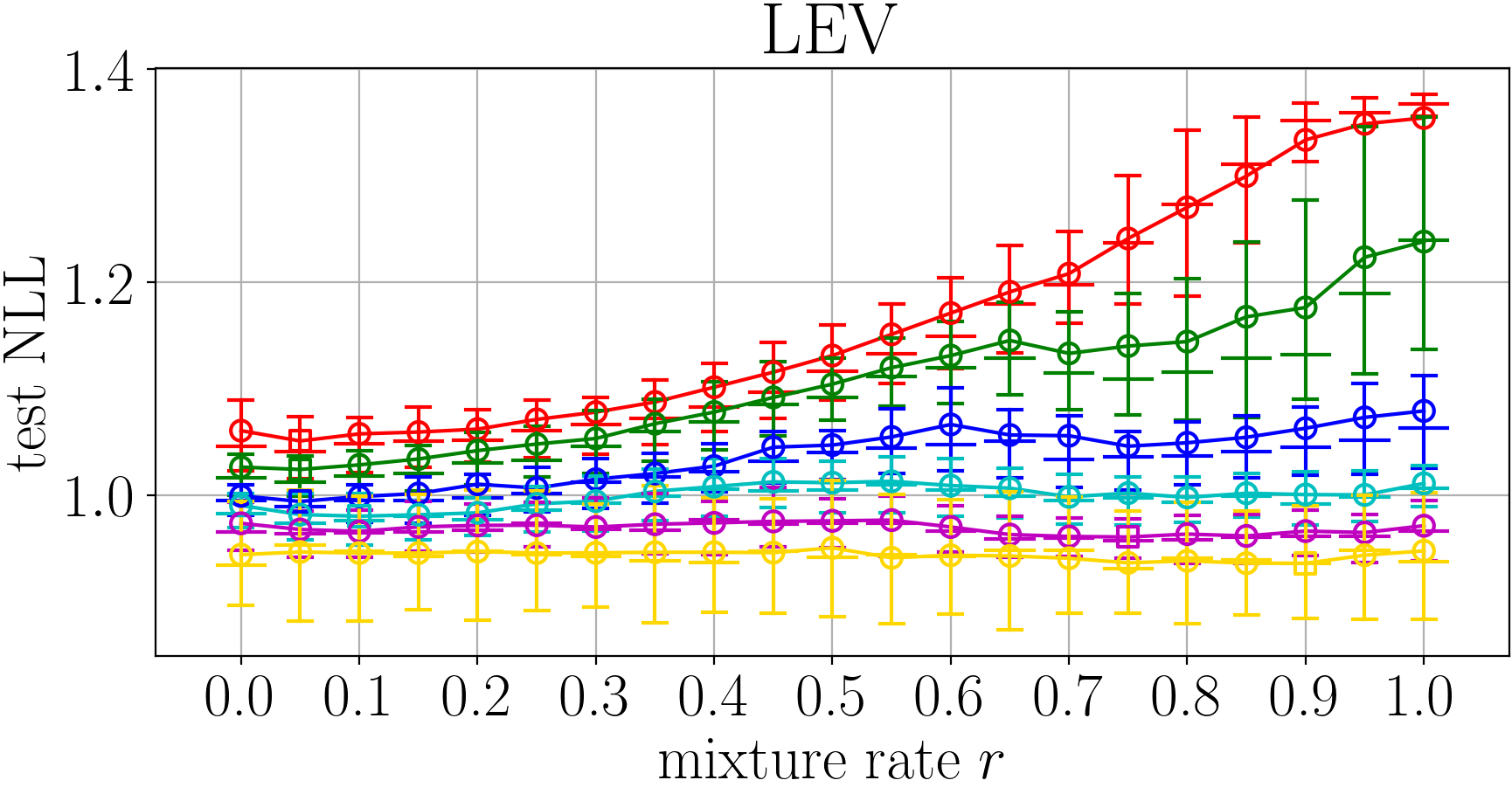}&
\includegraphics[height=1.54cm, bb=0 0 597 327]{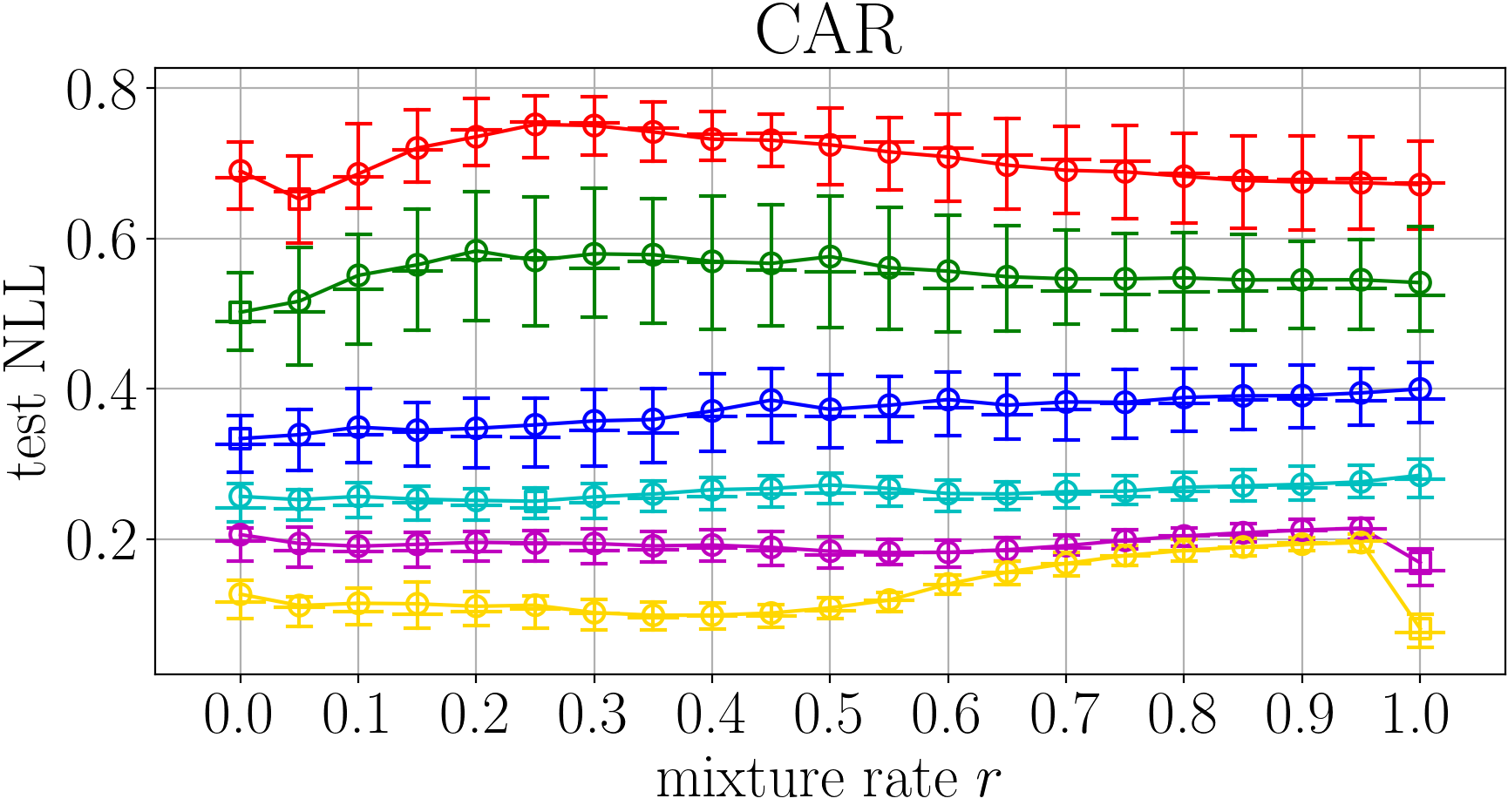}\\
\includegraphics[height=1.54cm, bb=0 0 597 327]{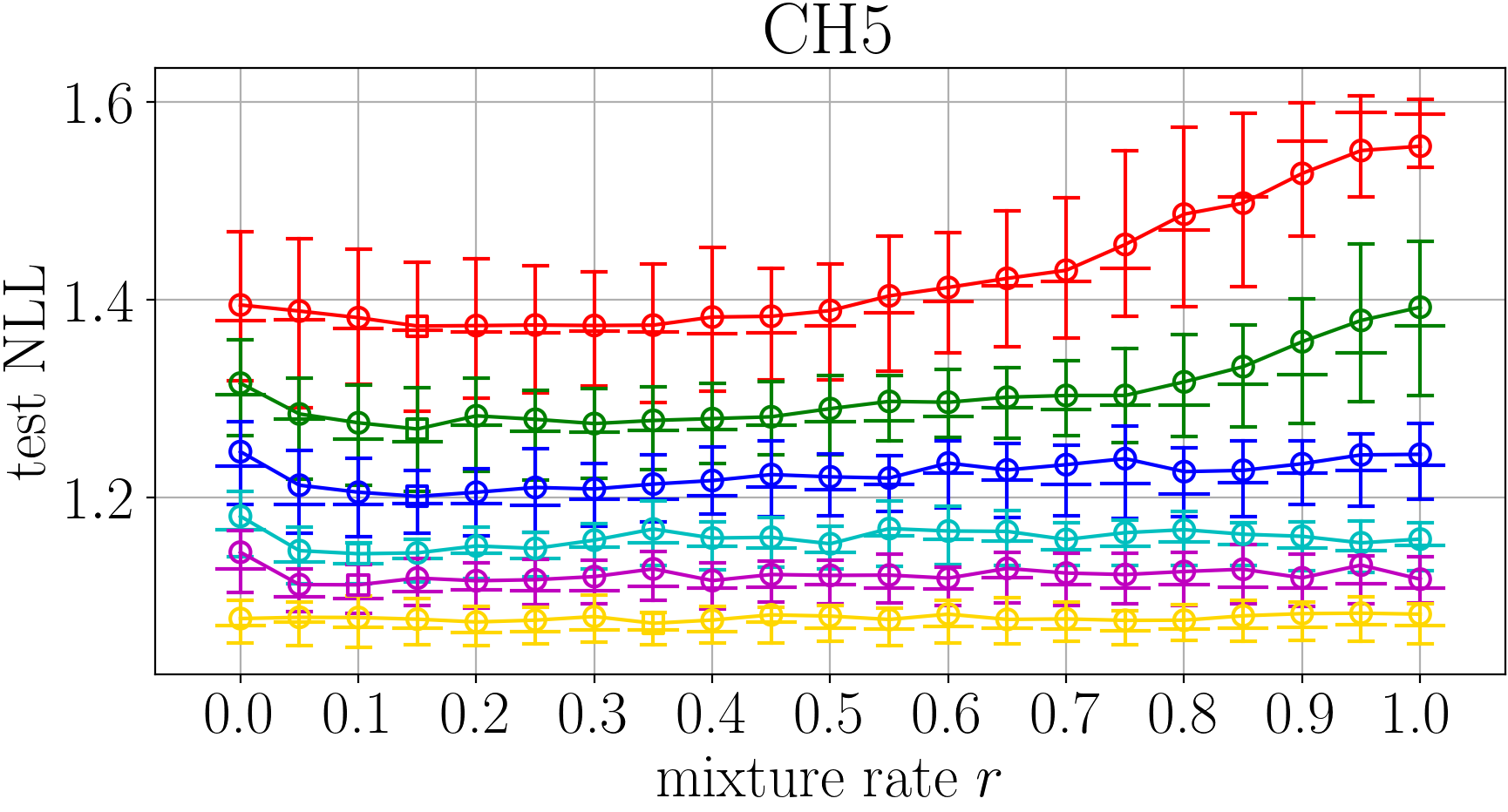}&
\includegraphics[height=1.54cm, bb=0 0 597 327]{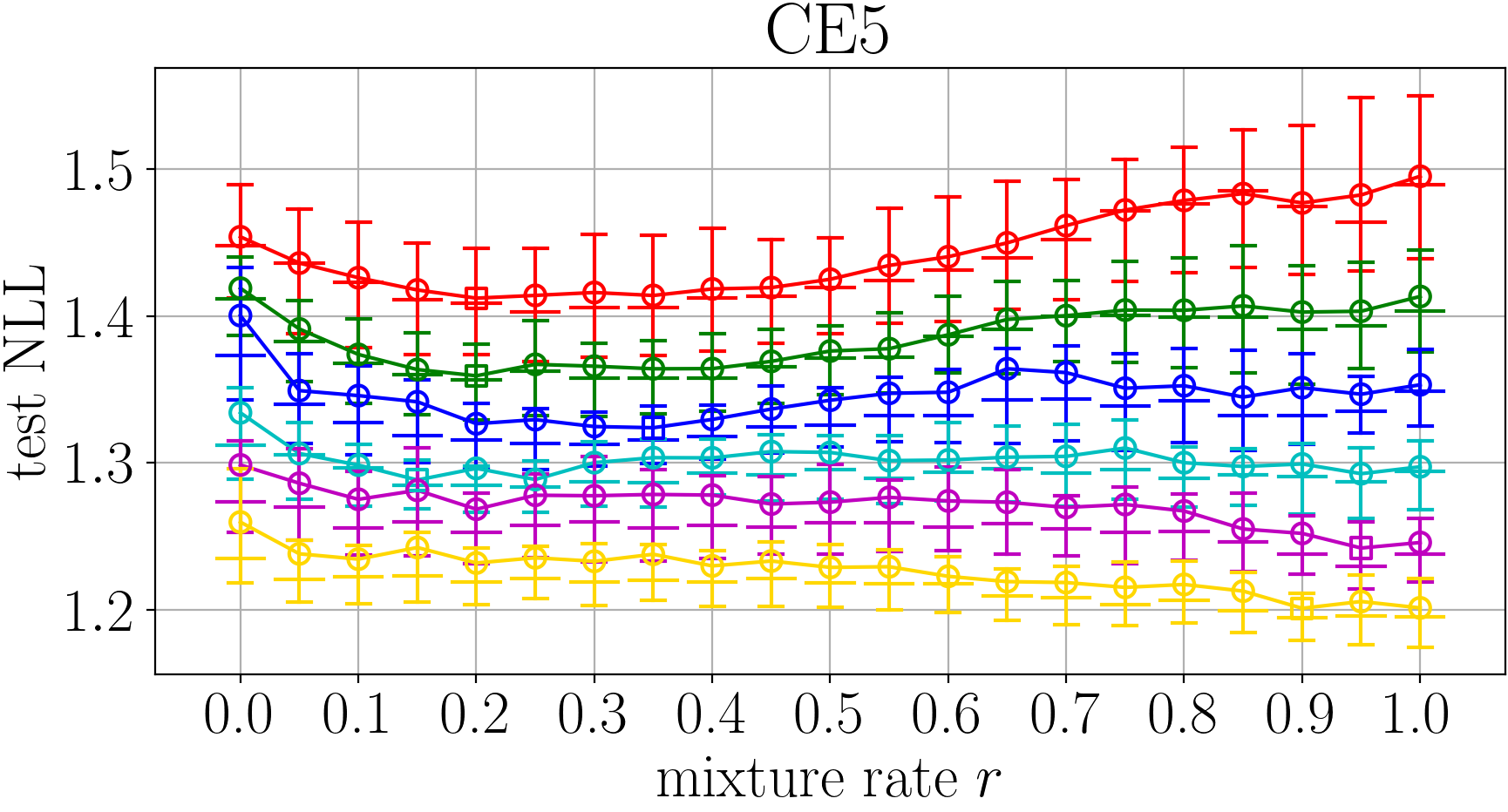}&
\includegraphics[height=1.54cm, bb=0 0 597 327]{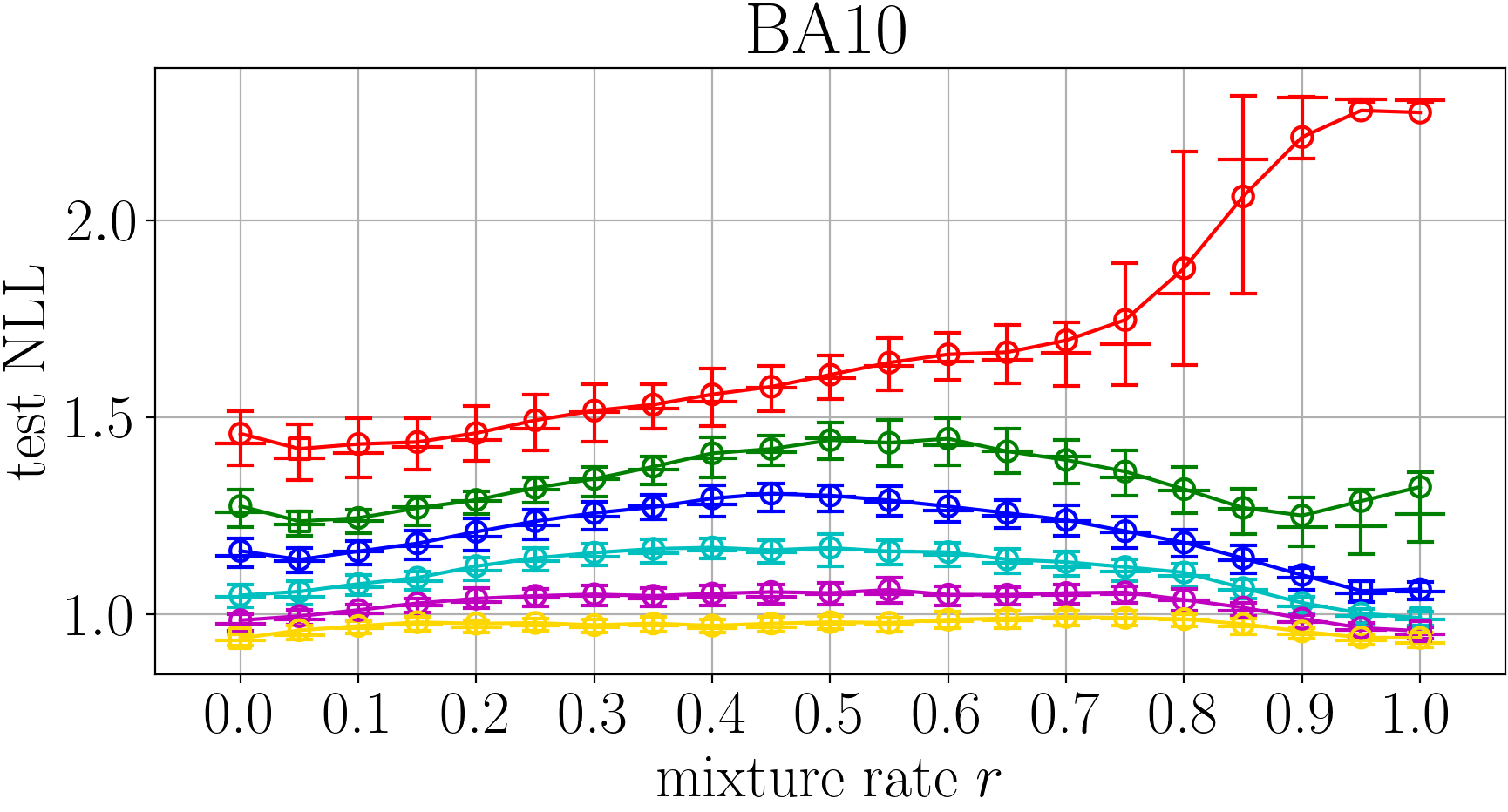}\\
\includegraphics[height=1.54cm, bb=0 0 597 327]{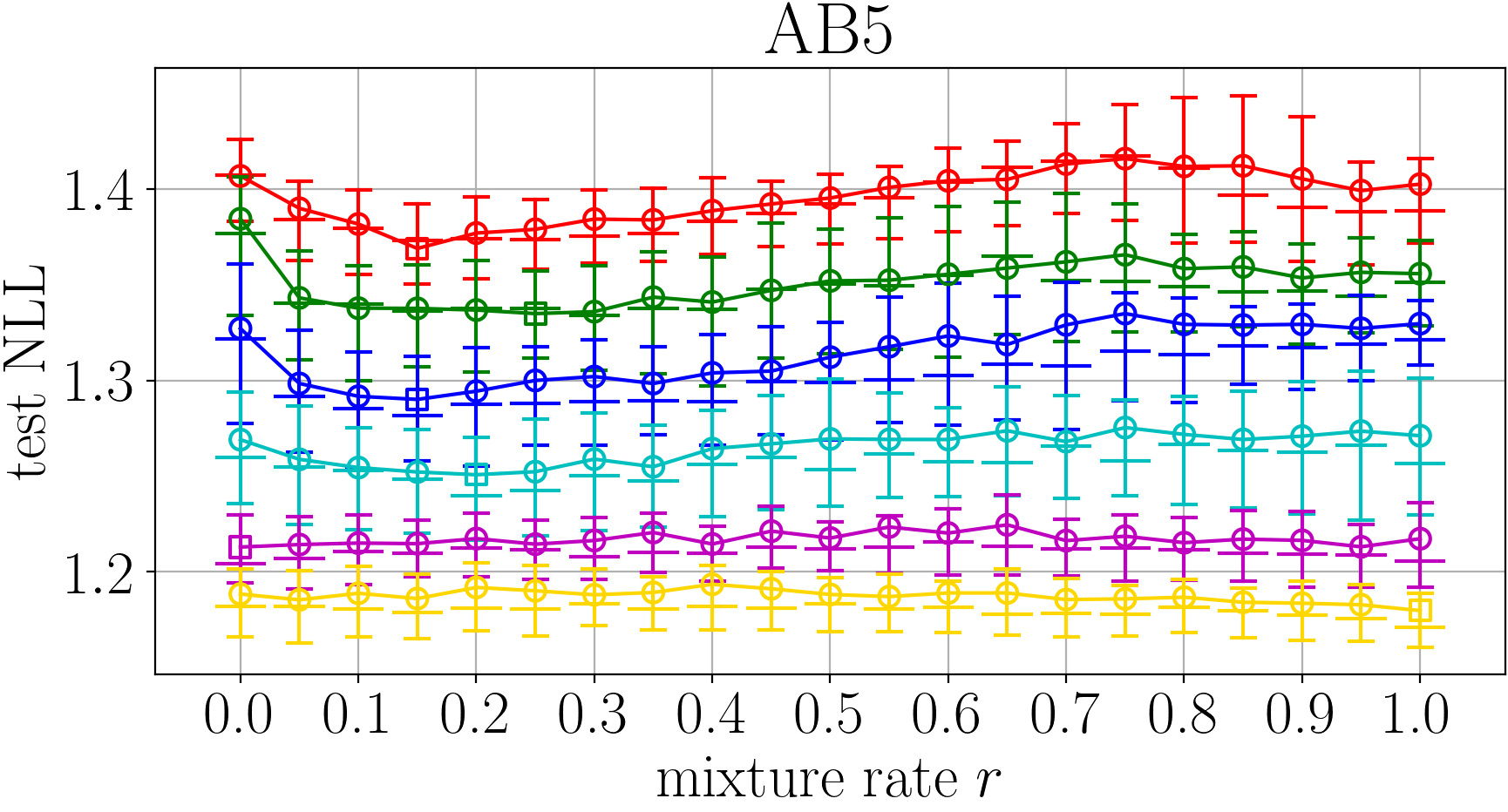}&
\includegraphics[height=1.54cm, bb=0 0 597 327]{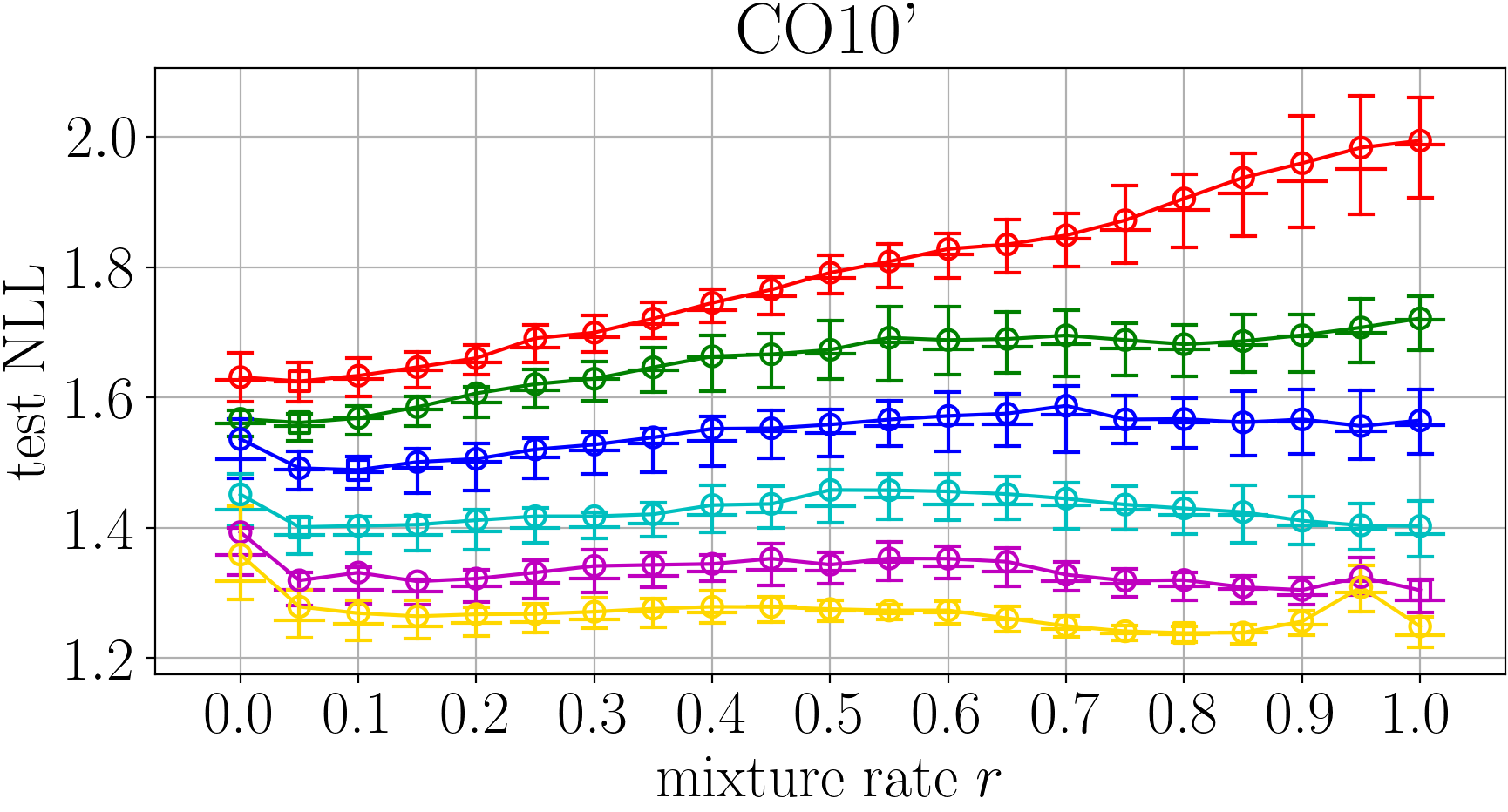}&
\includegraphics[height=1.54cm, bb=0 0 597 327]{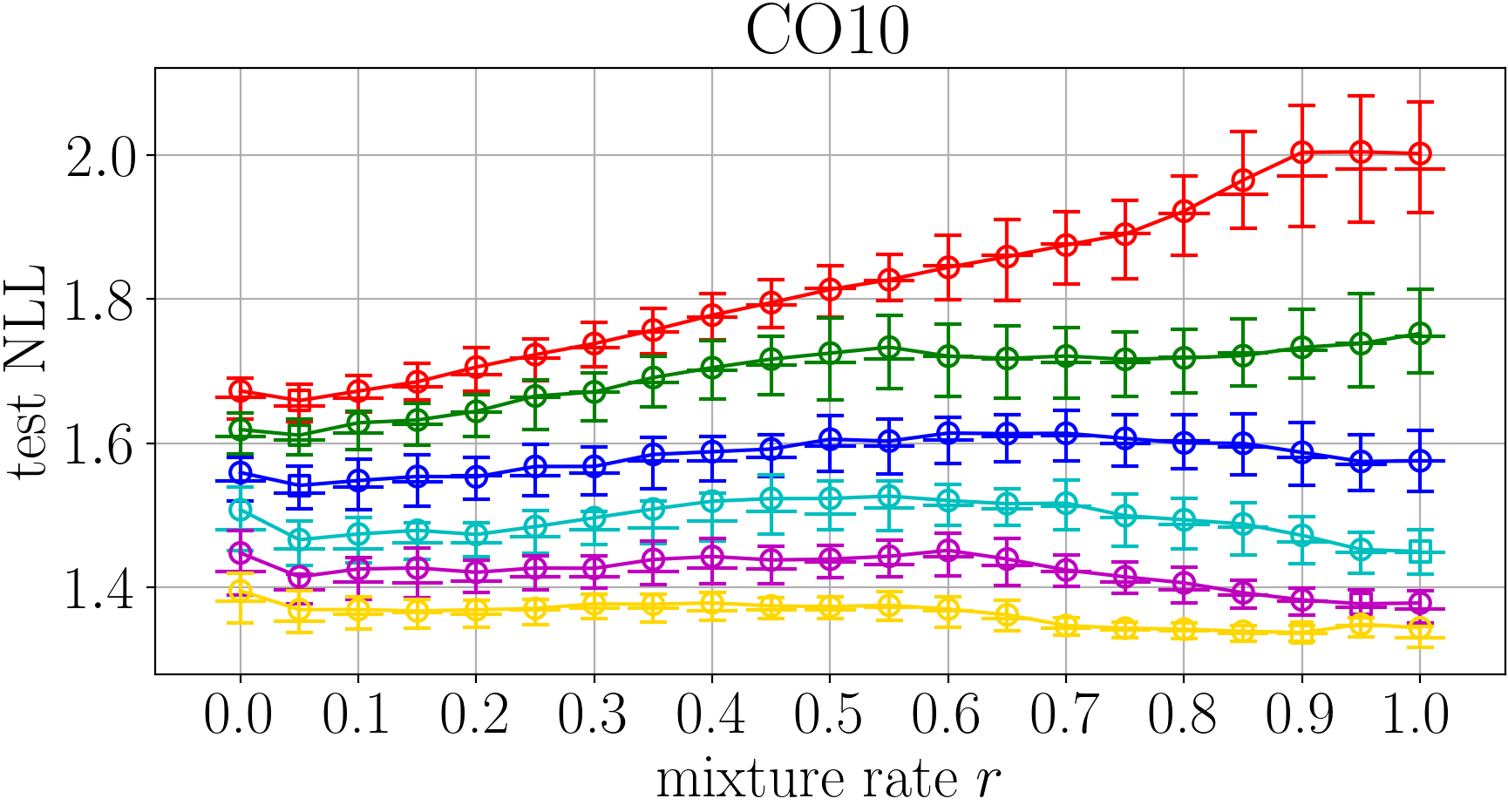}\\
\includegraphics[height=1.54cm, bb=0 0 597 327]{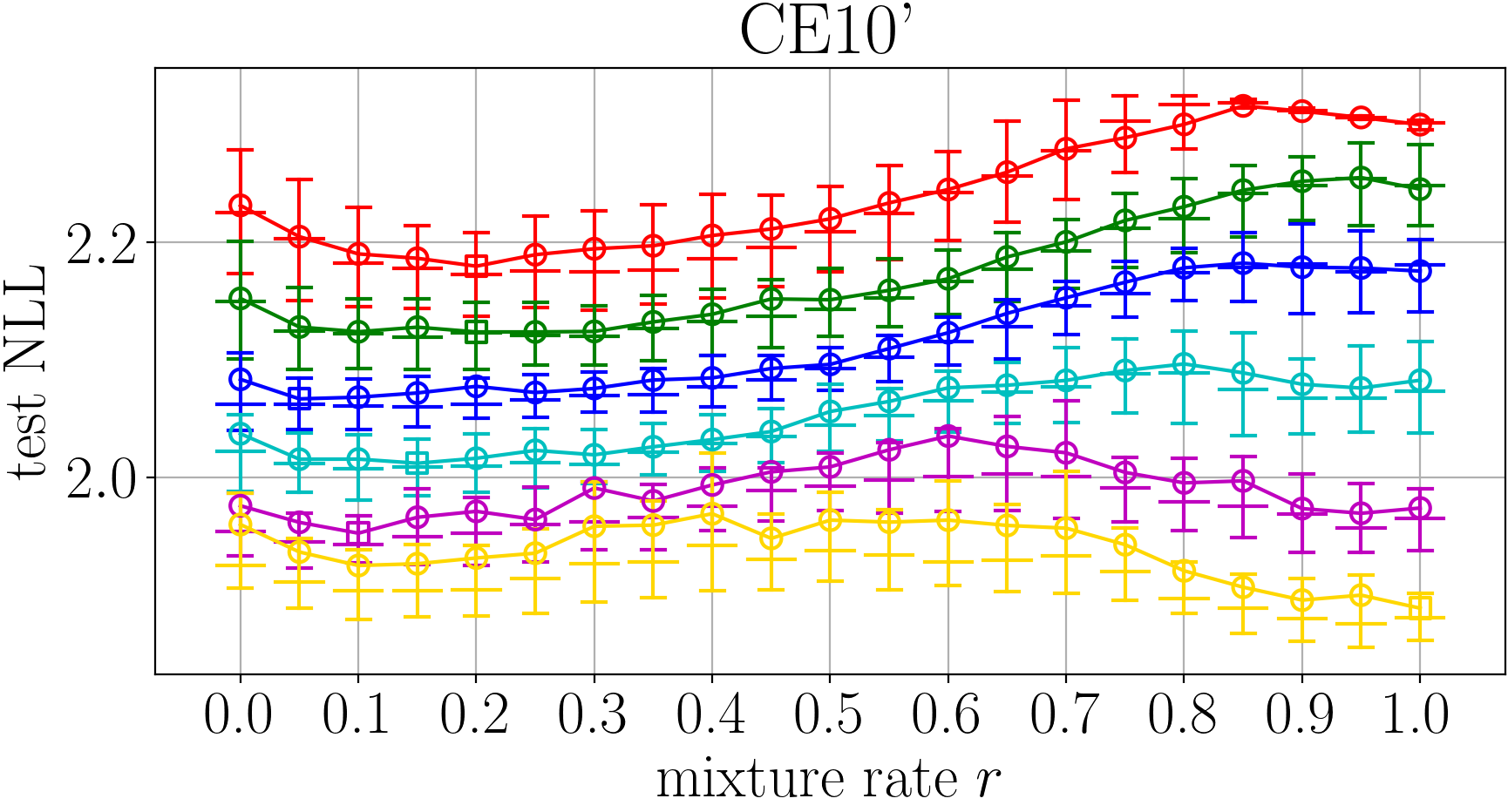}&
\includegraphics[height=1.54cm, bb=0 0 597 327]{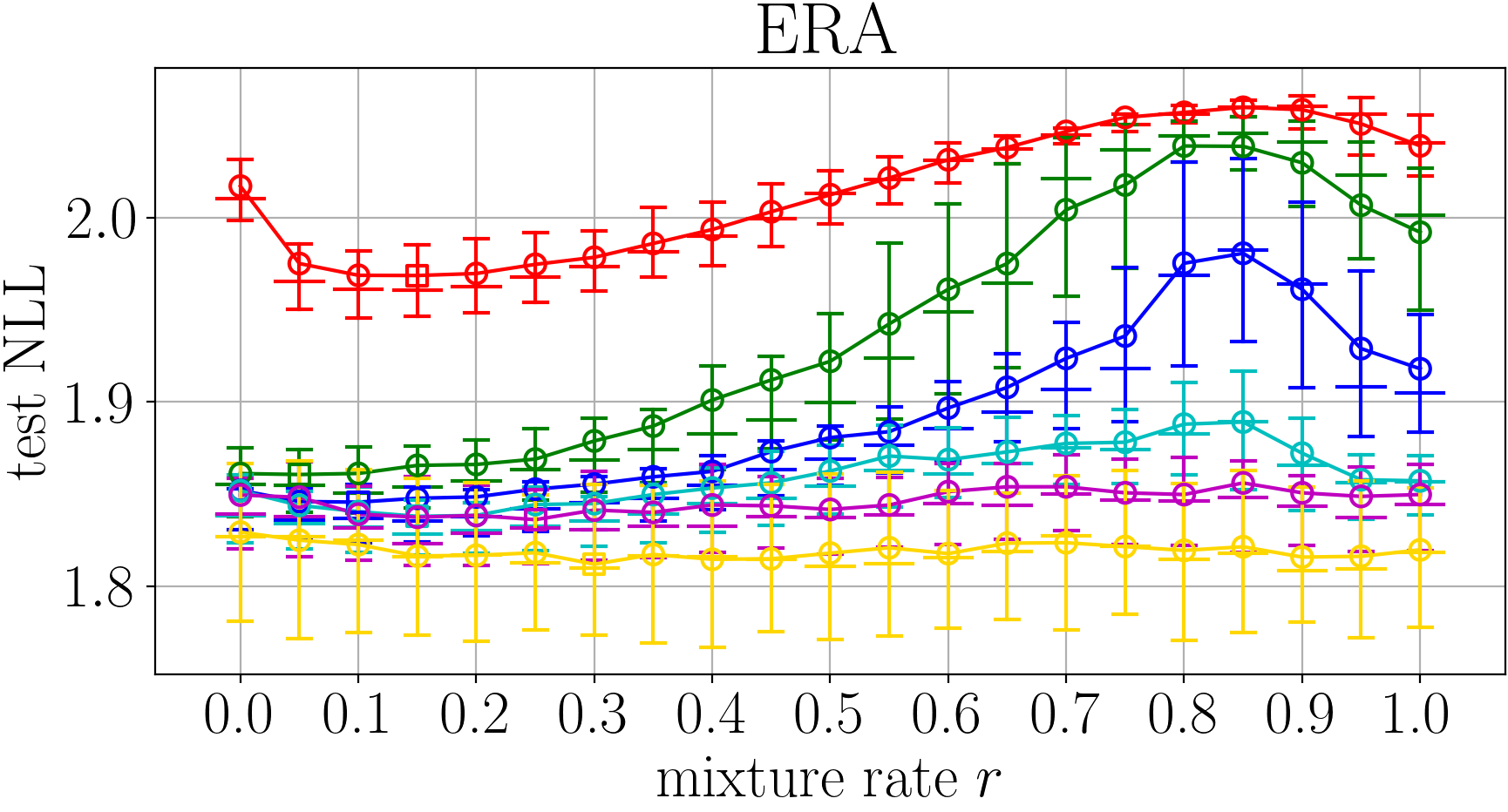}&
\includegraphics[height=1.54cm, bb=0 0 597 327]{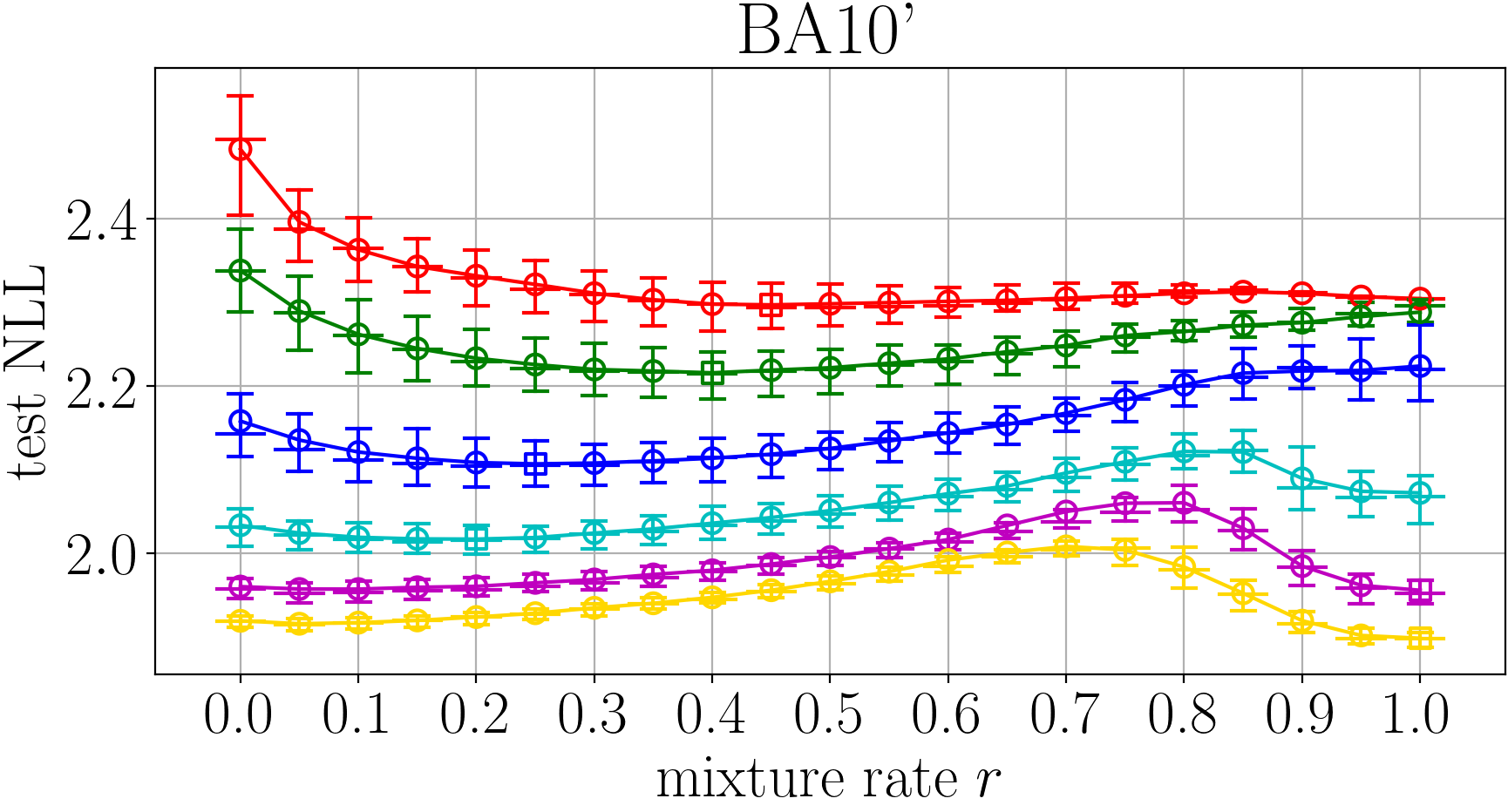}\\
\includegraphics[height=1.54cm, bb=0 0 597 327]{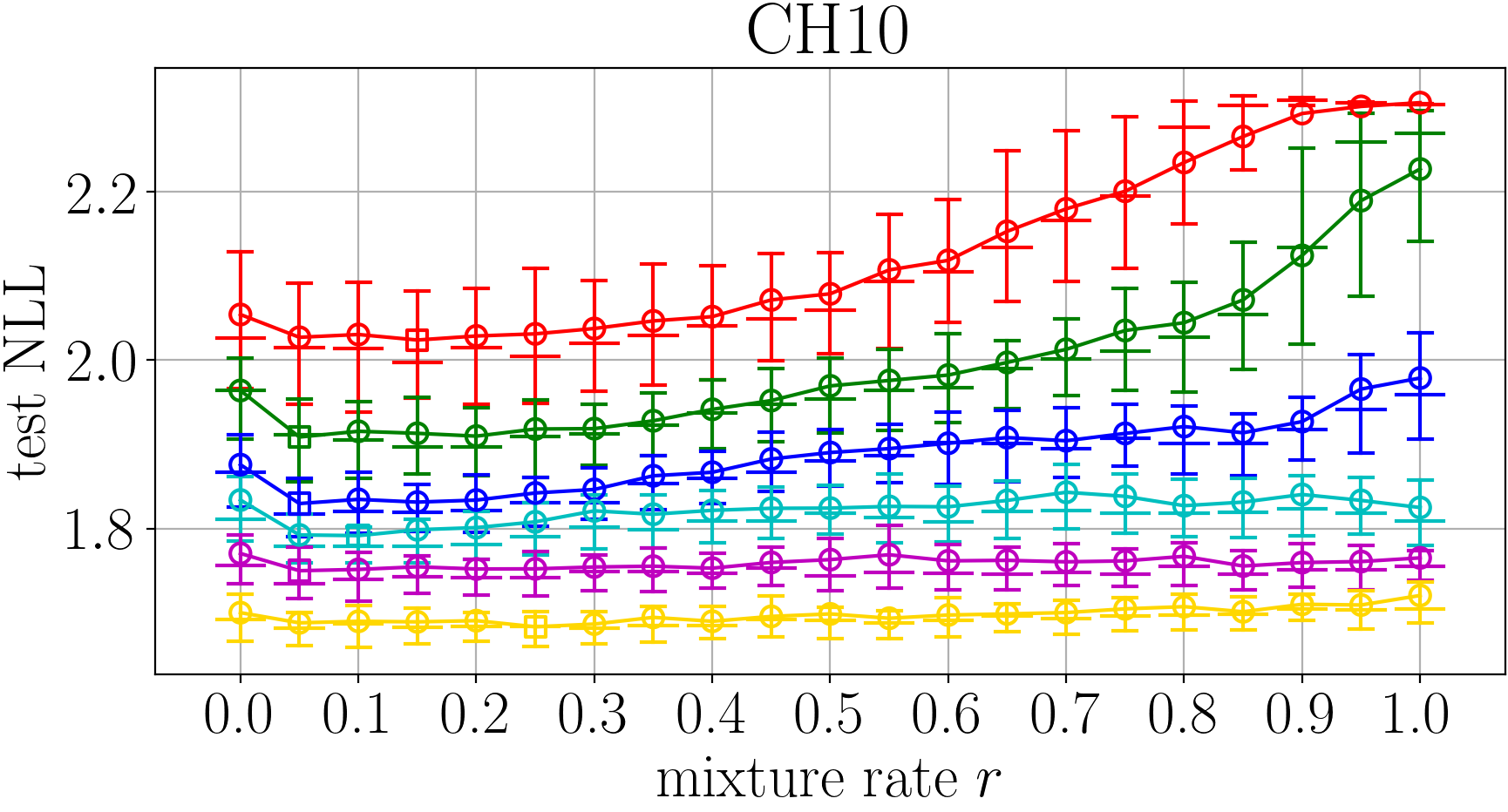}&
\includegraphics[height=1.54cm, bb=0 0 597 327]{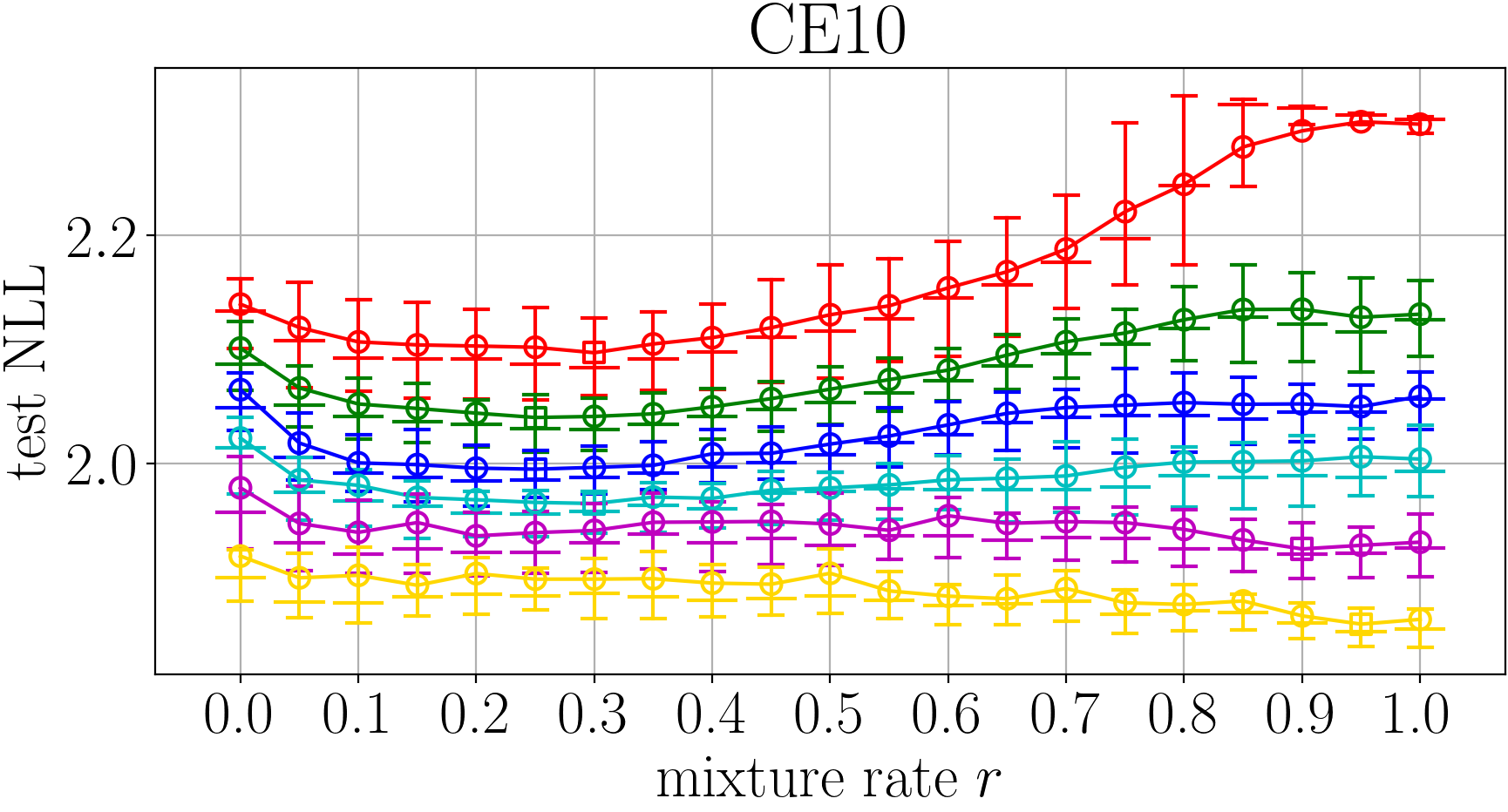}&
\includegraphics[height=1.54cm, bb=0 0 597 327]{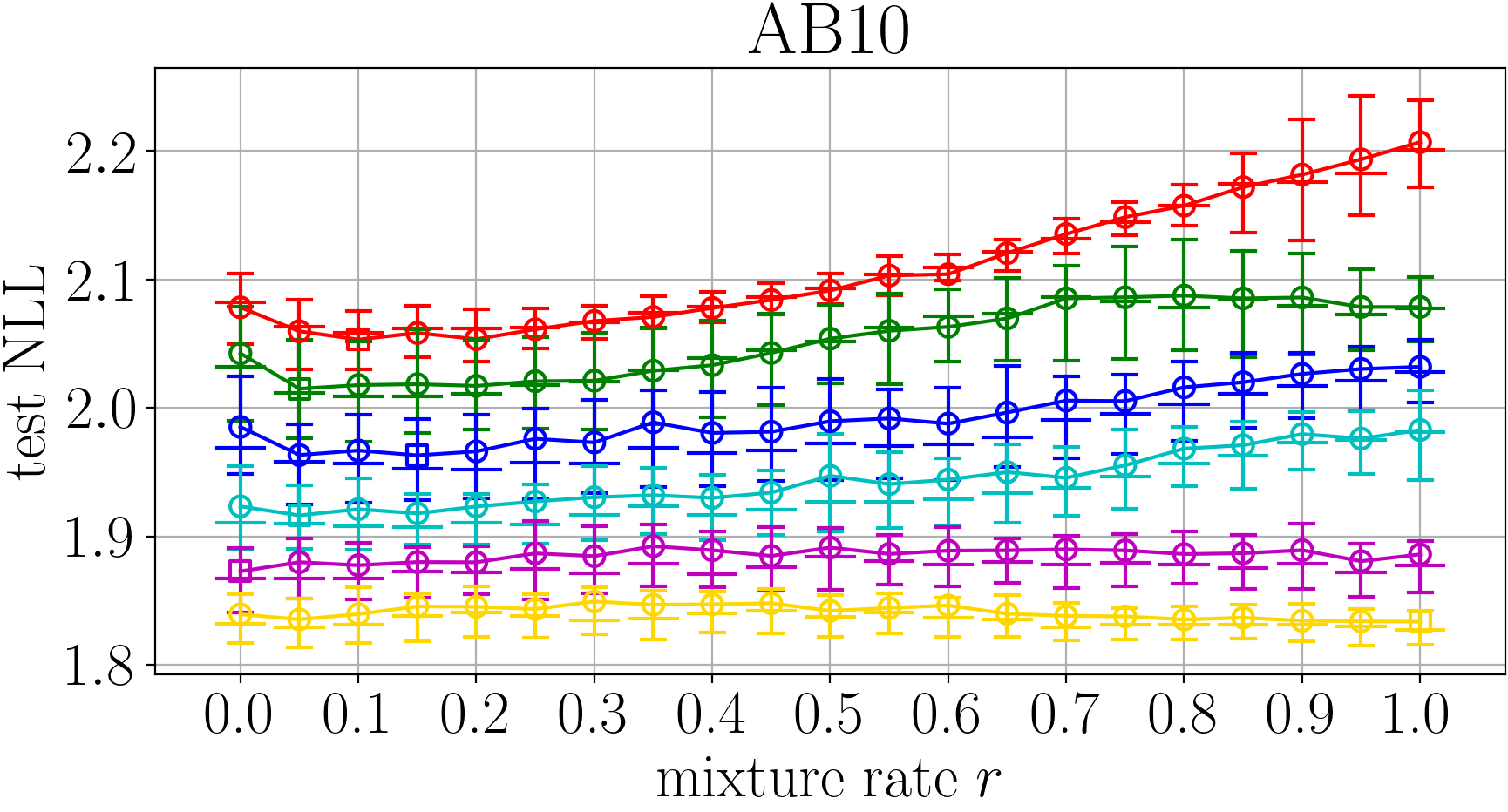}\\
\includegraphics[height=1.54cm, bb=0 0 597 327]{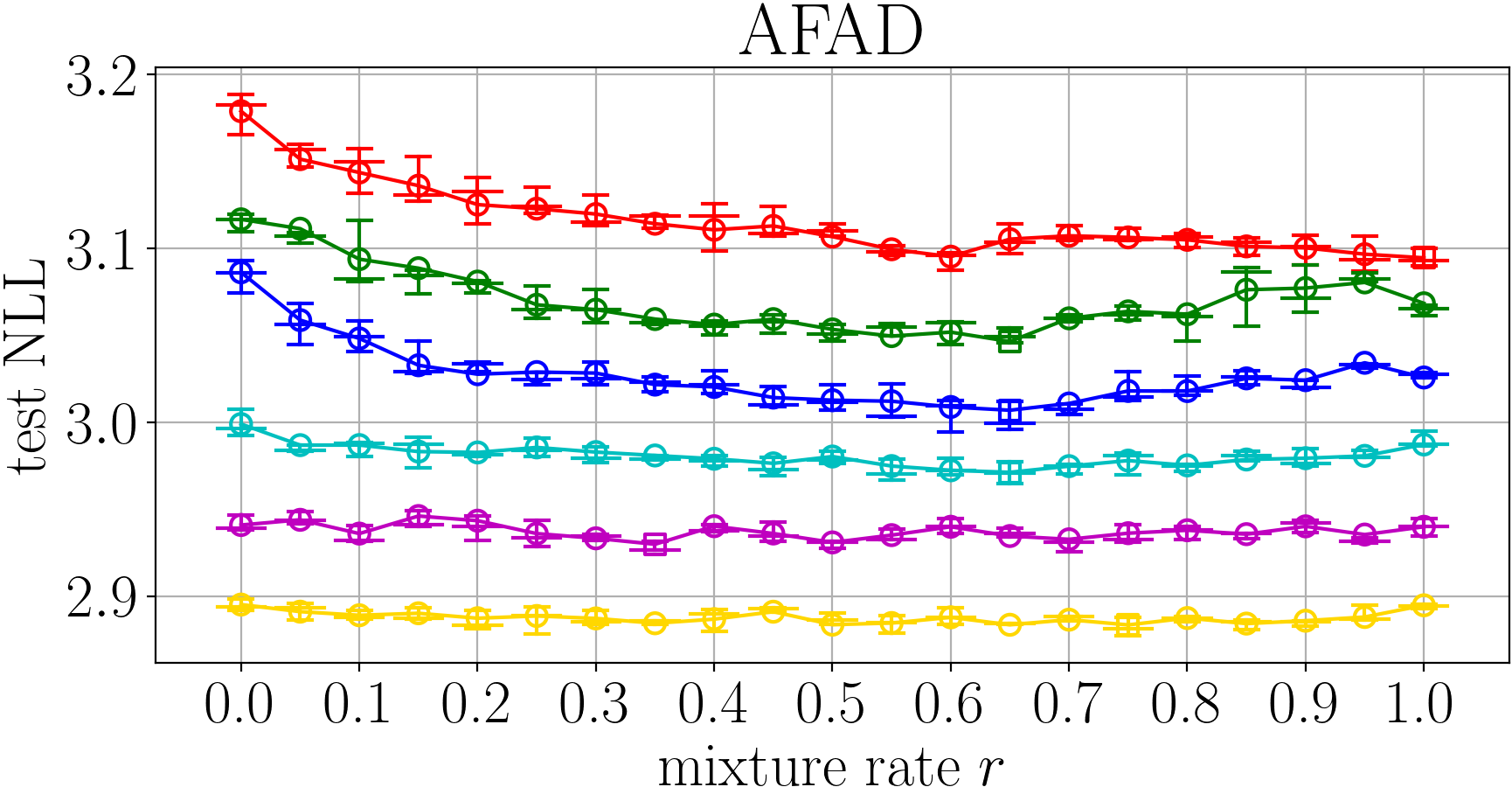}&
\includegraphics[height=1.54cm, bb=0 0 597 327]{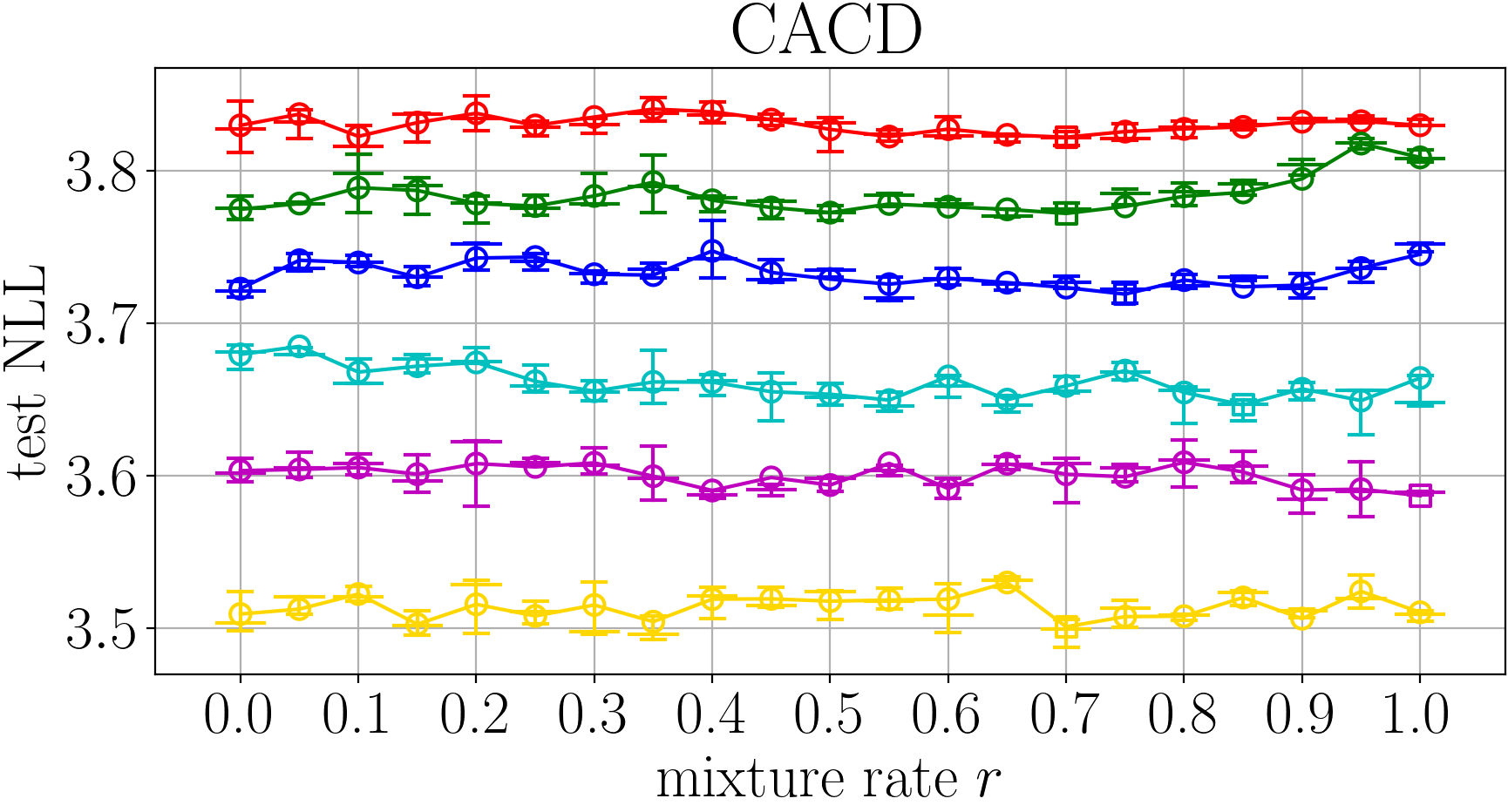}&
\includegraphics[height=1.54cm, bb=0 0 597 327]{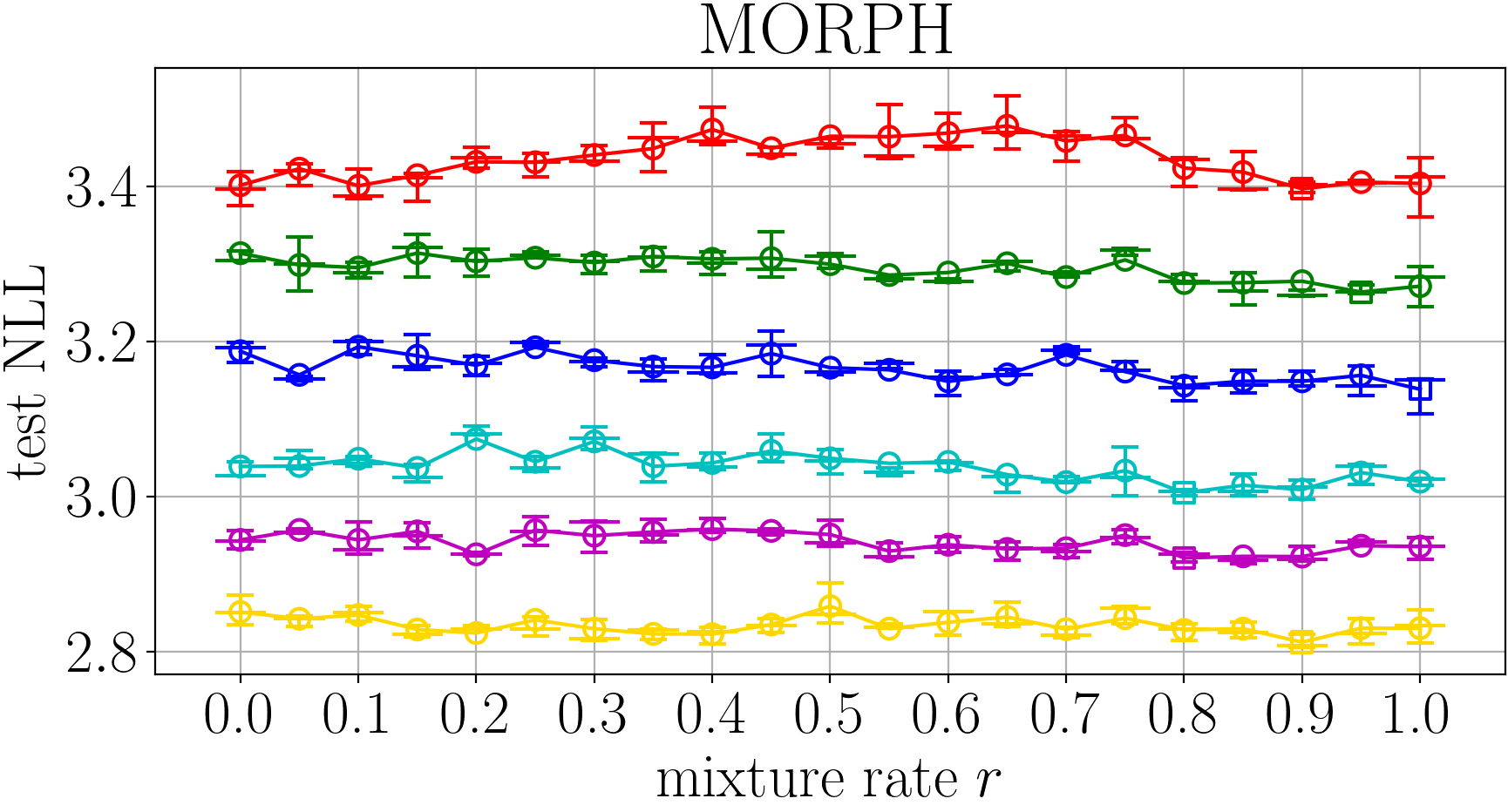}
\end{tabular}
\caption{%
Mean\&quantiles-plot\protect\footnotemark%
of test NLLs for the $\text{Mix}_{\mbox{\tiny\text{VSL,SL}}}$ 
model with the mixture rate $r=0.05,0.1,\ldots,0.95$,
and VSL ($r=0$) and SL ($r=1$) models for training data size $n_\tra$'s 
(red, green, blue, cyan, magenta, yellow from the smallest to the largest).
The square represents the best $r$.}
\label{fig:Res-Hypara-NLL}
\end{figure}

Figure~\ref{fig:Res-Hypara-NLL} shows 
a relationship of the training data size $n_\tra$,
mixture rate $r$, and NLL of 
the $\text{Mix}_{\mbox{\tiny\text{VSL,SL}}}$ 
model with $r=0.05, 0.1, \ldots, 0.95$,
VSL model ($r=0$), and SL model ($r=1$)
(those regarding the MZE, MAE, and MSE 
are similar and written in supplement).
As expected from the bias-variance trade-off
and the unimodality of many ordinal data, 
an intermediate $r$ was the best and the MAUL 
model improved the NLL for small $n_\tra$, 
while a larger $r$ was the best although 
the performance difference was small for large $n_\tra$.
Also, it is a useful finding that in many cases 
the prediction performance curve seemed to be continuous 
and had a single valley with respect to $r$;
we can trust this continuity to search for 
better $r$ efficiently through, for example, bisection method.

\begin{figure}[!t]
\centering%
\renewcommand{\arraystretch}{0.1}%
\renewcommand{\tabcolsep}{0pt}%
\begin{tabular}{C{2.95cm}C{2.95cm}C{2.95cm}}%
\includegraphics[height=1.54cm, bb=0 0 597 327]{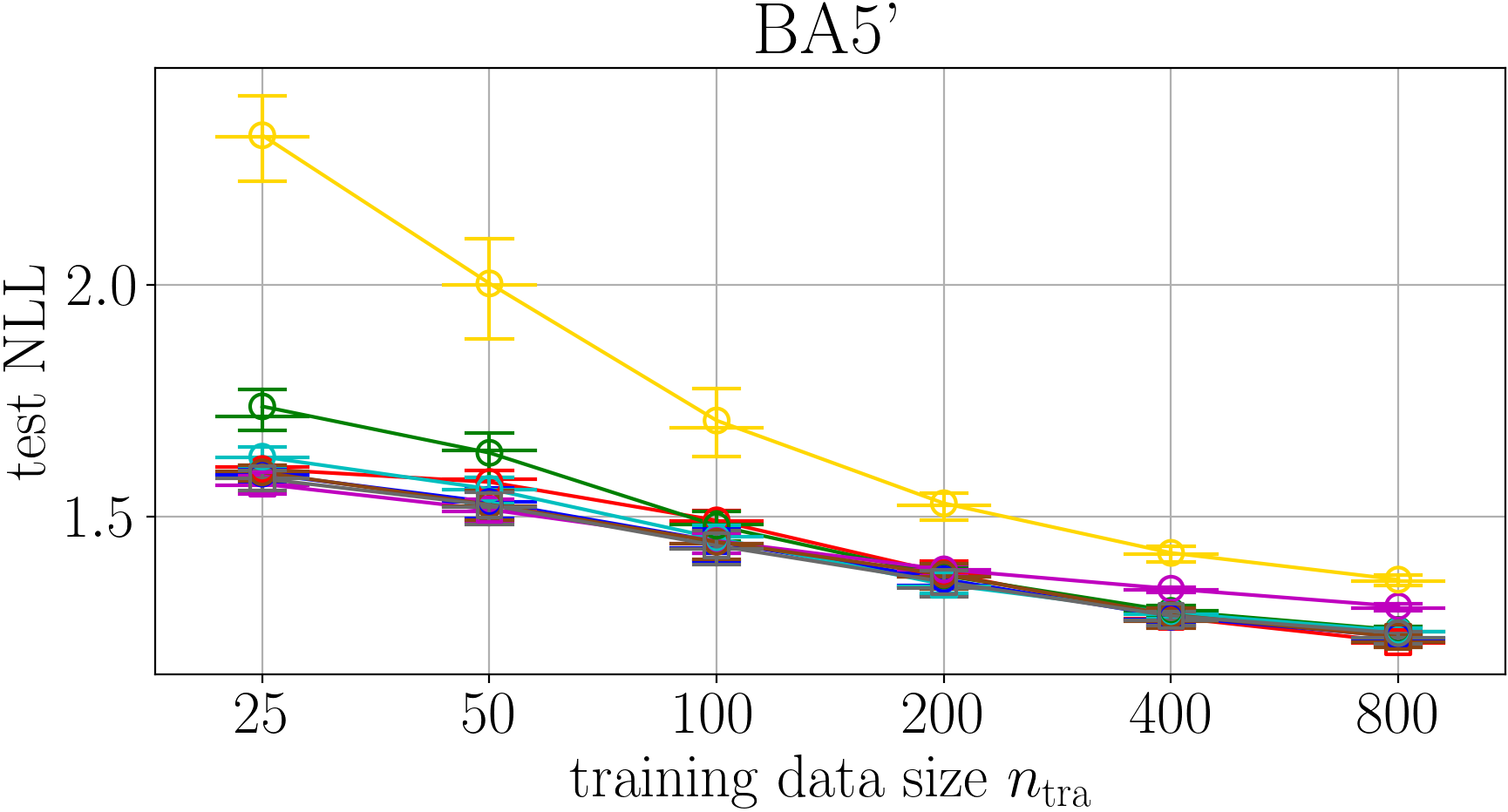}&
\includegraphics[height=1.54cm, bb=0 0 597 327]{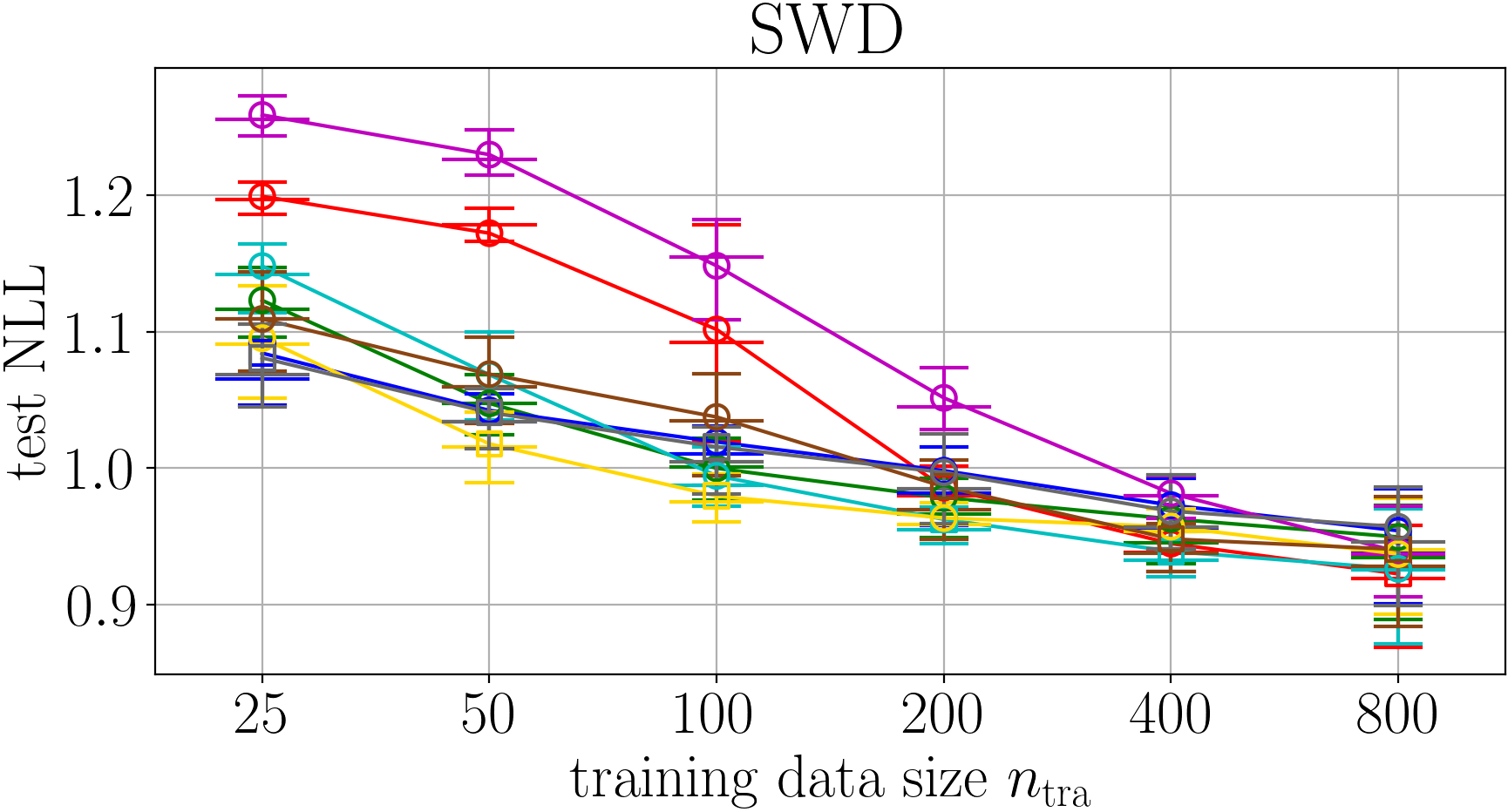}&
\includegraphics[height=1.54cm, bb=0 0 597 327]{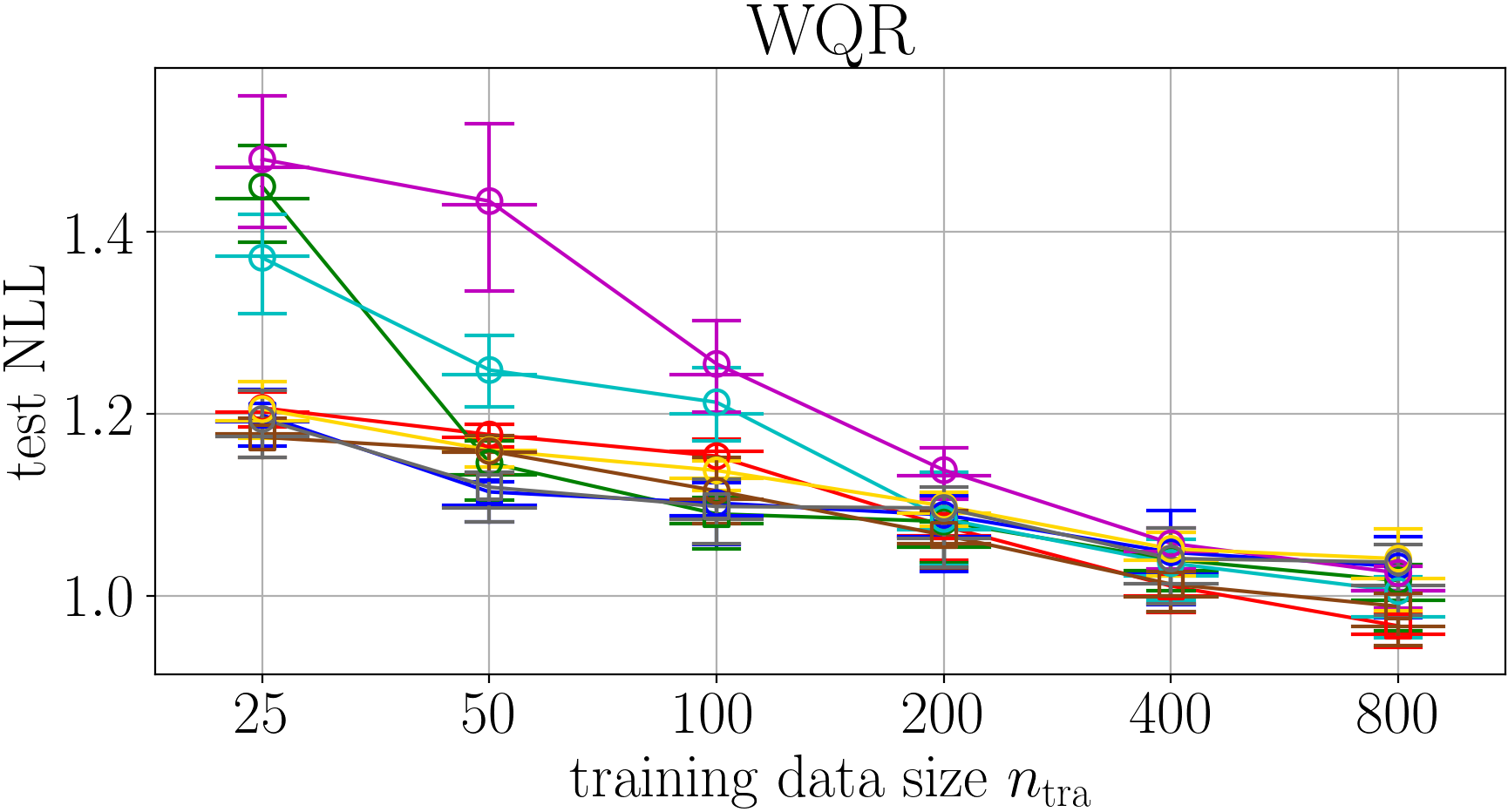}\\
\includegraphics[height=1.54cm, bb=0 0 597 327]{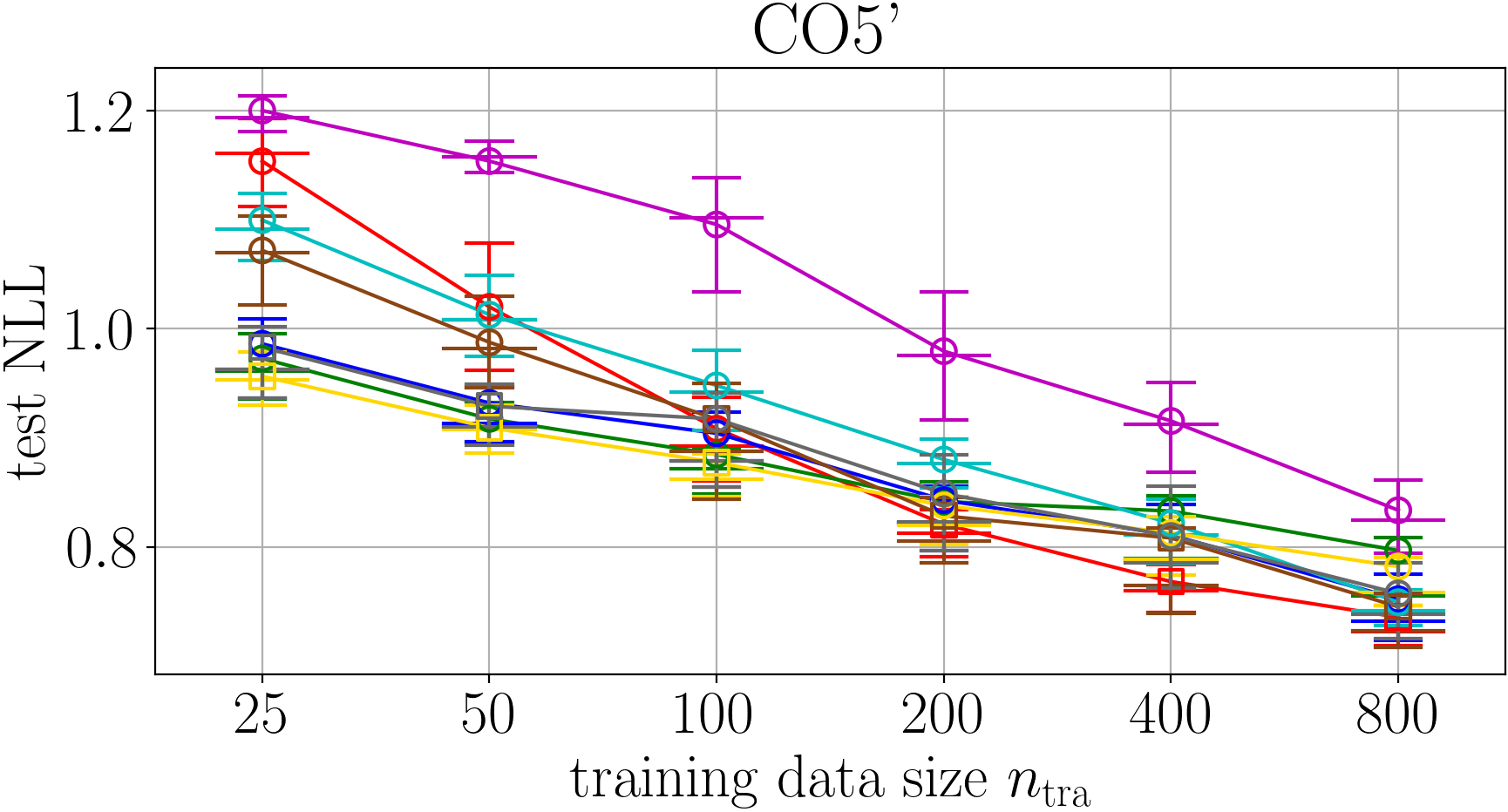}&
\includegraphics[height=1.54cm, bb=0 0 597 327]{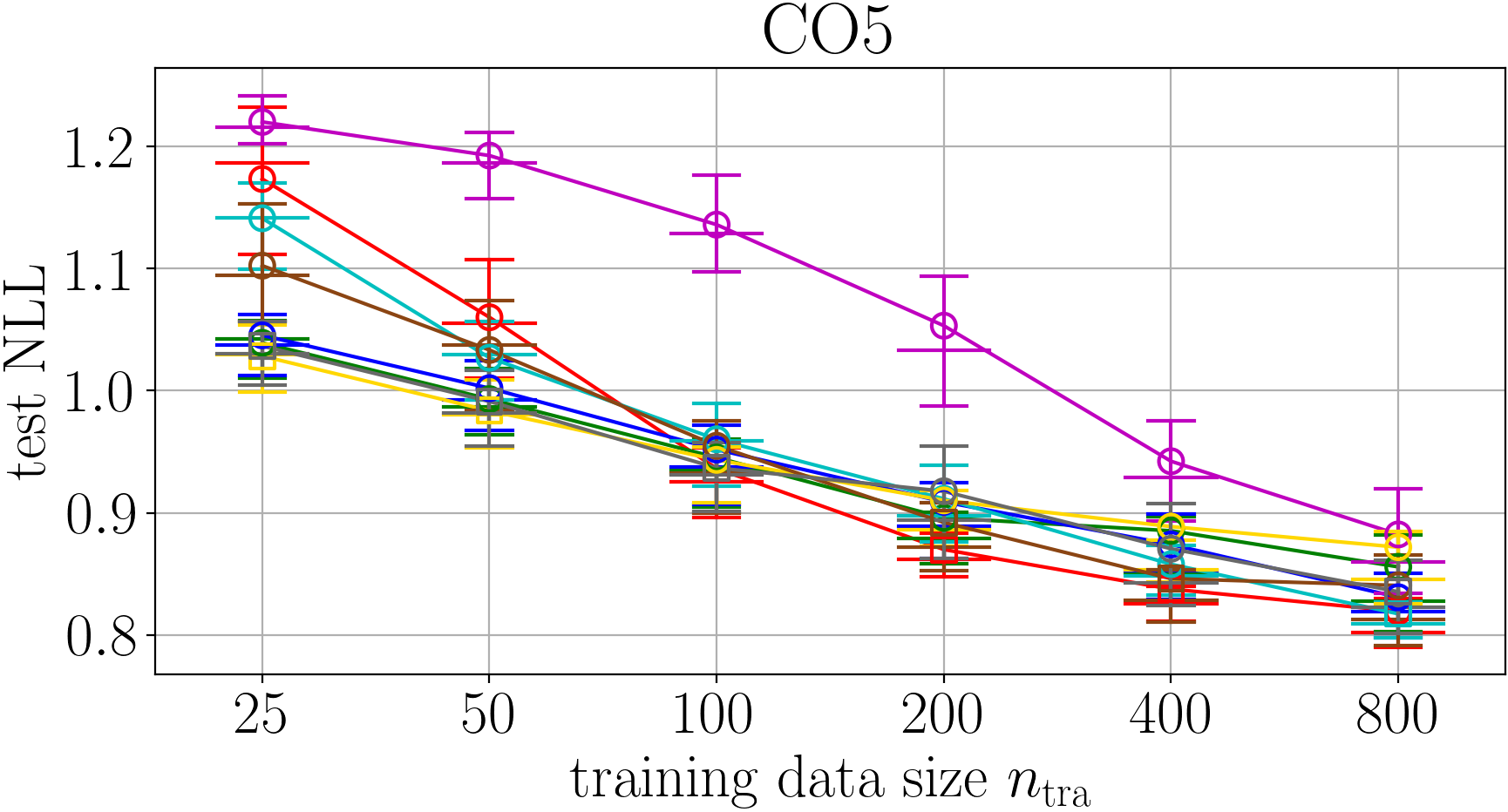}&
\includegraphics[height=1.54cm, bb=0 0 597 327]{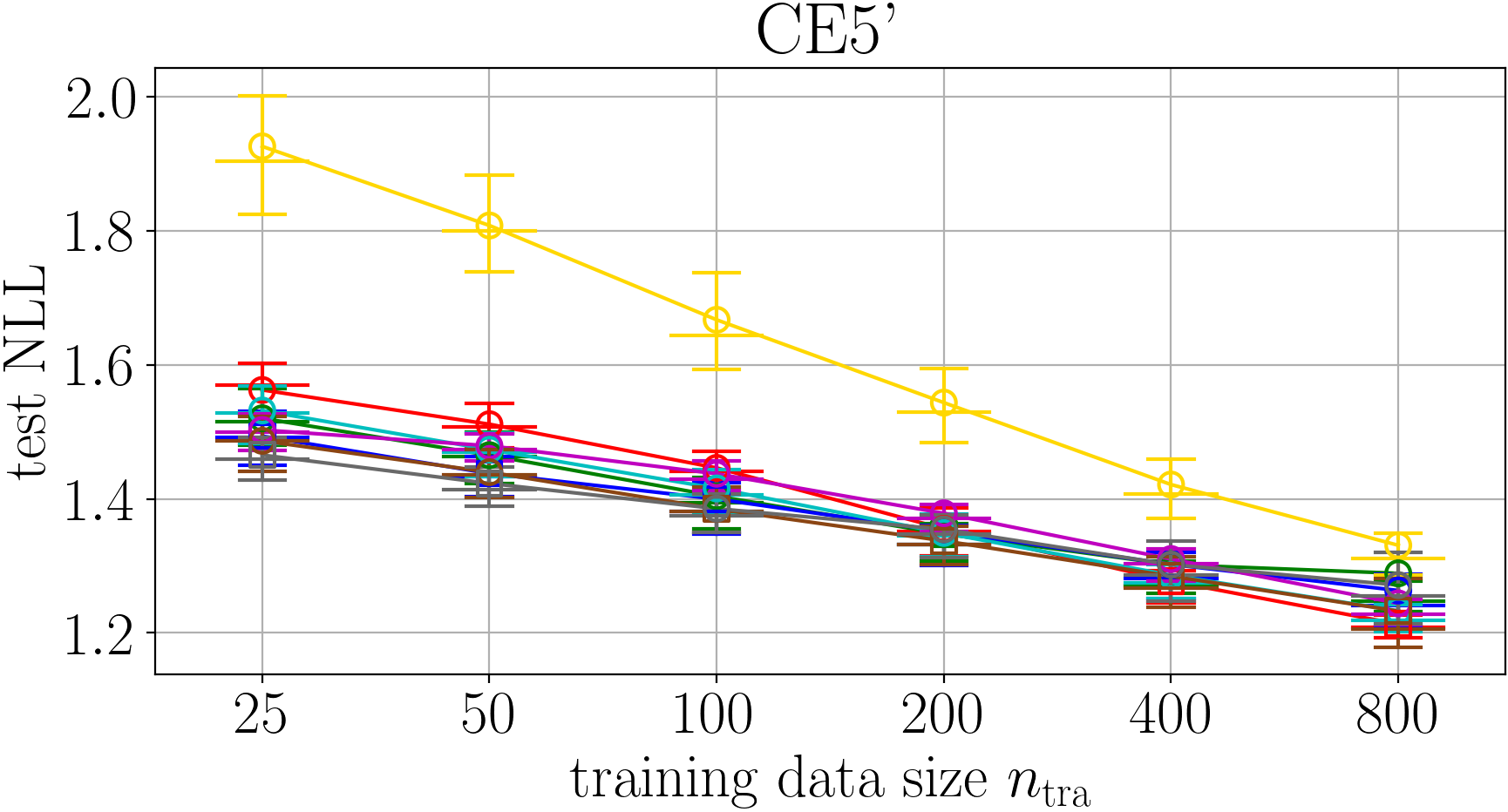}\\
\includegraphics[height=1.54cm, bb=0 0 597 327]{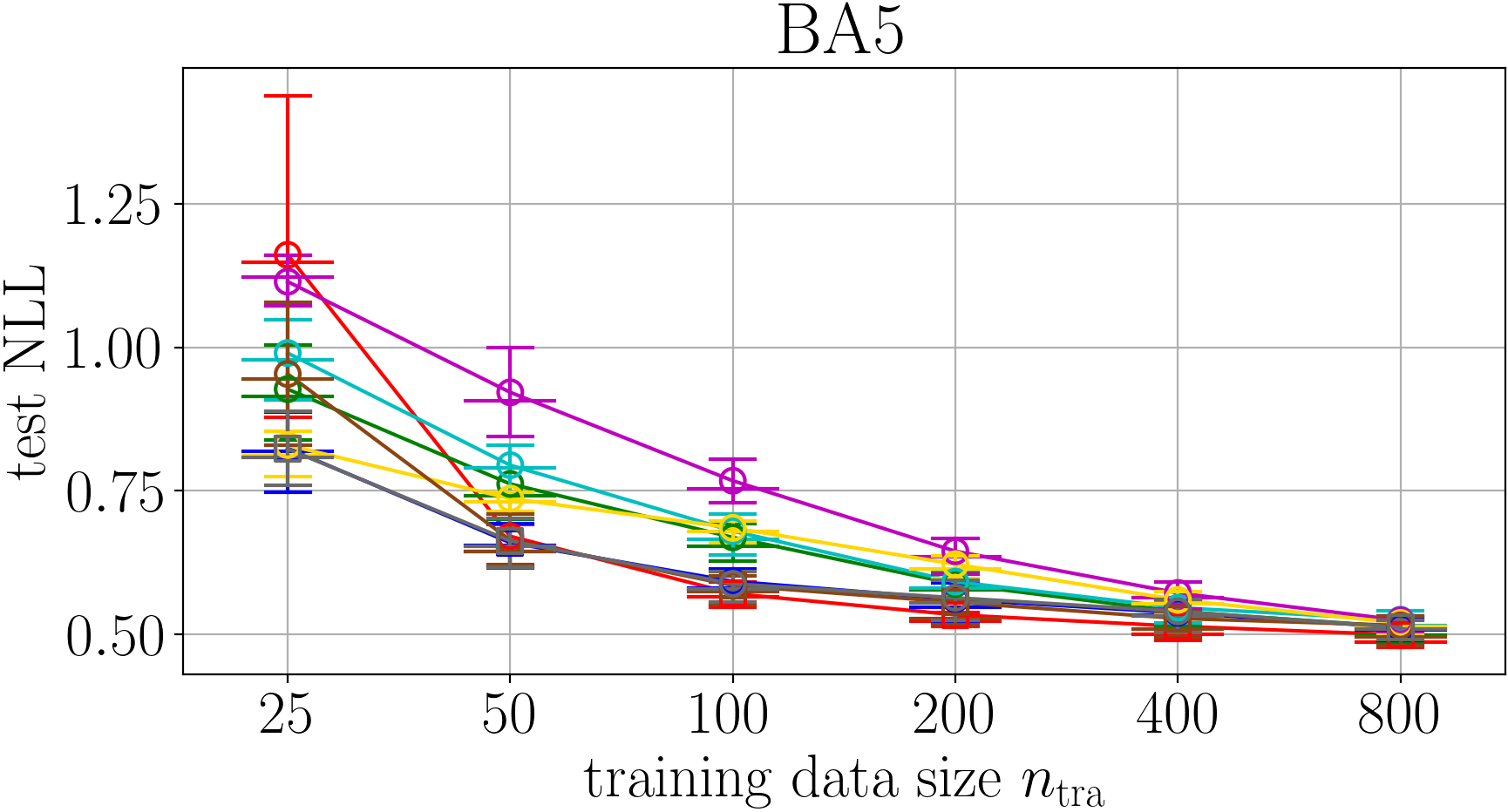}&
\includegraphics[height=1.54cm, bb=0 0 597 327]{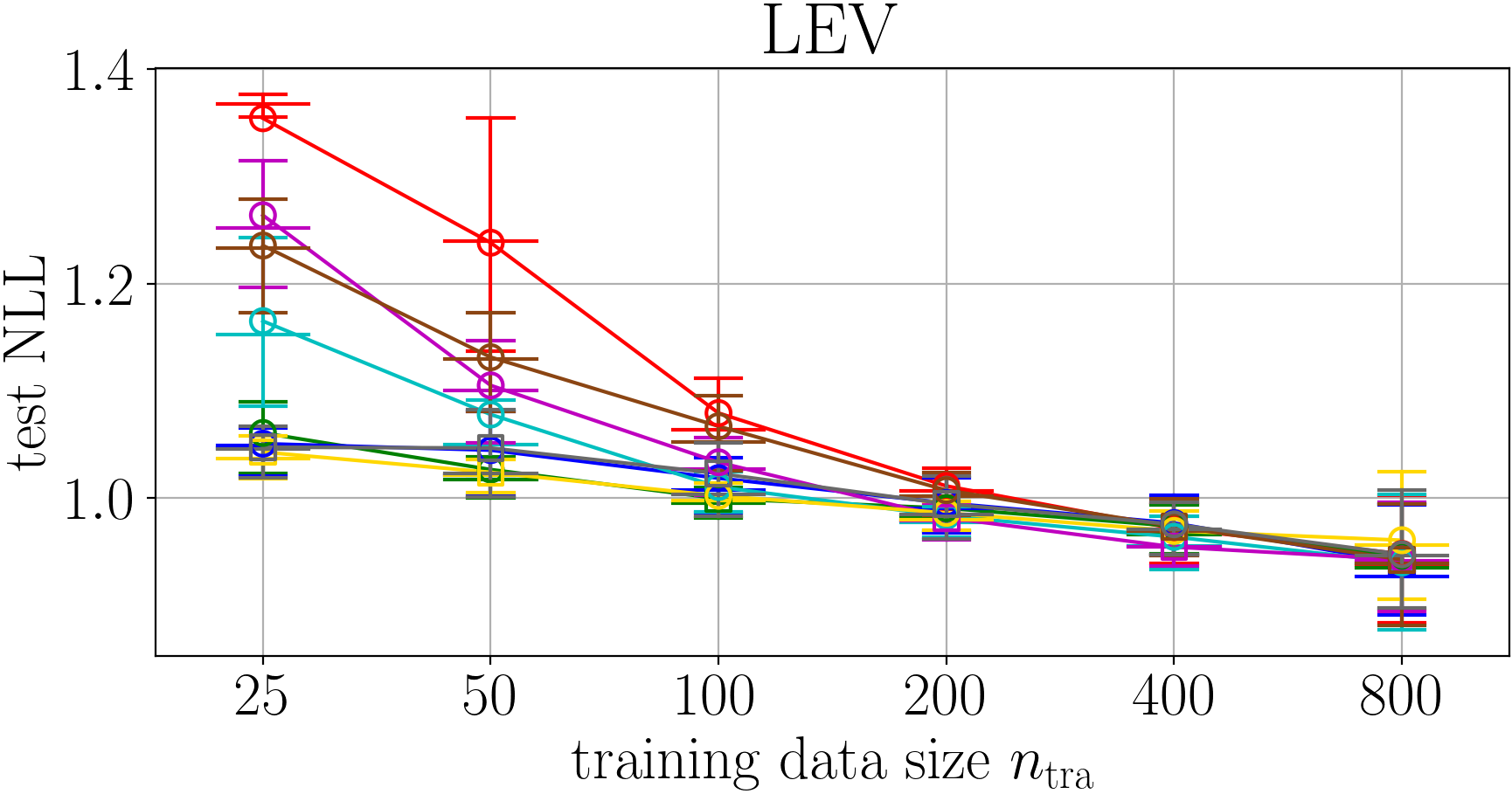}&
\includegraphics[height=1.54cm, bb=0 0 597 327]{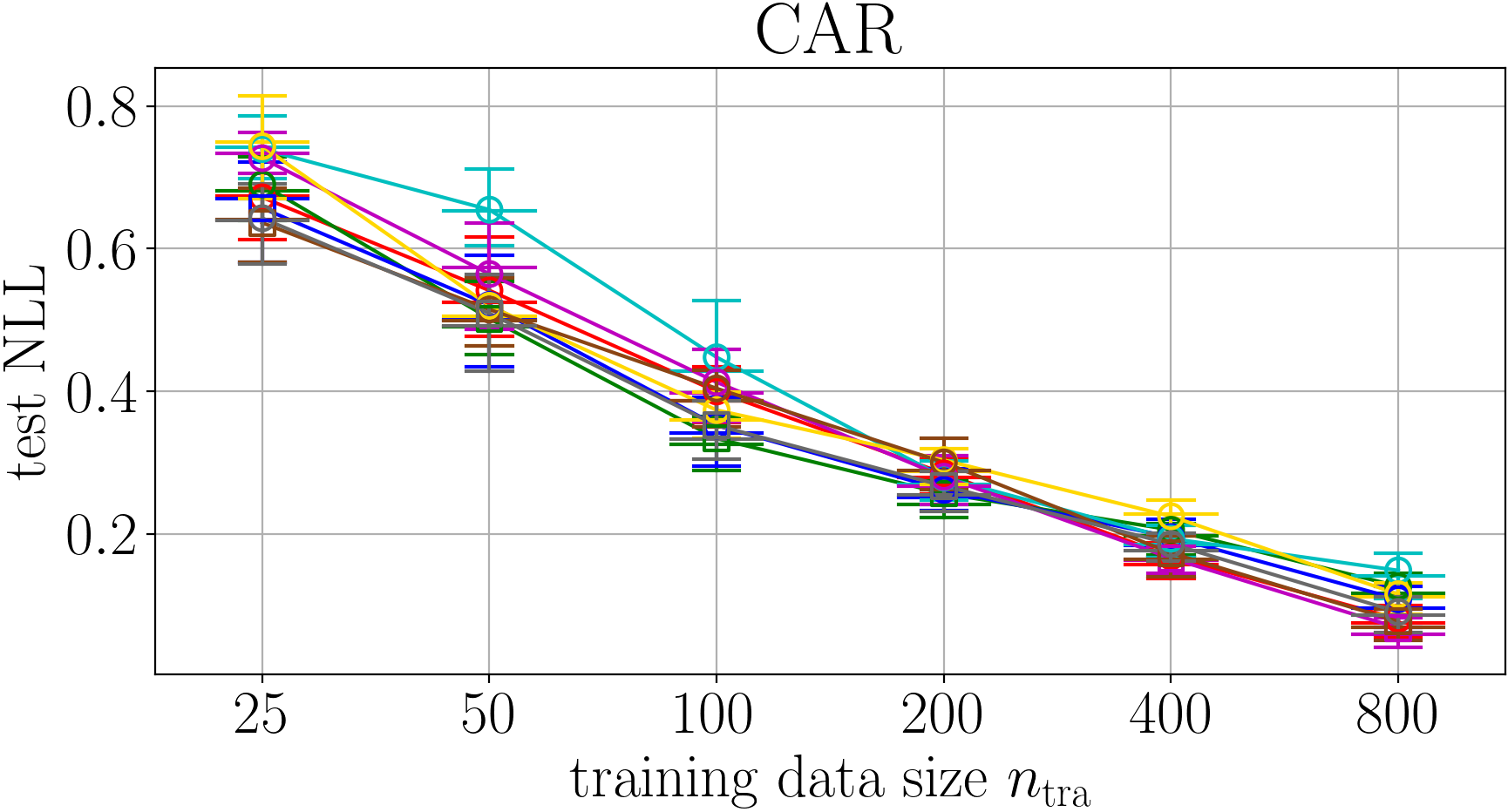}\\
\includegraphics[height=1.54cm, bb=0 0 597 327]{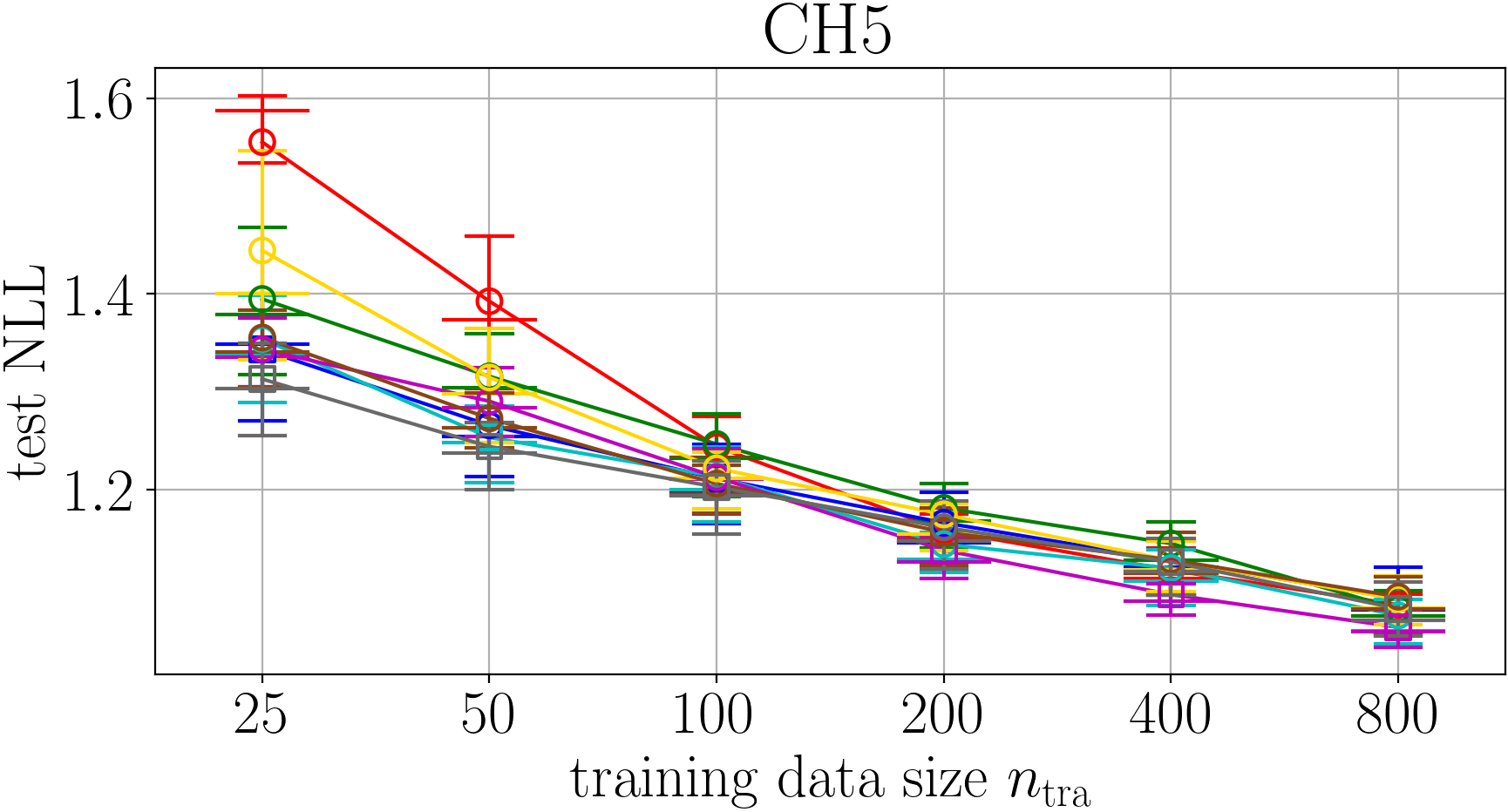}&
\includegraphics[height=1.54cm, bb=0 0 597 327]{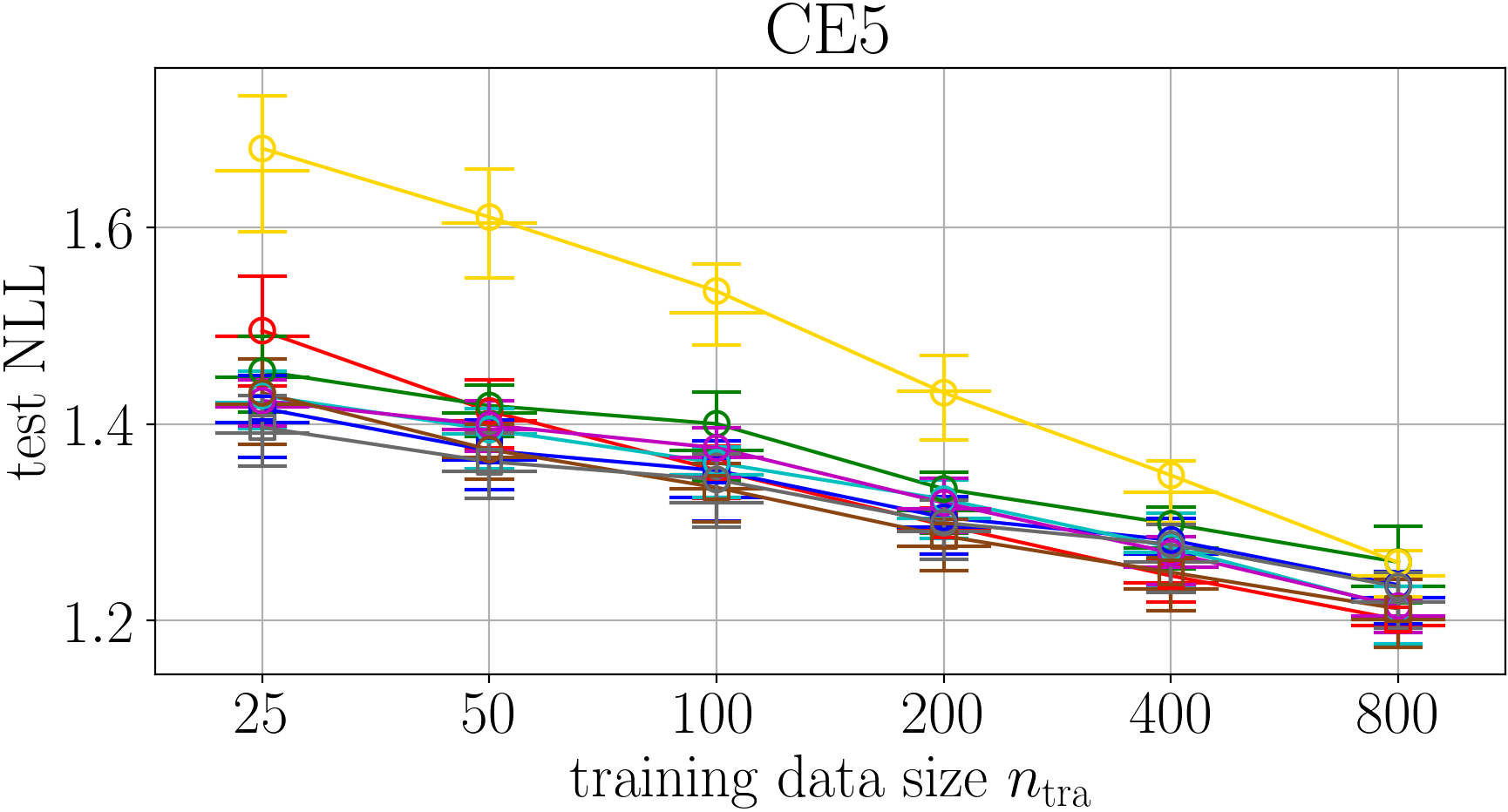}&
\includegraphics[height=1.54cm, bb=0 0 597 327]{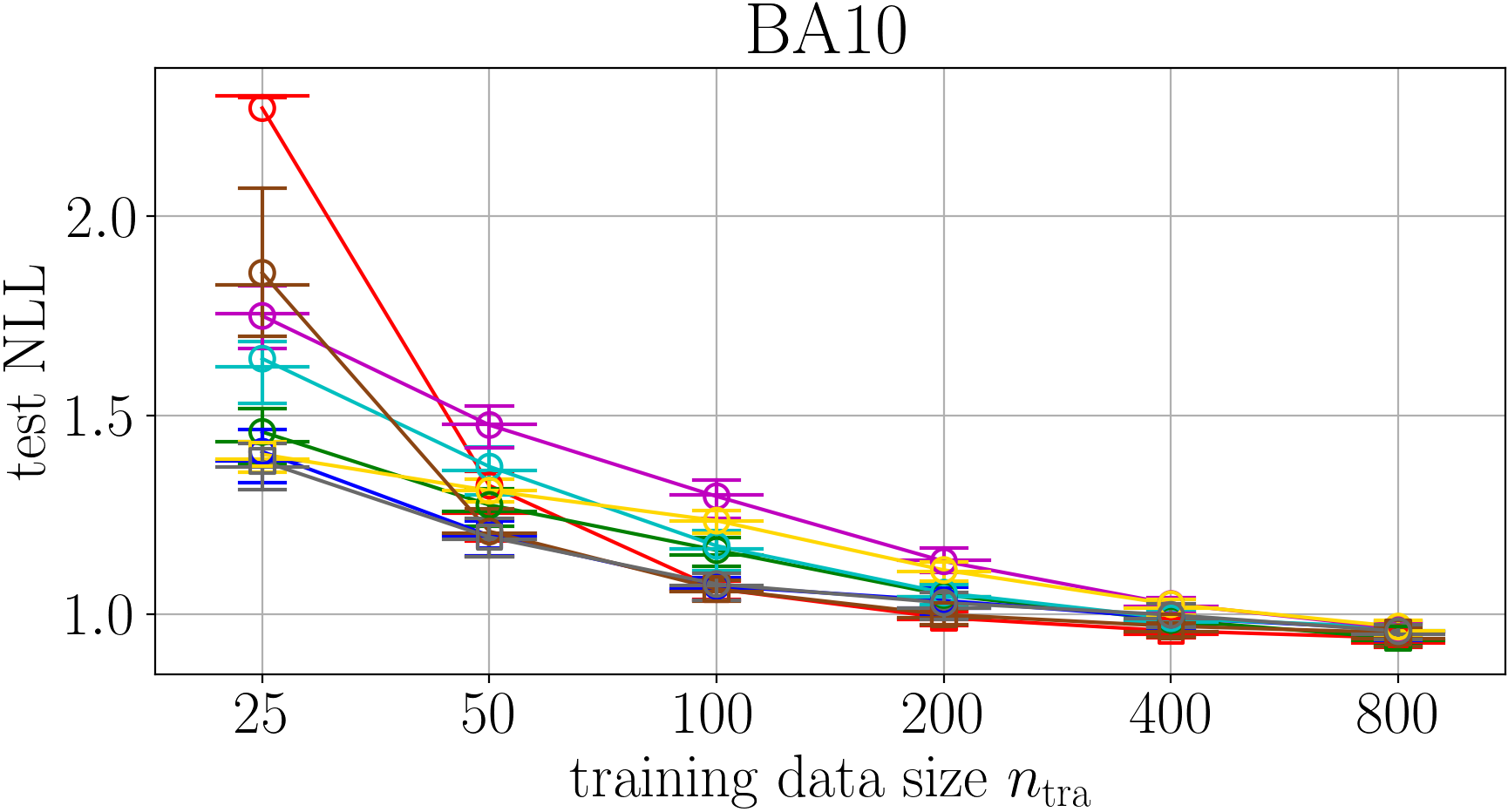}\\
\includegraphics[height=1.54cm, bb=0 0 597 327]{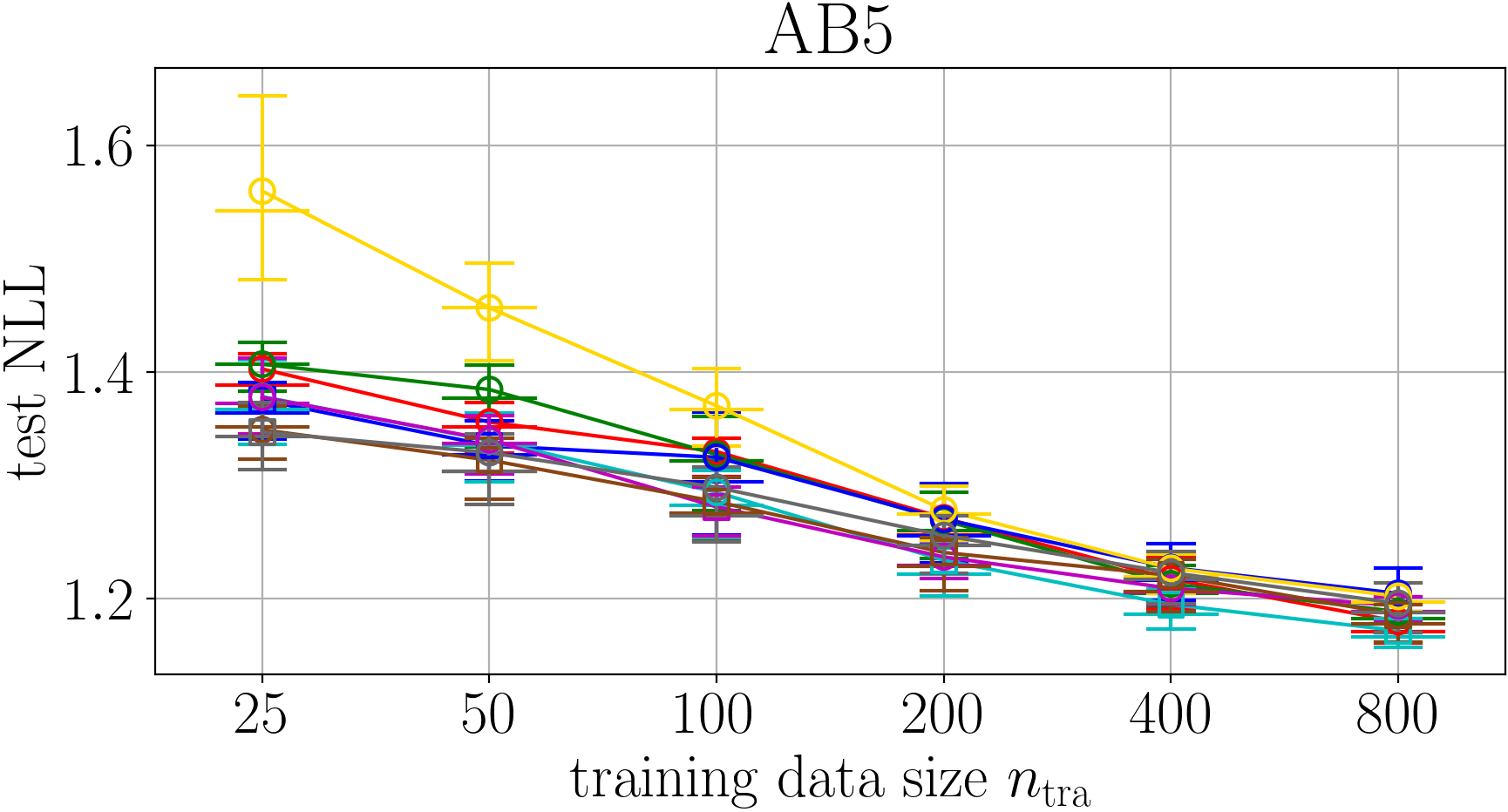}&
\includegraphics[height=1.54cm, bb=0 0 597 327]{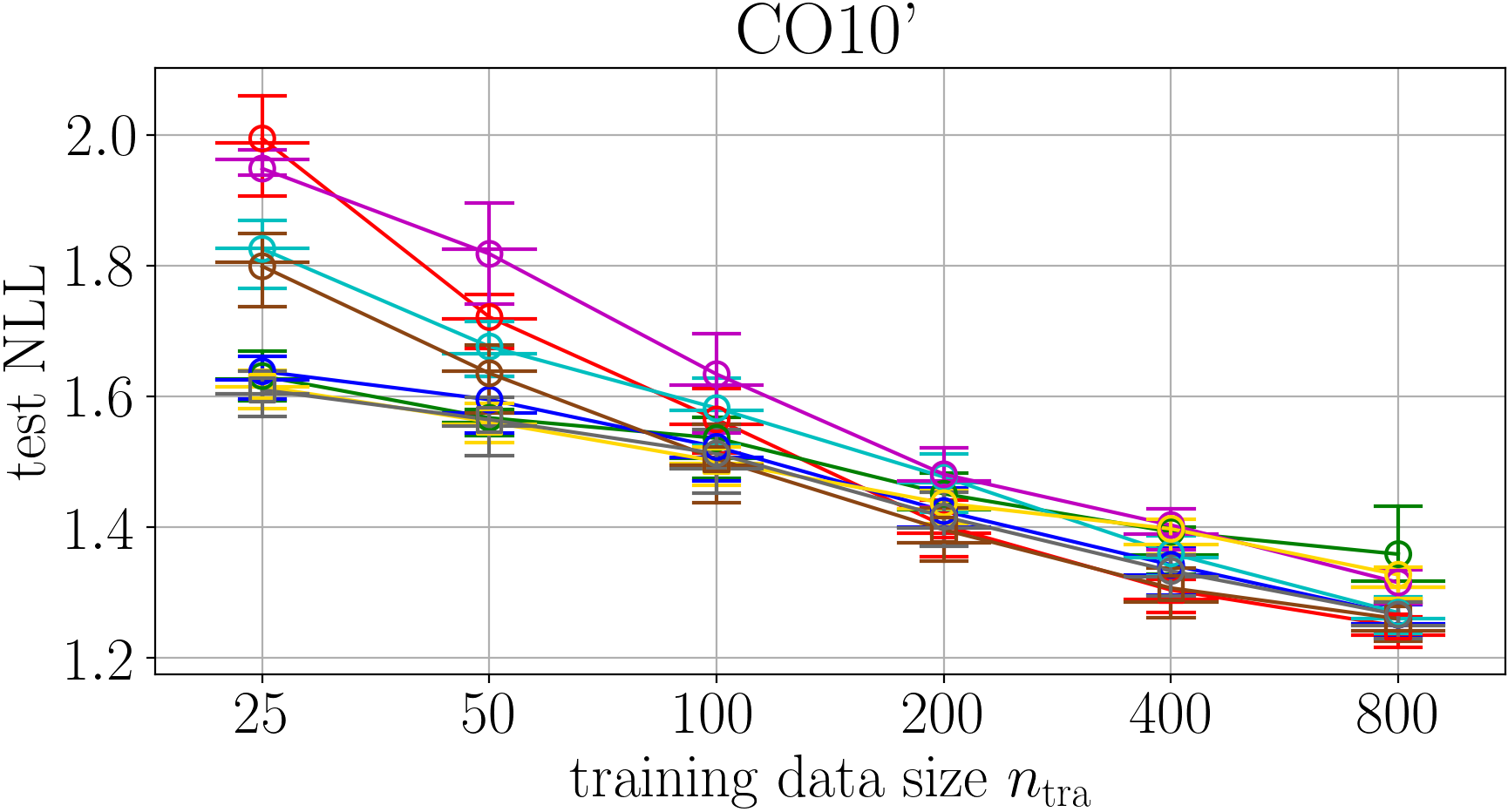}&
\includegraphics[height=1.54cm, bb=0 0 597 327]{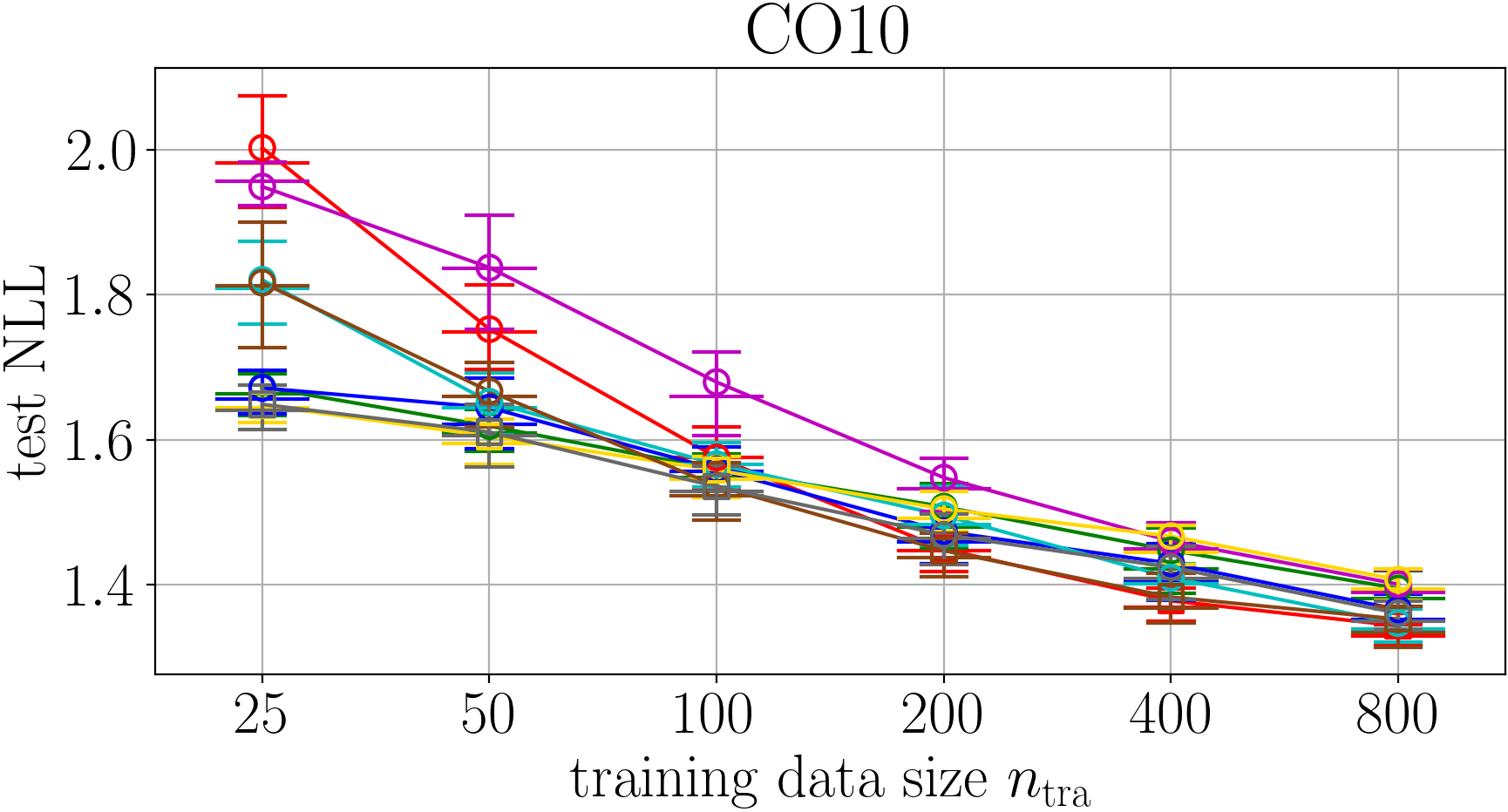}\\
\includegraphics[height=1.54cm, bb=0 0 597 327]{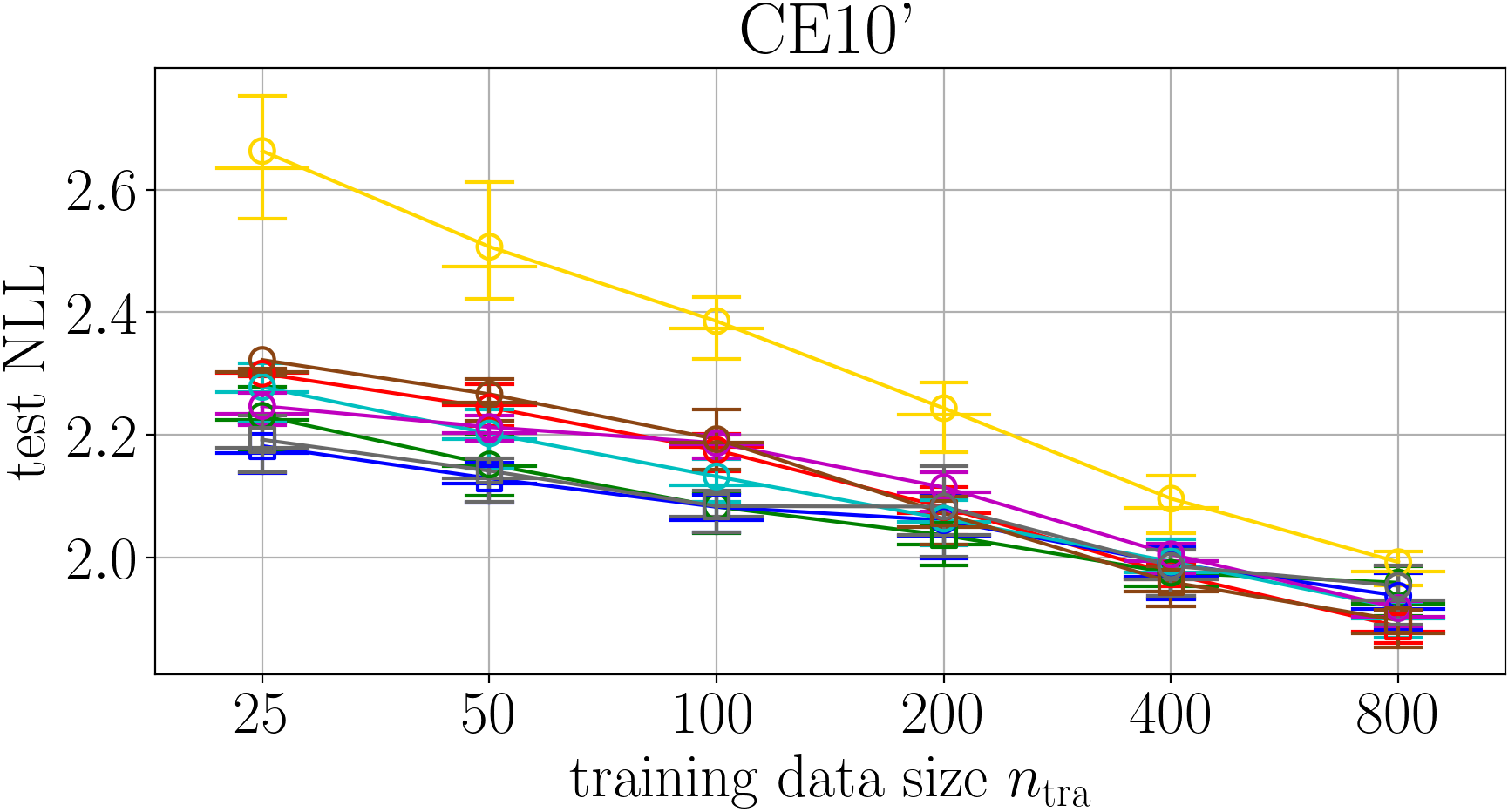}&
\includegraphics[height=1.54cm, bb=0 0 597 327]{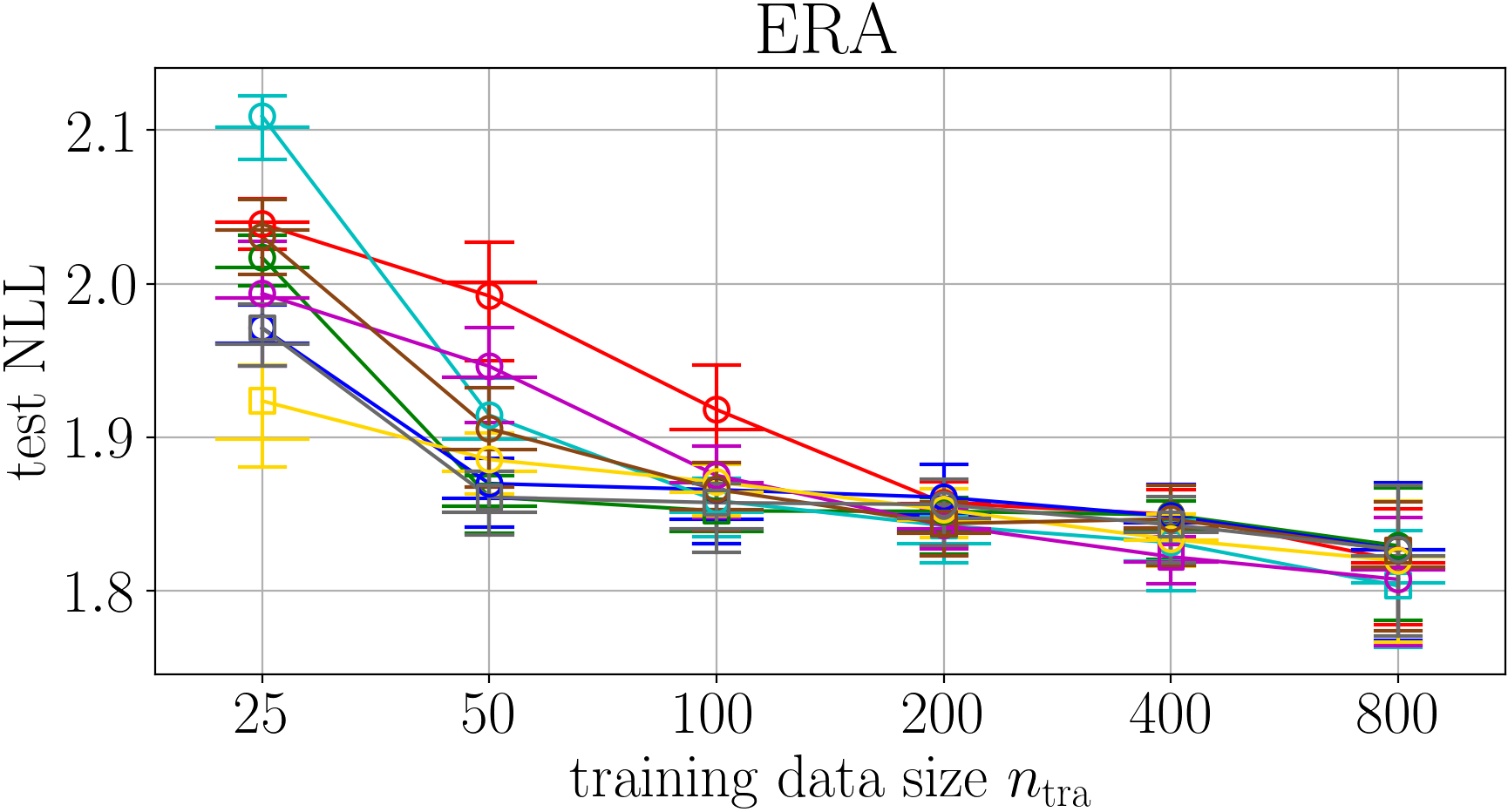}&
\includegraphics[height=1.54cm, bb=0 0 597 327]{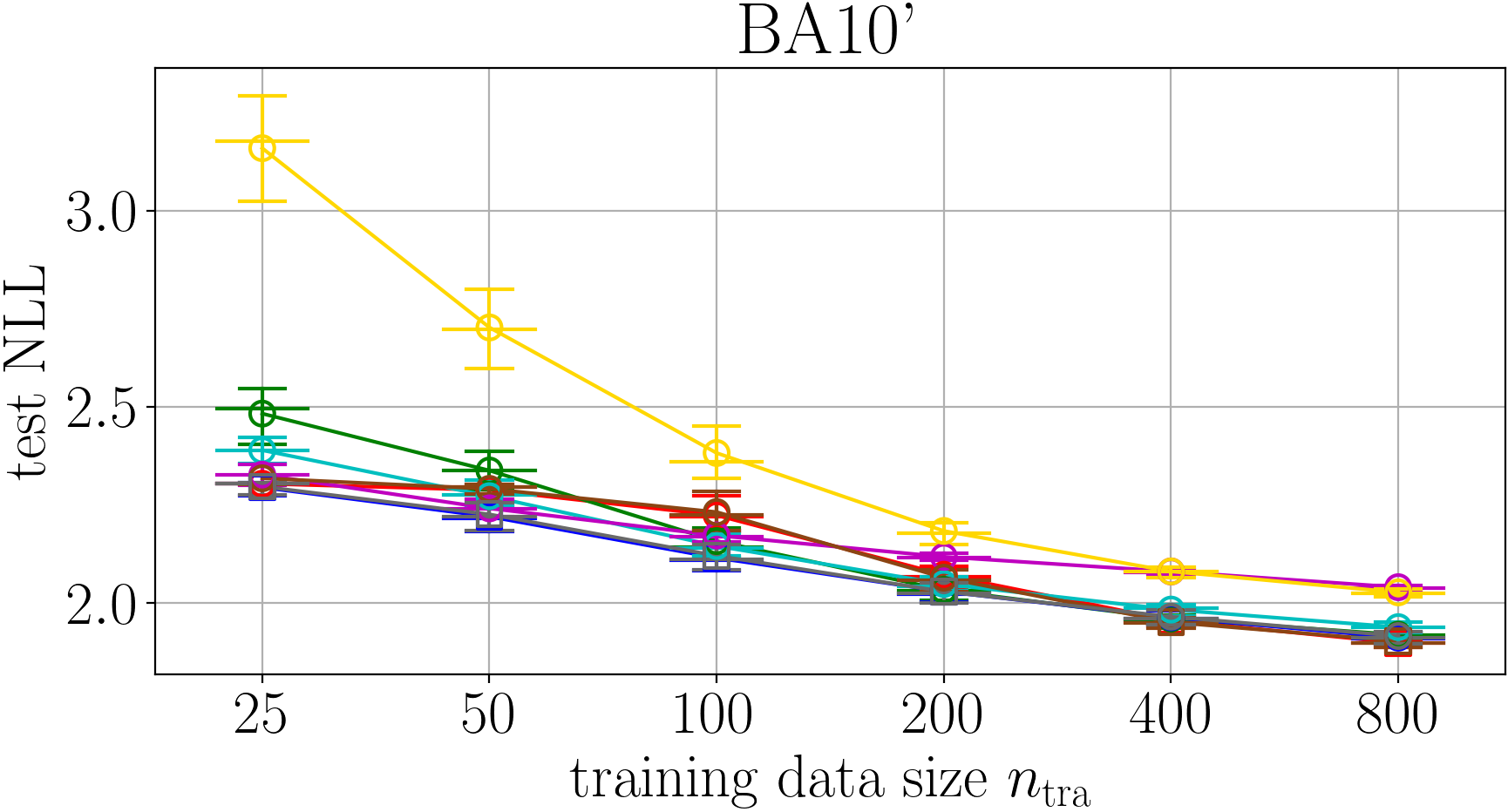}\\
\includegraphics[height=1.54cm, bb=0 0 597 327]{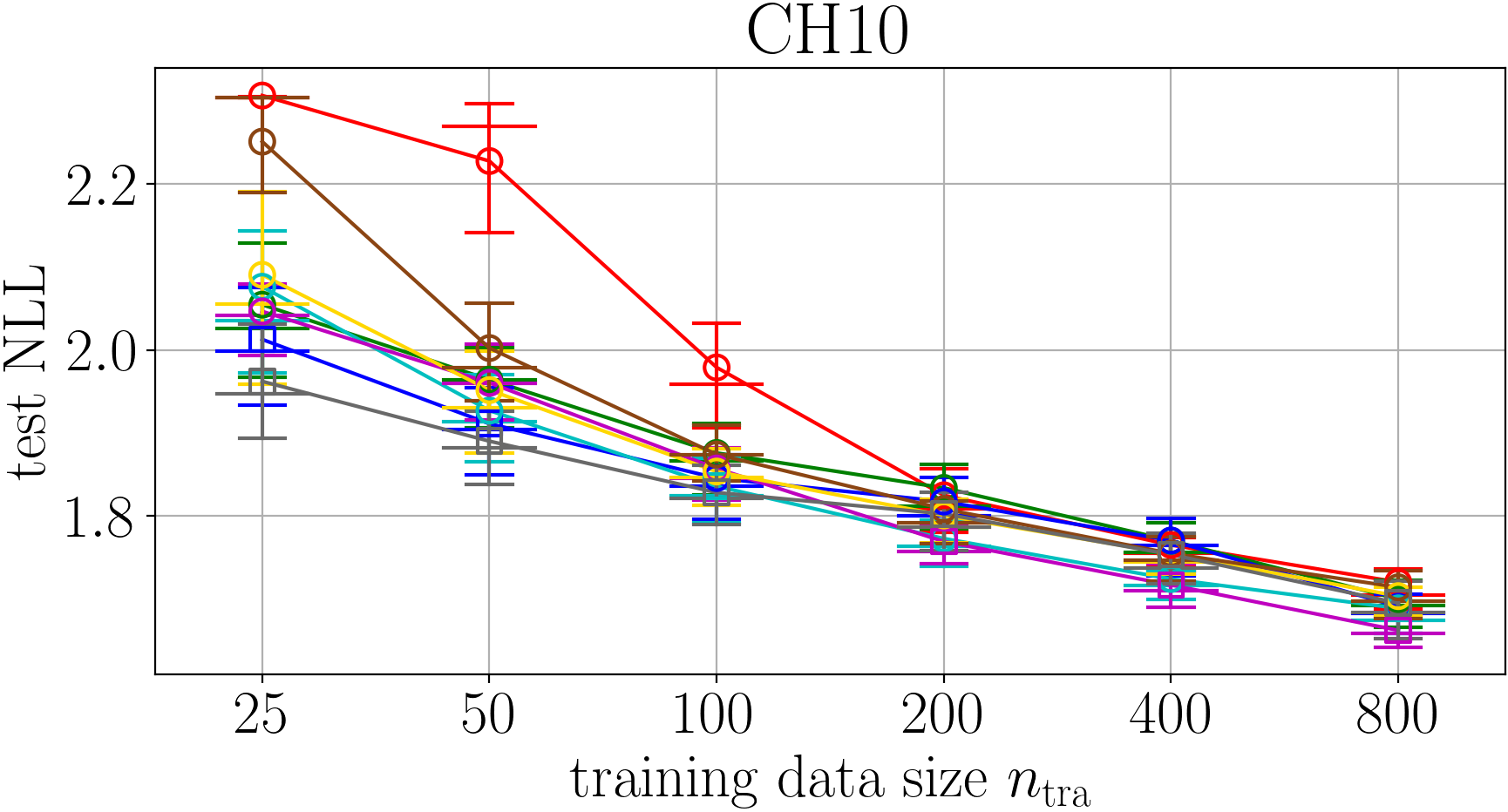}&
\includegraphics[height=1.54cm, bb=0 0 597 327]{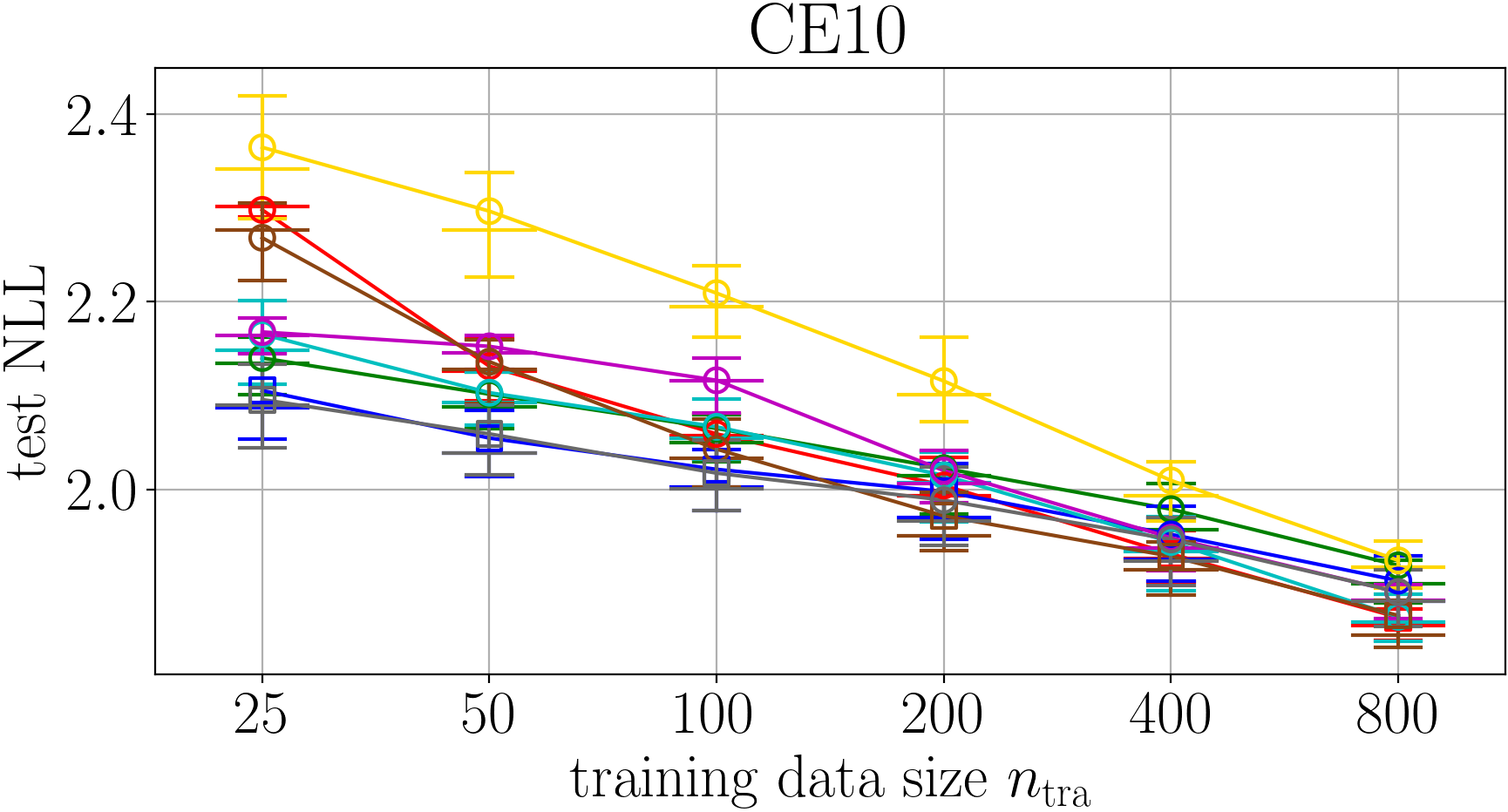}&
\includegraphics[height=1.54cm, bb=0 0 597 327]{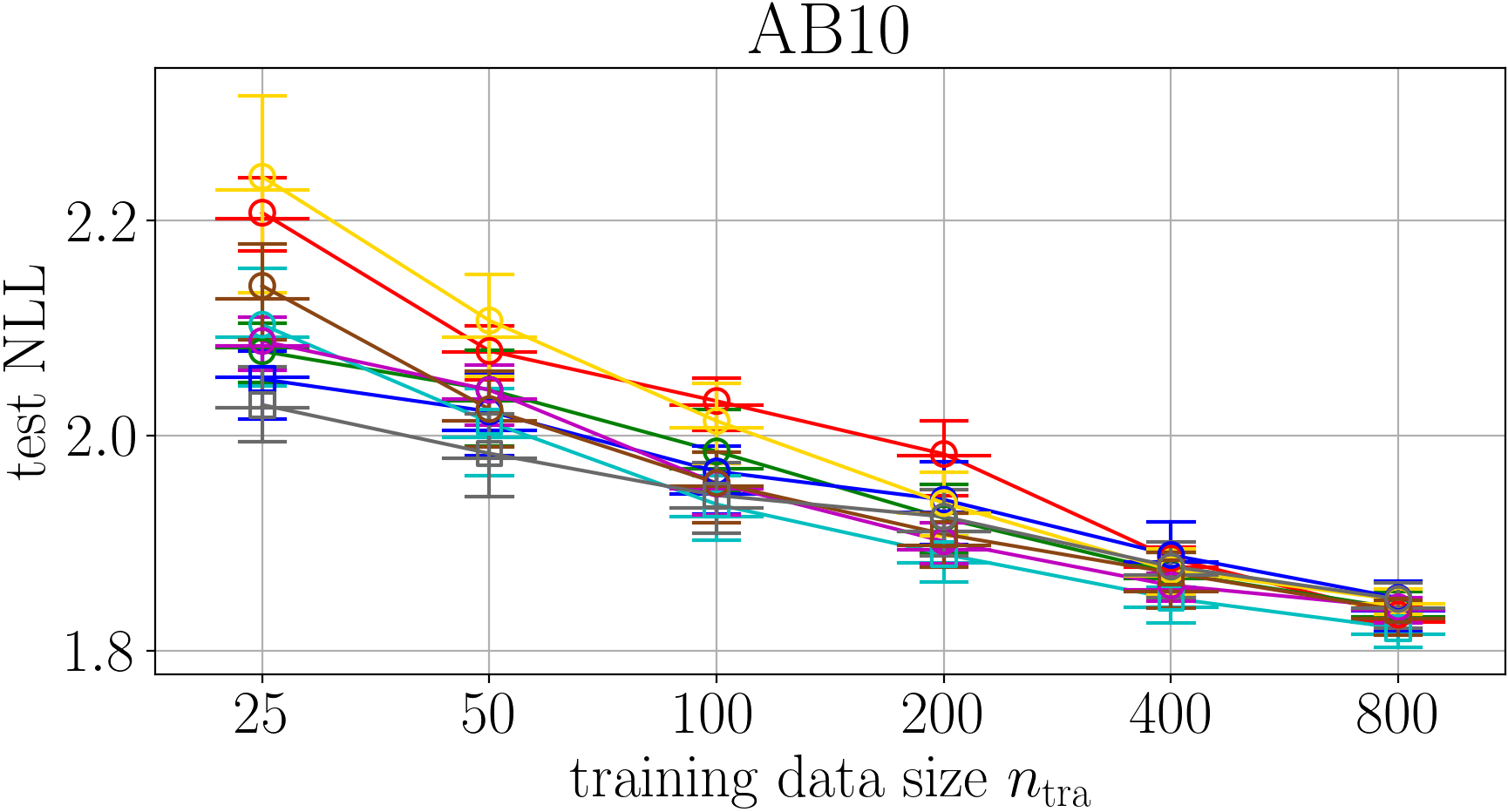}\\
\includegraphics[height=1.54cm, bb=0 0 597 327]{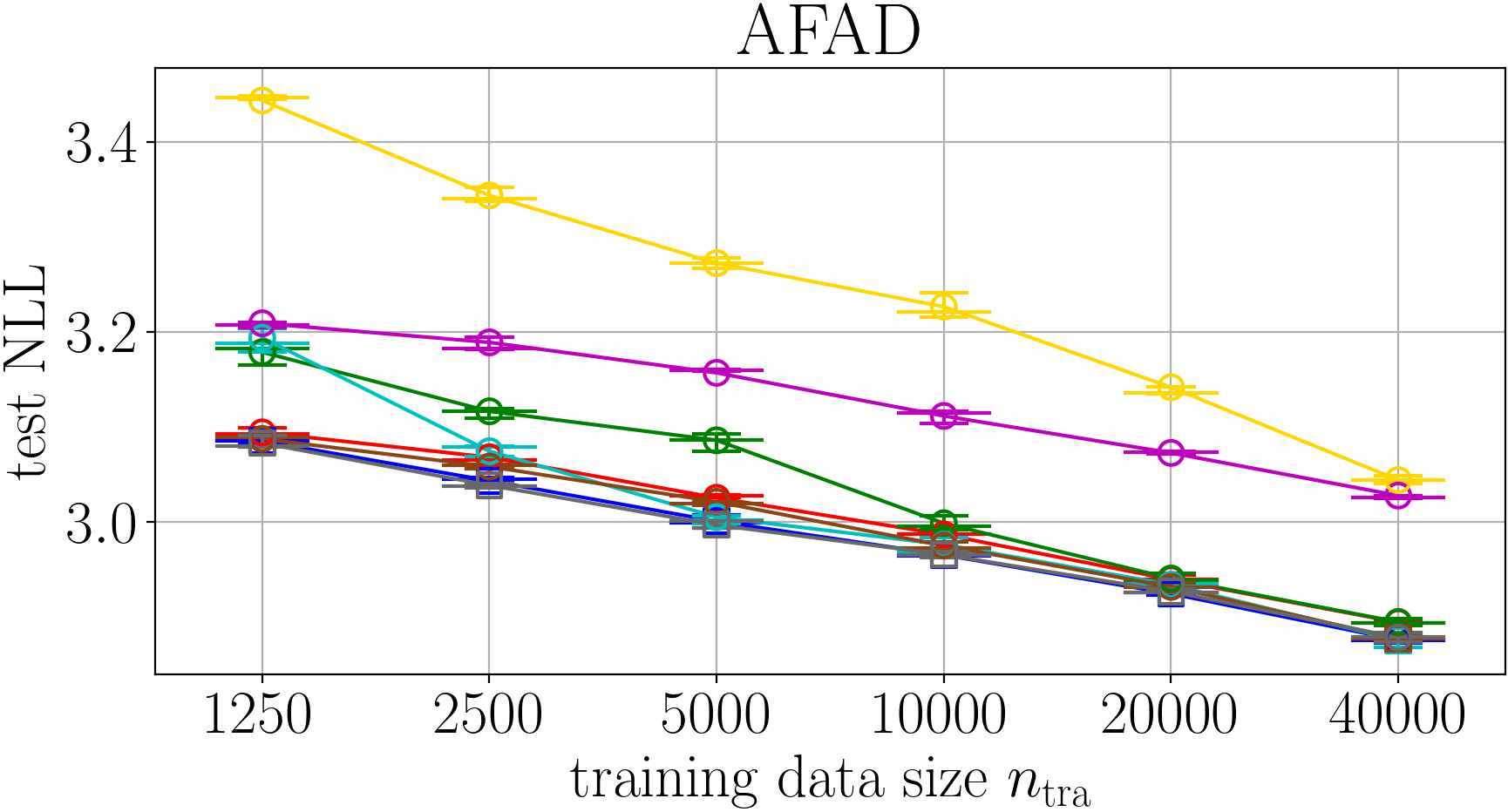}&
\includegraphics[height=1.54cm, bb=0 0 597 327]{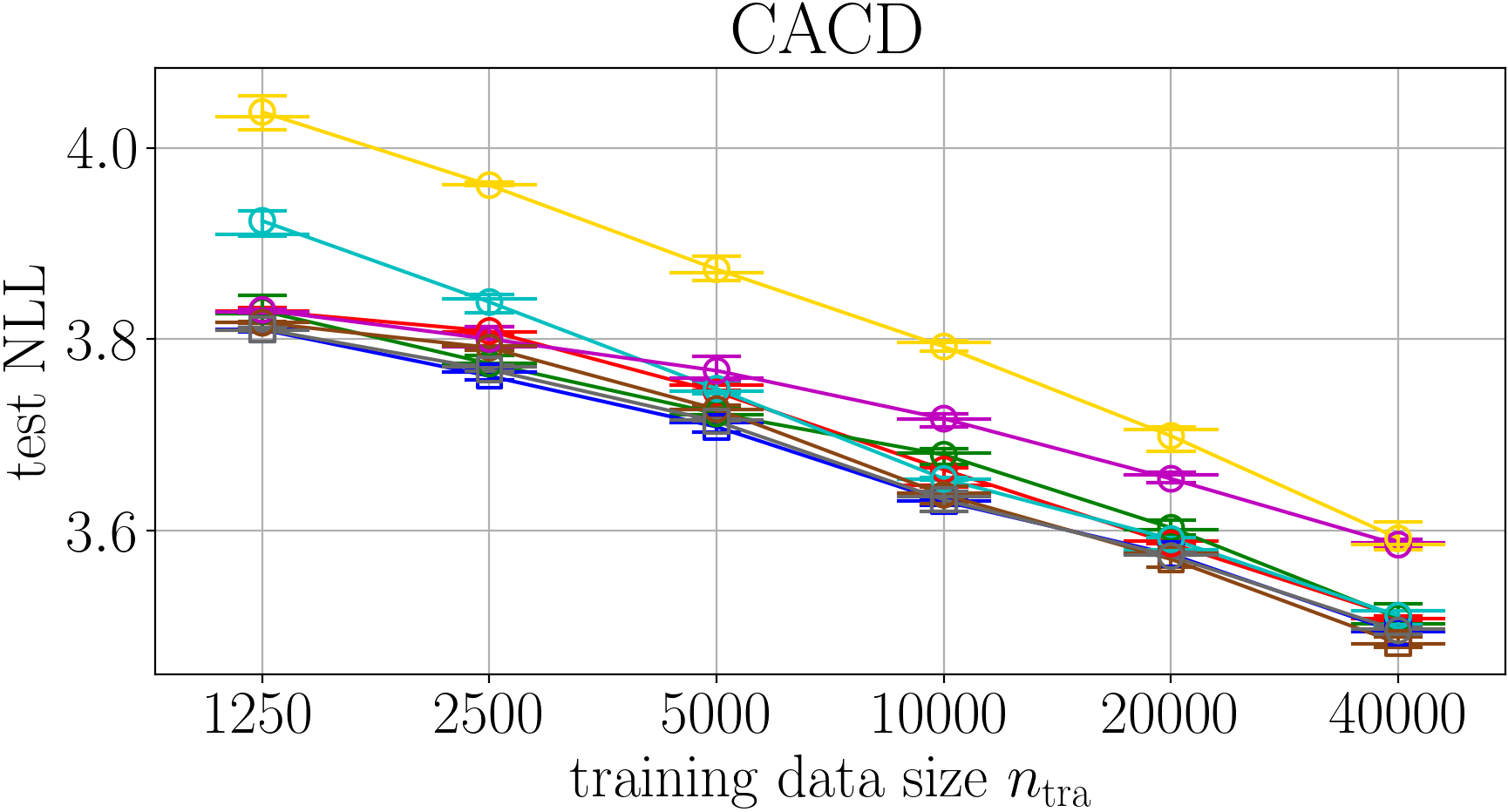}&
\includegraphics[height=1.54cm, bb=0 0 597 327]{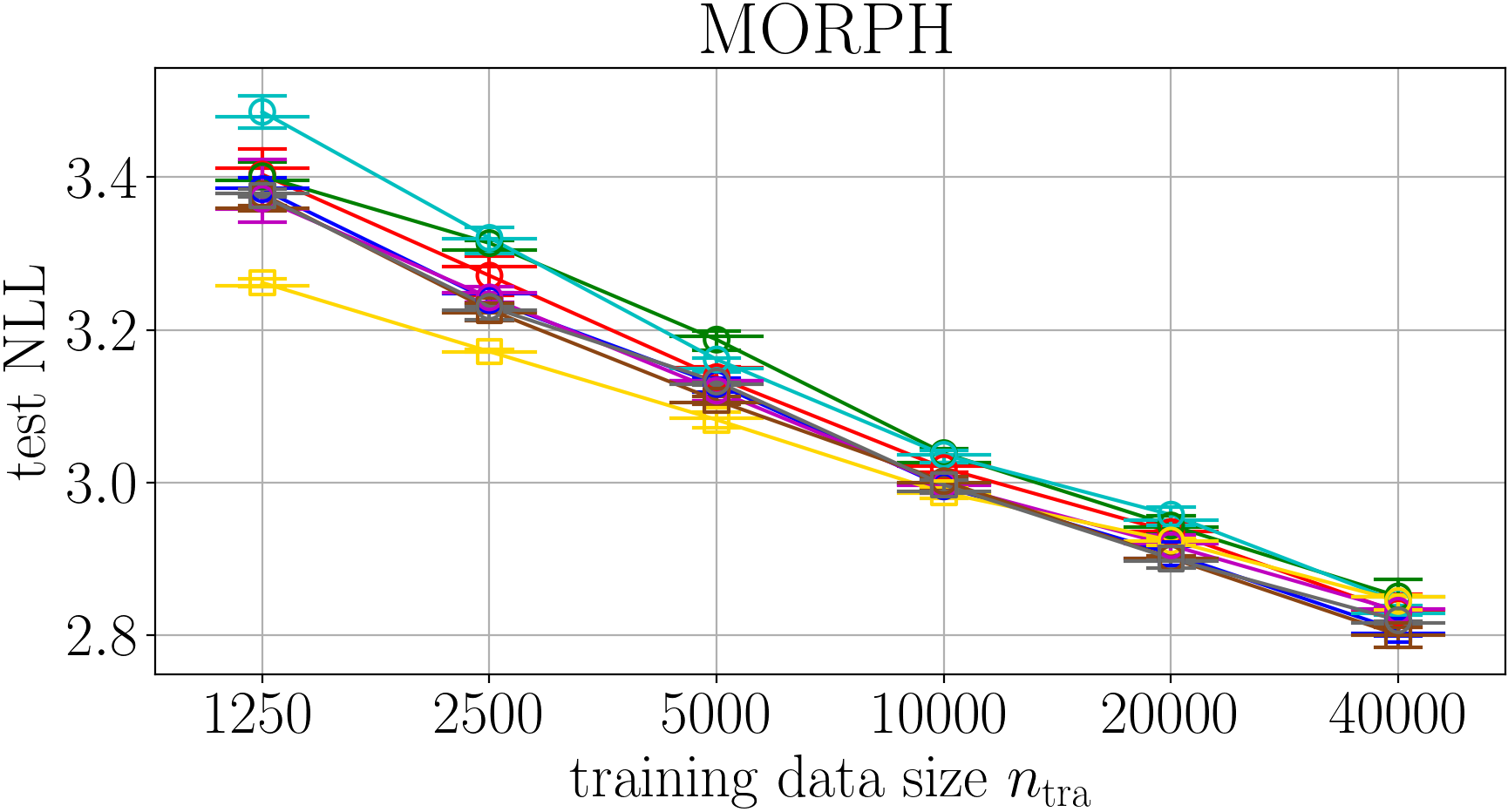}
\end{tabular}
\caption{%
Mean\&quantiles-plot of test NLLs for the SL (red), 
VSL (green), $\text{Mix}_{\mbox{\tiny\text{VSL,SL}}}$ (blue), 
CL (cyan), POCL (magenta), POVSL (yellow), $\text{SL}'$ (brown), 
and $\text{Mix}'_{\mbox{\tiny\text{VSL,SL}}}$ (gray) models.
The squares represent the best among the
SL, $\ldots,$ POVSL models and the better among the
$\text{SL}'$ and $\text{Mix}'_{\mbox{\tiny\text{VSL,SL}}}$ models.}
\label{fig:Res-NLL}
\end{figure}

\addtocounter{footnote}{-1}
\footnotetext{%
In mean\&quantiles-plot of Figures~\ref{fig:Res-Hypara-NLL} and \ref{fig:Res-NLL},
lower short/middle long/upper short bars represent .25/.5/.75 quantiles,
circle and square represents mean, vertical line connects .25 and .75 quantiles,
and polyline connects means.}

\begin{figure}[!t]
\centering%
\renewcommand{\arraystretch}{0.1}%
\renewcommand{\tabcolsep}{0pt}%
\begin{tabular}{C{2.95cm}C{2.95cm}C{2.95cm}}%
\includegraphics[height=1.54cm, bb=0 0 597 327]{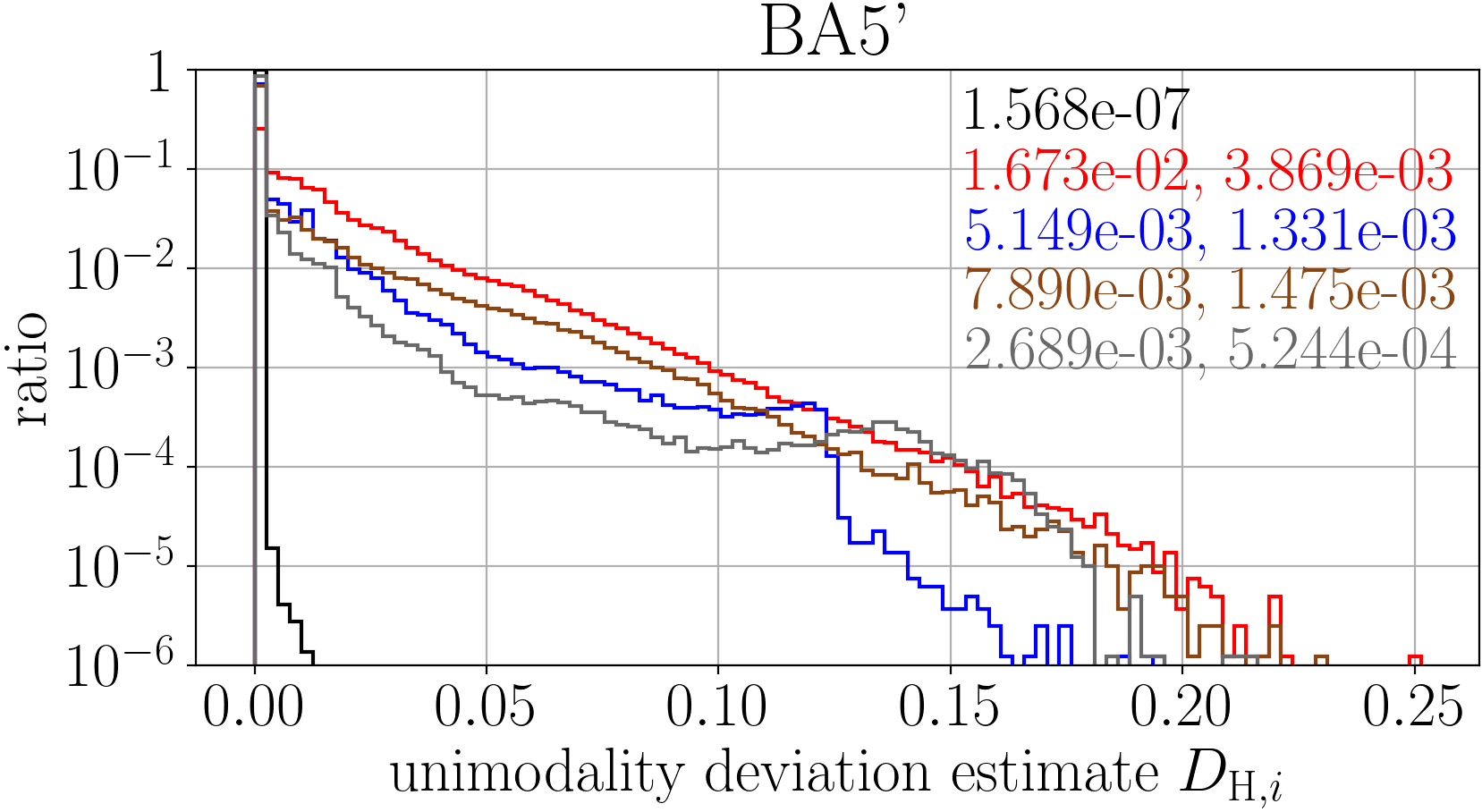}&
\includegraphics[height=1.54cm, bb=0 0 597 327]{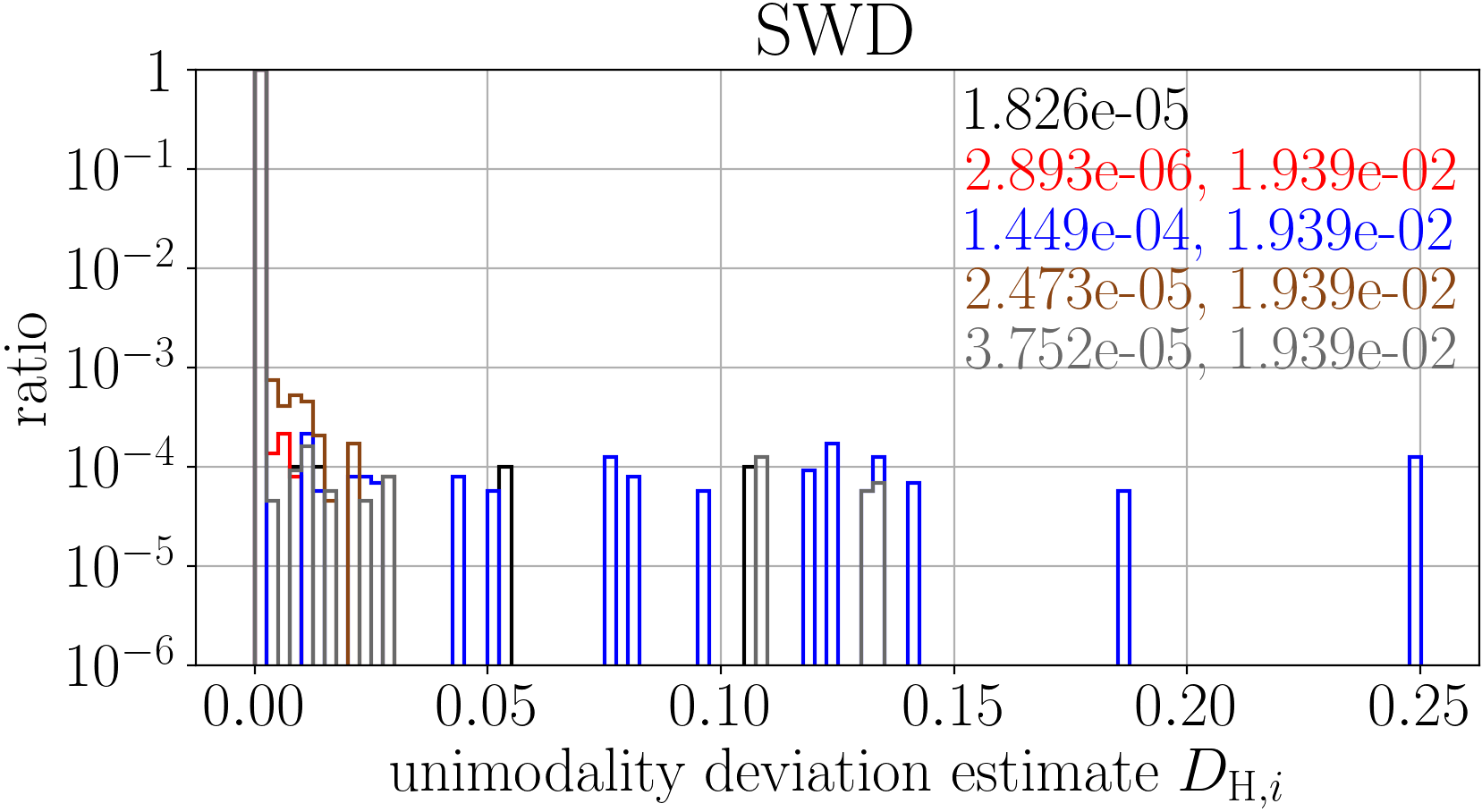}&
\includegraphics[height=1.54cm, bb=0 0 597 327]{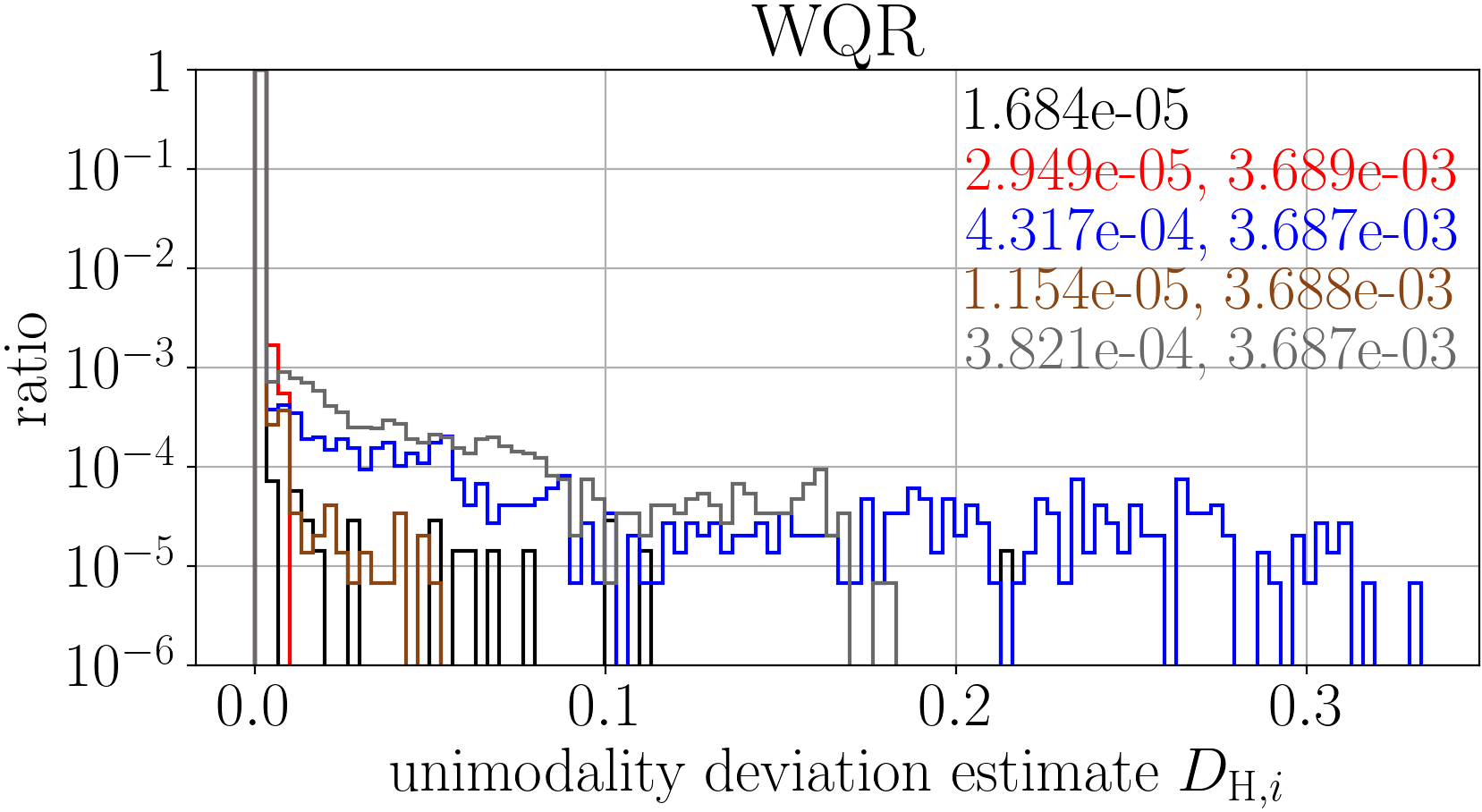}\\
\includegraphics[height=1.54cm, bb=0 0 597 327]{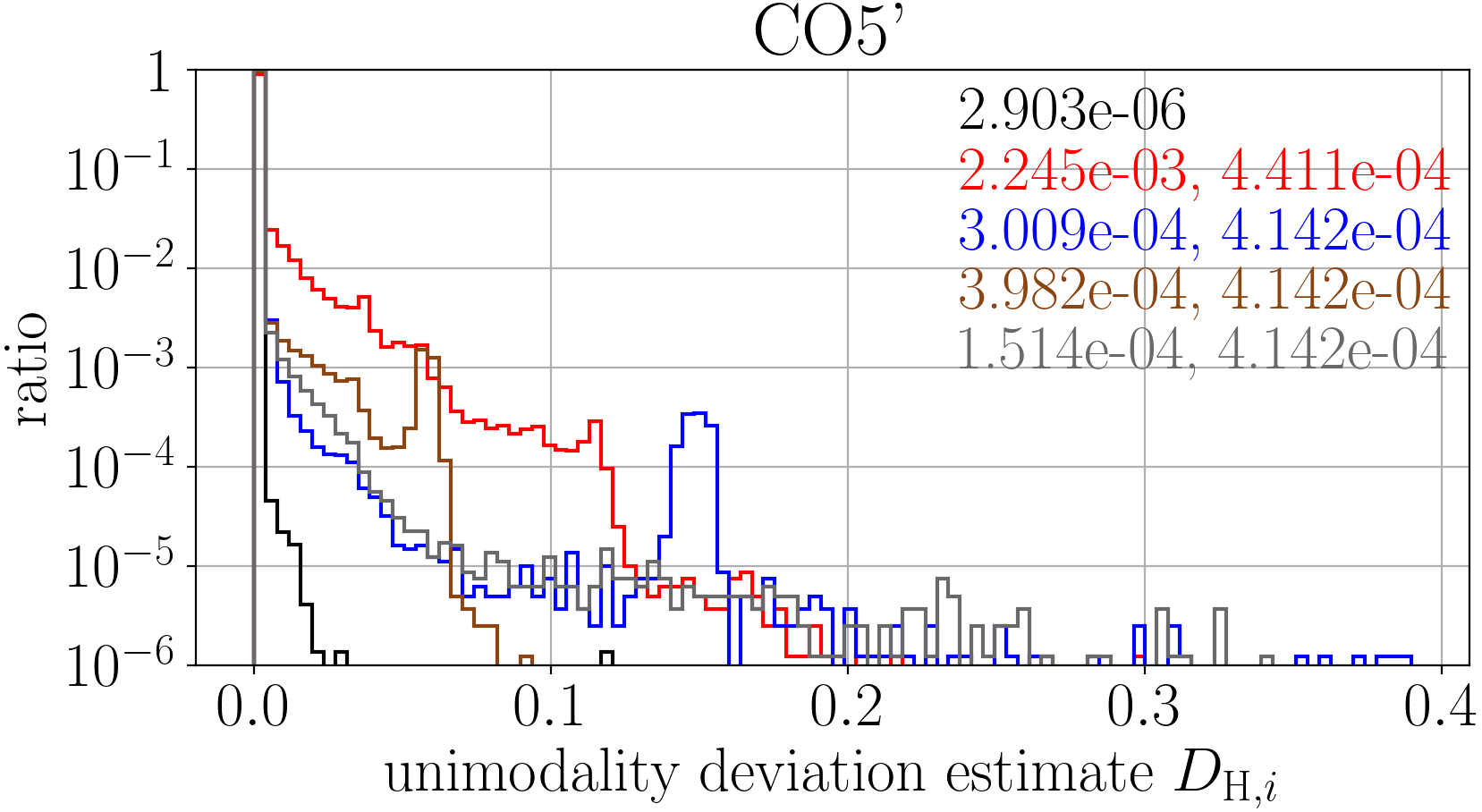}&
\includegraphics[height=1.54cm, bb=0 0 597 327]{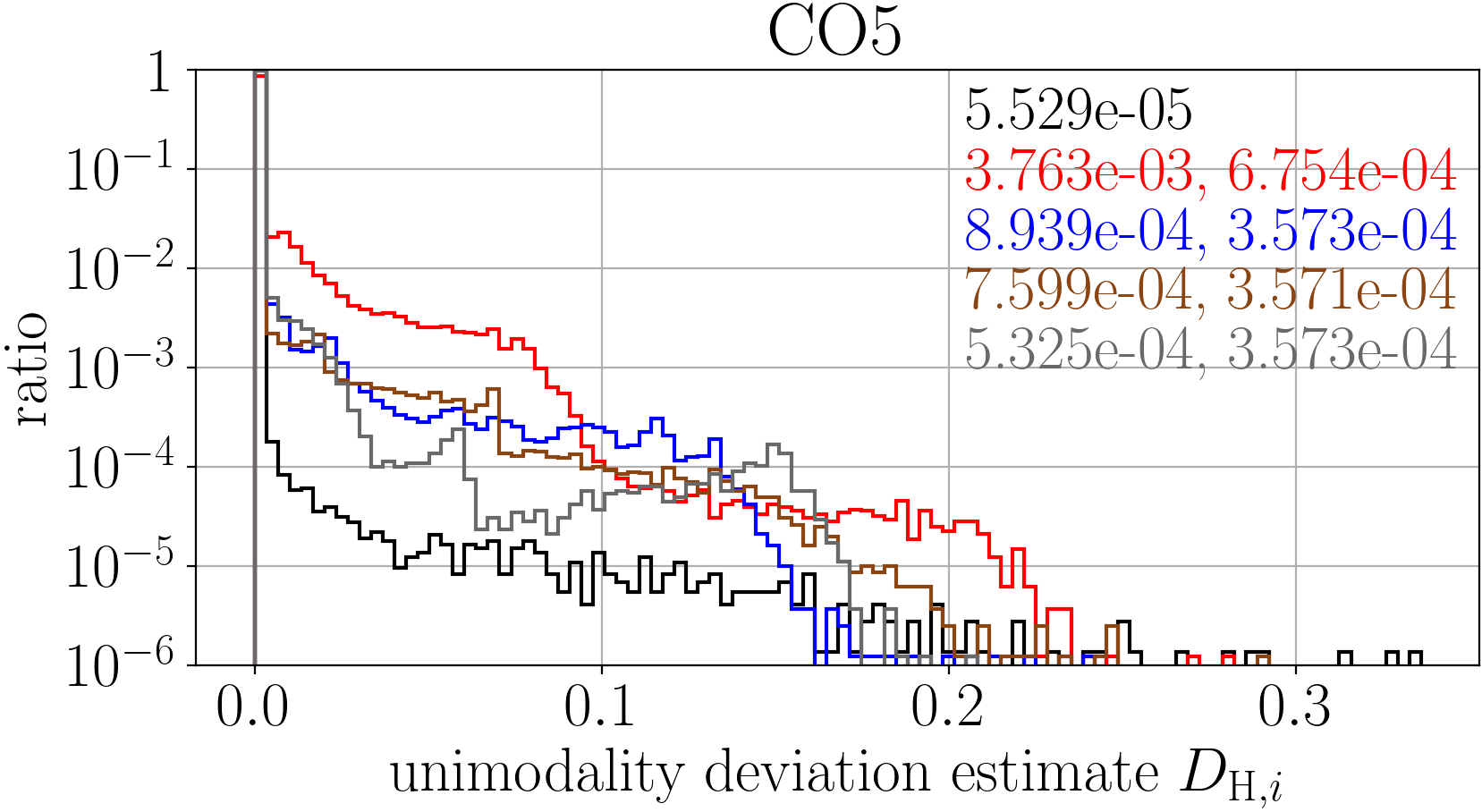}&
\includegraphics[height=1.54cm, bb=0 0 597 327]{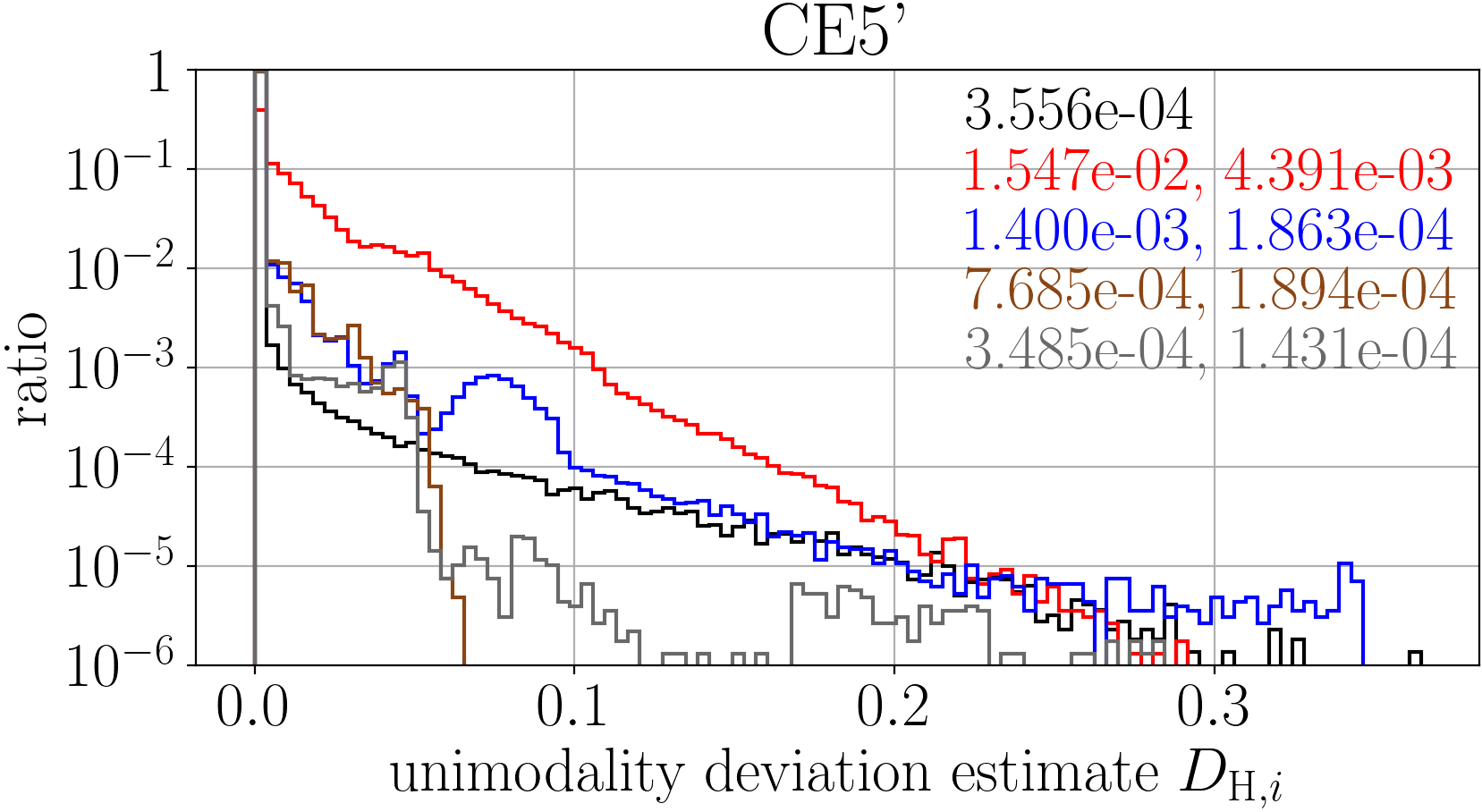}\\
\includegraphics[height=1.54cm, bb=0 0 597 327]{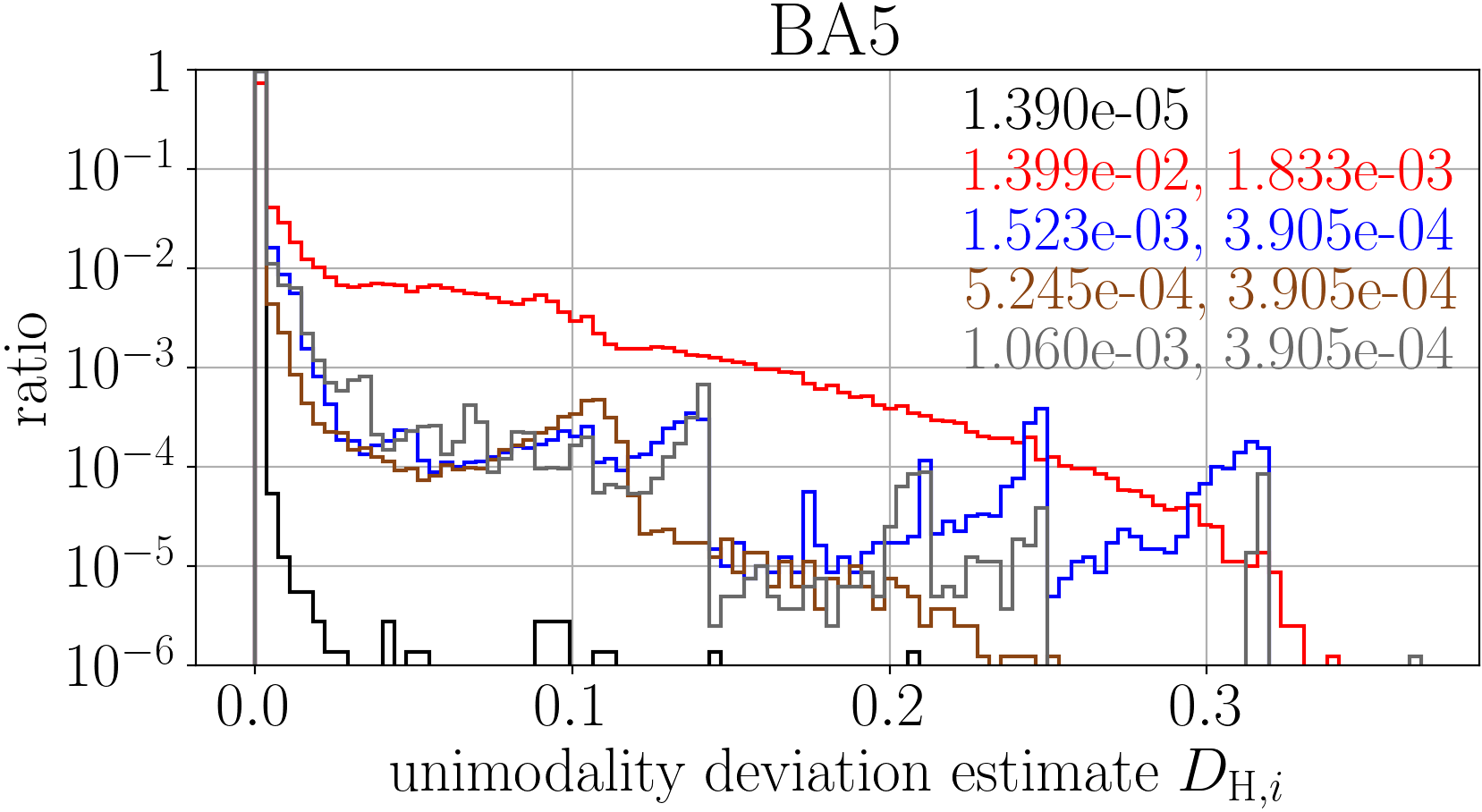}&
\includegraphics[height=1.54cm, bb=0 0 597 327]{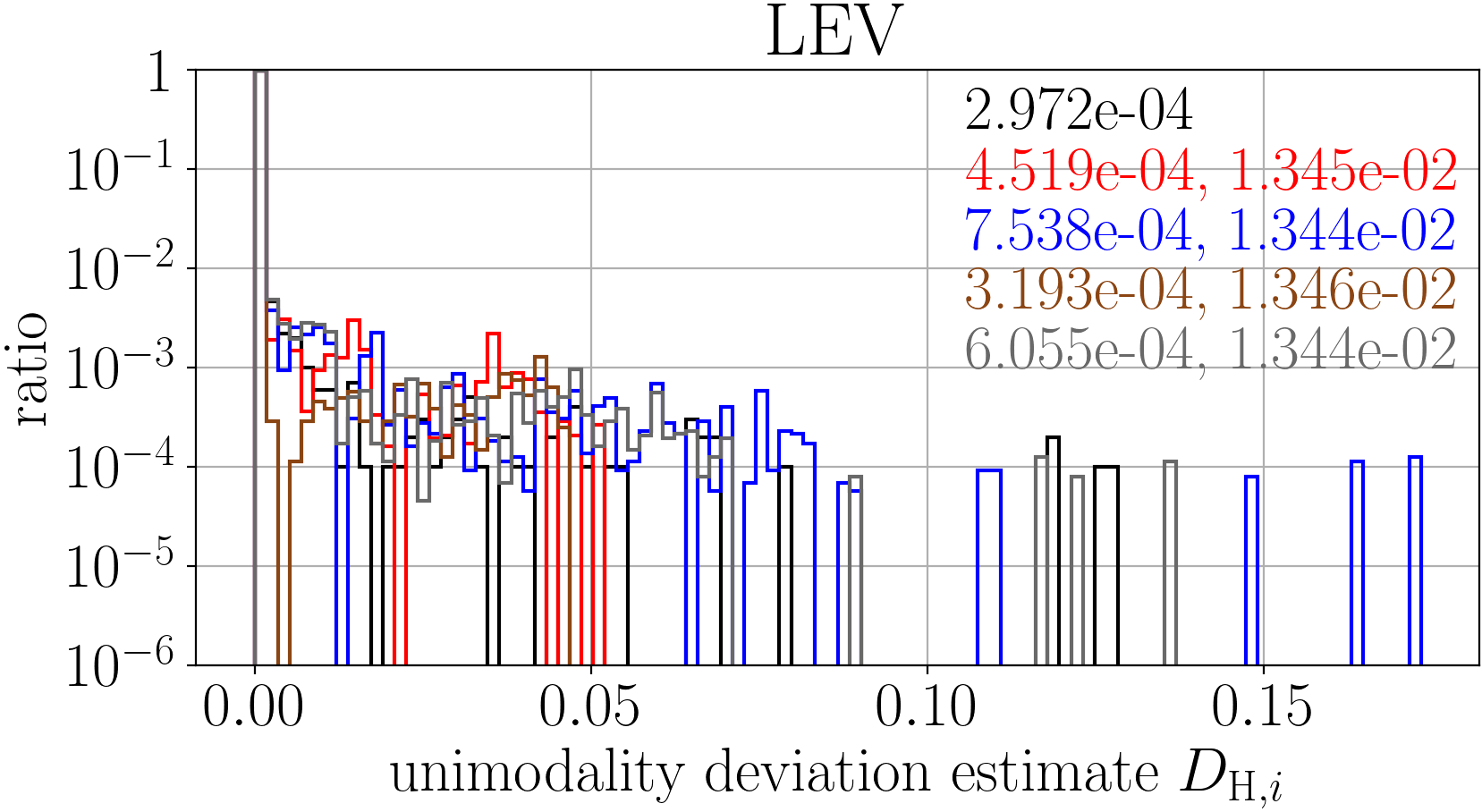}&
\includegraphics[height=1.54cm, bb=0 0 597 327]{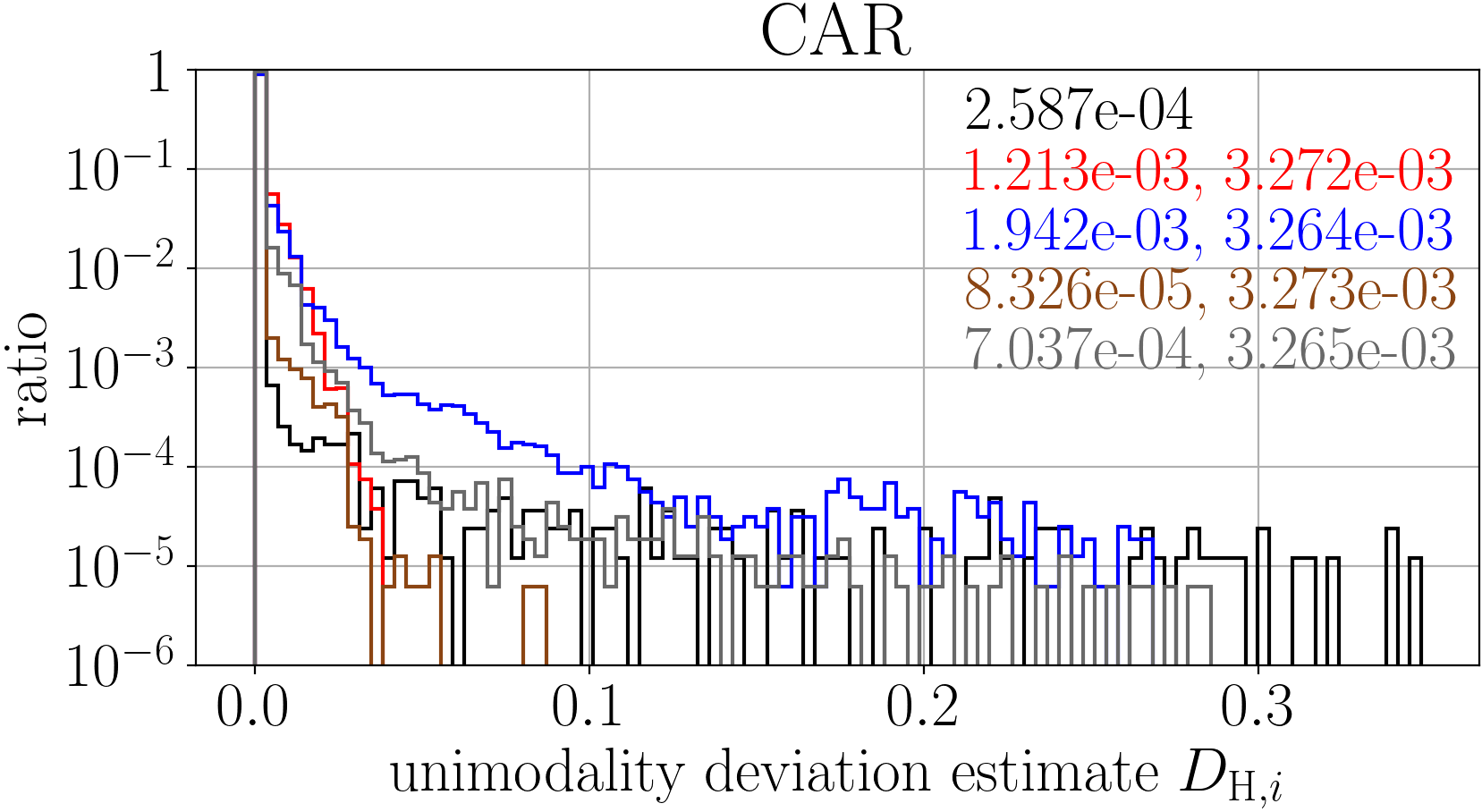}\\
\includegraphics[height=1.54cm, bb=0 0 597 327]{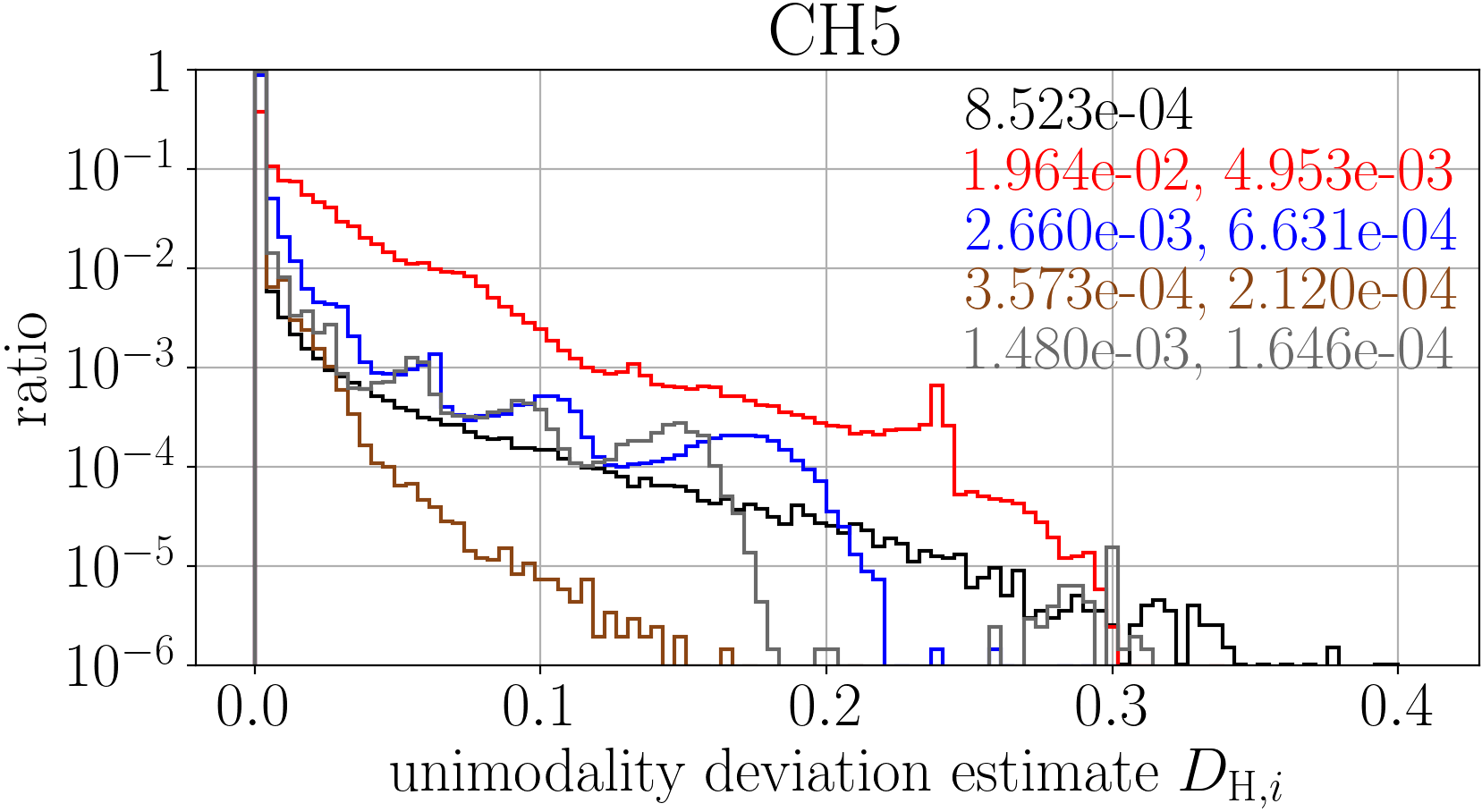}&
\includegraphics[height=1.54cm, bb=0 0 597 327]{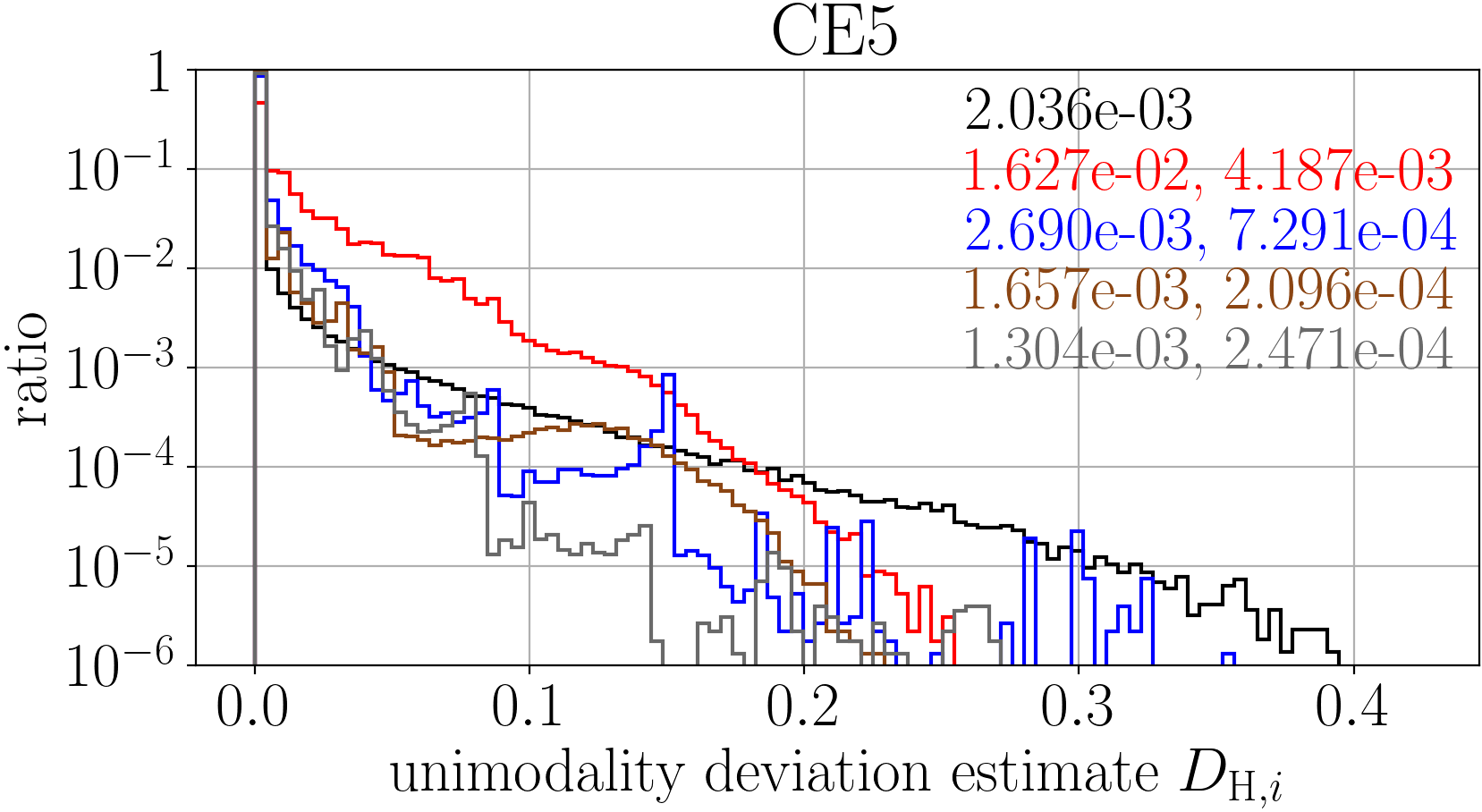}&
\includegraphics[height=1.54cm, bb=0 0 597 327]{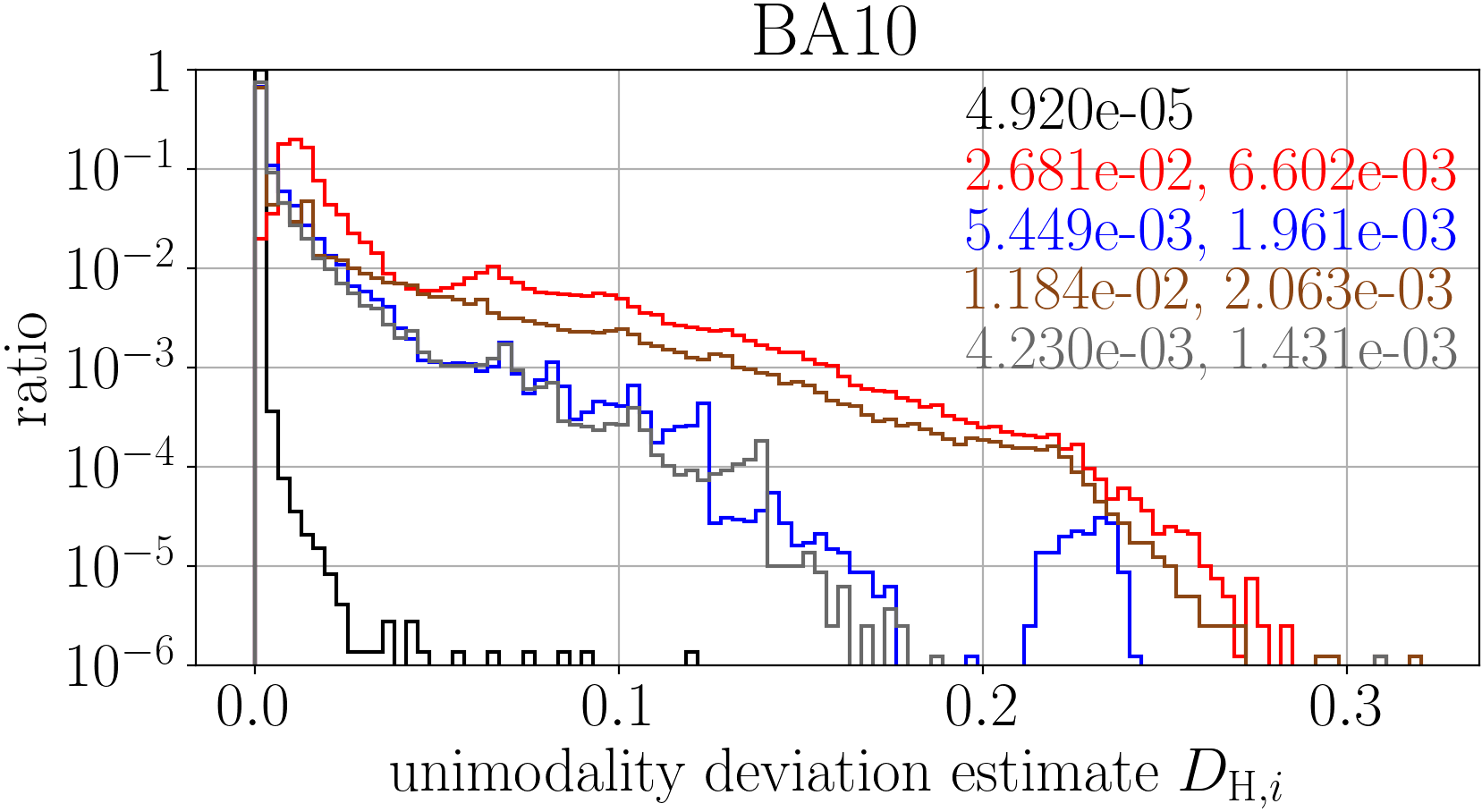}\\
\includegraphics[height=1.54cm, bb=0 0 597 327]{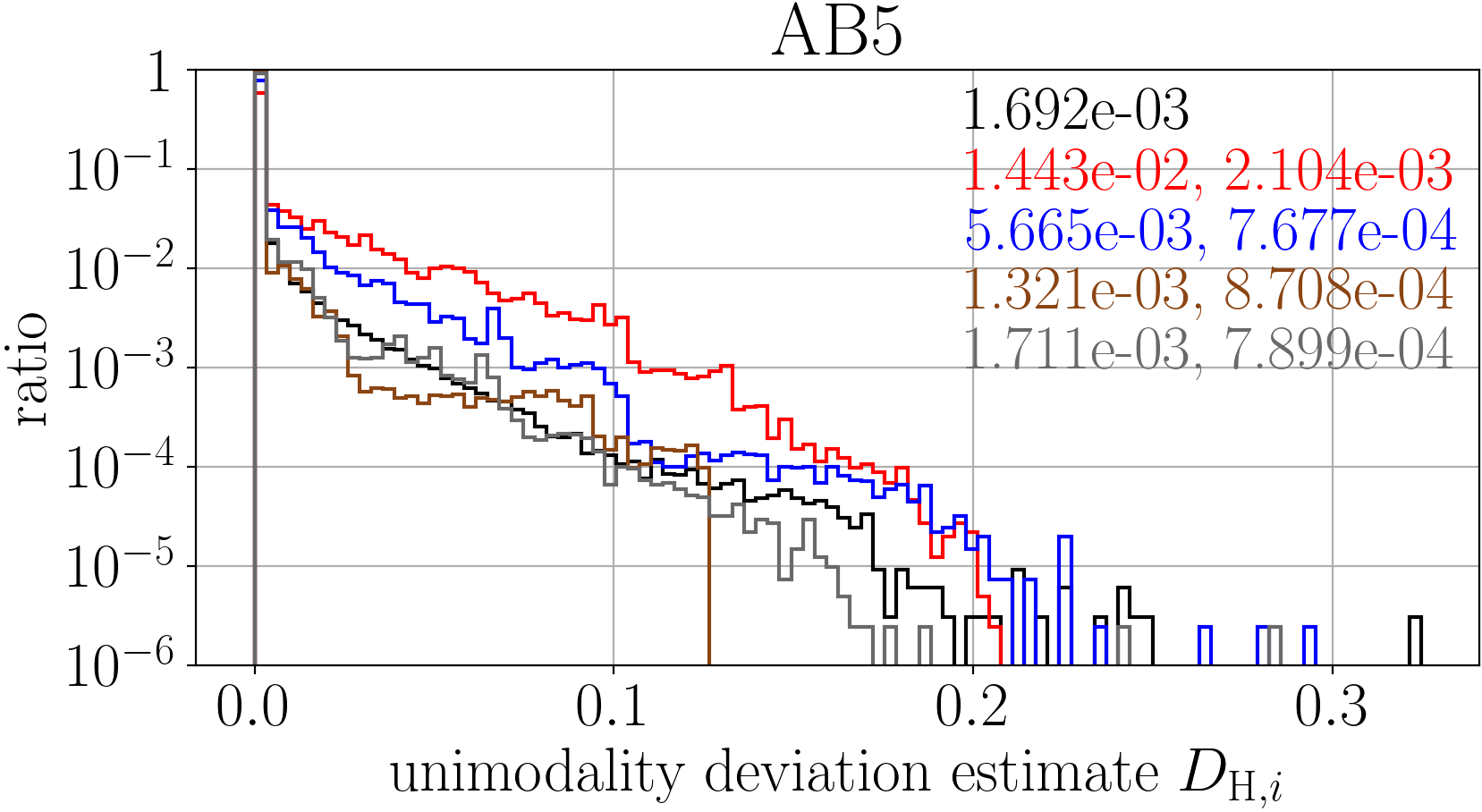}&
\includegraphics[height=1.54cm, bb=0 0 597 327]{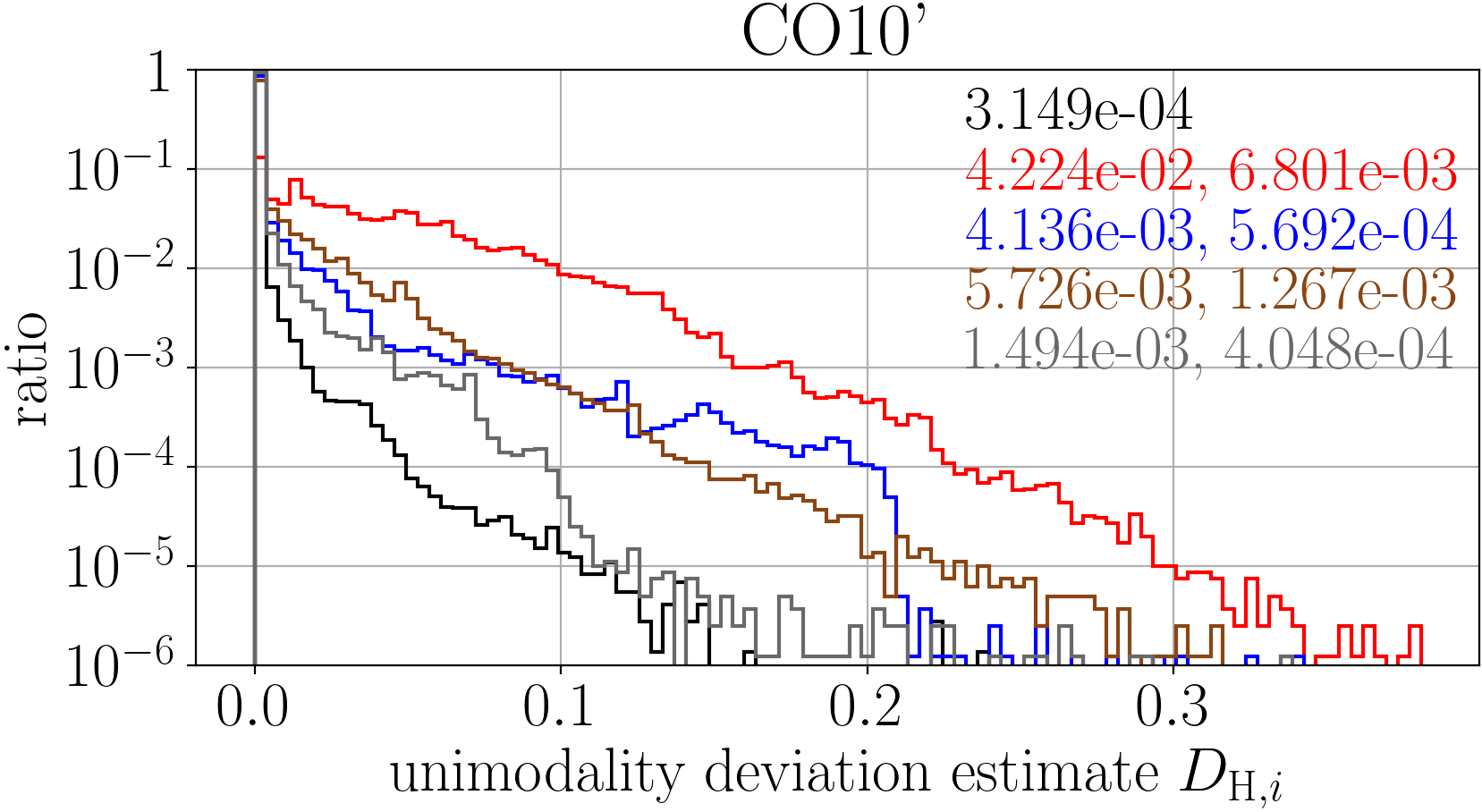}&
\includegraphics[height=1.54cm, bb=0 0 597 327]{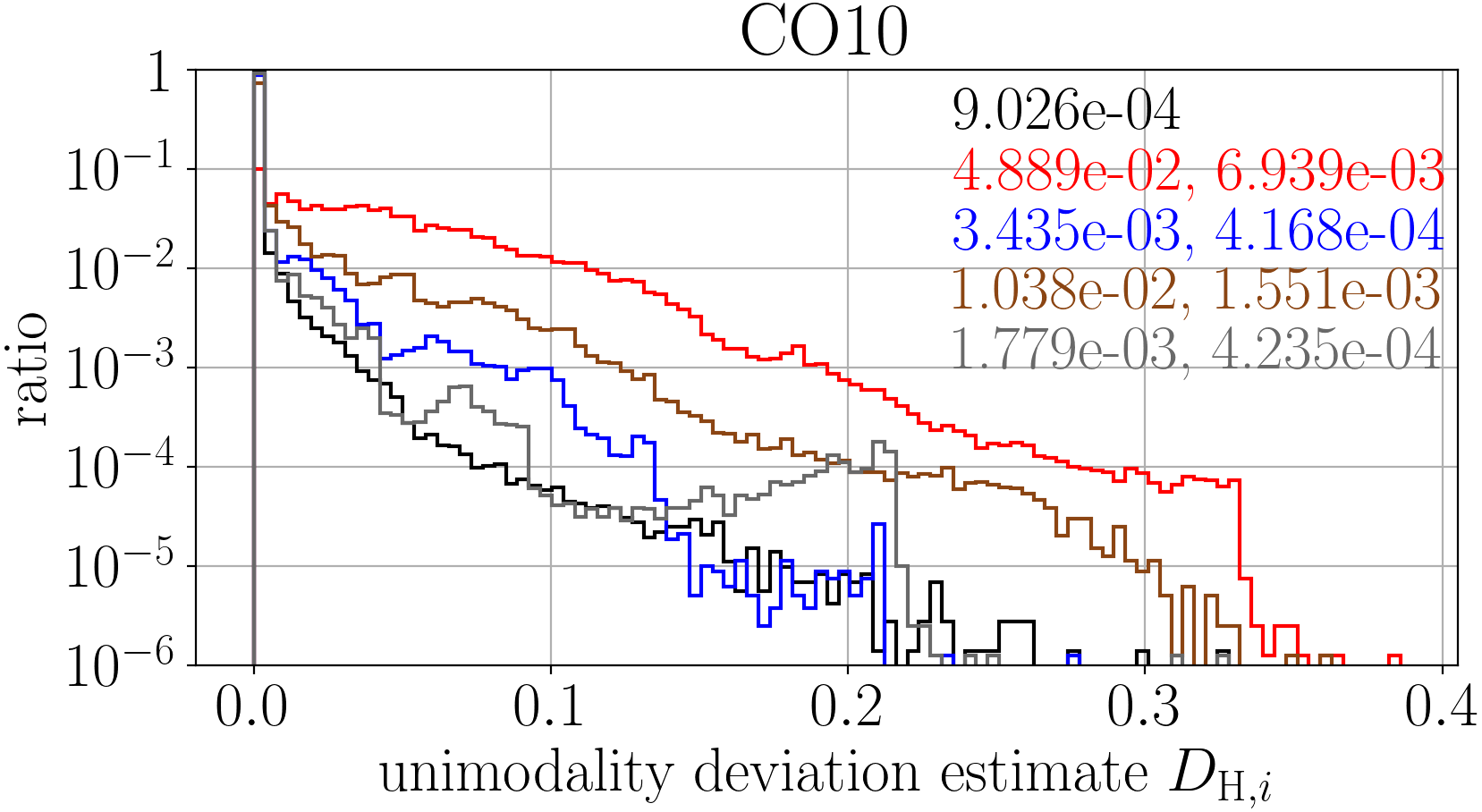}\\
\includegraphics[height=1.54cm, bb=0 0 597 327]{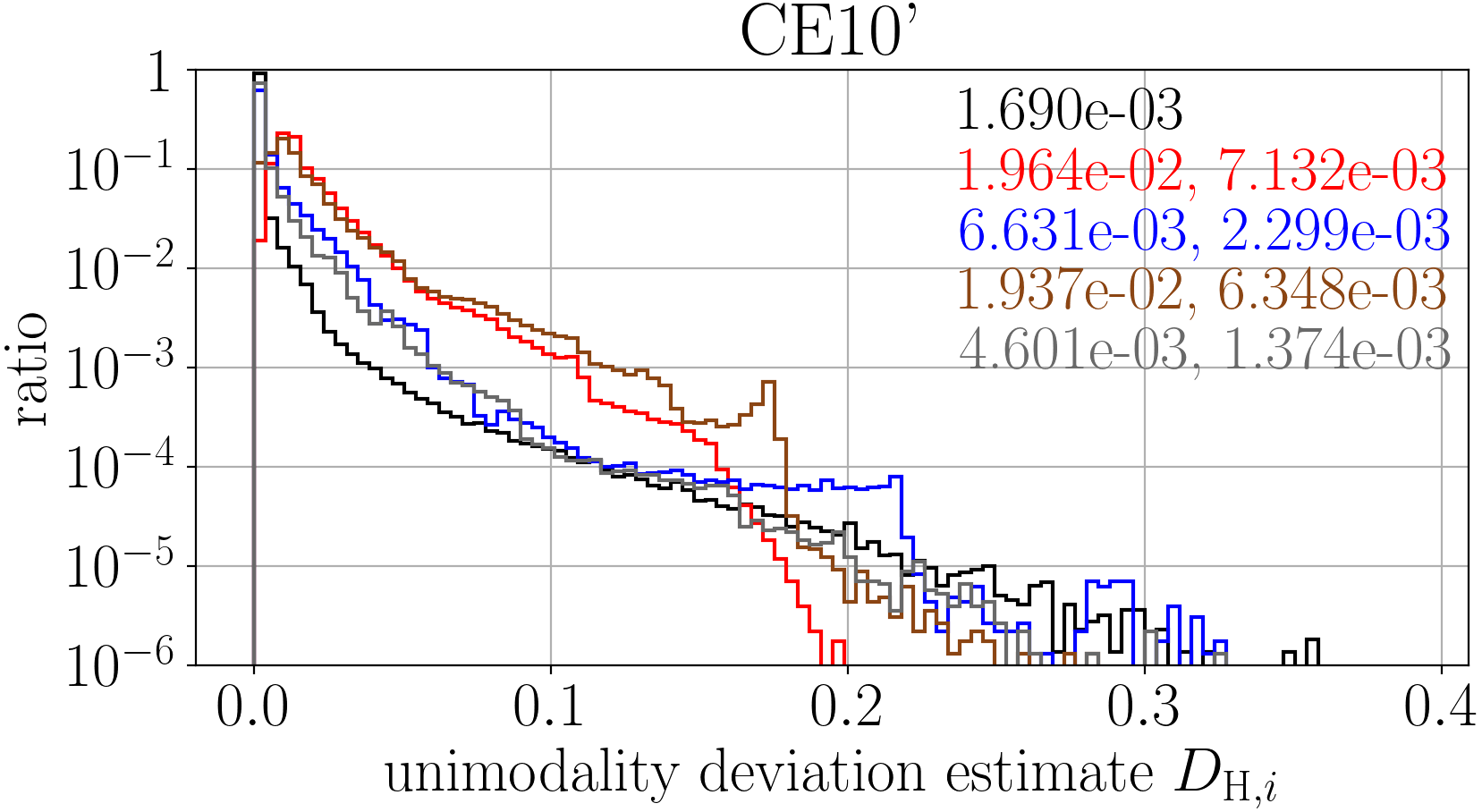}&
\includegraphics[height=1.54cm, bb=0 0 597 327]{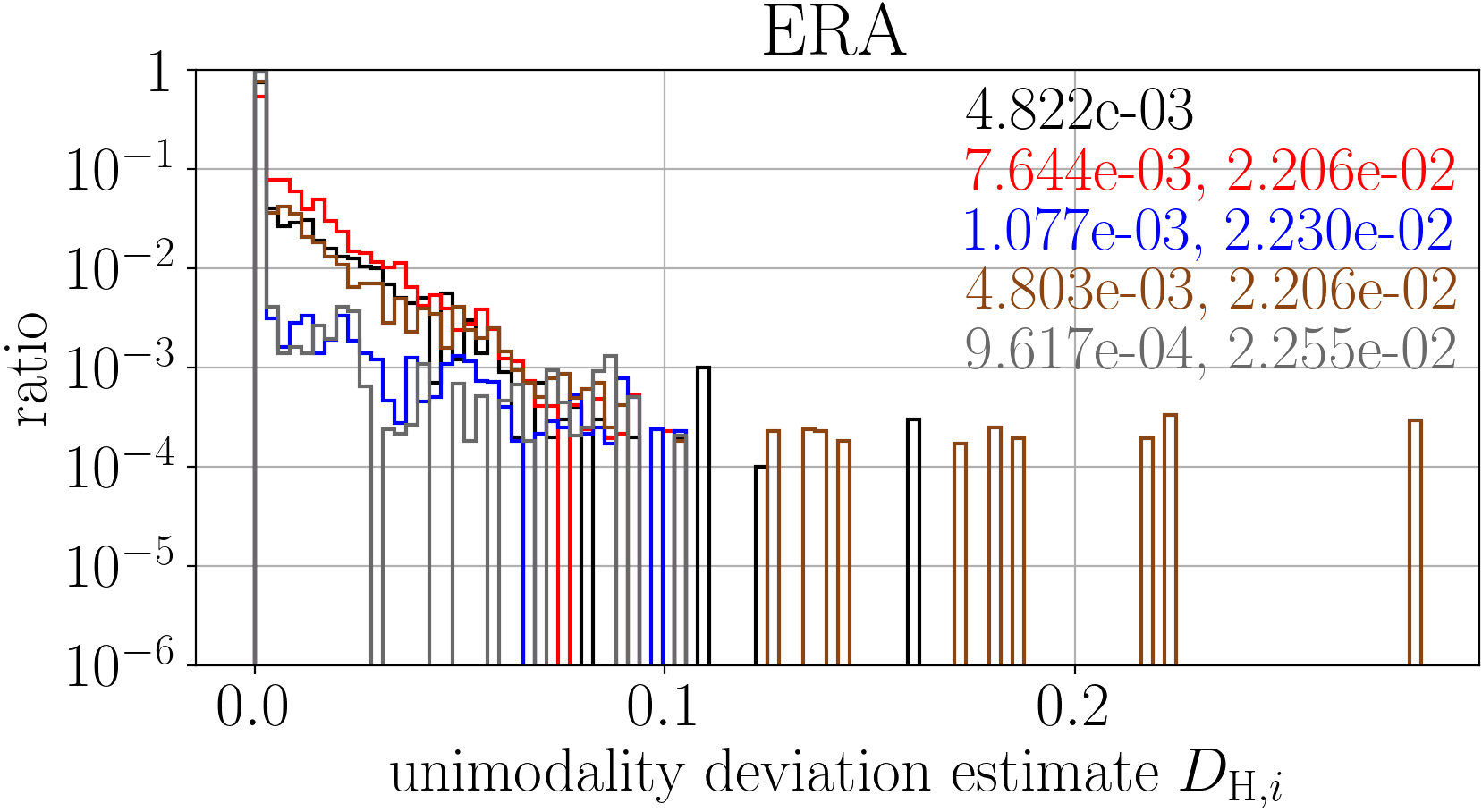}&
\includegraphics[height=1.54cm, bb=0 0 597 327]{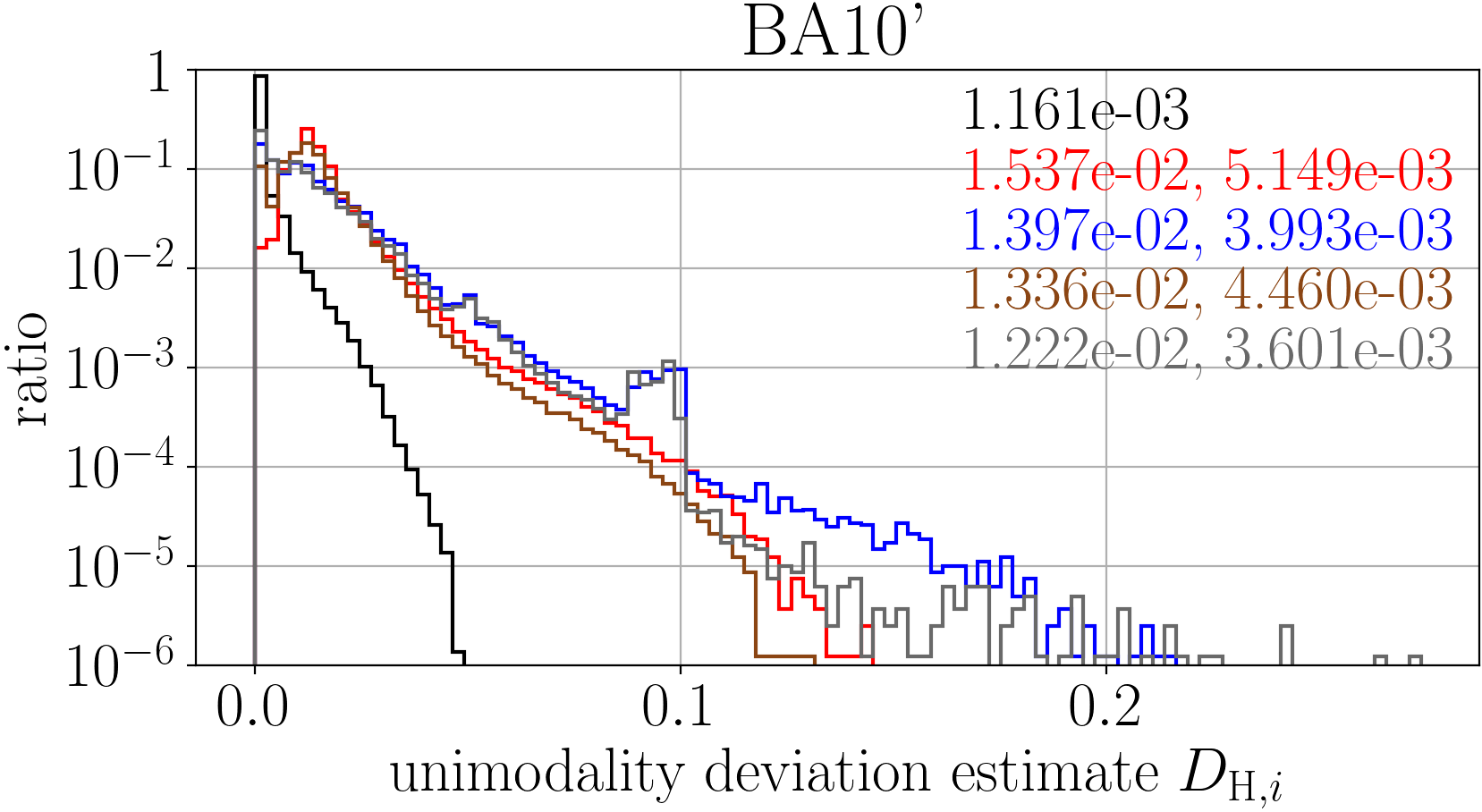}\\
\includegraphics[height=1.54cm, bb=0 0 597 327]{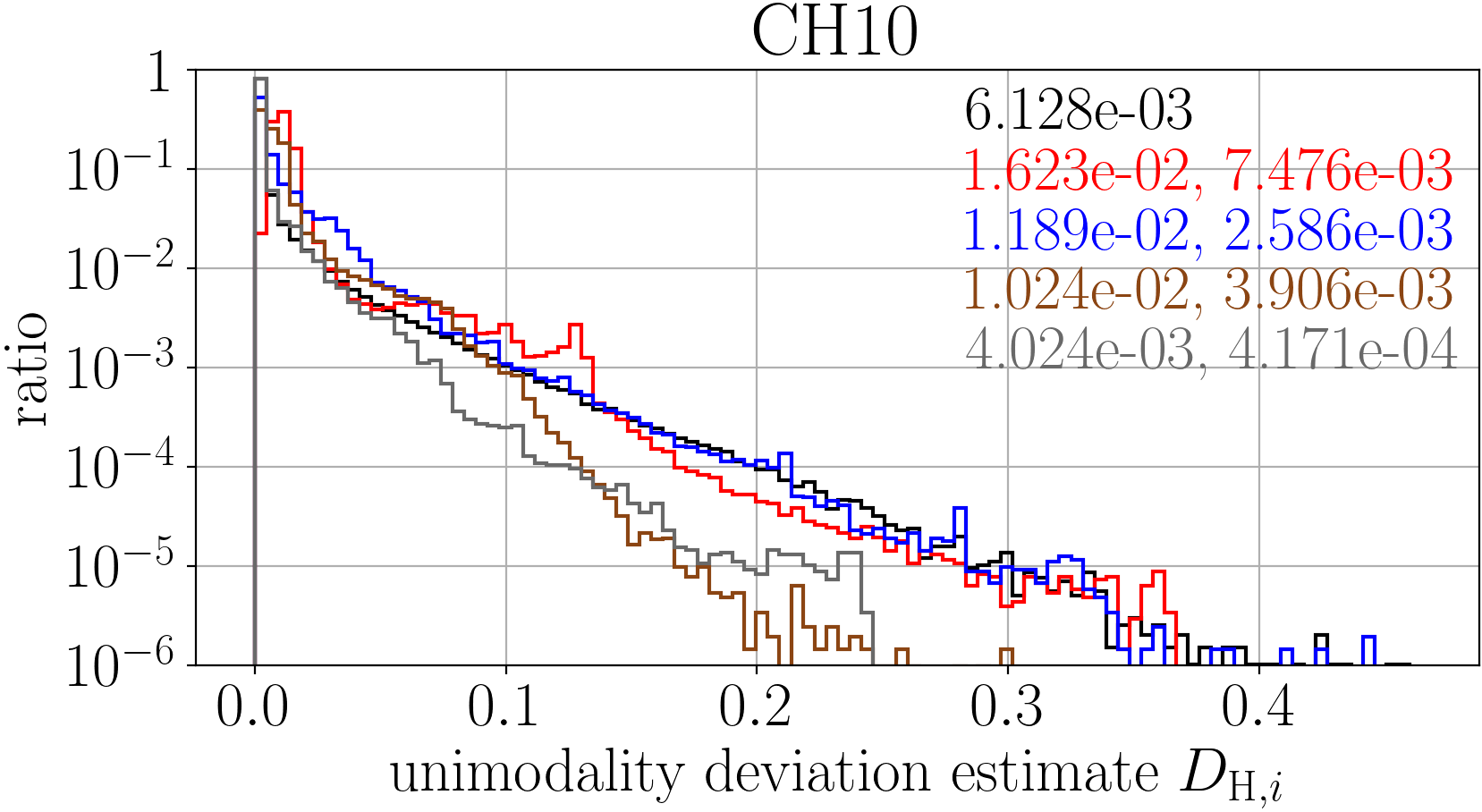}&
\includegraphics[height=1.54cm, bb=0 0 597 327]{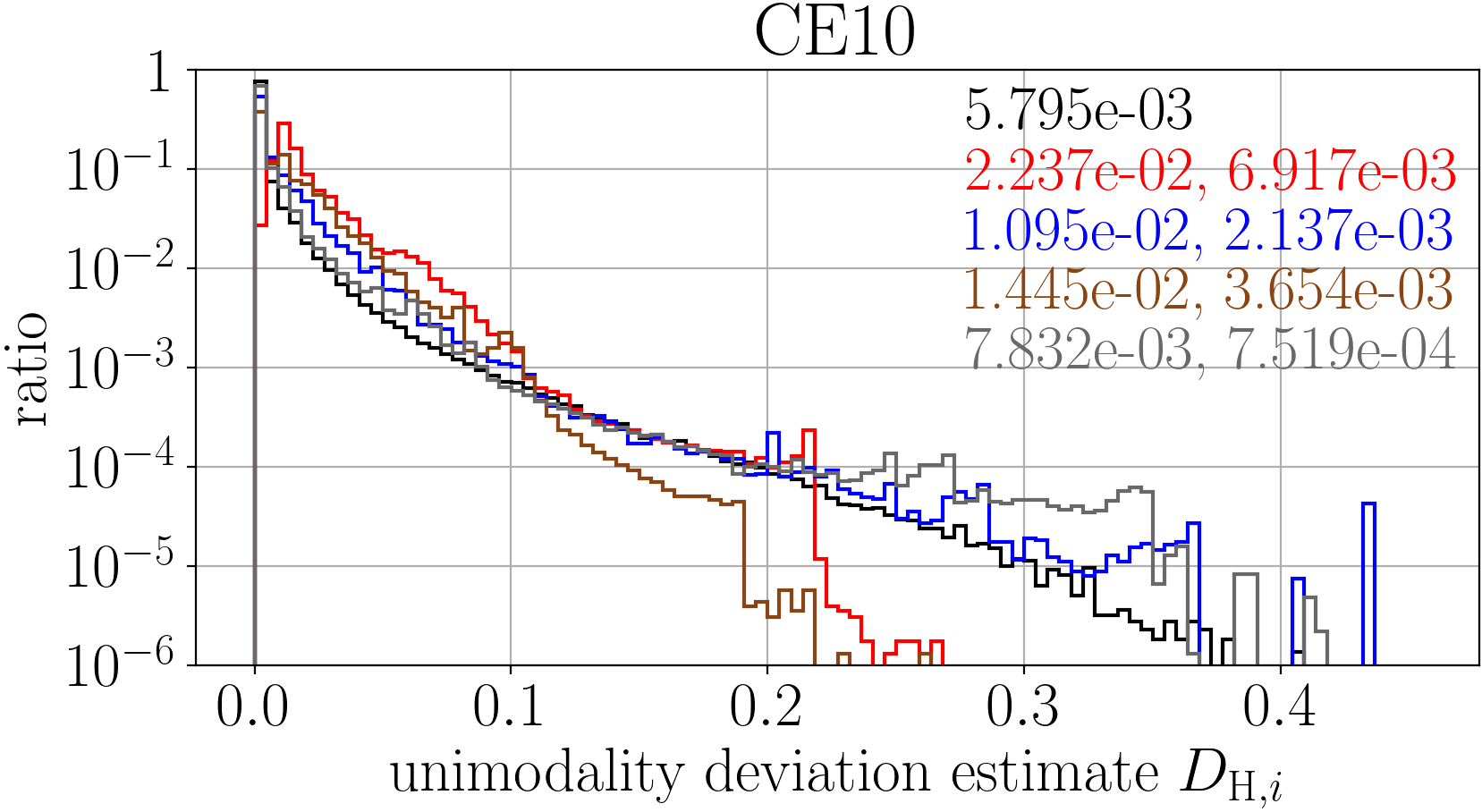}&
\includegraphics[height=1.54cm, bb=0 0 597 327]{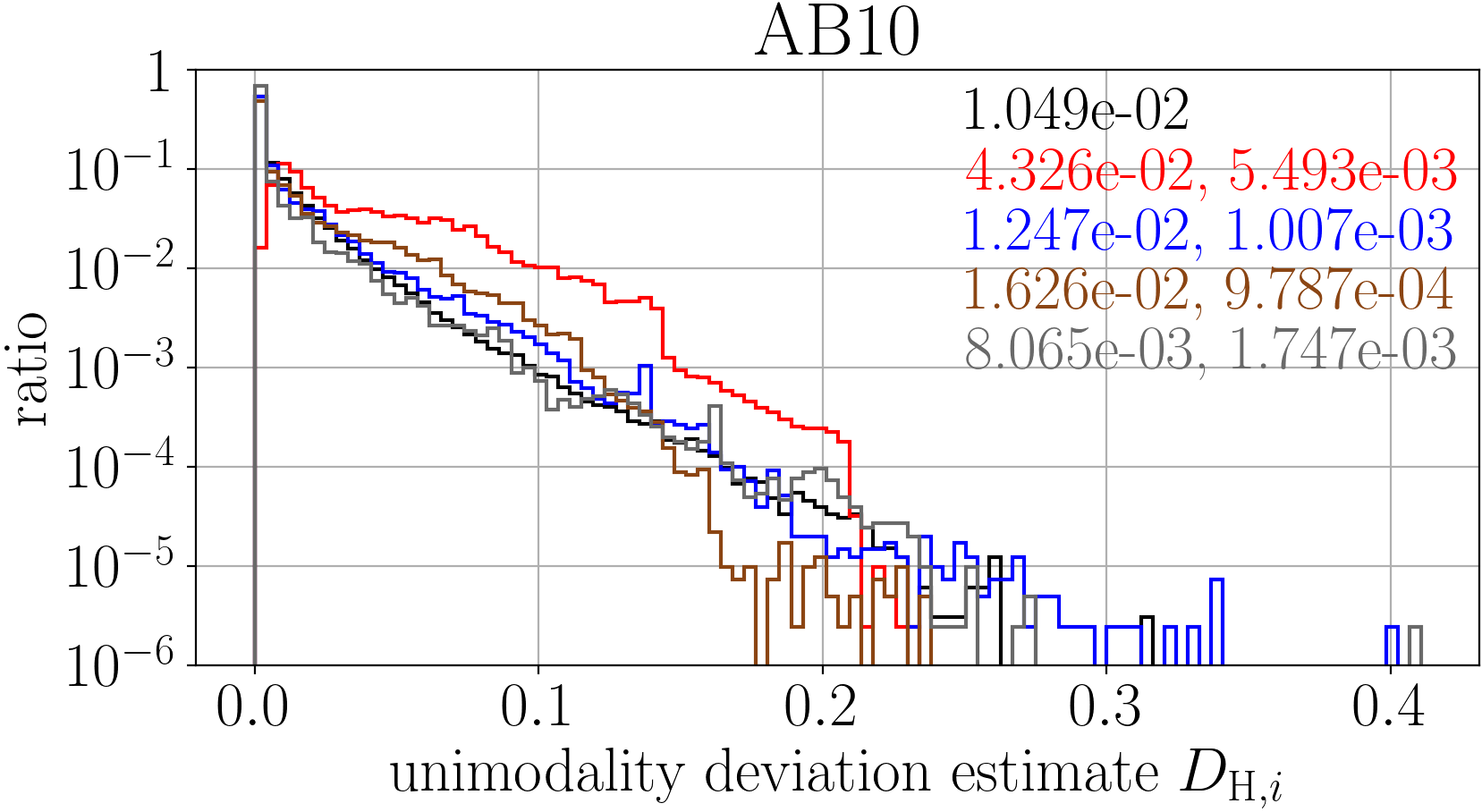}\\
\includegraphics[height=1.54cm, bb=0 0 597 327]{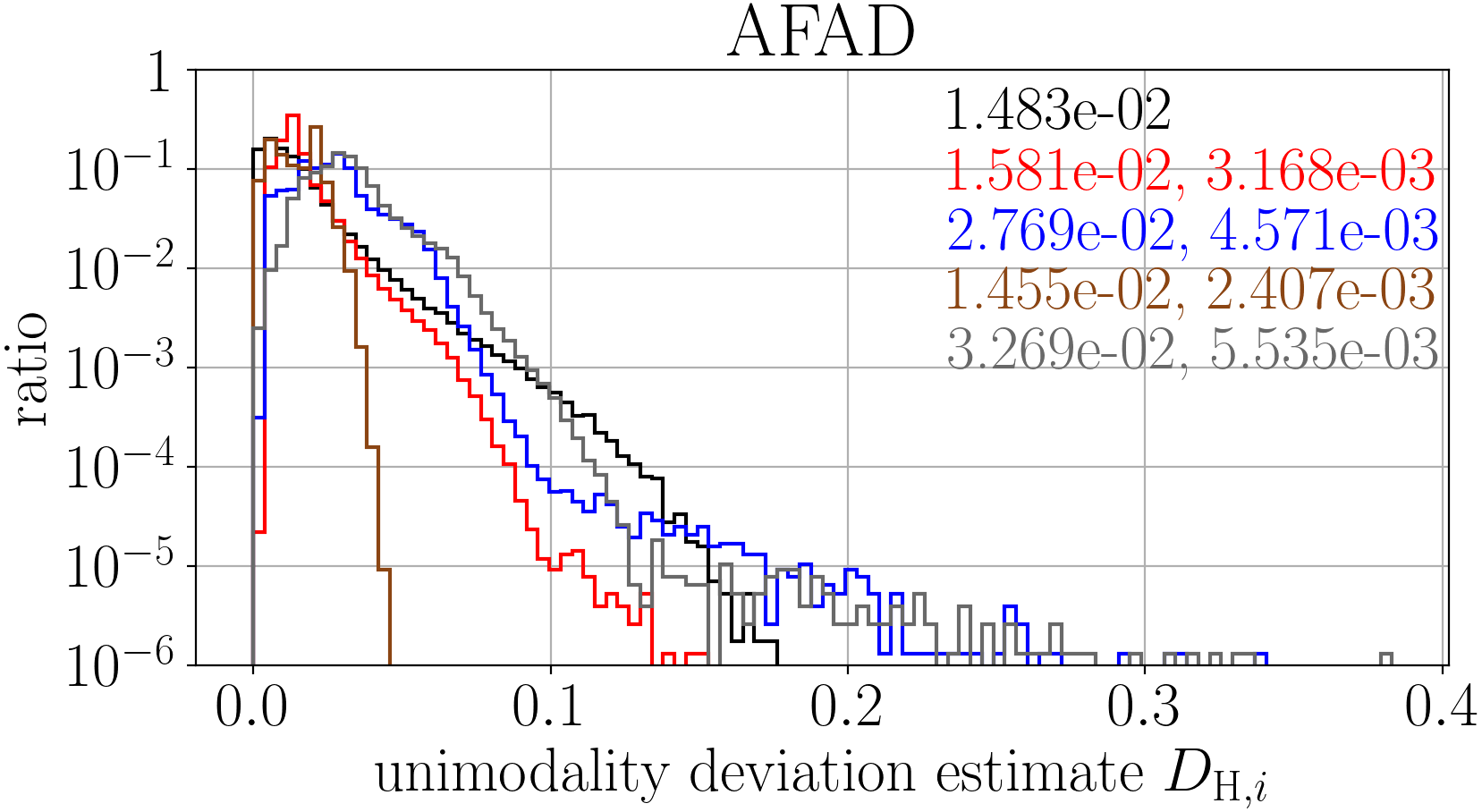}&
\includegraphics[height=1.54cm, bb=0 0 597 327]{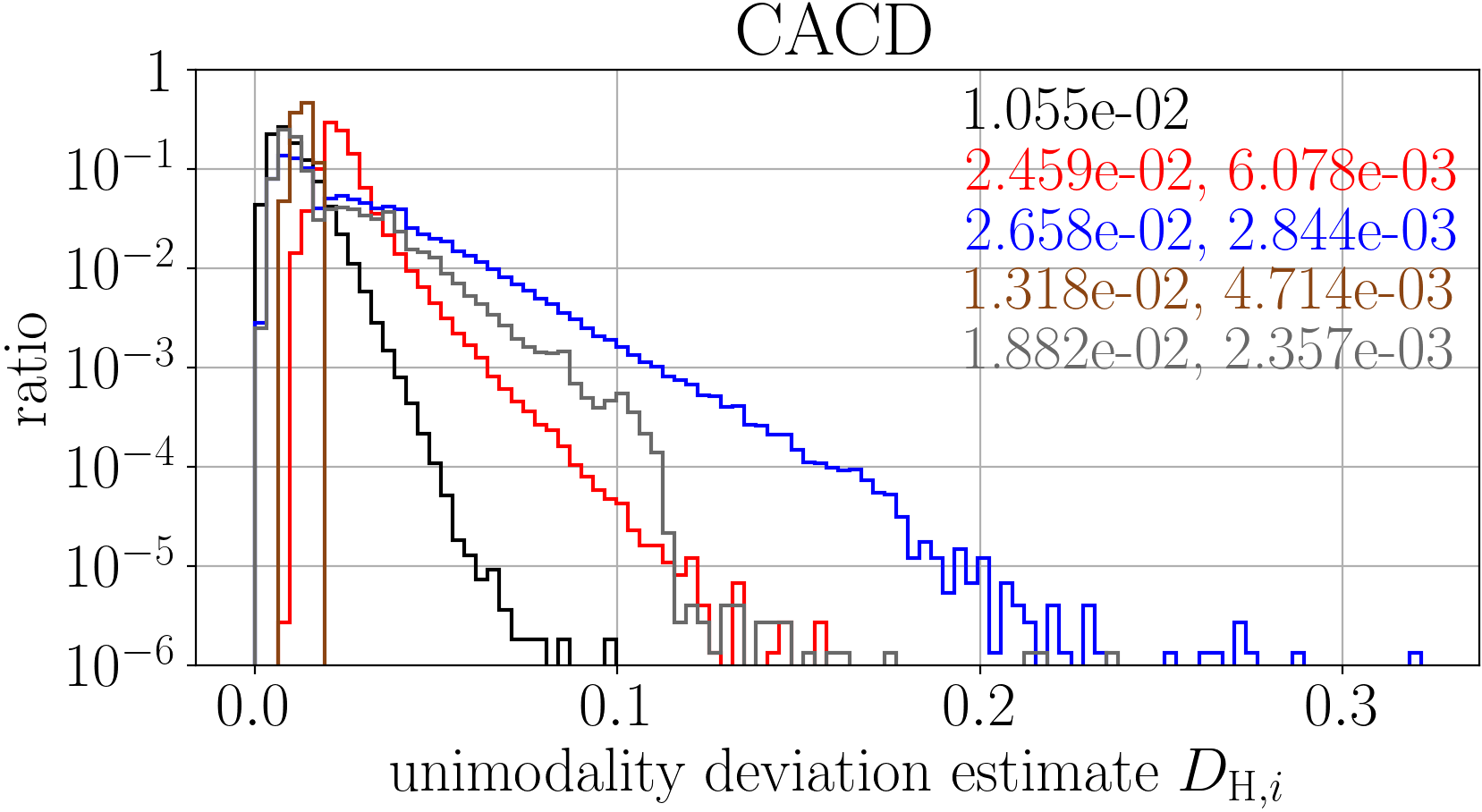}&
\includegraphics[height=1.54cm, bb=0 0 597 327]{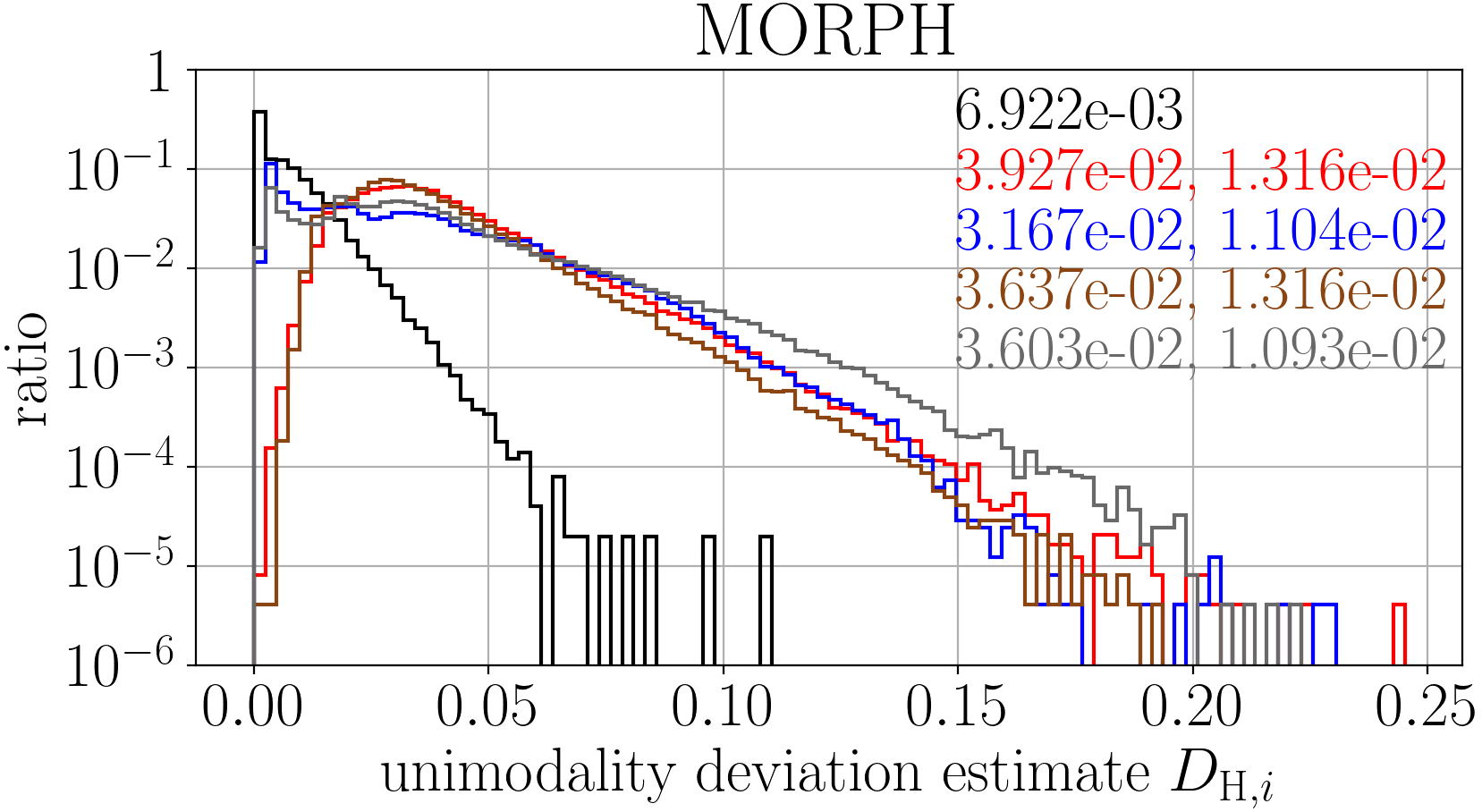}
\end{tabular}
\caption{%
Log-scaled histogram of 100-trial (for SB datasets) or 5-trial 
(for CV datasets) aggregation of estimates $D_{\rmH,i}$ of 
$D_\rmH((\Pr(Y=y|\bX=\bx_i))_{y\in[K]},\hat{\Delta}_{K-1})$
by the SL model with the largest $n_\tra$ (black) and 
SL (red), $\text{Mix}_{\mbox{\tiny\text{VSL,SL}}}$ (blue), 
$\text{SL}'$ (brown), $\text{Mix}'_{\mbox{\tiny\text{VSL,SL}}}$ 
(gray) models with the smallest $n_\tra$.
The mean of aggregation of estimates $D_{\rmH,i}$ (MDH)
and $L_1$ distance $\int_0^{\sqrt{2}}|f(u)-g(u)|\,du$ between 
the histogram $f$ of the SL model with the largest $n_\tra$ and 
the histogram $g$ of each model are shown 
in their color at the upper right corner of the figure
as `MDH, $L_1$ distance (except for $f$)'.}
\label{fig:Res-UD}
\end{figure}

\begin{table}[!t]
\centering%
\renewcommand{\tabcolsep}{0pt}%
\caption{%
A cell for `model A vs model B' and training data size $n_\tra$ shows 
the total (over 24 datasets) number of times that model A or model B 
wins the other model in Bonferroni correction 
with the significance level $0.05$ of Mann-Whitney U-test 
regarding the test error as `\#A wins, \#B wins'.
A model with larger win is better and highlighted with their color.}
\label{tab:Res-Errors}
{\footnotesize\renewcommand{\arraystretch}{0.75}%
\begin{tabular}{C{10pt}C{35pt}C{10pt}C{35pt}|C{20pt}C{20pt}C{20pt}C{20pt}C{20pt}C{20pt}}\toprule
\multicolumn{4}{c|}{Comparison$\backslash n_\tra$}&\multicolumn{6}{c}{smallest (left) to largest (right)}\\
\midrule
\multirow{8}{*}{\rotatebox{90}{NLL~~}}
& \tcb{$\text{Mix}_{\mbox{\tiny\text{VSL,SL}}}$}&vs&\tcr{SL}
& \tcb{18},0 & \tcb{19},0 & \tcb{14},0 & \tcb{7},3 & 0,\tcr{10} & 1,\tcr{14} \\
& \tcb{$\text{Mix}_{\mbox{\tiny\text{VSL,SL}}}$}&vs&\tcg{VSL}
& \tcb{14},0 & \tcb{12},0 & \tcb{8},0 & \tcb{5},0 & \tcb{3},1 & \tcb{7},1 \\
& \tcb{$\text{Mix}_{\mbox{\tiny\text{VSL,SL}}}$}&vs&\tcc{CL}
& \tcb{17},0 & \tcb{14},0 & \tcb{10},2 & 6,6 & 3,\tcc{5} & 3,\tcc{8} \\
& \tcb{$\text{Mix}_{\mbox{\tiny\text{VSL,SL}}}$}&vs&\tcm{POCL}
& \tcb{16},1 & \tcb{19},0 & \tcb{16},1 & \tcb{14},4 & \tcb{12},7 & \tcb{7},4 \\
& \tcb{$\text{Mix}_{\mbox{\tiny\text{VSL,SL}}}$}&vs&\tcy{POVSL}
& \tcb{11},2 & \tcb{13},4 & \tcb{12},3 & \tcb{12},1 & \tcb{11},0 & \tcb{11},0 \\
& \tcb{$\text{Mix}_{\mbox{\tiny\text{VSL,SL}}}$}&vs&\tco{$\text{SL}'$}
& \tcb{13},1 & \tcb{11},0 & \tcb{7},2 & 2,\tco{5} & 0,\tco{12} & 1,\tco{10} \\
& \tcb{$\text{Mix}_{\mbox{\tiny\text{VSL,SL}}}$}&vs&\tca{$\text{Mix}'_{\mbox{\tiny\text{VSL,SL}}}$}
& 0,\tca{6} & 0,\tca{2} & 0,\tca{1} & 0,0 & 0,0 & 0,0 \\
& \tco{$\text{SL}'$}&vs&\tca{$\text{Mix}'_{\mbox{\tiny\text{VSL,SL}}}$}
& 0,\tca{16} & 0,\tca{15} & 0,\tca{9} & 3,3 & \tco{8},0 & \tco{8},0 \\
\midrule
\multirow{8}{*}{\rotatebox{90}{MZE~~}}
& \tcb{$\text{Mix}_{\mbox{\tiny\text{VSL,SL}}}$}&vs&\tcr{SL}
& \tcb{19},0 & \tcb{17},0 & \tcb{10},0 & 1,\tcr{3} & 1,\tcr{3} & 4,4 \\
& \tcb{$\text{Mix}_{\mbox{\tiny\text{VSL,SL}}}$}&vs&\tcg{VSL}
& \tcb{5},0 & \tcb{7},0 & \tcb{10},0 & \tcb{11},0 & \tcb{8},0 & \tcb{6},0 \\
& \tcb{$\text{Mix}_{\mbox{\tiny\text{VSL,SL}}}$}&vs&\tcc{CL}
& \tcb{16},0 & \tcb{8},0 & \tcb{5},0 & \tcb{3},1 & \tcb{2},0 & 2,\tcc{3} \\
& \tcb{$\text{Mix}_{\mbox{\tiny\text{VSL,SL}}}$}&vs&\tcm{POCL}
& \tcb{18},1 & \tcb{17},0 & \tcb{16},0 & \tcb{14},1 & \tcb{14},1 & \tcb{10},2 \\
& \tcb{$\text{Mix}_{\mbox{\tiny\text{VSL,SL}}}$}&vs&\tcy{POVSL}
& \tcb{4},0 & \tcb{6},0 & \tcb{10},0 & \tcb{14},0 & \tcb{17},0 & \tcb{14},0 \\
& \tcb{$\text{Mix}_{\mbox{\tiny\text{VSL,SL}}}$}&vs&\tco{$\text{SL}'$}
& \tcb{15},0 & \tcb{7},0 & \tcb{3},0 & 0,\tco{1} & 1,\tco{3} & \tcb{4},3 \\
& \tcb{$\text{Mix}_{\mbox{\tiny\text{VSL,SL}}}$}&vs&\tca{$\text{Mix}'_{\mbox{\tiny\text{VSL,SL}}}$}
& 0,\tca{2} & 0,0 & \tcb{1},0 & 0,0 & 0,0 & 1,1 \\
& \tco{$\text{SL}'$}&vs&\tca{$\text{Mix}'_{\mbox{\tiny\text{VSL,SL}}}$}
& 0,\tca{15} & 0,\tca{10} & 0,\tca{4} & \tco{3},1 & \tco{3},0 & \tco{3},1 \\
\midrule
\multirow{8}{*}{\rotatebox{90}{MAE~~}}
& \tcb{$\text{Mix}_{\mbox{\tiny\text{VSL,SL}}}$}&vs&\tcr{SL}
& \tcb{20},0 & \tcb{16},0 & \tcb{10},0 & \tcb{3},2 & 0,\tcr{5} & 1,\tcr{5} \\
& \tcb{$\text{Mix}_{\mbox{\tiny\text{VSL,SL}}}$}&vs&\tcg{VSL}
& \tcb{3},0 & \tcb{4},0 & \tcb{2},0 & \tcb{5},0 & \tcb{4},0 & \tcb{7},0 \\
& \tcb{$\text{Mix}_{\mbox{\tiny\text{VSL,SL}}}$}&vs&\tcc{CL}
& \tcb{11},0 & \tcb{11},0 & \tcb{7},0 & \tcb{2},0 & 1,1 & 1,1 \\
& \tcb{$\text{Mix}_{\mbox{\tiny\text{VSL,SL}}}$}&vs&\tcm{POCL}
& \tcb{6},0 & \tcb{8},0 & \tcb{12},0 & \tcb{8},0 & \tcb{9},1 & \tcb{6},2 \\
& \tcb{$\text{Mix}_{\mbox{\tiny\text{VSL,SL}}}$}&vs&\tcy{POVSL}
& \tcb{6},1 & \tcb{4},1 & \tcb{5},1 & \tcb{8},1 & \tcb{9},0 & \tcb{11},0 \\
& \tcb{$\text{Mix}_{\mbox{\tiny\text{VSL,SL}}}$}&vs&\tco{$\text{SL}'$}
& \tcb{14},0 & \tcb{8},0 & \tcb{3},0 & 0,\tco{1} & 0,\tco{3} & 4,\tco{5} \\
& \tcb{$\text{Mix}_{\mbox{\tiny\text{VSL,SL}}}$}&vs&\tca{$\text{Mix}'_{\mbox{\tiny\text{VSL,SL}}}$}
& 0,0 & 0,0 & 0,0 & 0,0 & 0,0 & 0,0 \\
& \tco{$\text{SL}'$}&vs&\tca{$\text{Mix}'_{\mbox{\tiny\text{VSL,SL}}}$}
& 0,\tca{14} & 0,\tca{10} & 0,\tca{5} & 1,\tca{2} & \tco{1},0 & 1,\tca{3} \\
\midrule
\multirow{8}{*}{\rotatebox{90}{MSE~~}}
& \tcb{$\text{Mix}_{\mbox{\tiny\text{VSL,SL}}}$}&vs&\tcr{SL}
& \tcb{20},0 & \tcb{18},0 & \tcb{8},1 & 1,1 & 0,\tcr{6} & 0,\tcr{5} \\
& \tcb{$\text{Mix}_{\mbox{\tiny\text{VSL,SL}}}$}&vs&\tcg{VSL}
& \tcb{9},0 & \tcb{4},0 & \tcb{3},1 & \tcb{3},1 & \tcb{3},0 & \tcb{5},0 \\
& \tcb{$\text{Mix}_{\mbox{\tiny\text{VSL,SL}}}$}&vs&\tcc{CL}
& \tcb{13},0 & \tcb{10},0 & \tcb{4},0 & \tcb{2},0 & 1,\tcc{2} & 3,\tcc{5} \\
& \tcb{$\text{Mix}_{\mbox{\tiny\text{VSL,SL}}}$}&vs&\tcm{POCL}
& \tcb{8},0 & \tcb{7},0 & \tcb{7},1 & \tcb{5},2 & \tcb{6},2 & \tcb{4},2 \\
& \tcb{$\text{Mix}_{\mbox{\tiny\text{VSL,SL}}}$}&vs&\tcy{POVSL}
& \tcb{6},0 & \tcb{5},0 & \tcb{5},0 & \tcb{4},3 & \tcb{4},1 & \tcb{7},1 \\
& \tcb{$\text{Mix}_{\mbox{\tiny\text{VSL,SL}}}$}&vs&\tco{$\text{SL}'$}
& \tcb{13},1 & \tcb{7},1 & \tcb{3},2 & 0,\tco{2} & 0,\tco{2} & 2,2 \\
& \tcb{$\text{Mix}_{\mbox{\tiny\text{VSL,SL}}}$}&vs&\tca{$\text{Mix}'_{\mbox{\tiny\text{VSL,SL}}}$}
& 0,\tca{1} & 0,0 & 0,0 & 0,0 & 0,0 & 0,0 \\
& \tco{$\text{SL}'$}&vs&\tca{$\text{Mix}'_{\mbox{\tiny\text{VSL,SL}}}$}
& 0,\tca{15} & 1,\tca{9} & 0,\tca{3} & \tco{2},0 & \tco{2},0 & 1,\tca{2} \\
\bottomrule\end{tabular}}
\end{table}

Figure~\ref{fig:Res-UD} shows the evaluation of 
the UD for the estimated CPD based on the SL and 
$\text{Mix}_{\mbox{\tiny\text{VSL,SL}}}$ models.
Here, we consider the estimated CPD based on the SL 
model with the largest $n_\tra$ to be the true CPD
(since the underlying CPD for real-world data is unknown).
With the small $n_\tra$, the SL model seems to give 
a CPD estimate that largely deviates from `unimodal',
while the $\text{Mix}_{\mbox{\tiny\text{VSL,SL}}}$ model 
was able to reduce such deviation (see MDH in Figure~\ref{fig:Res-UD})
and yield a CPD estimate that is closer to the true CPD with respect to the UD
(see $L_1$ distance in Figure~\ref{fig:Res-UD}).
These results lend credence to the success of 
the $\text{Mix}_{\mbox{\tiny\text{VSL,SL}}}$ model 
for modeling of low-UD ordinal data.

Figure~\ref{fig:Res-NLL} and Table~\ref{tab:Res-Errors} 
show performance comparison of the 6 tried models.
Although the $\text{Mix}_{\mbox{\tiny\text{VSL,SL}}}$ model 
was inferior to the unconstrained SL and CL models
when the training data size is large enough,
it could work better when the training data size is small.
More importantly, the approximately unimodal 
$\text{Mix}_{\mbox{\tiny\text{VSL,SL}}}$ model
tended to perform better than the VSL, POCL, 
and POVSL models for tried $n_\tra$.
This result is likely because the approximately unimodal likelihood model 
can adequately represent many ordinal data and reduce the bias.

\section{Experiments II}
\label{sec:ExperimentsII}
\subsection{Experimental Purposes}
\label{sec:Purposes2}
As an approach to leverage the unimodality of ordinal data, 
we can employ UPRL that devises a learning procedure 
(see Section~\ref{sec:UPRL}), in addition to 
the approach of devising a likelihood model.
We were interested in comparing the approximately 
unimodal likelihood model with UPRL 
and in the compatibility of these approaches.
Thus, we performed experiments (Experiments II)
to study these aspects.

\subsection{Experimental Settings}
\label{sec:Settings2}
The experimental settings for Experiments II 
were almost identical to those for Experiments I 
(see Section~\ref{sec:Settings}).
In Experiments II, we tried the SL model and 
the $\text{Mix}_{\mbox{\tiny\text{VSL,SL}}}$ model
with the mixture rate $r=\bar{r}$ selected in 
the validation process regarding the NLL in Experiments I
(supplement shows similar results with $r=\bar{r}-0.05,\bar{r}+0.05$),
and trained these models by Adam optimization with 
the regularized NLL with the regularizer \eqref{eq:alb21} with $\delta=0.05$
and regularization parameter $\lambda=10^{-4},10^{-3.5},\ldots,10^4$
as the objective function.
We selected $\lambda$ via the validation process,
and we denote resulting models $\text{SL}'$ 
and $\text{Mix}'_{\mbox{\tiny\text{VSL,SL}}}$ models.

\subsection{Experimental Results}
\label{sec:Results2}
As Figure~\ref{fig:Res-UD} shows, UPRL was effective 
in providing a CPD estimate that is closer to be unimodal.
Also, as claimed in the UPRL studies 
\cite{belharbi2020,albuquerque2021ordinal,albuquerque2022quasi},
we can confirm in Figure~\ref{fig:Res-NLL} that 
the previous UPRL method, $\text{SL}'$ model, was better
than the $\text{SL}$ model for small-size training data.
On the other hand, in light of the purposes of Experiments II, 
the following findings are more important:
Figure~\ref{fig:Res-NLL} and Table~\ref{tab:Res-Errors} indicate 
that the $\text{Mix}_{\mbox{\tiny\text{VSL,SL}}}$ model outperformed 
the $\text{SL}'$ model, and the $\text{Mix}'_{\mbox{\tiny\text{VSL,SL}}}$ 
model outperformed the $\text{Mix}_{\mbox{\tiny\text{VSL,SL}}}$ 
model especially with respect to the NLL,
for small-size training data for which MAUL- or UPRL-based methods are effective.
While the MAUL model and UPRL each contributed 
to improve the predictive performance on their own, 
they performed even better when combined
(albeit at the expense of additional computational costs
for the sake of hyper-parameter search).

\section{Conclusion}
\label{sec:Conclusion}
In this study, we observed that many real-world ordinal data are 
not only high-UR \cite{yamasaki2022unimodal} but also low-UD 
(see Table~\ref{tab:Data-Prop} and Figure~\ref{fig:Data-Unimodality}).
On the ground of this observation, for statistical modeling 
of low-UD ordinal data, we proposed the MAUL model,
in which the deviation of a predicted CPD from the nearest unimodal CPD 
is guaranteed to be bounded from above by a value 
determined with a user-specified mixture rate of the model.
We expect that the MAUL model is effective with small-size training data 
where the variance dominates the prediction error of the CPD 
because it reduces the variance compared with unconstrained models 
thanks to the restricted representation ability, 
and keeps the bias small for low-UD data 
or rather reduces the bias compared with unimodal likelihood models.
Through numerical experiments, we verified that this likelihood model 
and associated OR method may provide superior prediction performance 
in the conditional probability estimation task of ordinal data 
and OR tasks especially when the training data size is small.

Some ordinal data may have a CPD that is 
unimodal or approximately unimodal at some points 
and bipolar (with large $\Pr(y=1|\bX=\bx)$ and 
$\Pr(y=K|\bX=\bx)$ and small others) at other points
as in \cite{chen1995response}.
The proposed MAUL model can potentially represent 
such data by using a large mixture rate, 
but which will diminish advantages of 
restricting the representation ability.
It is a possible direction for future research to
develop models appropriate for such data.
We are also interested in checking the unimodal hypothesis 
and in application of unimodal or approximately unimodal likelihood 
for various ordinal regression tasks.

\section*{Acknowledgment}
This work was supported by JSPS KAKENHI Grant Number JP24K23856.

\bibliographystyle{IEEEtran}
\bibliography{machine_learning, ordinal_regression, dataset, isotonic}


\textbf{Refer to the IEEE Publishing site for supplementary material of this paper.}

\end{document}